# Solving Combinatorial Optimization problems with Quantum inspired Evolutionary Algorithm Tuned using a Novel Heuristic Method

Nija Mani, Gursaran, and Ashish Mani

Nija Mani was with Department of Mathematics, Dayalbagh Educational Institute (Deemed University), Dayalbagh, Agra, India (e-mail: nijam@acm.org ).

Gursaran is with Department of Mathematics, Dayalbagh Educational Institute (Deemed University), Dayalbagh, Agra, India (e-mail: gursaran.db@gmail.com).

Ashish Mani was with University Science Instrumentation Centre, Dayalbagh Educational Institute (Deemed University), Dayalbagh, Agra, India. He is now with Department of Electrical and Electronics Engineering, Amity School of Engineering and Technology, Amity University Uttar Pradesh (e-mail: ashish.mani@ieee.org).

Abstract:

Quantum inspired Evolutionary Algorithms were proposed more than a decade ago and have been employed for solving a wide range of difficult search and optimization problems. A number of changes have been proposed to improve performance of canonical QEA. However, canonical QEA is one of the few evolutionary algorithms, which uses a search operator with relatively large number of parameters. It is well known that performance of evolutionary algorithms is dependent on specific value of parameters for a given problem. The advantage of having large number of parameters in an operator is that the search process can be made more powerful even with a single operator without requiring a combination of other operators for exploration and exploitation. However, the tuning of operators with large number of parameters is complex and computationally expensive. This paper proposes a novel heuristic method for tuning parameters of canonical QEA. The tuned QEA outperforms canonical QEA on a class of discrete combinatorial optimization problems which, validates the design of the proposed parameter tuning framework. The proposed framework can be used for tuning other algorithms with both large and small number of tunable parameters.

Keywords: Meta-heuristic, Orthogonal Arrays, Design of Experiments, Taguchi based method.

Quantum inspired Evolutionary Algorithms (QEA) are population based meta-heuristics that draw inspiration from quantum mechanical principles to improve search and optimization capabilities of Evolutionary Algorithms (EAs). QEA has been applied to solve a wide variety of problems ranging from Automatic Color Detection [1], Image Segmentation [2], Bandwidth [3], Circuit testing [4], Software Testing [5], Economic Dispatch [6], [7], Engineering Design Optimization [8], design of digital filters [9] and process optimization [10] etc.

The potential advantages of parallelism offered by quantum computing [11] and simultaneous evaluation of all possible represented states, have led to the development of approaches for integrating some aspects of quantum computing with evolutionary computation [12]. Most hybridizations have focused on designing algorithms that would run on conventional computers and not on quantum computers and are most appropriately classified as "quantum inspired". The first such attempt was made by Narayan and Moore [13] which used quantum parallel world interpretation to define a quantum inspired genetic algorithm to be run on a classical computer, subsequently, a number of other hybridizations also have been proposed. Han and Kim [14] proposed a popular model of quantum inspired evolutionary algorithm (QEA), that used a Q-bit as the smallest unit of information and a Q-bit individual as a string of Q-bits rather than binary, numeric or symbolic representations. Results of experiments showed that QEA performed well, even with a small population, without suffering from premature convergence as compared to the conventional genetic algorithm. Experimental studies have also been reported by Han and Kim to identify suitable parameter settings for the algorithm and enhancements have been proposed with new termination criterion and operators [15].

The QEA proposed by Han and Kim [14] is termed as Canonical QEA [14] as the basic structure proposed for QEA has remained unaltered in subsequent modifications [16]. The canonical QEA has three main constituents which differentiates it from other class of Evolutionary Algorithms. These are Q-bit representation, Measurement Operator for generating binary string from Q-bit string and Rotation Gate as Variation Operator to update the Q-bits. Further, it also has an island or coarse grained population model [17].

There have been many attempts to further improve canonical QEA by incorporating crossover and mutation operators in Li et al. [18]. The crossover and mutation operators are termed as quantum crossover and quantum mutation operators as they vary the Q-bit individuals instead of binary strings. Moreover, attempts have also been made by Zhang et al. [16] to modify canonical QEA by using a catastrophe operator and a novel update method for Q-gates. Further, hybridization of QEA has been attempted with Canonical Genetic Algorithm (CGAs) [19], immune algorithms [20] and particle swarm optimization (PSO) [21] to improve their performance. A number of modifications have also been proposed to improve the applicability and performance of canonical QEA, which range from modifications of existing operators like in case of real observation QEA [6], population structure as in case of vQEA [22] and variation operator [23] to introducing new variation operators like Quantum Crossover [24], Quantum Mutation [25] and Neighborhood operators [23], [26].

Most of the efforts on improving canonical QEA have focused on Variation operators i.e. either by modifying the existing rotation gate [26] or by introducing new type of operators through hybridization [21].

Canonical QEA has a well-designed Variation operator with eight rotation angle parameters to explore the genotype space comprehensively by obtaining suitable feedback from the objective function value. The rotation angles in variation operator, the migration condition and local neighborhood are taken to be design parameters and have to be chosen appropriately for the problem at hand. It is felt that these parameters may have a strong influence on the performance of a QEA, but studies on these have not been extensively reported in the literature. There are eleven design parameters in canonical QEA viz., eight rotation angles, population size, group size and global migration and they require fine tuning to improve their performance on specific problems. This has motivated investigation into parameter tuning of QEA and an attempt has been made to improve the performance of canonical QEA by effective and efficient tuning of parameters in Rotation Gate as well as other important parameters like number of groups, migration period and population size.

The motivation for this approach is twofold; first, the rotation gate has eight parameters, which are problem specific, so it is inherently difficult to set eight parameters by any ad-hoc mechanism. Second, the rotation gate with its eight parameters appears to be a powerful variation operator, at least in principle, which may provide for good search capability in wide variety of problems. Further, the rotation gate is mostly left unexplored in majority of efforts. These employ the same set of parameter values as suggested by Han and Kim [14] more than a decade ago.

Han and Kim [14] had intuitively assigned the value of eight rotation angles and then performed experiments by choosing three levels 0, $+0.005\pi$, and $-0.005\pi$ for eight rotation angle parameters on the knapsack problem of size 100. Though, the number of experiments were factorial i.e. $3^8$, however, the rest of the parameters like population size and migration periods, etc. were kept constant. After having verified the effect of rotation angles, the other parameters like population size and migration periods etc. were studied independently. One of the reasons for the ad-hoc tuning of parameters is the relatively large number of parameters, which need to be tuned.

Parameter Tuning is an important part in the designing process of an Evolutionary Algorithm as it affects efficacy of the search process. It is said that most of the effort in designing an EA for solving a set of problems is spent in parameter tuning [27]. It is a difficult optimization problem in itself as it is usually poorly structured, ill-defined and complex in nature [27], [28]. Further, the best set of parameter values can be guaranteed to be found only after exhaustive search in the entire parameter-space, however, such a strategy may not be practically feasible due to the huge amount of resources and time involved. Parameter tuning is essentially a problem of Design of Experiments (DOE) and analysis of results obtained from the experiments to arrive at the best parameter vector. Traditionally, there are four well known strategies for design of experiments viz., Ad-hoc, Factorial, Fractional Factorial and Random design of experiments [29]. In Ad-hoc experimentation, the design of experiments is guided by intuition and is subsequently verified by limited set of experiments. This technique has received wide popularity in EA due to two specific reasons. First, EAs are designed to give good solutions quickly for difficult optimization problems, thus if an EA can give better solutions than the existing ones, then it is considered as successful and acceptable. Therefore, there is no compulsion to study the behavior of EA in the entire parameter space as long as EA can find better solutions. Secondly, the structure of EA and the problem being solved, usually give some insight into the possible parameter values to an experienced EA designer. Thus, limited experiments using ad-hoc approach has been the most popular method amongst EA designers. However, there is lot of subjectivity in ad-hoc approach and as has been shown that a well-designed parameter tuning method can perform better than the claimed best EA [30], so parameter tuning should be done using a structured approach [27].

A number of attempts have been made for designing methods for parameter tuning in EAs [27], which have been developed to provide good parameter values within reasonable cost. These methods have been categorized as Sampling Methods, Model Based Methods, Screening Methods and Meta-Evolutionary Methods in [27]. They have been further divided into non-iterative and iterative sub categories, with iterative versions outperforming non-iterative versions. Similarly, they have also been categorized as Single Stage and Multi-Stage version, with multi-stage being more effective. These methods tend to use different designs of experiments during different stages, there by leading to a hybrid design of experiments.

Sampling methods use fractional factorial design of experiments to reduce the number of experiments as in case of the Taguchi's Orthogonal Arrays [31]. Statistical analysis is performed on the result of the experiments to compute the parameter vector that may give the best result. However, there is no guarantee of finding the optimal set of parameters as the emphasis is on reducing the effort. These methods require finding a set of levels for each parameter, which can be done in an ad-hoc manner or by adding an initialization stage to automatically find the levels with which they begin sampling [32], [33]. Further, they have been augmented by iterating the sampling process to refine the parameter vectors [32]. Calibra [34] uses full factorial design of experiments in its initial stage and fractional factorial design of experiment in its later stage.

Model based Methods generally use Sampling methods as a starting point to construct a model of the search process of EA [35]. The model helps in predicting the utility of the parameter vector in solving the problem. However, they appear more amenable in studying the nature of EA in terms of its characteristics like tuning ability and robustness to changes in problem specification. Single stage methods are generally unable to find the best parameter vector. The iterative multi stage methods [36] are computationally expensive and they rely heavily on the accuracy of the model. It is well known that there is always a trade-off between computational effort and accuracy of the model in such an effort. Further, if the objective is to find the best set of parameter values for solving a set of problems, then first finding an accurate model of a Stochastic process and subsequently using it to predict the best parameter vector appears to be indirect and errors often accumulate.

Screening methods try to reduce the number of experiments by using statistical measures to test only a small subset from a large set of parameter vectors [37], [38]. F-Race and its improved version, iterated F-Race, use the non-parametric Friedman test as a family-wise test, i.e. it checks whether there is evidence that at least one of the configurations is significantly different from the rest [39]. However, the implementation problem with such methods is due to the stochastic nature of EA. That is only after running the EA with a parameter vector for sufficient number of times with different set of random numbers, can any conclusion be drawn about its effectiveness. There are EAs which initially focus on exploration of the search space and only at later stages they exploit the good regions. Moreover, the EA instance with competing parameter vectors often performs differently with different set of random numbers, especially in case of difficult problems, therefore a fair comparison can be made only after each EA instance has been run to completion (i.e. same termination criteria) for about 30 to 50 times with different set of random numbers. Thus the idea of screening methods, which appear good in concept, would either not do just comparisons or not be able to reduce the effort, especially in case of difficult problems. However, there is reported evidence in favor of such methods and application of non-parametric statistical testing is a good approach to eliminate non-performers.

Meta Evolutionary Algorithms have also been used for parameter tuning as the nature of problem of finding high utility parameter vectors is similar to the problems solved by Evolutionary Algorithms. They have been quite successful as reported in [30]. However, the very idea of using Meta EA runs into the proverbial "Chicken or the Egg Causality Dilemma" i.e. an un-tuned EA is being used for parameter tuning of another EA or an EA tuned by some other mechanism is being used for parameter tuning of another EA. Thus, conceptually, it is a two level method, which is going to require double the effort in parameter tuning. However, examples of Meta EA like REVAC [40] have been successful in practice.

The generic process of solving the parameter tuning problem by the above discussed methods, in general, involves the following steps:

a) Selection of a set of Parameter Vectors
b) Experiment with selected Set of Parameter Vectors
c) Analysis of the Experimental Result
d) If Non Iterative OR Stopping_Criteria met, Output the Best Vector Found

e) If Multi_Stage & Stage_Transition_Criteria met, Change Selection and / or the Analysis process
f) If Iterative, Selection of a new set of Parameter Vectors based on Analysis of the Experimental Result
g) Go to (b)

The general conclusions that can be drawn from the reported efforts are that random initial selection of parameter set is more effective than a fixed set, Iterative and multi stage methods are better than non-iterative and single stage methods. An interesting observation is that iterative parameter tuning methods can be cast into meta-heuristic framework. The methods differ in statistical analysis and subsequent selection of parameter vectors. Further, there is no guarantee in any of the methods for arriving at the best parameter vector. All of them try to find a good parameter vector with respect to a given set of objectives while consuming minimum possible resources.

This paper proposes a novel parameter tuning method which integrates Taguchi's method in the meta-heuristic framework and explicitly divides the search for a parameter vector into exploration and exploitation stages. It does not require tuning of Meta EA parameters as Taguchi's method is being used for selection, analysis and variations. Further, it does not complicate the statistical analysis by trying to build regression models as it has a clear objective of finding the parameter vector that gives best results with respect to a set of well-defined objectives. It uses a modification of well-known Taguchi's method in selection of parameter vectors and statistical analysis and it can quickly find the main effect of parameters for the chosen levels. Therefore, the proposed method needs less effort to find a good parameter vector. It is a type of iterative Sampling method, which overcomes the limitation of existing sampling methods like Calibra [34] by explicitly dividing the search into two phases of exploration and exploitation and employing multiple levels of parameters and Taguchi's method in the first phase of iterative exploration instead of using full factorial design with just two levels in Calibra [34]. Further, it reduces effort during exploitation phase by testing around the best parameter vector. Of course, the proposed method does not guarantee finding the optimal parameter vector, but it can find near optimal parameter vectors by searching in a methodical way and consuming much less effort.

The objective of parameter tuning in this paper is to find the best set of parameter vector values for solving a class of problems while spending only a reasonable amount of resources. This would then corroborate the fact that EAs find good solutions (not necessarily optimal) in reasonable amount of time and an EA can be tuned to find good solutions for a particular class of problems. In this paper, the focus is also on finding near optimal solution in reasonable amount of time with the robustness to the randomness essential in stochastic search process of QEA. Robustness to changes in problem specification is handled indirectly by running the algorithm on the several problems of same as well as different class. The parameter set for the same set of problems may be taken to be same where as it can be different for a different set of problems. However, they can all be arrived at by following the same procedure for parameter tuning.

The paper is further organized as follows. Section 2 discusses QEA's basic structure. Parameter Tuning method is proposed in Section 3 along with a discussion on Orthogonal Arrays. Section 4 presents the experimental testing and analysis of the proposed Parameter Tuning method on QEA. Conclusions are drawn in Section 5 giving directions for future research endeavors.

## I. QUANTUM-INSPIRED EVOLUTIONARY ALGORITHMS

Canonical QEA maintains a population of individuals in quantum bits or Q-bits. A Q-bit coded individual can probabilistically represent a linear superposition of states in the search space. Thus it has better characteristics of population diversity than other representations [14]. A Q-bit is represented as follows:

$$q_i = \begin{bmatrix}\alpha_i \\ \beta_i\end{bmatrix} \tag{1}$$

where $|\alpha_i|^2$ is probability of Q-bit, qi to be in state 0, $|0\rangle$ and $|\beta_i|^2$ is probability of Q-bit, $q_i$ to be in state 1, $|1\rangle$ and

$$|\alpha_i|^2 + |\beta_i|^2 = 1. \tag{2}$$

$\alpha_i$ and $\beta_i$ are real numbers for QEA implementations in this paper. Each individual is represented by a set of Q-bits in a string as:

$$Q(t) = \begin{bmatrix}\alpha_1 & \alpha_2 \ldots & \alpha_n \\ \beta_1 & \beta_{2\ldots} & \beta_n\end{bmatrix} \tag{3}$$

such that $|\alpha_i|^2 + |\beta_i|^2 = 1$, where i = 1 to n.

Measurement is the process of generating binary strings from the Q-bit string, Q. To observe the Q-bit string (Q), a string consisting of random numbers between 0 and 1 (R) is generated. The $i^{th}$ element of binary string, $b_i$, is set to zero if the $i^{th}$ random number, $r_i$, is less than $|\alpha_i|^2$ and one otherwise. In every iteration, it is possible to generate more than one solution strings from the Q by generating a new string of Random numbers as given above. The fitness values of each of these strings can be computed and the solution with the best fitness is identified.

A quantum gate or Q-Gate is utilized for updating the elements of a Q-bit string so that they move towards the best solution. Thus, there is a higher probability of generating solution strings, which are similar to the best solution in subsequent iterations. One such Q-Gate is Rotation gate, which is unitary in nature and updates the Q-bit as follows:

$$\begin{bmatrix}\alpha_i^{t+1}\\ \beta_i^{t+1}\end{bmatrix} = \begin{bmatrix}\cos(\Delta\theta_i) & -\sin(\Delta\theta_i)\\ \sin(\Delta\theta_i) & \cos(\Delta\theta_i)\end{bmatrix}\begin{bmatrix}\alpha_i^{t}\\ \beta_i^{t}\end{bmatrix} \quad (4)$$

where $\alpha_i^{t+1}$ and $\beta_i^{t+1}$ denote probabilities of $i^{th}$ Q-bit in $(t+1)^{th}$ iteration. $\Delta\theta_i$ is the angle of rotation, which is depicted in Fig. 1.

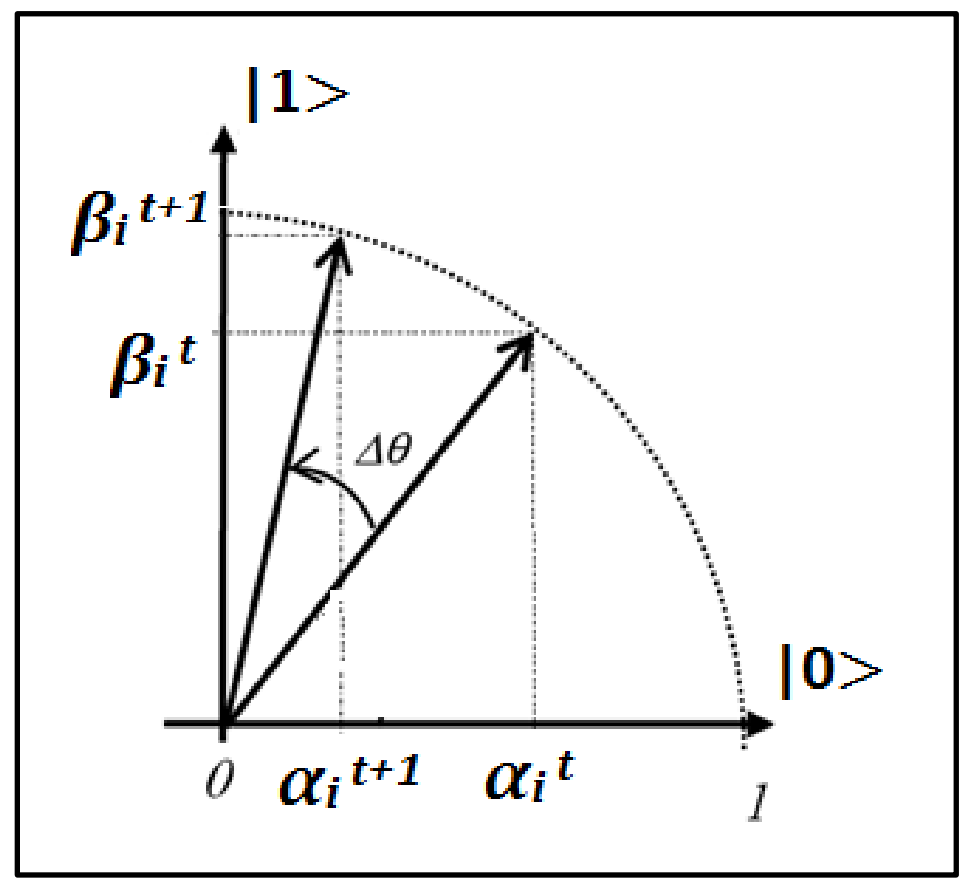

Fig 1. Effect of Quantum Rotation Operator on Q-bit

The quantum Rotation gate also requires an attractor [22] towards which the Q-bit would be rotated. It further takes into account the relative current fitness level of the individual and the attractor and also their binary bit values for determining the magnitude and direction of rotation. The magnitude of rotation is a tunable parameter and is selected from a set of eight rotation angles viz., $\theta_1$, $\theta_2$,.. $\theta_8$. The value of the rotation angles are problem dependent and require tuning [14]. The selection of eight rotation angles viz., $\theta_1$, $\theta_2$,.. $\theta_8$ is made from the lookup Table 1.

The $\Delta\theta_i$ has been made a function of the $i^{th}$ bit, $b_i$ of the best solution $B_j$, found so far till $j^{th}$ iteration, the $i^{th}$ bit $x_i$ of the current binary solution and the condition that the Q-bit, $q_i$, $|q_i\rangle = \alpha_i|0\rangle + \beta_i|1\rangle$), associated with $x_i$ should rotate towards the corresponding basis state $|0\rangle$ or $|1\rangle$ to increase the probability of $q_i$ so that $x_i$ in the next iteration has better probability of collapsing to $b_i$. Let us consider the following example, if $b_i$ and $x_i$ are 0 and 1, respectively, and if objective function value of the best solution, f(B) is better than the objection function value of the current solution, f(X), (i.e. f(X) < f(B) for maximization problems and f(X) > f(B) for minimization problems ), then:

(i) If the Q-bit is located in the first or the third quadrant in Fig. 2, the value of $\Delta\theta_i$ is set to a negative value so that the probability of $q_i$ to collapse to the state **|0⟩ is increased.**

(ii) If the Q-bit is located in the second or the fourth quadrant, in Fig. 2, the value of $\Delta\theta_i$ is set to a positive value so that the probability of $q_i$ to collapse to the state **|0⟩ is increased**.

If $b_i$ and $x_i$ are 1 and 0, respectively, and if f(B) is better than f(X), then:

(i) If the Q-bit is located in the first or the third quadrant, in Fig. 2, the value of $\Delta\theta_i$ is set to a positive value so that the probability of $q_i$ to collapse to the state **|1⟩ is increased.**

(ii) if the Q-bit is located in the second or the fourth quadrant, in Fig. 2, the value of $\Delta\theta_i$ is set to a negative value so that the probability of $q_i$ to collapse to the state **|1⟩ is increased.**

Han and Kim had suggested that if it is ambiguous to select a positive or a negative number for the values of the angle parameters, it is recommended to set the values to 0 [14]. However, the Fig. 2, suggests that there should be no ambiguity in selection of values for any of the possible cases listed in the Table 1. For example, the case were $b_i$ and $x_i$ are 0 and 0, respectively, and if f(B) is not better than f(X), then:

(i) If the Q-bit is located in the first or the third quadrant, in Fig. 2, the value of $\Delta\theta_i$ is set to a negative value so that the probability of $q_i$ to collapse to the state **|0⟩ is increased as it is the desired state through which it has found better solution.**

(ii) If the Q-bit is located in the second or the fourth quadrant, in Fig. 2, the value of $\Delta\theta_i$ is set to a positive value so that the probability of $q_i$ to collapse to the state **|0⟩ is increased as it is the desired state through which it has found better solution.**

Similarly, the case were $b_i$ and $x_i$ are 0 and 0, respectively, and if f(B) is better than f(X), then:

(i) If the Q-bit is located in the first or third quadrant, in Fig. 2, the value of $\Delta\theta_i$ is set to a negative value so that the probability of $q_i$ to collapse to the state **|0⟩ is increased as it is in desired state through which f(B) has better solution.**

(ii) if the Q-bit is located in the second or fourth quadrant, in Fig. 2, the value of $\Delta\theta_i$ is set to a positive value so that the probability of $q_i$ to collapse to the state **|0⟩ is increased as it is in desired state through which f(B) has better solution.**

Han and Kim had recommended [14] to set all the angles zero except $\theta_3 = 0.01\pi$ and $\theta_5 = -0.01\pi$. The magnitude of $\Delta\theta_i$ has an effect on the speed of convergence, but if it is too big, the solutions may diverge or converge prematurely to a local optimum. The values from .001 $\pi$ to .05 $\pi$ are recommended for the magnitude of $\Delta\theta_i$, although they depend on the problems. The sign of $\Delta\theta_i$ determines the direction of convergence.

TABLE 1
LOOKUP TABLE FOR ROTATION ANGLES

| $x_i$ | $b_i$ | f(B) better than f(X) | $\Delta\theta_i$ | $\alpha_i$ | $\beta_i$ | Sign | Han & Kim, [Han2002] | Remarks |
|---|---|---|---|---|---|---|---|---|
| 0 | 0 | True | $\theta_1$ | + | + | - | 0 (so no change in state vector) | State vector should be rotated slightly towards \|0⟩ as $b_i$=0, so in next iteration also, $i^{th}$ Q-bit, $q_i$ should have an improved probability of collapsing \|0⟩. |
| | | | | - | + | + | | |
| | | | | - | - | - | | |
| | | | | + | - | + | | |
| 0 | 0 | False | $\theta_2$ | + | + | - | 0 (so no change in state vector) | State vector should be rotated slightly towards \|0⟩ as with $x_i$=0 it has found a better solution so in next iteration also, $i^{th}$ Q-bit, $q_i$ should have an improved probability of collapsing \|0⟩. |
| | | | | - | + | + | | |
| | | | | - | - | - | | |
| | | | | + | - | + | | |
| 0 | 1 | True | $\theta_3$ | + | + | + | $0.01\pi$ | State vector should be rotated adequately towards \|1⟩ as $b_i$=1 so in next iteration, $i^{th}$ Q-bit, $q_i$ should have an improved probability of collapsing \|1⟩. |
| | | | | - | + | - | | |
| | | | | - | - | + | | |
| | | | | + | - | - | | |
| 0 | 1 | False | $\theta_4$ | + | + | - | 0 (so no change in state vector) | State vector should be rotated slightly towards \|0⟩ as with $x_i$=0, it has found a better solution so in the next iteration also, $i^{th}$ Q-bit, $q_i$ should have an improved probability of collapsing \|0⟩. |
| | | | | - | + | + | | |
| | | | | - | - | - | | |
| | | | | + | - | + | | |
| 1 | 0 | True | $\theta_5$ | + | + | - | $0.01\pi$ | State vector should be rotated adequately towards \|0⟩ as $b_i$=0 so in next iteration $i^{th}$ Q-bit, $q_i$ should have an improved probability of collapsing \|0⟩. |
| | | | | - | + | + | | |
| | | | | - | - | - | | |
| | | | | + | - | + | | |
| 1 | 0 | False | $\theta_6$ | + | + | + | 0 (so no change in state vector) | State vector should be rotated slightly towards \|1⟩ as with $x_i$=1, it has found a better solution so in next iteration also $i^{th}$ Q-bit, $q_i$ should have an improved probability of collapsing \|1⟩. |
| | | | | - | + | - | | |
| | | | | - | - | + | | |
| | | | | + | - | - | | |
| 1 | 1 | True | $\theta_7$ | + | + | + | 0 (so no change in state vector) | State vector should be rotated slightly towards \|1⟩ as $b_i$=1 so in next iteration |
| | | | | - | + | - | | |

| | | | | - | - | + | | also this i[th] Q-bit, $q_i$ should have an improved probability of collapsing $\lvert 1\rangle$. |
|---|---|---|---|---|---|---|---|---|
| | | | | + | - | - | | |
| 1 | 1 | False | $\theta_8$ | + | + | + | 0 (so no change in state vector) | State vector should be rotated slightly towards $\lvert 1\rangle$ as with $x_i$=1 it has found a better solution so in the next iteration also i[th] Q-bit, $q_i$ should have an improved probability of collapsing $\lvert 1\rangle$. |
| | | | | - | + | - | | |
| | | | | - | - | + | | |
| | | | | + | - | - | | |

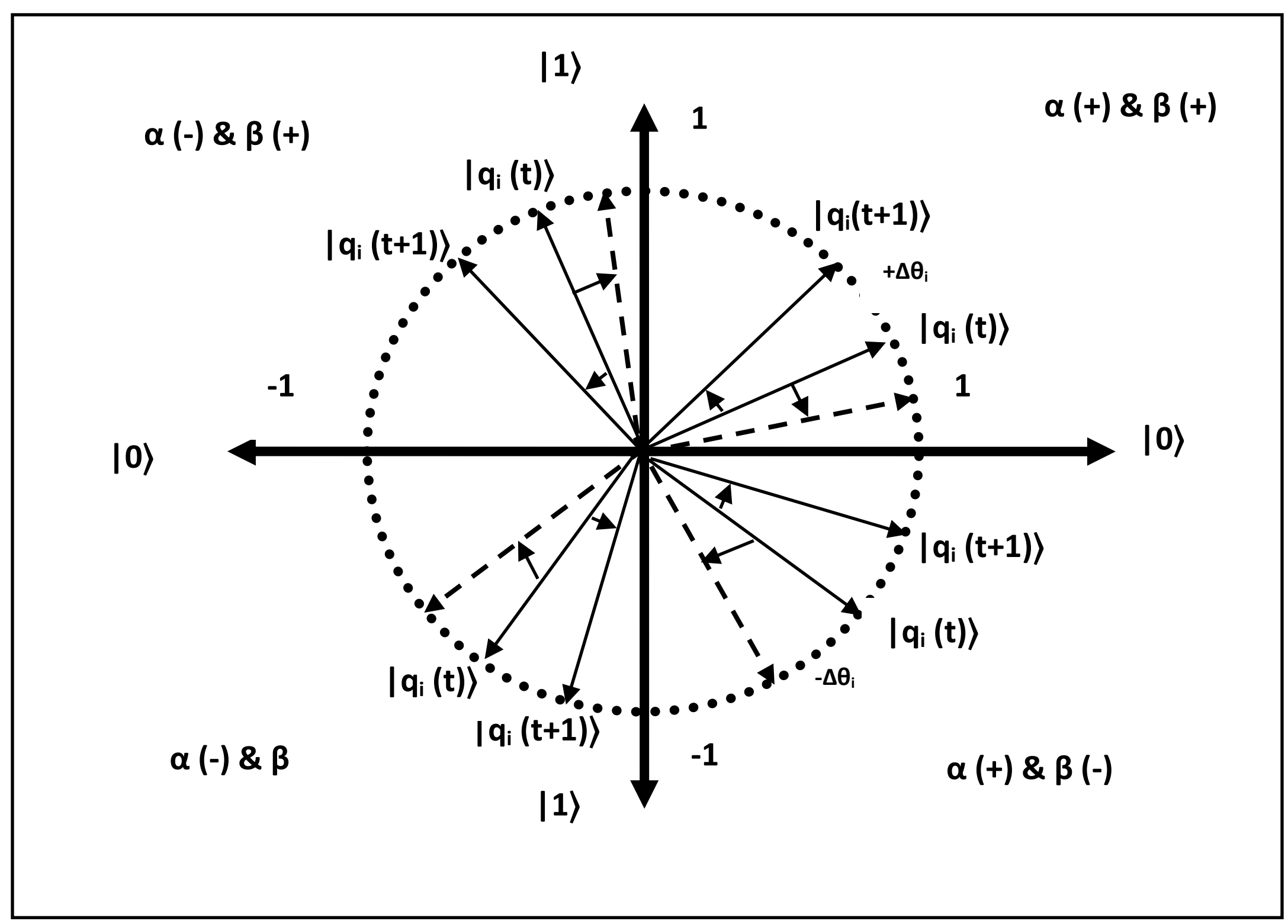

Fig 2. Four Quadrant Rotation of Q-bit

There are two methods to initially set all the Q-bits viz., the random initialization and Equal Probability initialization. In random initialization, Q-bits are assigned values between -1 to +1 by generating them randomly, while taking into account the normalization criteria described in eq. 5. The second method of initialization is done so that observation results in either 0 or 1 with equal probability by setting the values of $\alpha_i$ and $\beta_i$ to 0.707.

$$\left|Q\right\rangle = \sum_{k=1}^{2^m} \frac{1}{\sqrt{2^m}} \left|X_k\right\rangle \tag{5}$$

The termination condition is usually based on the number of generations, number of fitness evaluations and convergence of search or a combination of them. A measure of diversity in population can be made out of real-valued Q-bit strings. If the Q-bits have a value of α as 0.707, the diversity can be considered to be highest, whereas the diversity can be considered least when the value of α is near the extremes, i.e. 0 or 1. Hence level of convergence can be considered as number of Q-bits which have reached very close to 0 or 1. If this number is equal to number of elements in Q, then the chances of the measurement process generating diverse solutions becomes very low and it may be said that the search has converged. In this work, the stopping criterion is taken as the completion of a maximum number of generations.

The structure of the canonical Quantum-inspired Evolutionary Algorithm is shown by flow chart in Fig. 3 and it works as follows [14]:

```
a) t = 0; Population Size = N, Group Size = GS;
b) initialize Q_1(t)…Q_N(t), divide into M (= N/GS) groups (QG_1… QG_M);
c) make P_1(t)…P_N(t) by observing the states of Q_1(t)…Q_N(t) respectively;
d) if repair required then repair P_i(t), i = 1 .. N ;
e) evaluate P_1(t)…P_N(t) & store in OP_1(t).. OP_N(t);
f) store the global, Local and individual best solutions into GB(t), LB_j(t) & IB_i(t)
   respectively, i = 1 .. N, j = 1 .. M;

while (termination condition is not met) {

g) t = t + 1;
for each individual_i i = 1 .. N {
h) determine Attractor A_i(t)based on migration condition:
      Global: A_i(t) = GB(t-1);
      Local: A_i(t) = LB_j(t-1) where i^th individual belongs to j^th Group;
      No-Migration: A_i(t) = IB_i(t-1);
i) apply Q-gate(s) on Q_i(t-1) to update to Q_i(t);
j) make P_i(t) by observing the states of Q_i(t);
k) if repair required then repair P_i(t);
l) evaluate P_i(t) & store result into OP_i(t);
m) store better solution among IB_i(t-1) and OP_i(t) into IB_i(t);
n) store the better solution among LB_i(t-1) and Best_individual_in_Group_j(t) into
   NB_i(t), where i^th individual belongs to j^th Group ;
o) store the global best solution GB(t-1) among IB_i(t) into GB(t); }}
```

In step a), initialize the population size N, size of the Group to GS. In step b), the qubit register $Q(t)$ containing Q-bit strings $Q_1(t)$ … $Q_N(t)$ are initialized randomly and divided into M groups ($QG_1$… $QG_M$), where no. of groups $M = N/GS$. In step c), the binary solutions represented by $P_1(t)$ … $P_N(t)$ are constructed by measuring the states of $Q_1(t)$…$Q_N(t)$ respectively. In step d), if repairing is required in binary solutions $P_i(t)$ then repairing is performed. In step e) binary solution is evaluated to give a measure of its fitness $OP_i(t)$, where $OP_i(t)$ represents the objective function value . In step f), the initial global, neighborhood and individual best solutions are then selected among the binary solutions $OP_i(t)$, and stored into $GB(t)$, $LB_i(t)$, $IB_i(t)$ respectively, Local best solution is determined from the individuals in the Group. In step h), the attractor $A_i(t)$ for the $i^{th}$ individual is determined according to the migration strategy. If the migration is global, then global best is assigned as the attractor, whereas if the migration is local then local group best is assigned as the attractor and if no migration is there then individual best becomes the attractor. In step i), update $Q_i(t-1)$ to $Q_i(t)$ using Q-Gates, which is quantum rotation gate described earlier in Section 2. In step j), the binary solutions in $P_i(t)$ are formed by measuring the states of $Q_i(t)$ as in step c). In step k), if the repair is required then it is performed as in step d) and in step l), each binary solution is evaluated for the fitness as in step e). In step m), n) and o), the global, local and individual best solutions are selected and stored into $GB(t)$, $LB_i(t)$ and $IB_i(t)$ respectively based on a comparison between previous and current best solutions.

In the Fig. 3, flow chart depicts the working of canonical QEA, three different lines are used to identify information flow of Q-bit string, Binary String and the fitness function. The broken / dash & dot line indicates that information flowing is the Q-bit string, whereas dot line indicates that information flowing is binary string. The solid line shows that information flowing is objective function value. $QG_1$ to $QG_M$ are M groups in which N individuals are assigned. Measurement operator creates N binary bit strings from N individual Q-bit strings. The rest is as explained in the algorithm.

## II. PROPOSED PARAMETER TUNING METHOD

The proposed parameter tuning method has been developed by integrating Taguchi method into metaheuristic framework. It is an iterative multistage technique that utilizes the strength of Taguchi's method [31] while avoiding its well-known disadvantage of aliasing in presence of interaction between parameters. The metaheuristic framework is required for exploring the parameter space as Taguchi's method can perform local optimization only and that too if there are no interactions amongst parameters. If there are interactions amongst parameters then Taguchi's method fails due to aliasing but metaheuristic framework helps in taking the search forward by selecting the parameter set with the best result in the experiments performed so far i.e. an elitist selection is made and the search proceeds to the next iteration or the stage depending on the prevailing conditions.

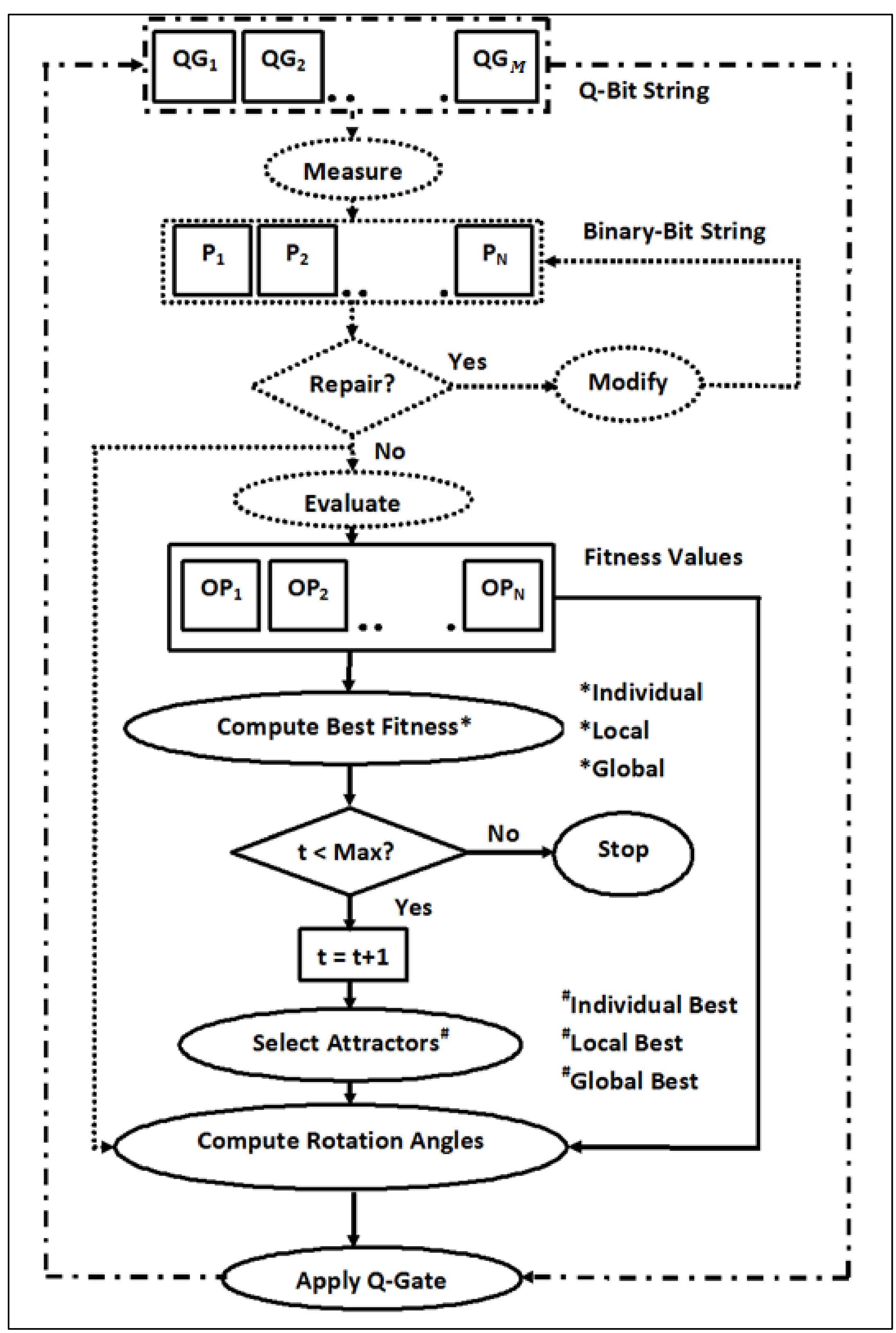

Fig 3. Flow Chart of QEA

The following notation is introduced in order to describe the proposed parameter tuning method:

**N1:** Maximum Number of iterations that can be run during Stage 1.

**N2:** Maximum Number of iterations that can be run during Stage 2.

**NWI1:** Maximum Number of consequent iterations that can be run without observing any improvement in objective function value during Stage 1.

**NWI2:** Maximum Number of consequent iterations that can be run without observing any improvement in objective function value during Stage 2.

**NP:** Number of parameters to be tuned.

**$UL_j$:** Upper limit on the value of parameter j (for j = 1…NP).

**$LL_j$:** Lower limit on the value of parameter j (for j = 1…NP).

**NL1:** Number of levels in Stage 1 of each parameter.

**NL2:** Number of levels in Stage 2 of each parameter.

**OA1:** Orthogonal Array for Stage 1.

**OA2:** Orthogonal Array Stage 2.

**$PVA_i$:** Parameter Vector obtained from the Analysis of results of Experiments designed by OA at $i^{th}$ iteration.

**$PVB_i$:** Parameter Vector which performed best in Experiments designed by OA at $i^{th}$ iteration.

**$OFV_PVA_i$:** Objective Function Value with $PVA_i$ at the $i^{th}$ iteration.

**$OFV_PVB_i$:** Objective Function Value with $PVB_i$ at the $i^{th}$ iteration.

**$PIVOT_i$:** Best performing parameter vector known till $i^{th}$ iteration.

**$OFV_PIVOT_i$:** Objective Function Value with Best performing parameter vector known till $i^{th}$ iteration.

**i:** Iteration Counter.

**I:** No. of iterations since no improvement in $OFV_PIVOT_i$ has been observed.

Framework of proposed tuning method is as follows:

```
1. Set N1, N2, NWI1, NWI2, NP, NL1, NL2, OA1 & OA2;
2. For each tuned parameter, Set LLj and ULj, where j = 1 to NP;
3. Initialize I = 0; i =0;
   /* EXPLORATION STAGE */
4. For each parameter, Randomly Distributed 'NL1' levels between LLj and ULj;
   Do {
5. Perform experiments according to OA1.
6. Compute PVAi using Taguchi Method.
```

7. Compute OFV_PVA$_i$.
8. Determine PVB$_i$ and OFV_ PVB$_i$.
9. If i = 0 then
If OFV_PVB$_i$ better than OFV_ PVA$_i$ then
OFV_PIVOT$_i$ = OFV_PVB$_i$
PIVOT$_i$ = PVB$_i$
I=0;
Else

OFV_PIVOT$_i$ = OFV_PVA$_i$
PIVOT$_i$ = PVA$_i$
I=0;
Else If OFV_PVB$_i$ better than OFV_ PVA$_i$ and OFV_PIVOT$_{i-1}$ then
OFV_PIVOT$_i$ = OFV_PVB$_i$;
PIVOT$_i$ = PVB$_i$;
I=0;
Else if OFV_PVA$_i$ better than OFV_PIVOT$_{i-1}$ then

OFV_PIVOT$_i$ = OFV_PVA$_i$;
PIVOT$_i$ = PVA$_i$;
I=0;
Else
PIVOT$_i$ = PIVOT$_{i-1}$;
I++;
10. Use the PIVOT$_i$ to generate randomly ('NL1' – 1) other levels. (Assign (NL1 – 1) / 2 levels between LL$_j$ & PIVOT$_{j,i}$ and (NL1 – 1) / 2 levels between UL$_j$ & PIVOT$_{j,I}$ for the j$^{th}$ parameter in the PIVOT$_i$ vector) to be used in the next iteration, j = 1 .. NP;
11. ++i;
} While (i != N1 && I < NWI1)
/* EXPLOITATION STAGE */
12. PIVOT $_{-1}$ = PIVOT$_i$; OFV_PIVOT$_{-1}$ = OFV_PIVOT$_i$; I = 0; i = 0;
13. Use the PIVOT$_{i-1}$ to generate randomly ('NL2' – 1) other levels. Assign the (NL2 – 1)/2 levels between (PIVOT$_{j,i-1}$ – 0.1*(PIVOT$_{j,i-1}$ - LLj)) & PIVOT$_{j,i-1}$ and other (NL2 – 1)/2 levels between PIVOT$_{j,i-1}$ & PIVOT$_{j,i-1}$ + 0.1* (ULj – PIVOT$_{j,i-1}$))), j = 1 .. NP.
Do {

14. Perform experiments according to OA2.
15. Compute PVA$_i$ using Taguchi Method.
16. Compute OFV_PVA$_i$.
17. Determine PVB$_i$ and OFV_ PVB$_i$.
18. If OFV_PVB$_i$ better than OFV_ PVA$_i$ and OFV_PIVOT$_{i-1}$ then
OFV_PIVOT$_i$ = OFV_PVB$_i$;
PIVOT$_i$ = PVB$_i$;
I=0;
Else if OFV_PVA$_i$ better than OFV_PIVOT$_{i-1}$ then

OFV_PIVOT$_i$ = OFV_PVA$_i$;
PIVOT$_i$ = PVA$_i$;
I=0;
Else
OFV_PIVOT$_i$ = OFV_PIVOT$_{i-1}$;
PIVOT$_i$ = PIVOT$_{i-1}$;
++I;
19. If (I == 0)
Generate uniformly distributed ('NL2' – 2) other levels between PIVOT$_{j,i}$ and PIVOT$_{j,i-1}$ ,j=1…NP;
Else

Use the PIVOT$_{j,i}$ to generate randomly ('NL2' – 1) other levels, j = 1..NP;

**Assign the (NL2 – 1)/2 levels between ($PIVOT_{j,i}$ – (0.1/i)*($PIVOT_{j,i}$ - LLj)) & $PIVOT_{j,i}$ and other (NL2 – 1)/2 levels between $PIVOT_{j,I}$ & $PIVOT_{j,i}$ + (0.1/i)*(ULj – $PIVOT_{j,i}$))), j = 1..NP;**

**20. ++i;**
**} While (i != N2 && I < NWI2)**
**21. OUTPUT $PIVOT_{i-1}$ as Parameter Value.**

The proposed method has two explicit iterative stages i.e. exploration and exploitation. First of all initialization of parameters controlling the parameter tuning method like N1, N2, NWI1, NWI2, NP, NL1, NL2, OA1 & OA2 are specified. In step 2, lower and upper limits of all the parameters being tuned are specified. In exploration stage, first of all, each parameter is randomly initialized to NL1 different levels within their respective range. In step 5), the experiments are performed according to orthogonal array OA1, i.e. OA1 is used for determining the parameter values of the EAs being tuned on a specific problem, for performing the experiments. The results are analyzed by using Taguchi's method for finding the parameter vector, $PVA_i$ that gives the best result for the given problem. Analysis of the result is performed by using the best objective function value obtained from the individual experiment (i.e. thirty runs minimum for each experiment) as the search is for finding the $PVA_i$. In step 7), the EA being tuned is executed using $PVA_i$ to find the $OFV_PVA_i$. In step 8), the $PVB_i$ and $OFV_PVB_i$ are determined from experiments performed using OA1 for cross checking Taguchi's method and is also the elitist selection for implementing meta-heuristic framework, if Taguchi's method fails. In step 9), $OFV_PVA_i$ and $OFV_PVB_i$ are compared and the better of the two becomes $PIVOT_i$. In step 10), $PIVOT_i$ is used for generating new levels for parameters of EA being tuned for further experiments in exploration stage, till the termination criteria of maximum number of iterations N1 or maximum number of iteration since no improvement is observed in $PIVOT_i$ , NWI1 is met.

The exploitation stage begins with the selection of NL2 levels for every parameter, out of which, one of the level is taken as the best $PIVOT_i$ from the exploration stage. In step 14), the experiments are performed according to orthogonal array OA2, i.e. OA2 is used for determining the parameter values of the EA being tuned on a specific problem, for performing the experiments. The results are analyzed by using Taguchi's method for finding the set of parameter vector, $PVA_i$ that gives the best result for the given problem. Analysis of the result is performed by using the best objective function value obtained from the individual experiment (i.e. thirty runs minimum for each experiment) as the search is for finding the $PVA_i$. In step 16), the EA being tuned is executed using $PVA_i$ to find the $OFV_PVA_i$. In step 17), the $PVB_i$ and $OFV_PVB_i$ are determined from experiments performed using OA2 for cross checking Taguchi's method and is also the elitist selection for implementing meta-heuristic framework, if Taguchi's method fails. In step 18), $OFV_PVA_i$ and $OFV_PVB_i$are compared and the better of the two becomes $PIVOT_i$. In step 19), if there has been improvement in $PIVOT_i$ from previous iteration then $PIVOT_i$ and $PIVOT_{i-1}$ is used for generating new levels otherwise, new levels are generated randomly in the vicinity of $PIVOT_i$, by perturbing it on the either side by (10% / i) of its Euclidian distance from the extremes randomly, for the parameters of EA being tuned for further experiments in exploitation stage, till the termination criteria of maximum number of iterations N2or maximum number of iteration since no improvement is observed in $PIVOT_i$ , NWI2 is met.

In this work, N1 = 3, N2 =3, NWI1 = 2, NWI2 = 2, NL1 = 5, NL2 = 3, OA1 is OA(50, $2^1$ x $5^{11}$,2) i.e. **L-50 Table** [41], [42] with fifty experiments, one parameter with two levels and eleven parameters with five levels each and its strength is 2. OA2 is OA(27, $3^{13}$,2) i.e. **L-27 Table** [41], [42] with 27 experiments, 13 parameters with three levels each and its strength is 2. The rest of the parameters of proposed parameter tuning method are problem dependent.

## III. TESTING

The canonical QEA has been fine-tuned by the proposed parameter tuning method on a suite of benchmark test problems, which have been used in many studies. This serves two purposes, firstly, it validates the proposed parameter tuning framework and secondly, it helps in further improving the performance of canonical QEA.

The process followed for testing involves fine-tuning the eleven parameters of canonical QEA with a single instance of the problem and subsequently using the same parameter values to solve some instances of the problem. Further, a comparison is made with the canonical QEA [14] to study the efficacy of the proposed technique. The computational time required in parameter tuning depends on the maximum number of function evaluation in each run, the number of runs in each experiment, number of experiment to be performed in each iteration (OA1 for exploration and OA2 for exploitation stage) and the number of iterations used in Exploration Stage (N1) and Exploitation Stage (N2). The maximum number of function evaluations may be limited to five hundred thousand. It can be more or less depending on the problem and can be determined with some random experiments. One can start with a certain value and then change it according to the performance. If the optimal is known then, the run can be terminated upon reaching the optima, thus saving the computational resource. The number of independent runs should be minimum thirty as QEA are stochastic in nature. The number of experiments is a function of the orthogonal array selected for a particular stage. L50 with fifty experiments, five levels for eleven parameters and two levels for one parameter has been used for Exploration stage and L27 with twenty seven experiments and three levels for eleven parameters has been used for Exploitation Stage. In our experience, three to five iterations were sufficient in exploration stage. The exploitation stage, at times, ended in a single iteration but required up to five iterations in some other cases. Thus, the availability of computational resources and complexity of problem should help in deciding the effort to be consumed in Exploration and Exploitation Stage. Further, experiments can be designed with larger size of orthogonal Arrays to accommodate more number of parameters as well as more number of levels. Thus, the proposed framework can be used for a very quick parameter tuning as well as for a relative exhaustive parameter tuning also.

In this work, `N1 = 3, N2 =3, NWI1 = 2, NWI2 = 2, NL1 = 5, NL2 = 3, OA1 is` $OA(50, 2^1 \times 5^{11}, 2)$ `i.e.` **L-50 Table** [41], [42] with fifty experiments, one parameter with two levels and eleven parameters with five levels each and its strength is 2. `OA2 is` $OA(27, 3^{13}, 2)$ `i.e.` **L-27 Table** [41], [42] with 27 experiments, 13 parameters with three levels each and its strength is 2. The rest of the parameters of proposed Tuning method are problem dependent.

The robustness of the algorithm to changes in parameter values [27] have been studied from the data collected during the tuning process. It has been studied both for large and small variations in parameter values [27]. The data for the first case i.e. large variation has been collected from the first iteration of the Exploration Stage. The solutions of all the experiments in the first iteration of the exploration stage have been used for computing the deviation from the average solution of the QEA. This measure can statistically indicate the robustness of the QEA in the entire parameter space for a given problem through sampling. The data for the second case i.e. small variation has been collected from the first iteration of the Exploitation Stage. The solutions of all the experiments in the first iteration of the exploitation stage have been used for computing the deviation from the best solution of the QEA. This measure can statistically indicate the robustness of the QEA against minor variations in the parameters for a given problem through sampling.

The problem suite used for testing has Massively Multimodal Deceptive Problems (MMDP), COUNTSAT Problems, Knapsack problems and P-Peaks Problems. The eleven parameters of QEA i.e. eight rotation angles (θ1 to θ8), migration period, group size / No. of groups (i.e. population = No. of groups * group size) and population size have been fine-tuned using the proposed framework. Subsequently a comparison has been made between the fine-tuned QEA and the canonical QEA with parameters given in Table 2 as they have been widely used in literature [14], [16]. The stopping criterion is same for both the QEAs, which is maximum number of function evaluations.

### *A. Massively Multimodal Deceptive Problem (MMDP) (MMDP)* [43], [44]

MMDP with size K = 40 was used for parameter tuning of canonical QEA. The initial range of values for parameters in QEA used for tuning is given in Table 3. The parameter range for magnitude of rotation angles (θ1, θ2, θ4, θ6, θ7 & θ8) is 0 to 0.05π as they change by small magnitude as compared to θ3 & θ5, whose range is from 0 to 0.5π, which is very large as compared to the range suggested by [14]. The direction of rotation depends on the sign of α, β and relative fitness as per Table 1. The range for population size is 5 to 200, and covers the values for similar parameter for most studies in Evolutionary Algorithms. The range for no. of groups is 1 to 20, which is twice big as the value suggested in [14]. The range for global migration is 1 to 500, which is again five times the value suggested by [14]. The change in value of each parameter during the tuning process is depicted in Table 4 and shown in Fig. 4 to Fig. 14. There were three rounds for exploration stage and one round for exploitation stage. It can be observed that after the third round, all the thirty independent runs of QEA of the Tuning experiment had reached the optimal, so it was decided to stop further exploration and start the exploitation so as to further search within the vicinity of the Best Parameter Set found so far and improve the convergence rate. After the first round of the exploitation stage, the convergence could be achieved within twenty generations as shown in Fig. 15, so it was decided to stop further tuning.

TABLE 2
PARAMETER SETTING FOR QEA

| Parameters | Canonical QEA |
|---|---|
| θ1 | 0 |
| θ2 | 0 |
| θ3 | 0.01π |
| θ4 | 0 |
| θ5 | 0.01π |
| θ6 | 0 |
| θ7 | 0 |
| θ8 | 0 |

| Population Size | 50 |
|---|---|
| Group size / No. of Groups | 5 / 10 |
| Global Migration Period (Generations) | 100 |

TABLE 3

INITIAL RANGE OF PARAMETERS OF CANONICAL QEA

| **Parameter** | **θ1 (* π)** | **θ2 (* π)** | **θ3 (* π)** | **θ4 (* π)** | **θ5 (* π)** | **θ6 (* π)** | **θ7 (* π)** | **θ8 (* π)** | **Pop size** | **No. of Groups** | **Global Migration** |
|---|---|---|---|---|---|---|---|---|---|---|---|
| Lower Limit | 0 | 0 | 0 | 0 | 0 | 0 | 0 | 0 | 5 | 1 | 1 |
| Upper Limit | 0.05 | 0.05 | 0.5 | 0.05 | 0.5 | 0.05 | 0.05 | 0.05 | 200 | 20 | 500 |

TABLE 4

BEST PARAMETER VECTOR (PIVOT) DURING TUNING PROCESS

| **Iter. No.** | **Best Pivot Parameter Value** | | | | | | | | | | | **Av. OFV** |
|---|---|---|---|---|---|---|---|---|---|---|---|---|
| | **θ1 (* π)** | **θ2 (* π)** | **θ3 (* π)** | **θ4 (* π)** | **θ5 (* π)** | **θ6 (* π)** | **θ7 (* π)** | **θ8 (* π)** | **Pop. Size** | **No. of Grp.** | **Glb. Mig.** | |
| Explor. – 1 | 0.001 | 0.032 | 0.066 | 0.021 | 0.176 | 0.027 | 0.034 | 0.02 | 36 | 6 | 137 | 33.1 |
| Explor. – 2 | 0.0002 | 0.032 | 0.276 | 0.0431 | 0.019 | 0.0069 | 0.034 | 0.033 | 120 | 3 | 5 | 34.0 |
| Explor. – 3 | 0.00004 | 0.0282 | 0.223 | 0.0484 | 0.0022 | 0.018 | 0.035 | 0.033 | 33 | 3 | 3 | 40.0 |
| Expltt. – 4 | **0.000147** | **0.0282** | **0.205** | **0.0485** | **0.002** | **0.0205** | **0.035** | **0.033** | **28** | **4** | **6** | **40.0** |

The parameter value for $\theta1$ initially decreased during exploration and then increased a little during exploitation stage. The parameter value for $\theta2$ initially remained constant then decreased during exploration and then remained constant during exploitation stage. The parameter value for $\theta3$ initially increased and then decreased during exploration and then again decreased a little during exploitation stage. The parameter value for $\theta4$ initially increased during exploration and then increased minutely during exploitation stage. The parameter value for $\theta5$ initially decreased during exploration and then decreased a little during exploitation stage. The parameter value for $\theta6$ initially decreased and then increased during exploration and then increased a little during exploitation stage. The parameter value for $\theta7$ initially remained constant and then increased minutely during exploration and then remained constant during exploitation stage. The parameter value for $\theta8$ initially increased and then remained constant during exploration and then remained during exploitation stage. The parameter value for Population Size initially increased then decreased during exploration and then decreased slightly during exploitation stage. The parameter value for no. of groups initially decreased during exploration and then increased a little during exploitation stage. The range of parameter value for Global Migration had to be reset to 0 to 10 as the optimal was reached by Best performing experiment in twenty iterations. Thus, the Fig. 15 is a semi log graph, which shows that parameter value for global migration initially decreased during exploration and subsequently increased during exploitation stage. The final value of each parameter is given in Table 4 in the last row. The convergence graph given in Fig. 16 shows fast convergence to optimal within ten generations.

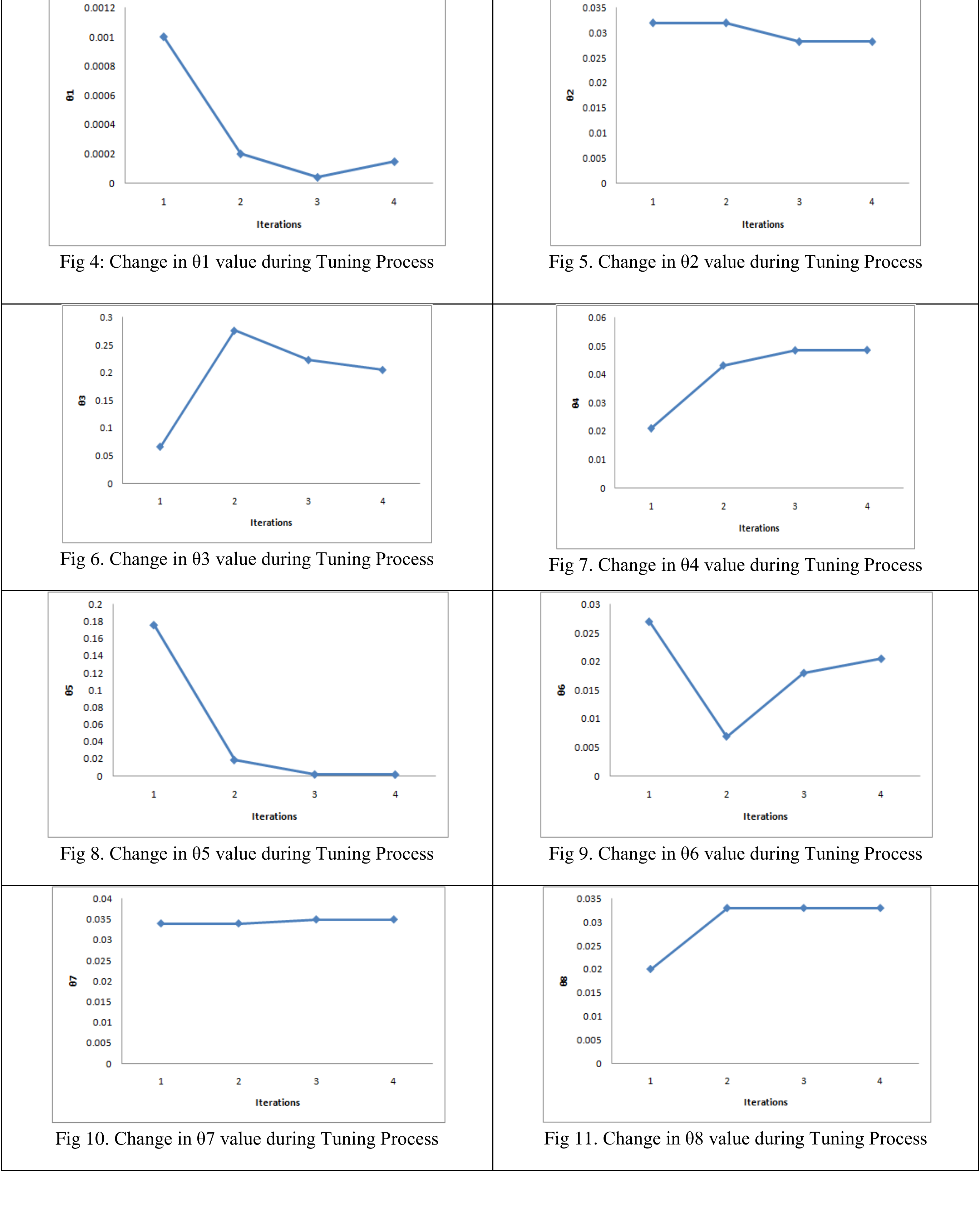

Fig 4: Change in θ1 value during Tuning Process

Fig 5. Change in θ2 value during Tuning Process

Fig 6. Change in θ3 value during Tuning Process

Fig 7. Change in θ4 value during Tuning Process

Fig 8. Change in θ5 value during Tuning Process

Fig 9. Change in θ6 value during Tuning Process

Fig 10. Change in θ7 value during Tuning Process

Fig 11. Change in θ8 value during Tuning Process

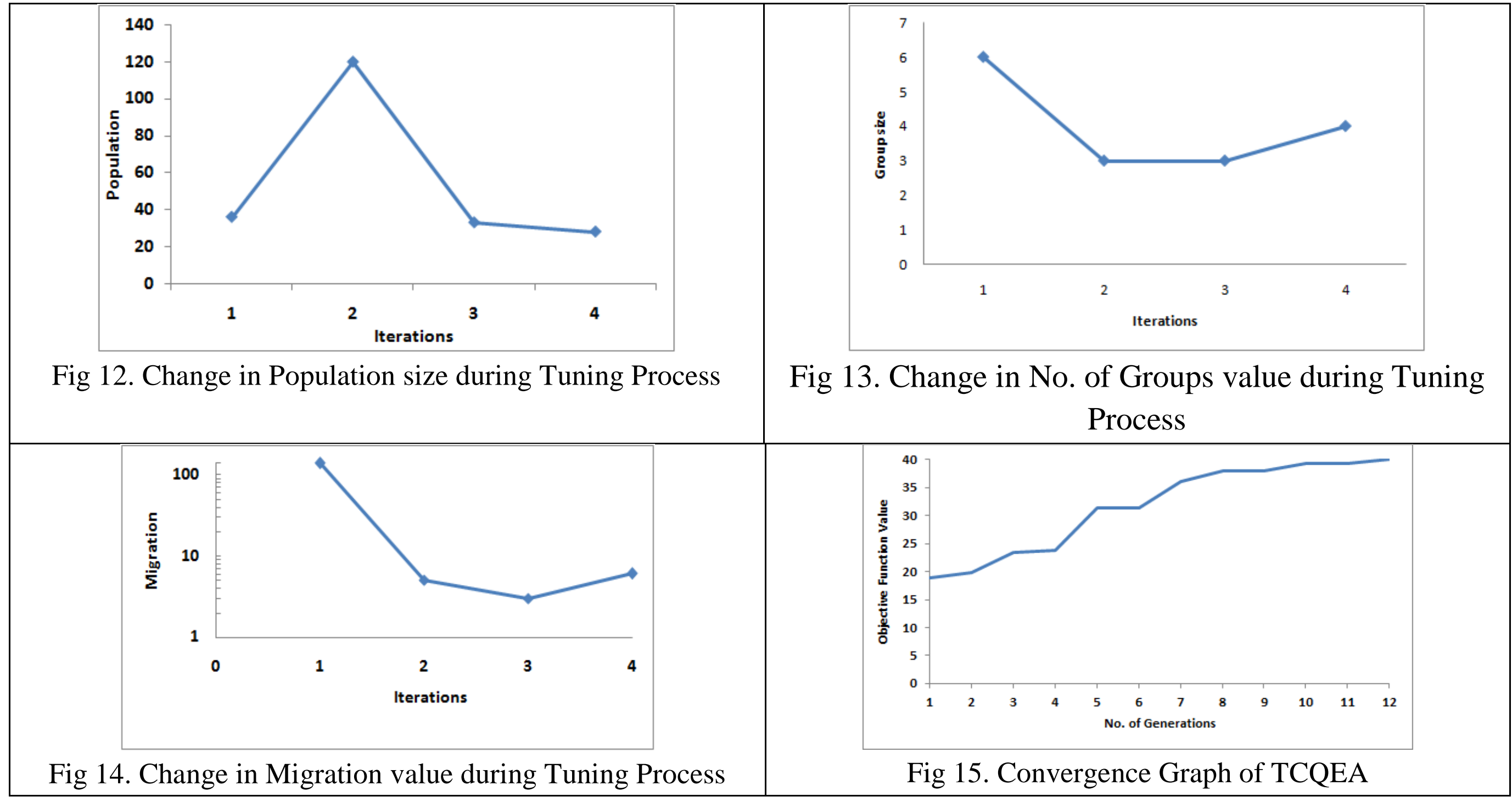

Fig 12. Change in Population size during Tuning Process

Fig 13. Change in No. of Groups value during Tuning Process

Fig 14. Change in Migration value during Tuning Process

Fig 15. Convergence Graph of TCQEA

The deviation from optimal for large variation in parameter setting, which is computed from the average results of fifty experiments from the first iteration of exploration stage is 7.54 whereas the deviation from optimal for small variation in parameter setting, which is computed from the average results of twenty seven experiments from the first iteration of exploitation stage is zero. Therefore, the tuned QEA is robust to small variation in parameter set, however, it is relatively unstable for large variation in parameter set, thus justifying the effort put in tuning the parameter set of QEA for MMDP. Thus, the deviation from the known optimal at the end of the iteration can be used for deciding the need for continuing the search in that stage. However, if the Optimal is unknown then this strategy cannot be applied with the same confidence, but in case of real world problems, often a best known solution is available, so in such cases, it may be used in place of the Optimal.

A comparative study was done between parameter Tuned QEA (TCQEA) and Canonical QEA (UCQEA) with Parameters given in Table 2 on MMDP with a set of twenty problems of size k varying from 10 to 200 at an interval of 10. The results are given in Table 5. Thirty independent runs were made on each problem size and the comparison has been made on Best, Median, Worst, Mean, percentage of runs in which optimal was achieved i.e. percentage of Success runs, and Average number of function evaluations (NFE). The Canonical QEA was not able to reach optimal value any runs for any of the twenty problems whereas the Tuned-QEA was able to reach 100% success in nineteen out of twenty problems. It has 93.3% of success run in the last problem of the benchmark suite. The performance of Tuned QEA was also good on speed of convergence as indicated by average NFE and the convergence graphs shown in figure, which also compared the speed of convergence of Tuned QEA to Canonical QEA.

The convergence graphs have been plotted between objective function value and number of generations for both Tuned-QEA (TCQEA) and Canonical QEA (UCQEA) for all the problem instances of MMDP for the median run as shown in Fig. 17 to Fig. 36. The convergence graph clearly establishes the superiority of Tuning as the Tuned QEA is able to reach the optimal in less than twenty generations in all the graphs whereas the Canonical QEA is not able to reach the optimal even in two thousand generations. The distance from optimal has increased as the size of the problem has increased in case of Canonical QEA but Tuned QEA has performed consistently well for this class of problems.

TABLE 5

COMPARATIVE STUDY BETWEEN TCQEA AND UCQEA ON MMDP INSTANCES

| Problem | Algo | Best | Median | Worst | Mean | % Success Runs | Average NFE |
|---|---|---|---|---|---|---|---|
| K=10 | UCQEA | 9.64 | 9.28 | 8.56 | 9.17 | 0.00 | 500050 |
| | **TCQEA** | **10.00** | **10.00** | **10.00** | **10.00** | **100.00** | **182** |
| K=20 | UCQEA | 18.56 | 16.59 | 15.33 | 16.67 | 0.00 | 500050 |

| | **TCQEA** | **20.00** | **20.00** | **20.00** | **20.00** | **100.00** | **234** |
|---|---|---|---|---|---|---|---|
| K=30 | UCQEA | 25.69 | 23.89 | 22.45 | 24.06 | 0.00 | 500050 |
| | **TCQEA** | **30.00** | **30.00** | **30.00** | **30.00** | **100.00** | **259** |
| K=40 | UCQEA | 33.53 | 31.01 | 29.58 | 31.21 | 0.00 | 500050 |
| | **TCQEA** | **40.00** | **40.00** | **40.00** | **40.00** | **100.00** | **280** |
| K=50 | UCQEA | 42.09 | 38.49 | 35.98 | 38.94 | 0.00 | 500050 |
| | **TCQEA** | **50.00** | **50.00** | **50.00** | **50.00** | **100.00** | **297** |
| K=60 | UCQEA | 48.14 | 45.26 | 43.47 | 45.56 | 0.00 | 500050 |
| | **TCQEA** | **60.00** | **60.00** | **60.00** | **60.00** | **100.00** | **280** |
| K=70 | UCQEA | 57.42 | 53.11 | 49.15 | 53.26 | 0.00 | 500050 |
| | **TCQEA** | **70.00** | **70.00** | **70.00** | **70.00** | **100.00** | **345** |
| K=80 | UCQEA | 63.83 | 59.87 | 57.00 | 60.15 | 0.00 | 500050 |
| | TCQEA | **80.00** | **80.00** | **80.00** | **80.00** | **100.00** | **360** |
| K=90 | UCQEA | 72.39 | 67.36 | 65.20 | 67.64 | 0.00 | 500050 |
| | TCQEA | **90.00** | **90.00** | **90.00** | **90.00** | **100.00** | **389** |
| K=100 | UCQEA | 78.79 | 74.84 | 71.61 | 75.50 | 0.00 | 500050 |
| | TCQEA | **100.00** | **100.00** | **100.00** | **100.00** | **100.00** | **415** |
| K=110 | UCQEA | 87.36 | 81.79 | 79.45 | 82.16 | 0.00 | 500050 |
| | TCQEA | **110.00** | **110.00** | **110.00** | **110.00** | **100.00** | **441** |
| K=120 | UCQEA | 93.40 | 90.35 | 85.50 | 90.18 | 0.00 | 500050 |
| | TCQEA | **120.00** | **120.00** | **120.00** | **120.00** | **100.00** | **469** |
| K=130 | UCQEA | 100.17 | 96.93 | 93.34 | 96.93 | 0.00 | 500050 |
| | TCQEA | **130.00** | **130.00** | **130.00** | **130.00** | **100.00** | **495** |
| K=140 | UCQEA | 106.57 | 103.70 | 100.10 | 103.84 | 0.00 | 500050 |
| | TCQEA | **140.00** | **140.00** | **140.00** | **140.00** | **100.00** | **553** |
| K=150 | UCQEA | 115.86 | 112.44 | 107.59 | 111.95 | 0.00 | 500050 |
| | TCQEA | **150.00** | **150.00** | **150.00** | **150.00** | **100.00** | **658** |
| K=160 | UCQEA | 123.70 | 119.03 | 113.63 | 119.18 | 0.00 | 500050 |
| | TCQEA | **160.00** | **160.00** | **160.00** | **160.00** | **100.00** | **713** |
| K=170 | UCQEA | 130.82 | 125.97 | 122.56 | 126.13 | 0.00 | 500050 |
| | TCQEA | **170.00** | **170.00** | **170.00** | **170.00** | **100.00** | **873** |
| K=180 | UCQEA | 137.95 | 133.28 | 128.96 | 133.25 | 0.00 | 500050 |
| | TCQEA | **180.00** | **180.00** | **180.00** | **180.00** | **100.00** | **921** |
| K=190 | UCQEA | 144.35 | 140.22 | 136.45 | 140.16 | 0.00 | 500050 |
| | TCQEA | **190.00** | **190.00** | **190.00** | **190.00** | **100.00** | **1139** |
| K=200 | UCQEA | 151.84 | 147.88 | 145.01 | 147.99 | 0.00 | 500050 |
| | TCQEA | **200.00** | **200.00** | **199.00** | **199.96** | **93.30** | **3338** |

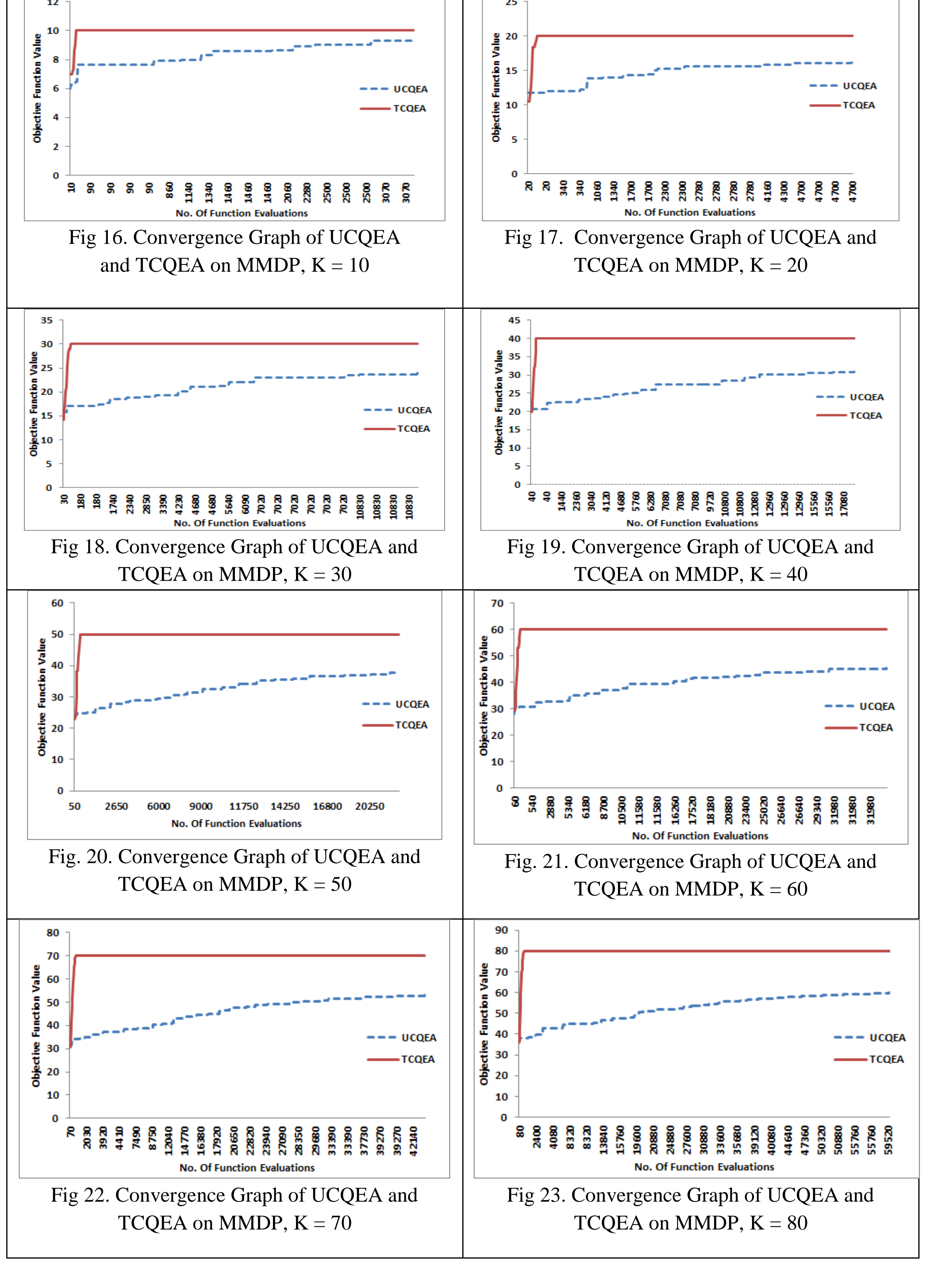

Fig 16. Convergence Graph of UCQEA and TCQEA on MMDP, K = 10

Fig 17. Convergence Graph of UCQEA and TCQEA on MMDP, K = 20

Fig 18. Convergence Graph of UCQEA and TCQEA on MMDP, K = 30

Fig 19. Convergence Graph of UCQEA and TCQEA on MMDP, K = 40

Fig. 20. Convergence Graph of UCQEA and TCQEA on MMDP, K = 50

Fig. 21. Convergence Graph of UCQEA and TCQEA on MMDP, K = 60

Fig 22. Convergence Graph of UCQEA and TCQEA on MMDP, K = 70

Fig 23. Convergence Graph of UCQEA and TCQEA on MMDP, K = 80

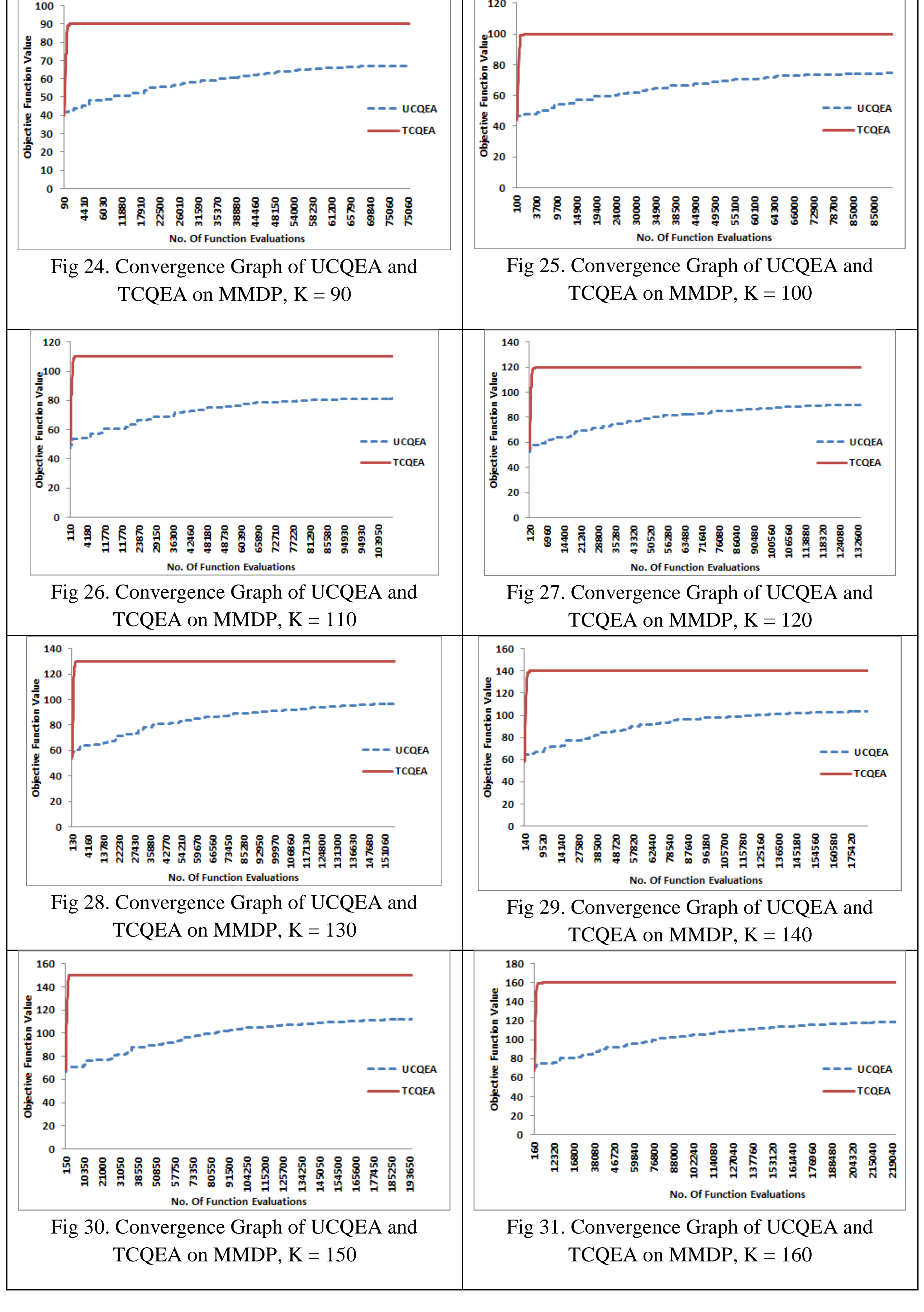

Fig 24. Convergence Graph of UCQEA and TCQEA on MMDP, K = 90

Fig 25. Convergence Graph of UCQEA and TCQEA on MMDP, K = 100

Fig 26. Convergence Graph of UCQEA and TCQEA on MMDP, K = 110

Fig 27. Convergence Graph of UCQEA and TCQEA on MMDP, K = 120

Fig 28. Convergence Graph of UCQEA and TCQEA on MMDP, K = 130

Fig 29. Convergence Graph of UCQEA and TCQEA on MMDP, K = 140

Fig 30. Convergence Graph of UCQEA and TCQEA on MMDP, K = 150

Fig 31. Convergence Graph of UCQEA and TCQEA on MMDP, K = 160

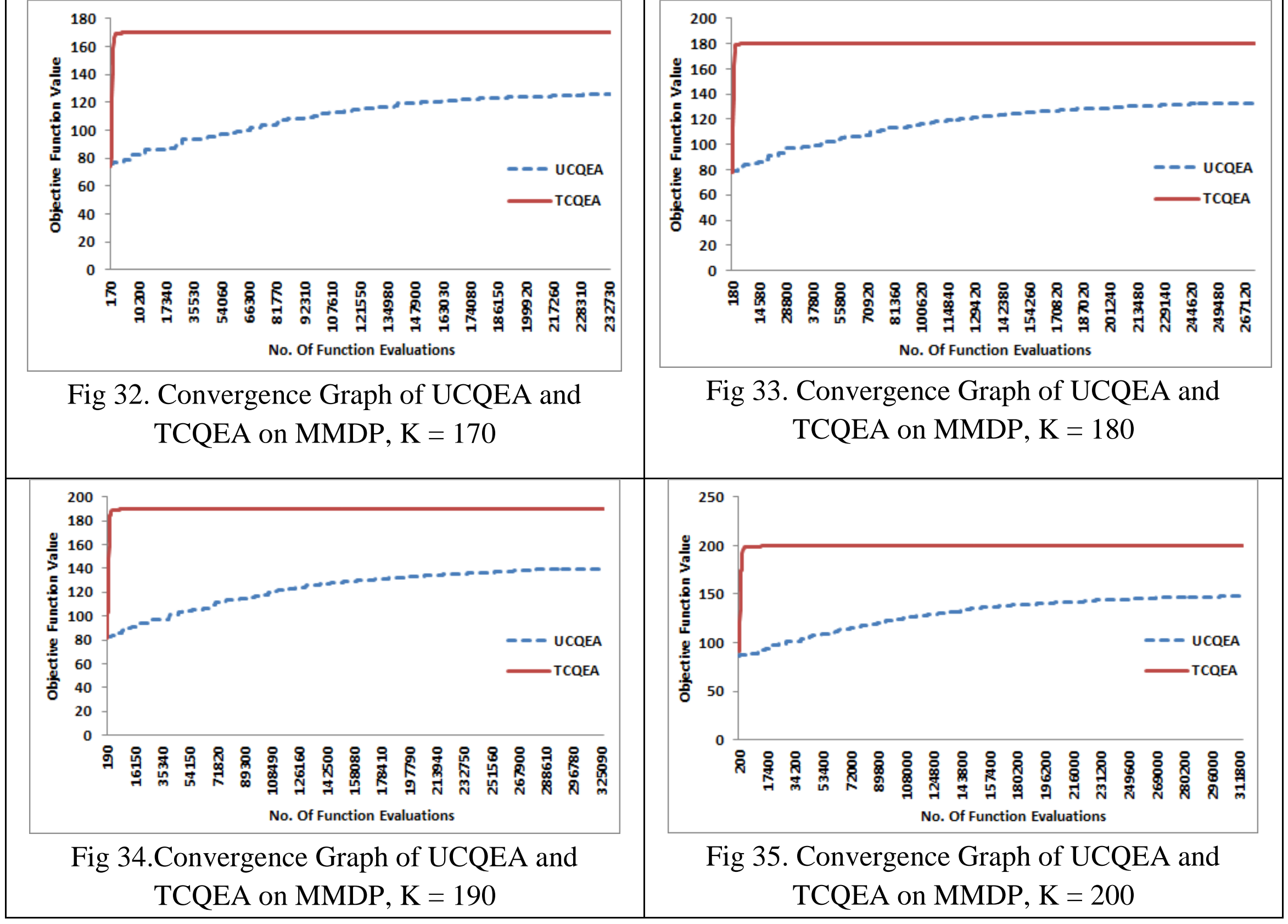

Fig 32. Convergence Graph of UCQEA and TCQEA on MMDP, K = 170

Fig 33. Convergence Graph of UCQEA and TCQEA on MMDP, K = 180

Fig 34.Convergence Graph of UCQEA and TCQEA on MMDP, K = 190

Fig 35. Convergence Graph of UCQEA and TCQEA on MMDP, K = 200

The performance of Tuned QEA is superior to Canonical QEA for all instances of MMDP used in this work as indicated by the Table 5 and Fig. 16 to 35. This indicates the success of the proposed tuning method for tuning QEA on problems like MMDP.

In order to confirm the findings in Table 5, multi-problem non-parametric Wilcoxon's Signed Rank Test [Der2011] was performed on average objective function value (OFV) and average NFE of Tuned QEA and Canonical QEA on all the instances of MMDP at a significance level of 5%. In case of comparison on average OFV, the null hypothesis for comparison was average OFV of Tuned QEA $\mu_1$ is less than or equal to the average OFV of Canonical QEA $\mu_2$ i.e. $H_0$: $\mu_1 \leq \mu_2$ and the alternate hypothesis is H1: $\mu_1 > \mu_2$. The result of test is presented in Table 6, which shows that null hypothesis can be safely rejected as Wilcoxon's Signed Rank test statistic is zero and less than the critical value of 60 at significance level of α = 5%. This indicates the success of the proposed tuning method for tuning QEA on problems like MMDP.

**Table 6: Wilcoxon's Signed Rank Test on MMDP Instances (OFV)**

| | TCQEA | UCQEA |
|---|---|---|
| | **$X_1$** | **$X_2$** |
| 1 | 10 | 9.17 |
| 2 | 20 | 16.67 |
| 3 | 30 | 24.06 |
| 4 | 40 | 31.1 |
| 5 | 50 | 38.5 |
| 6 | 60 | 45.56 |
| 7 | 70 | 53.26 |
| 8 | 80 | 60.15 |
| 9 | 90 | 67.64 |
| 10 | 100 | 75.5 |
| 11 | 110 | 82.16 |

| | |
|---|---|
| Σ(+) | 210 |
| Σ(−) | 0 |
| *n* | 20 |

| Null Hypothesis | **Test Statistic** *T* | **Critical Value** At an α of 5% |
|---|---|---|
| $H_0$: $\mu_1$ <= $\mu_2$ | 0 | 60 |

| | | | |
|---|---|---|---|
| 12 | 120 | 90.18 | |
| 13 | 130 | 96.93 | |
| 14 | 140 | 103.84 | |
| 15 | 150 | 111.95 | |
| 16 | 160 | 119.18 | |
| 17 | 170 | 126.13 | |
| 18 | 180 | 133.25 | |
| 19 | 190 | 140.16 | |
| 20 | 200 | 147.99 | |

In case of comparison on average NFE, the null hypothesis for comparison was average NFE of Tuned QEA $\mu_1$ is greater than or equal to the average NFE of Canonical QEA $\mu_2$ i.e. $H_0$: $\mu_1 \geq \mu_2$ and the alternate hypothesis is H1: $\mu_1 < \mu_2$. The result of test is presented in Table 7, which shows that null hypothesis can be safely rejected as Wilcoxon's Signed Rank test statistic is zero and less than the critical value of 60 at significance level of $\alpha$ = 5%. This indicates the success of the proposed tuning method for tuning QEA on problems like MMDP.

**Table 7: Wilcoxon's Signed Rank Test on MMDP Instances (NFE)**

| | TCQEA | UCQEA |
|---|---|---|
| | $X_1$ | $X_2$ |
| 1 | 182 | 500050 |
| 2 | 234 | 500050 |
| 3 | 259 | 500050 |
| 4 | 280 | 500050 |
| 5 | 297 | 500050 |
| 6 | 280 | 500050 |
| 7 | 345 | 500050 |
| 8 | 360 | 500050 |
| 9 | 389 | 500050 |
| 10 | 415 | 500050 |
| 11 | 441 | 500050 |
| 12 | 469 | 500050 |
| 13 | 495 | 500050 |
| 14 | 553 | 500050 |
| 15 | 658 | 500050 |
| 16 | 713 | 500050 |
| 17 | 873 | 500050 |
| 18 | 921 | 500050 |
| 19 | 1139 | 500050 |
| 20 | 3338 | 500050 |

| | |
|---|---|
| $\Sigma(+)$ | 0 |
| $\Sigma(-)$ | 210 |
| $n$ | 20 |

| Null Hypothesis | Test Statistic $T$ | Critical Value At an $\alpha$ of 5% |
|---|---|---|
| $H_0$: $\mu_1$ >= $\mu_2$ | 0 | 60 |

The same set of parameters has been also used for solving a instances of a well-known problems known as COUNTSAT.

B. *COUNTSAT problem* [43]

It is an instance of the MAXSAT problem. In COUNTSAT, the value of a given solution is the number of satisfied clauses (among all the possible Horn clauses of three variables) by an input composed by n boolean variables. It is easy to check that the optimum is obtained when the value of all the variables is 1 i.e. s= n.

In this study we consider the instance of n = 20 to 1000 variables, and thus, the value of the optimal solution varies from 6860 to 997003000.

$$f_{\text{COUNTSAT}}(s) = s + n.(n-1).(n-2) - 2.(n-2).\binom{s}{2} + 6.\binom{s}{3}$$

$$\text{For (n=20)} \quad = s + 6840 - 18.s.(s-1) + s.(s-1).(s-2) \qquad (7)$$

For (n=1000) $= s + 997002000 - 998.s.(s-1) + s.(s-1).(s-2)$

A comparative study performed between parameter Tuned QEA (TCQEA) and Canonical QEA (UCQEA) with Parameters given in Table 2 on COUNTSAT with a set of 21 problems of size 20 to 1000. The results are given in Table 8. Thirty independent runs were made on each problem size and the comparison has been made on Best, Median, Worst, Mean, percentage of runs in which optimal was achieved i.e. percentage of Success runs, and Average number of function evaluations (NFE). The Canonical QEA was able to reach optimal value till problem size 700, but was not able to find the optimal for rest of the problem instances whereas the Tuned QEA was able to reach 100% success in all the problem instances. The performance of Tuned QEA was also good on speed of convergence as indicated by average NFE and the convergence graphs shown in fig. 36 to 56, which also compared the speed of convergence of Tuned QEA to Canonical QEA.

The convergence graphs have been plotted between objective function value and number of generations for both Tuned-QEA (TCQEA) and Canonical QEA (UCQEA) for all the problem instances of COUNTSAT for the median run. The convergence graph clearly establishes the superiority of Tuning as the Tuned QEA is able to reach the optimal in less than 30 generations in all the graphs whereas the Canonical QEA is much slower and takes much larger number of generations to reach near the optimal, which increase with the size of the problem. The distance from optimal has increased as the size of the problem has increased in case of Canonical QEA but Tuned QEA has performed consistently well for this class of problems.

TABLE 8
COMPARATIVE STUDY BETWEEN UCQEA AND TCQEA ON COUNTSAT PROBLEM INSTANCES

| Prob. | Algo | Best | Worst | Average | Median | % Success Runs | Std | Avg. NFE |
|---|---|---|---|---|---|---|---|---|
| K=20 | UCQEA | 6860 | 6860 | 6860 | 6860 | 100 | 0 | 1152 |
| | **TCQEA** | **6860** | **6860** | **6860** | **6860** | **100** | **0** | **106** |
| K=50 | UCQEA | 117650 | 117650 | 117650 | 117650 | 100 | 0 | 4907 |
| | **TCQEA** | **117650** | **117650** | **117650** | **117650** | **100** | **0** | **177** |
| K=100 | UCQEA | 970300 | 970300 | 970300 | 970300 | 100 | 0 | 10462 |
| | **TCQEA** | **970300** | **970300** | **970300** | **970300** | **100** | **0** | **226** |
| K=150 | UCQEA | 3307950 | 3307801 | 3307925 | 3307950 | 83 | 56 | 20863 |
| | **TCQEA** | **3307950** | **3307950** | **3307950** | **3307950** | **100** | **0** | **255** |
| K=200 | UCQEA | 7880600 | 7880401 | 7880580 | 7880600 | 90 | 61 | 22300 |
| | **TCQEA** | **7880600** | **7880600** | **7880600** | **7880600** | **100** | **0** | **298** |
| K=250 | UCQEA | 15438250 | 15435052 | 15438102 | 15438250 | 80 | 584 | 29220 |
| | **TCQEA** | **15438250** | **15438250** | **15438250** | **15438250** | **100** | **0** | **327** |
| K=300 | UCQEA | 26730900 | 26670765 | 26727928 | 26730900 | 87 | 11303 | 30092 |
| | **TCQEA** | **26730900** | **26730900** | **26730900** | **26730900** | **100** | **0** | **347** |
| K=350 | UCQEA | 42508550 | 42297626 | 42487362 | 42508550 | 80 | 52963 | 34312 |
| | **TCQEA** | **42508550** | **42508550** | **42508550** | **42508550** | **100** | **0** | **364** |
| K=400 | UCQEA | 63521200 | 63258981 | 63512459 | 63521200 | 97 | 47874 | 34483 |
| | **TCQEA** | **63521200** | **63521200** | **63521200** | **63521200** | **100** | **0** | **382** |
| K=450 | UCQEA | 90518850 | 89462956 | 90414627 | 90518850 | 83 | 259748 | 39618 |
| | **TCQEA** | **90518850** | **90518850** | **90518850** | **90518850** | **100** | **0** | **401** |
| K=500 | UCQEA | 124251500 | 121775681 | 123726361 | 124251500 | 70 | 762261 | 44297 |
| | **TCQEA** | **124251500** | **124251500** | **124251500** | **124251500** | **100** | **0** | **431** |
| K=550 | UCQEA | 165469150 | 163088041 | 165117481 | 165469150 | 67 | 681032 | 46058 |
| | **TCQEA** | **165469150** | **165469150** | **165469150** | **165469150** | **100** | **0** | **432** |
| K=600 | UCQEA | 214921800 | 209881401 | 213598979 | 214385692 | 37 | 1524388 | 48732 |
| | **TCQEA** | **214921800** | **214921800** | **214921800** | **214921800** | **100** | **0** | **468** |
| K=650 | UCQEA | 273359450 | 267304186 | 271874229 | 272729917 | 33 | 1881364 | 49330 |
| | **TCQEA** | **273359450** | **273359450** | **273359450** | **273359450** | **100** | **0** | **478** |
| K=700 | UCQEA | 341532100 | 324622561 | 338459233 | 340076761 | 3 | 3598957 | 50003 |
| | **TCQEA** | **341532100** | **341532100** | **341532100** | **341532100** | **100** | **0** | **515** |
| K=750 | UCQEA | 419072236 | 406621801 | 416371450 | 417418345 | 0 | 3027881 | 50050 |
| | **TCQEA** | **420189750** | **420189750** | **420189750** | **420189750** | **100** | **0** | **535** |
| K=800 | UCQEA | 508810386 | 498371881 | 504276211 | 504764499 | 0 | 2502890 | 50050 |
| | **TCQEA** | **510082400** | **510082400** | **510082400** | **510082400** | **100** | **0** | **588** |
| K=850 | UCQEA | 607691092 | 584540601 | 601343410 | 603543166 | 0 | 6311747 | 50050 |
| | **TCQEA** | **611960050** | **611960050** | **611960050** | **611960050** | **100** | **0** | **647** |
| K=900 | UCQEA | 717888937 | 698472201 | 712273067 | 714460559 | 0 | 5259800 | 50050 |
| | **TCQEA** | **726572700** | **726572700** | **726572700** | **726572700** | **100** | **0** | **644** |
| K=950 | UCQEA | 843268921 | 793964756 | 834497691 | 838219561 | 0 | 10389589 | 50050 |
| | **TCQEA** | **854670350** | **854670350** | **854670350** | **854670350** | **100** | **0** | **843** |
| K=1000 | UCQEA | 979662826 | 948748605 | 969462614 | 972362722 | 0 | 8750511 | 50050 |
| | **TCQEA** | **997003000** | **997003000** | **997003000** | **997003000** | **100** | **0** | **881** |

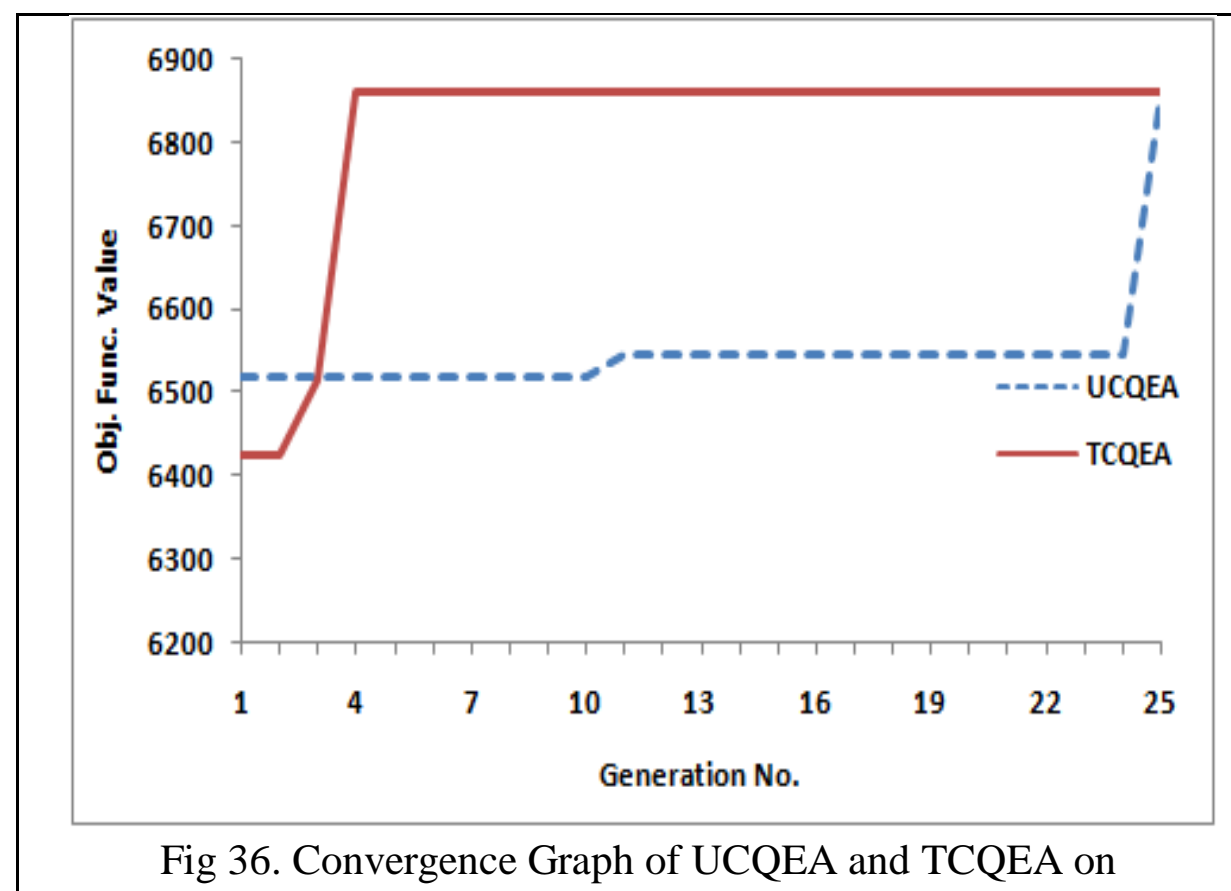

Fig 36. Convergence Graph of UCQEA and TCQEA on COUNTSAT Problem size, K = 20

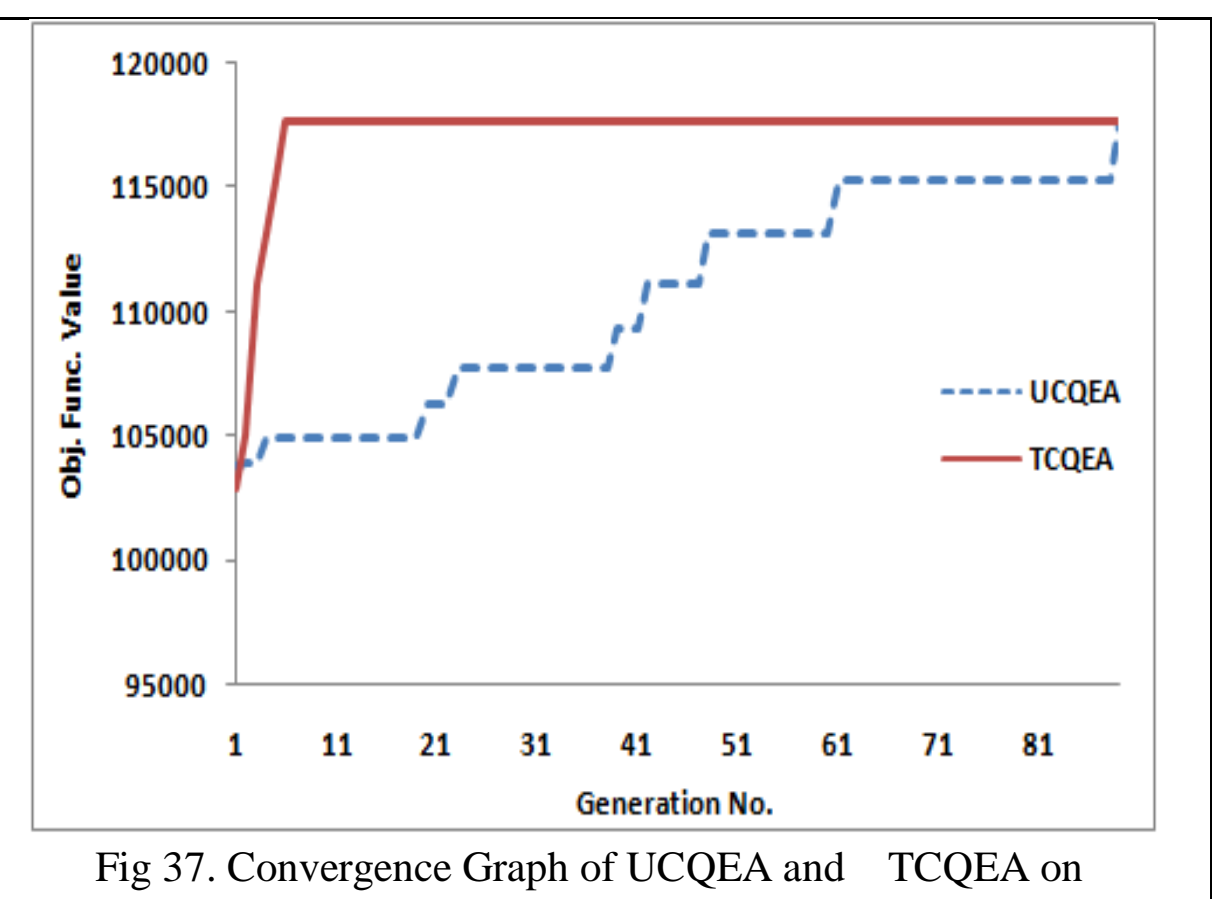

Fig 37. Convergence Graph of UCQEA and TCQEA on COUNTSAT Problem size, K =50

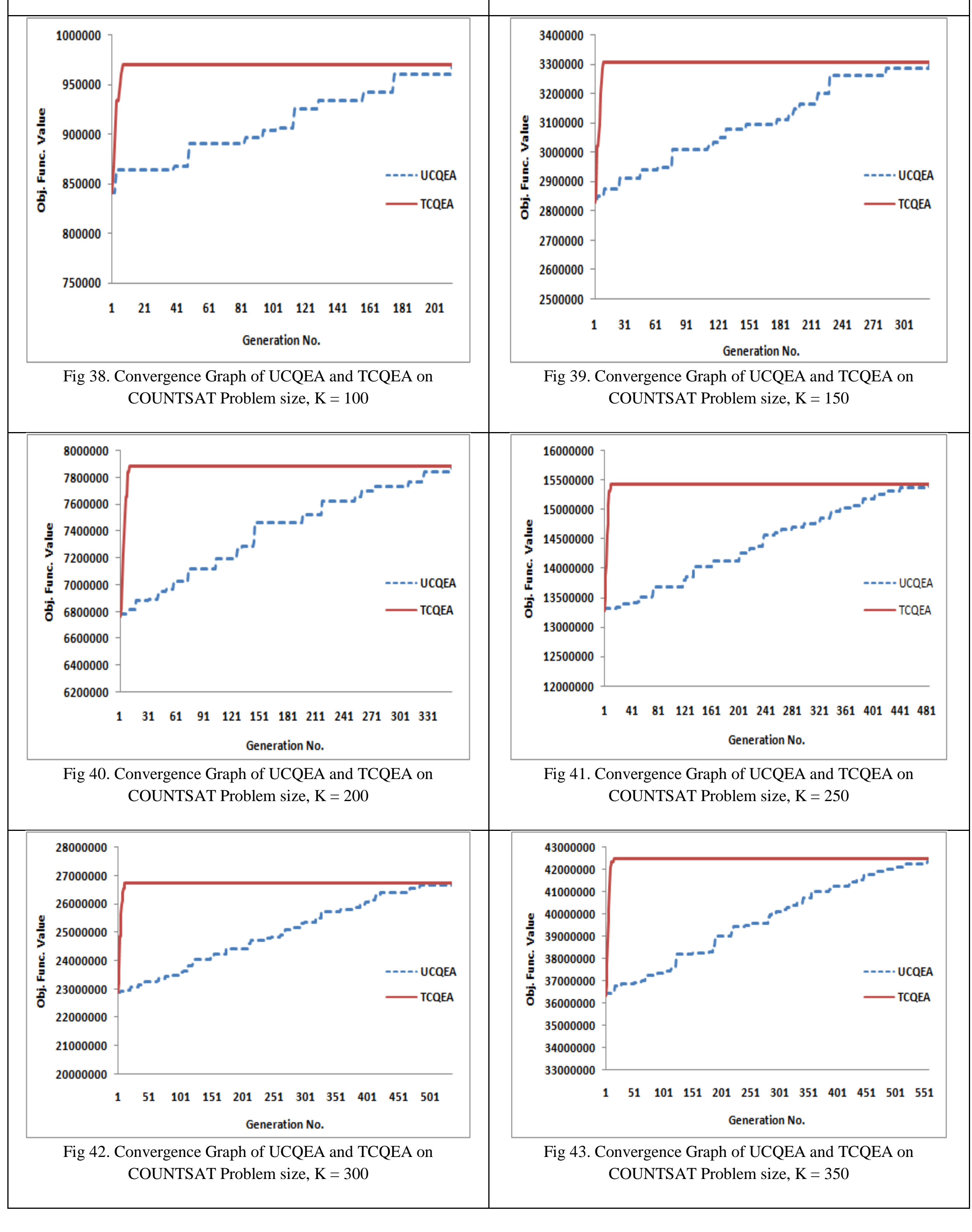

Fig 38. Convergence Graph of UCQEA and TCQEA on COUNTSAT Problem size, K = 100

Fig 39. Convergence Graph of UCQEA and TCQEA on COUNTSAT Problem size, K = 150

Fig 40. Convergence Graph of UCQEA and TCQEA on COUNTSAT Problem size, K = 200

Fig 41. Convergence Graph of UCQEA and TCQEA on COUNTSAT Problem size, K = 250

Fig 42. Convergence Graph of UCQEA and TCQEA on COUNTSAT Problem size, K = 300

Fig 43. Convergence Graph of UCQEA and TCQEA on COUNTSAT Problem size, K = 350

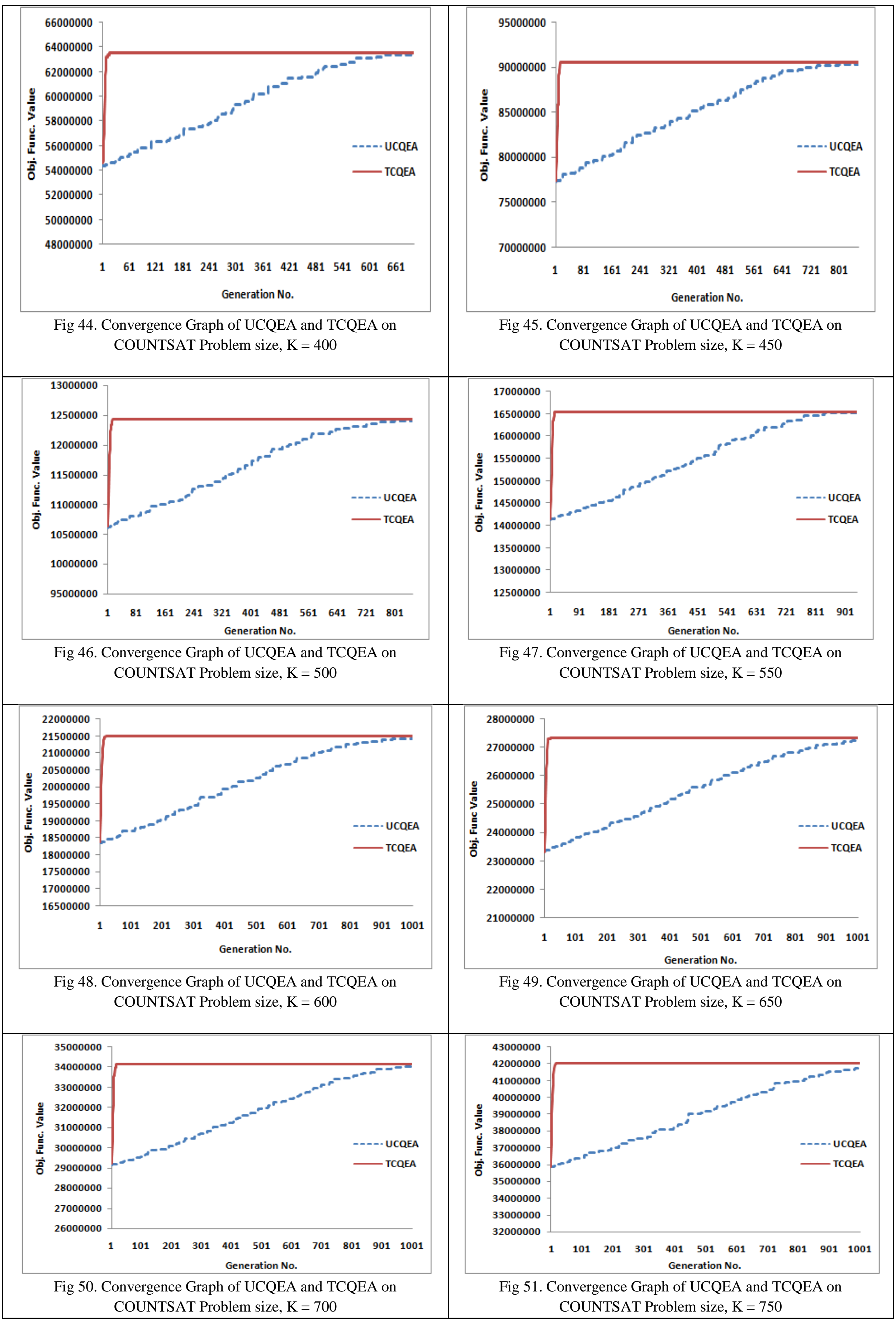

Fig 44. Convergence Graph of UCQEA and TCQEA on COUNTSAT Problem size, K = 400

Fig 45. Convergence Graph of UCQEA and TCQEA on COUNTSAT Problem size, K = 450

Fig 46. Convergence Graph of UCQEA and TCQEA on COUNTSAT Problem size, K = 500

Fig 47. Convergence Graph of UCQEA and TCQEA on COUNTSAT Problem size, K = 550

Fig 48. Convergence Graph of UCQEA and TCQEA on COUNTSAT Problem size, K = 600

Fig 49. Convergence Graph of UCQEA and TCQEA on COUNTSAT Problem size, K = 650

Fig 50. Convergence Graph of UCQEA and TCQEA on COUNTSAT Problem size, K = 700

Fig 51. Convergence Graph of UCQEA and TCQEA on COUNTSAT Problem size, K = 750

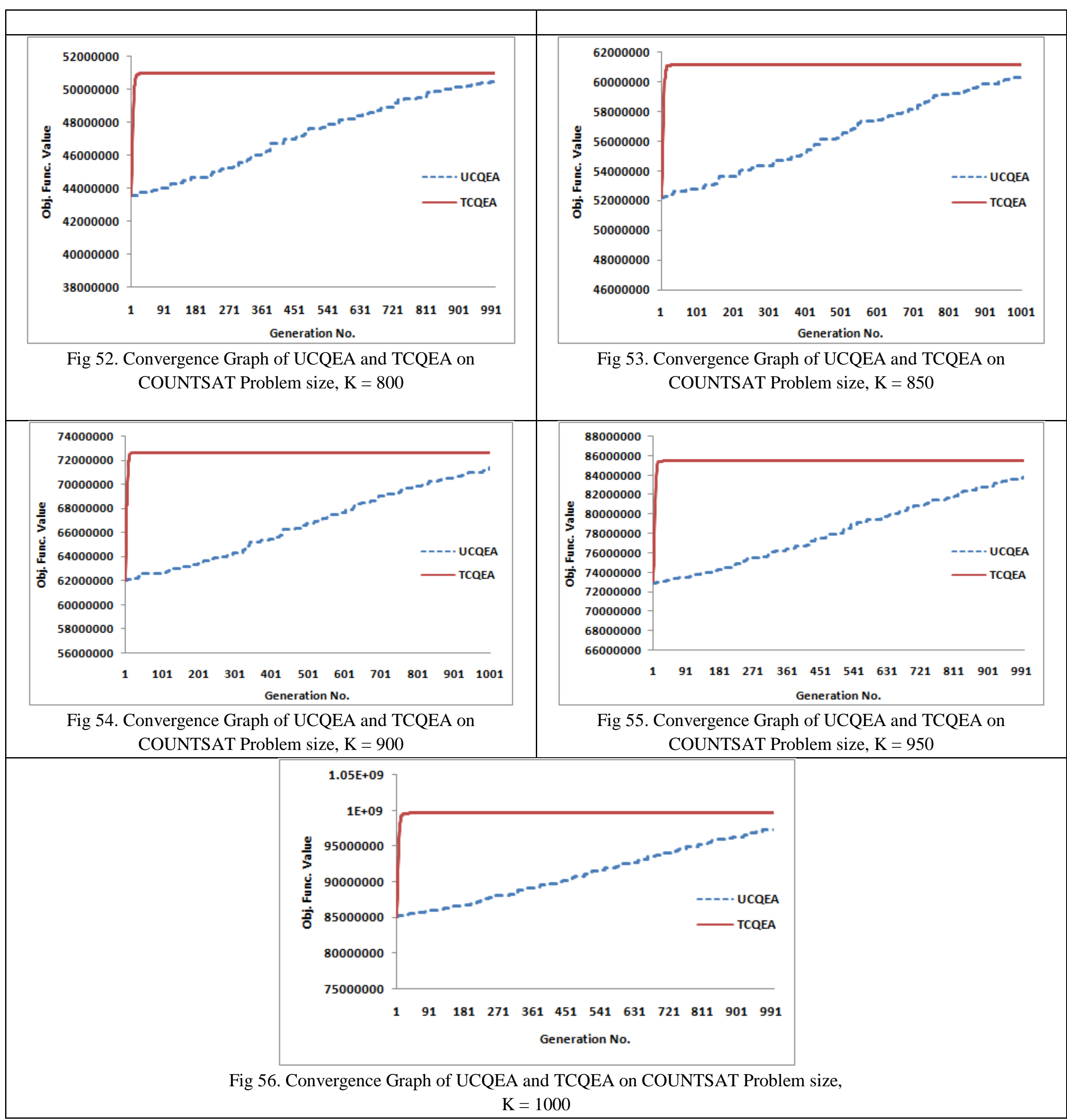

Fig 52. Convergence Graph of UCQEA and TCQEA on COUNTSAT Problem size, K = 800

Fig 53. Convergence Graph of UCQEA and TCQEA on COUNTSAT Problem size, K = 850

Fig 54. Convergence Graph of UCQEA and TCQEA on COUNTSAT Problem size, K = 900

Fig 55. Convergence Graph of UCQEA and TCQEA on COUNTSAT Problem size, K = 950

Fig 56. Convergence Graph of UCQEA and TCQEA on COUNTSAT Problem size, K = 1000

The performance of Tuned QEA is superior to Canonical QEA for all instances of COUNTSAT Problem used in this work as indicated by the Table 8 and Fig. 36 to 56. This indicates the success of the proposed tuning method for problems like COUNTSAT problem.

In order to confirm the findings in Table 8, multi-problem non-parametric Wilcoxon's Signed Rank Test [Der2011] was performed on average objective function value (OFV) and average NFE of Tuned QEA and Canonical QEA on all the instances of COUNTSAT at a significance level of 5%. In case of comparison on average OFV, the null hypothesis for comparison was average OFV of Tuned QEA $\mu_1$ is less than or equal to the average OFV of Canonical QEA $\mu_2$ i.e. $H_0$: $\mu_1 \leq \mu_2$ and the alternate hypothesis is H1: $\mu_1 > \mu_2$. The result of test is presented in Table 9, which shows that null hypothesis can be safely rejected as Wilcoxon's Signed Rank test statistic is zero and less than the critical value of 47 at significance level of $\alpha = 5\%$. This indicates the success of the proposed tuning method for tuning QEA on problems like COUNTSAT.

In case of comparison on average NFE, the null hypothesis for comparison was average NFE of Tuned QEA $\mu_1$ is greater than or equal to the average NFE of Canonical QEA $\mu_2$ i.e. $H_0$: $\mu_1 \geq \mu_2$ and the alternate hypothesis is $\mu_1 < \mu_2$. The result of test is presented in Table 10, which shows that null hypothesis can be safely rejected as Wilcoxon's

Signed Rank test statistic is zero and less than the critical value of 68 at significance level of α = 5%. This indicates the success of the proposed tuning method for tuning QEA on problems like COUNTSAT.

**Table 9: Wilcoxon's Signed Rank Test on COUNTSAT Instances (OFV)**

| | TCQEA | UCQEA |
|---|---|---|
| | $X_1$ | $X_2$ |
| 1 | 6860 | 6860 |
| 2 | 117650 | 117650 |
| 3 | 970300 | 970300 |
| 4 | 3307950 | 3307925 |
| 5 | 7880600 | 7880580 |
| 6 | 15438250 | 15438102 |
| 7 | 26730900 | 26727928 |
| 8 | 42508550 | 42487362 |
| 9 | 63521200 | 63512459 |
| 10 | 90518850 | 90414627 |
| 11 | 124251500 | 123726361 |
| 12 | 165469150 | 165117481 |
| 13 | 214921800 | 213598979 |
| 14 | 273359450 | 271874229 |
| 15 | 341532100 | 338459233 |
| 16 | 420189750 | 416371450 |
| 17 | 510082400 | 504276211 |
| 18 | 611960050 | 601343410 |
| 19 | 726572700 | 712273067 |
| 20 | 854670350 | 834497691 |
| 21 | 997003000 | 969462614 |

| | |
|---|---|
| Σ(+) | 171 |
| Σ(–) | 0 |
| *n* | 18 |

| Null Hypothesis | Test Statistic *T* | Critical Value At an α of 5% |
|---|---|---|
| $H_0$: $\mu_1 <= \mu_2$ | 0 | 47 |

**Table 10: Wilcoxon's Signed Rank Test on COUNTSAT Instances (NFE)**

| | TCQEA | UCQEA |
|---|---|---|
| | $X_1$ | $X_2$ |
| 1 | 106 | 1152 |
| 2 | 177 | 4907 |
| 3 | 226 | 10462 |
| 4 | 255 | 20863 |
| 5 | 298 | 22300 |
| 6 | 327 | 29220 |
| 7 | 347 | 30092 |
| 8 | 364 | 34312 |
| 9 | 382 | 34483 |
| 10 | 401 | 39618 |
| 11 | 431 | 44297 |
| 12 | 432 | 46058 |
| 13 | 468 | 48732 |
| 14 | 478 | 49330 |
| 15 | 515 | 50003 |
| 16 | 535 | 50050 |
| 17 | 588 | 50050 |
| 18 | 647 | 50050 |

| | |
|---|---|
| Σ(+) | 0 |
| Σ(–) | 231 |
| *N* | 21 |

| Null Hypothesis | Test Statistic *T* | Critical Value At an α of 5% |
|---|---|---|
| $H_0$: $\mu_1 >= \mu_2$ | 0 | 68 |

| 19 | 644 | 50050 | |
|---|---|---|---|
| 20 | 843 | 50050 | |
| 21 | 881 | 50050 | |

Therefore, the same set of parameter has been used for solving instances of two well-known benchmark problems viz., MMDP and COUNTSAT problem. However, it was found that the parameter vector did not perform well on 0-1 Knapsack problem instances, so it was decided to again tune the QEA using the proposed method for 0-1 knapsack problem.

*C. 0-1 Knapsack problems*

The 0-1 knapsack problem is a profit maximization problem, in which there are n items of different profit and weight available for selection. The selection is made to maximize the profit while keeping the weight of the selected items below the capacity of the knapsack. It is formulated as follows:

Given a set of n items and a knapsack of Capacity C, select a subset of the items to maximize the profit f(x):

$$f(x) = \sum p_i x_i \quad (8)$$

subject to the condition

$$\sum w_i x_i < C \quad (9)$$

where $x_i = (x_1 \ldots. x_n)$, $x_i$ is 0 or 1, $p_i$ is the profit of item *i*, $w_i$ is the weight of item *i*. If the *i*th item is selected for the knapsack, $x_i = 1$, else $x_i = 0$.

Eleven groups of randomly generated instances of (KP) which have been constructed to test the canonical and Tuned-QEA. In all instances the weights are uniformly distributed in a given interval. The profits are expressed as a function of the weights, yielding the specific properties of each group [46].

i. **Uncorrelated data instances**: The profits, *pj* and weights, *wj* of the items are chosen randomly in [1, 1000] so there is no correlation between the profit and weight of an item. These instances represent situations where it can be safely assumed that there is no correlation between weight and profits of the items and are generally easy to solve, as there is a large variation between the profits and weights.

ii. **Weakly correlated instances**: The weights *wj* of the items are chosen randomly in [1, 1000] and the profits *pj* are function of *wj* i.e. *pj* lies in [ *wj* – 100, *wj* + 100] such that $pj \geq 1$. Weakly correlated instances have relatively high correlation between the profit and weight of an item as the profit differs from the weight by only a few percent. These instances are quite realistic in management, as the return of an investment is mostly proportional to the sum invested with some random variations.

iii. **Strongly correlated instances**: The weights *wj* of the items are distributed in [1, 1000] and profit $pj = wj + 100$. These instances correspond to real-life situations where the return is proportional to the investment plus some fixed charge for each project. The strongly correlated instances are mostly hard to solve as they are *ill-conditioned* and Sorting is not of much help.

iv. **Inverse strongly correlated instances**: The profits *pj* of the items are distributed in [1, 1000] and weight $wj = pj + 100$. These instances are similar to strongly correlated instances, however, the fixed charge is negative.

v. **Almost strongly correlated instances**: The weights *wj* of the items are distributed in [1, 1000] and the profits *pj* in [*wj* + 98, *wj* + 102]. These instances are type of fixed-charge problem with some randomness and have the properties of both strongly and weakly correlated instances.

vi. **Subset sum instances**: The weights, *wj,* of the items are randomly distributed in [1, 1000] and the profit $pj = wj$. These instances represent situation in which the profit of each item is equal (or proportional) to the weight so the goal is to obtain a filled knapsack.

vii. **Uncorrelated instances with similar weights**: The weights of the items, *wj,* are distributed uniformly in [100 000, 100 100] and the profits, *pj* in [1, 1000]. These instances are similar to uncorrelated data instances but all the items have similar weights with large difference in profits.

viii. **Spanner instances** span *(v, m)*: These instances are constructed from spanner set i.e. all the items are multiples of a very small set of items. The spanner instances span *(v, m)* are defined by the size, *v*, of the spanner set, the multiplier limit, *m*, and the distribution (uncorrelated, weakly correlated, strongly correlated, etc.) of the items in the spanner set. The instances used in this work are generated as follows: A set of $v=2$ items is generated with weights in the interval [1, 1000], and profits according to the strongly correlated distribution. The items (*pk*, *wk*) in the spanner set are normalized by dividing the profits and weights with $m + 1$. The *n* items are then constructed, by repeatedly choosing an item (*pk*, *wk*) from the spanner set, and a multiplier, *a,* randomly generated in the interval [1, 10]. The constructed item has profit and weight (*a*pk, a*wk*). Computational experiments have showed that the instances became harder to solve for smaller spanner sets [46], so the instances with strongly correlated span (2, 10) have been used in this work.

ix. **multiple strongly correlated instances** mstr( $k1, k2, d$) : They are constructed as a combination of two sets of strongly correlated instances, which have profits $pj := wj + ki$ where $ki$, i = 1,2 is different for the two sets. The multiple strongly correlated instances mstr( $k1, k2, d$) have been generated in this work as follows: The weights of the $n$ items are randomly distributed in [1, 1000]. If the weight $wj$ is divisible by $d=6$, then we set the profit $pj := wj + k1$ otherwise set it to $pj := wj + k2$. The weights $wj$ in the first group (i.e. where $pj = wj + k1$) will all be multiples of $d$, so that using only these weights can at most use $d[c/d]$of the capacity, therefore, in order to obtain a completely filled knapsack, some of the items from the second distribution will also be included. Computational experiments have shown that very difficult instances could be obtained with the Parameters mstr(300, 200, $6$) [46].

x. **profit ceiling instances** pceil($d$) : These instances have profits of all items as multiples of a given parameter $d$. The weights of the $n$ items are randomly distributed in [1, 1000], and their profits are set to $pj = d[wj/d]$. The parameter $d$ has been experimentally chosen as $d$=3, as this resulted in sufficiently difficult instances [46].

xi. **circle instances** circle($d$) : These instances have the profit of their items as function of the weights form an arc of a circle (actually an ellipsis). The weights are uniformly distributed in [1, 1000] and for each weight $wi$ the corresponding profit is chosen as $pi = d\sqrt{2000^2 - (wi - 2000)^2}$. Experimental results have showed in [Pis2004] that difficult instances appeared by choosing $d$= 2/3 which was chosen for testing in this work.

The set of parameter vector that was used for successfully solving instances of two well-known benchmark problems viz., MMDP and COUNTSAT problem, did not perform well on 0-1 Knapsack problem instances, so it was decided to again tune the QEA using the proposed method on Strongly Correlated instance with weight of the items, $w_i$, uniformly distributed between [1, 1000] with $p_i = w_i + 100$ and capacity, $C$, as half the total weight of all the available 1000 items.

The initial range of values for parameters in QEA used for tuning is given in Table 11. The parameter range for magnitude of rotation angles for (θ1 to θ8, except θ3 & θ5) is 0.0 to 0.001π, and for θ3 & θ5 is 0.0 to 0.05 π, which is large as compared to the range suggested by [14]. The direction of rotation depends on the sign of α, β and relative fitness as per Table 1. The range for population size is 5 to 100. The range for no. of groups is 1 to 10. The range for global migration is 1 to 200, which is two times the value suggested by [14]. The maximum number of generations was limited to ten thousand. The change in value of each parameter during the tuning process is depicted in Table 12 and shown in Fig. 57 to 67. There were four rounds of exploration and three rounds of exploitation. It can be observed that between round two and round three of the exploration stages, there was slight decrease in the best objective function value. On further, exploration in round four, there was only slight increase in the best objective function value, so it was decided to stop further exploration and start the exploitation so as to further search within the vicinity of the Best parameter values found so far. After the third round of the exploitation stage, there was no improvement in the best objective function value, so it was decided to stop further tuning.

The parameter value for θ1 and θ2 has changed during exploration stage but remained constant during exploitation stage. The parameter value for θ3 changed during exploration and first two round of exploitation stage but did not change in the last round. The parameter value for θ4 initially remained unchanged during first two rounds of exploration and then changed during last round of exploration and first round of exploitation stage, but remained unchanged in the last two rounds of exploitation. The parameter value for θ5 initially remained unchanged then decreased a little before increasing during exploration. It kept decreasing during exploitation stage. The parameter value for θ6 initially remained unchanged, then kept changing during exploration and exploitation stage. The parameter value for θ7 remained almost unchanged. The parameter value for θ8 decreased during initial part of exploration and then increased before becoming constant during exploitation stage. The parameter value for Population Size kept changing during exploration and initial round of exploitation stage, but did not change in the last round. The parameter value for no. of groups initially increased and then remained unchanged during exploration and then increased in first round of exploitation stage. The parameter value for migration changed during exploration and remained unchanged during first two rounds of exploitation stage, but increased in the last round. The final value of each parameter is given in Table 12 in the last row. The convergence graph given in Fig. 68 shows fast convergence to near optimal within 1500 generations.

TABLE 11
INITIAL RANGE OF PARAMETERS

| **Parameter** | **θ1 (* π)** | **θ2 (* π)** | **θ3 (* π)** | **θ4 (* π)** | **θ5 (* π)** | **θ6 (* π)** | **θ7 (* π)** | **θ8 (* π)** | **Pop size** | **No. of Groups** | **Global Migration** |
|---|---|---|---|---|---|---|---|---|---|---|---|
| Lower Limit | 0 | 0 | 0 | 0 | 0 | 0 | 0 | 0 | 5 | 1 | 1 |
| Upper Limit | 0.001 | 0.001 | 0.05 | 0.001 | 0.05 | 0.001 | 0.001 | 0.001 | 100 | 10 | 200 |

TABLE 12
BEST PARAMETER VECTOR (PIVOT) DURING TUNING PROCESS

| **Iter. No.** | **Best Pivot Parameter Value** | | | | | | | | | | | **Best OFV** |
|---|---|---|---|---|---|---|---|---|---|---|---|---|
| | **$\theta 1$ ($* \pi$)** | **$\theta 2$ ($* \pi$)** | **$\theta 3$ ($* \pi$)** | **$\theta 4$ ($* \pi$)** | **$\theta 5$ ($* \pi$)** | **$\theta 6$ ($* \pi$)** | **$\theta 7$ ($* \pi$)** | **$\theta 8$ ($* \pi$)** | **Pop. Size** | **No. of Grp** | **Glb. Mig.** | |
| Explor. – 1 | 0.0009 | 0.0005 | 0.0438 | 0.0006 | 0.0112 | 0.0007 | 0.0005 | 0.0007 | 20 | 3 | 55 | 314869.9892 |
| Explor. – 2 | 0.0002 | 0.0007 | 0.0404 | 0.0006 | 0.0112 | 0.0007 | 0.0005 | 0.0005 | 77 | 7 | 132 | 315169.9709 |
| Explor. – 3 | 0.0004 | 0.0004 | 0.0473 | 0.0006 | 0.0079 | 0.0004 | 0.0005 | 0.0003 | 52 | 8 | 134 | 315169.9404 |
| Explor. – 4 | 0.00058 | 0.0004 | 0.02475 | 0.0003 | 0.0273 | 0.00073 | 0.00045 | 0.00088 | 80 | 8 | 117 | 315169.9770 |
| Exploit. – 5 | 0.00036 | 0.00026 | 0.01949 | 0.00042 | 0.02108 | 0.00086 | 0.00070 | 0.00088 | 68 | 8 | 119 | 315369.9891 |
| Exploit. – 6 | 0.00036 | 0.00026 | 0.01423 | 0.0003 | 0.01485 | 0.00073 | 0.00070 | 0.00088 | 80 | 10 | 119 | 315469.9830 |
| Exploit. – 7 | 0.00035 | 0.00026 | 0.01423 | 0.0003 | 0.01405 | 0.00070 | 0.00067 | 0.00088 | 80 | 10 | 197 | 315469.9830 |

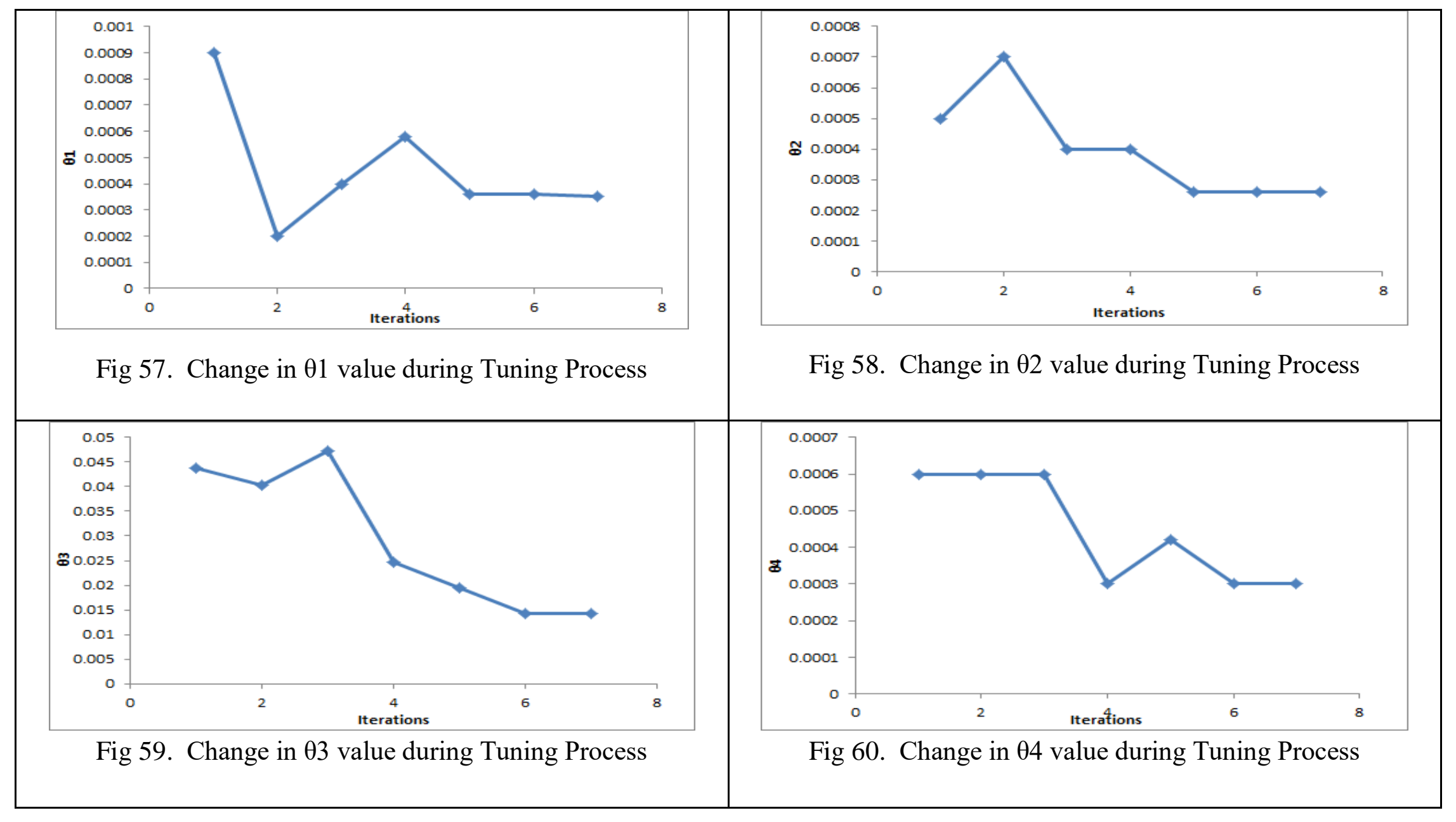

Fig 57. Change in θ1 value during Tuning Process

Fig 58. Change in θ2 value during Tuning Process

Fig 59. Change in θ3 value during Tuning Process

Fig 60. Change in θ4 value during Tuning Process

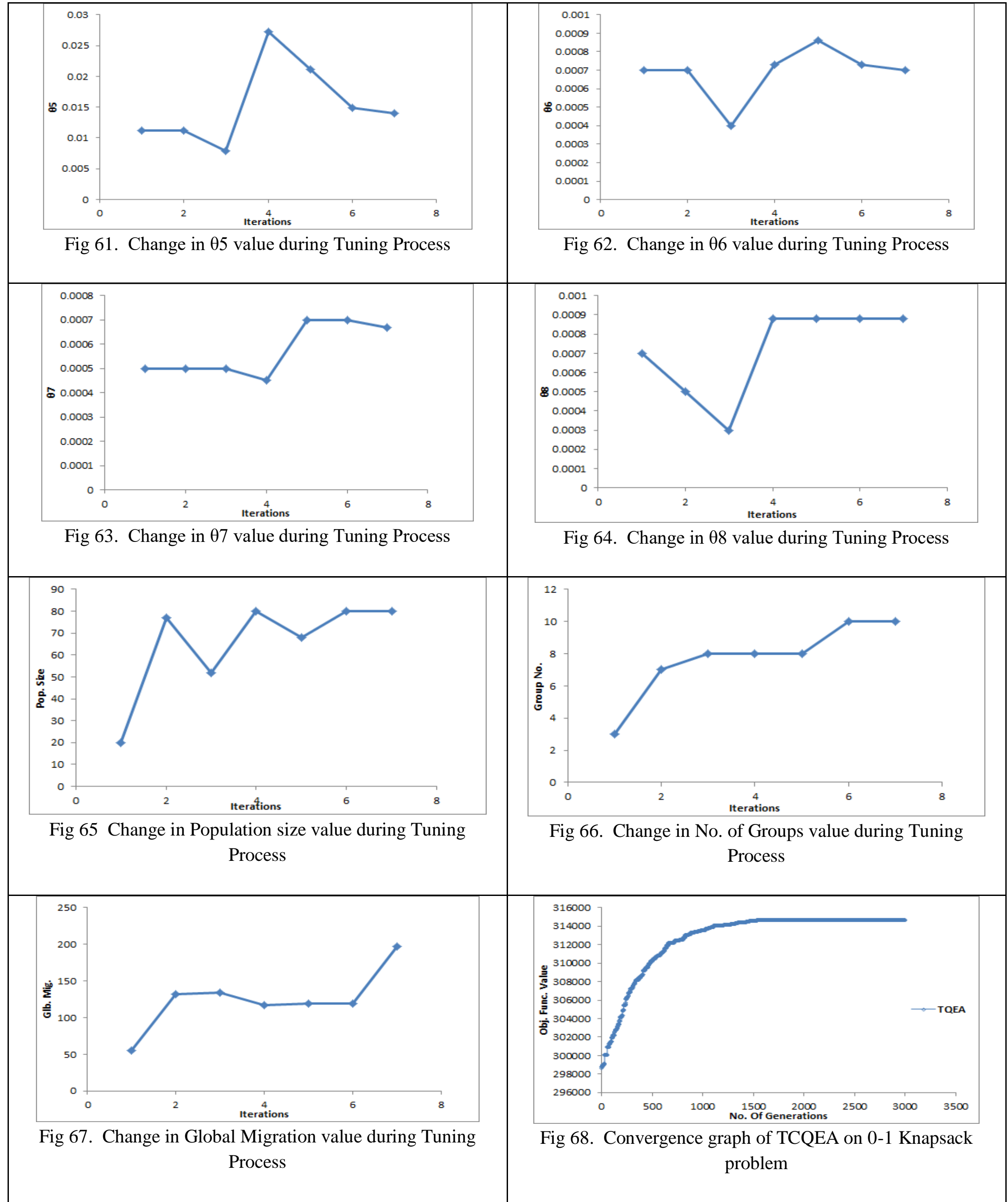

Fig 61. Change in θ5 value during Tuning Process

Fig 62. Change in θ6 value during Tuning Process

Fig 63. Change in θ7 value during Tuning Process

Fig 64. Change in θ8 value during Tuning Process

Fig 65 Change in Population size value during Tuning Process

Fig 66. Change in No. of Groups value during Tuning Process

Fig 67. Change in Global Migration value during Tuning Process

Fig 68. Convergence graph of TCQEA on 0-1 Knapsack problem

The deviation from best objective function value found so far, for large variation in parameter setting, which is computed from the best results of fifty runs from the first iteration of exploration stage is ***737.35*** whereas the deviation from optimal for small variation in parameter setting, which is computed from the best results of twenty seven runs from the first iteration of exploitation stage is ***263.6***. Therefore, the tuned QEA is relatively robust to small variations in parameter vector, however, it is quite unstable for large variation in parameter set, thus, justifying the effort put in tuning the parameter set of QEA for 0-1 Knapsack problem.

A comparative study was performed between parameter Tuned QEA (with population size as fifty and no. of groups as ten) and Canonical QEA with Parameters given in Table 2 on eleven instances of 0-1 knapsack problem instances. The population size of both the algorithms, Tuned QEA and Canonical QEA, was made equal to make the comparison fair between them. The results are given in Tables 13 to 111. Seven different problems were randomly created by varying the number of items for each of the eleven instances from 100 to 10,000. Each of these seven problems further had five instances each, generated by changing the capacity of the knapsack from 1% to 50% of the total capacity of all the items. Therefore, 35 problem instances were solved for each of the eleven instances, so total number of problems instances solved with same set of parameters is 385.

Thirty independent runs were made on each problem and the comparison has been made on Best, Worst, Average, Median of objective function value and Average number of generations. The maximum number of generations was limited to one thousand. The Tuned QEA (TCQEA) has either been able to match the performance of Canonical QEA (UCQEA) or beat it in all the 385 instances, when compared on the objective function value. The canonical QEA was able to match Tuned QEA only in small size knapsack problems on objective function value, but when compared on average generations, the Tuned-QEA was found to be much faster than canonical QEA, therefore either canonical QEA loses or is found to be relatively inefficient as compared to tuned QEA.

***1)*** *Uncorrelated Data Instances:*

The result of comparative study between Tuned QEA (TCQEA) and Canonical QEA (UCQEA) are given in Tables 13 to 19. The performance of TCQEA and UCQEA are similar when the no. of items to choose is 100 and the capacity of knapsack is 1% of the total capacity. However, with the increase in number of items and the capacity of knapsack, the performance of TCQEA has improved over UCQEA in all the instances. The performance of TCQEA was also good on speed of convergence as indicated by average generations and the convergence graphs shown in Fig. 69 to 74, which also compared the speed of convergence of UCQEA and TCQEA.

The convergence graphs have been plotted between objective function value and number of generations for both TCQEA and UCQEA for problem instances having No. of Items as 200 and 5000 and capacity as 1%, 5% and 10% of the total weight of items available for selection. The convergence graph clearly establishes the superiority of Tuning as the TCQEA is faster than UCQEA in all the graphs. The difference in performance between TCQEA and UCQEA increases with the capacity size and number of items in the knapsack problem.

TABLE 13

COMPARATIVE STUDY BETWEEN UCQEA AND TCQEA ON 0-1 KNAPSACK PROBLEM WITH NO. OF ITEMS 100 ON UNCORRELATED DATA INSTANCES

| **% of Total Weight** | **Canonical QEA** | | | | | | **Tuned-QEA** | | | | | |
|---|---|---|---|---|---|---|---|---|---|---|---|---|
| | **Best** | **Worst** | **Average** | **Median** | **Std** | **Av. Gen** | **Best** | **Worst** | **Average** | **Median** | **Std** | **Av. Gen** |
| 1% | 507.89 | 499.36 | 502.31 | 502.29 | **1.80** | 191866.67 | **517.85** | **502.41** | **510.97** | **512.95** | 4.60 | **137082.00** |
| 5% | **2496.59** | 2461.61 | 2482.80 | 2481.55 | 7.60 | 186953.33 | 2491.60 | **2475.98** | **2483.80** | **2485.43** | **5.78** | **138791.40** |
| 10% | **4938.22** | 4898.21 | 4920.25 | 4920.50 | 9.69 | **142001.67** | 4938.19 | **4903.21** | **4925.10** | **4927.70** | **8.36** | 188205.60 |
| 20% | 9786.41 | **9756.41** | 9772.07 | **9776.25** | 8.54 | **194318.33** | **9789.48** | 9756.20 | **9772.62** | 9771.91 | **8.27** | 198214.50 |
| 50% | 24261.10 | **24236.34** | 24252.62 | **24255.87** | 6.38 | 196908.33 | **24266.15** | 24231.35 | **24253.74** | 24255.28 | **6.34** | **171847.50** |

TABLE 14

COMPARATIVE STUDY BETWEEN UCQEA AND TCQEA ON 0-1 KNAPSACK PROBLEM WITH NO. OF ITEMS 200 ON UNCORRELATED DATA INSTANCES

| **% of Total Weight** | **Canonical QEA** | | | | | | **Tuned-QEA** | | | | | |
|---|---|---|---|---|---|---|---|---|---|---|---|---|
| | **Best** | **Worst** | **Average** | **Median** | **Std** | **Av. Gen** | **Best** | **Worst** | **Average** | **Median** | **Std** | **Av. Gen** |
| 1% | **1102.11** | 1053.54 | 1077.07 | 1078.13 | 11.57 | 343616.67 | 1098.55 | **1063.50** | **1084.50** | **1083.49** | **9.59** | **212731.20** |
| 5% | 5223.04 | 5172.83 | 5201.03 | 5200.33 | **10.85** | **230640.00** | **5228.04** | **5173.01** | **5205.08** | **5207.76** | 15.35 | 241008.90 |
| 10% | 10331.10 | 10271.12 | 10305.53 | 10305.94 | **13.39** | **247538.33** | **10340.70** | **10276.11** | **10311.11** | **10311.08** | 14.05 | 305507.40 |
| 20% | 20482.24 | 20432.21 | 20464.63 | **20467.20** | **11.41** | 298075.00 | **20497.26** | **20437.15** | **20466.52** | 20464.51 | 13.05 | **272794.50** |
| 50% | 50850.74 | 50810.73 | 50834.47 | 50835.44 | **9.76** | 269708.33 | **50855.76** | **50815.72** | **50837.51** | **50840.51** | 9.99 | **224403.30** |

TABLE 15

COMPARATIVE STUDY BETWEEN UCQEA AND TCQEA ON 0-1 KNAPSACK PROBLEM WITH NO. OF ITEMS 500 ON UNCORRELATED DATA INSTANCES

| % of Total Weight | Canonical QEA | | | | | | Tuned-QEA | | | | | |
|---|---|---|---|---|---|---|---|---|---|---|---|---|
| | Best | Worst | Average | Median | Std | Av. Gen | Best | Worst | Average | Median | Std | Av. Gen |
| 1% | 2631.25 | **2571.23** | 2602.28 | 2603.73 | **16.63** | 390638.33 | **2655.76** | 2556.16 | **2608.91** | **2611.28** | 24.55 | **326587.80** |
| 5% | **12706.48** | **12616.53** | **12657.26** | 12658.99 | **21.45** | **416428.33** | 12701.51 | 12601.45 | 12656.91 | **12660.85** | 24.26 | **371804.40** |
| 10% | 25143.04 | 25048.00 | 25104.36 | 25103.00 | 22.75 | 425583.33 | **25158.05** | **25078.03** | **25113.44** | **25110.47** | **21.75** | **420499.20** |
| 20% | **49946.07** | **49841.10** | **49895.12** | **49899.64** | **22.34** | 448365.00 | 49931.11 | 49840.91 | 49887.68 | 49885.62 | 26.49 | **428679.90** |
| 50% | 123990.07 | 123925.31 | **123962.55** | **123965.24** | **14.77** | 432330.00 | **124000.31** | **123935.23** | 123961.90 | 123961.07 | 17.16 | **362910.90** |

TABLE 16

COMPARATIVE STUDY BETWEEN UCQEA AND TCQEA ON 0-1 KNAPSACK PROBLEM WITH NO. OF ITEMS 1000 ON UNCORRELATED DATA INSTANCES

| % of Total Weight | Canonical QEA | | | | | | Tuned-QEA | | | | | |
|---|---|---|---|---|---|---|---|---|---|---|---|---|
| | Best | Worst | Average | Median | Std | Av. Gen | Best | Worst | Average | Median | Std | Av. Gen |
| 1% | 5233.84 | **5133.33** | 5185.33 | 5180.89 | **25.93** | **450305.00** | **5248.25** | 5121.15 | **5193.84** | **5197.73** | 31.68 | 405474.30 |
| 5% | **25446.03** | **25277.00** | **25359.71** | **25356.99** | 30.71 | 474241.67 | 25376.97 | 25256.99 | 25333.62 | 25341.06 | **28.18** | **466111.80** |
| 10% | **50408.93** | 50263.95 | **50351.14** | 50348.74 | **34.69** | **478390.00** | 50400.43 | **50273.48** | 50341.67 | **50348.97** | 34.80 | 478783.80 |
| 20% | 100173.00 | **100047.85** | **100112.08** | **100111.67** | **34.56** | 491388.33 | **100182.87** | 100022.97 | 100109.90 | 100110.41 | 39.47 | **466085.40** |
| 50% | 248944.97 | **248869.64** | **248905.32** | **248912.39** | **21.40** | 479845.00 | **248954.75** | 248834.98 | 248890.56 | 248887.31 | 30.75 | **441361.80** |

TABLE 17

COMPARATIVE STUDY BETWEEN UCQEA AND TCQEA ON 0-1 KNAPSACK PROBLEM WITH NO. OF ITEMS 2000 ON UNCORRELATED DATA INSTANCES

| % of Total Weight | Canonical QEA | | | | | | Tuned-QEA | | | | | |
|---|---|---|---|---|---|---|---|---|---|---|---|---|
| | Best | Worst | Average | Median | Std | Av. Gen | Best | Worst | Average | Median | Std | Av. Gen |
| 1% | 10480.13 | **10341.12** | 10404.56 | 10402.48 | 34.87 | 486423.33 | **10482.77** | 10351.11 | **10418.22** | **10411.76** | **34.40** | **482552.40** |
| 5% | **51157.67** | **50971.89** | **51085.41** | **51094.35** | **44.08** | **496360.00** | 51103.69 | 50909.93 | 51037.56 | 51045.01 | 50.84 | 497102.10 |
| 10% | 101595.65 | **101461.34** | **101537.05** | **101543.73** | **37.11** | 495966.67 | **101606.22** | 101422.20 | 101514.86 | 101518.64 | 57.63 | **493762.50** |
| 20% | 202225.04 | **202012.11** | 202114.84 | **202110.81** | **51.73** | **496485.00** | **202247.18** | 201933.87 | **202114.87** | 202105.43 | 80.51 | 497577.30 |
| 50% | 503210.64 | **503031.93** | **503118.04** | **503116.78** | 45.90 | 493806.67 | **503201.96** | 503020.75 | 503114.95 | 503114.84 | **45.70** | **491366.70** |

TABLE 18

COMPARATIVE STUDY BETWEEN UCQEA AND TCQEA ON 0-1 KNAPSACK PROBLEM WITH NO. OF ITEMS 5000 ON UNCORRELATED DATA INSTANCES

| % of Total Weight | Canonical QEA | | | | | | Tuned-QEA | | | | | |
|---|---|---|---|---|---|---|---|---|---|---|---|---|
| | Best | Worst | Average | Median | Std | Av. Gen | Best | Worst | Average | Median | Std | Av. Gen |
| 1% | 25806.44 | 25670.00 | 25752.66 | 25750.26 | **37.40** | 497273.33 | **25864.24** | **25678.16** | **25774.55** | **25779.08** | 39.19 | **495181.50** |
| 5% | 127217.29 | **126962.57** | **127083.98** | **127079.93** | **74.64** | **497981.67** | **127230.99** | 126905.62 | 127068.43 | 127065.11 | 90.41 | 498316.50 |
| 10% | 253198.19 | **252902.52** | 253011.73 | 253003.24 | **79.37** | 498426.67 | **253360.55** | 252886.99 | **253086.30** | **253059.34** | 124.34 | **497966.70** |
| 20% | 504556.96 | 504101.91 | 504405.49 | 504423.50 | **99.99** | **497416.67** | **504909.64** | **504200.93** | **504612.95** | **504613.31** | 159.22 | 497768.70 |
| 50% | 1257566.72 | 1257159.23 | 1257350.48 | 1257341.09 | 115.53 | 498088.33 | **1257654.00** | **1257256.55** | **1257471.27** | **1257488.44** | **110.49** | **497412.30** |

TABLE 19

COMPARATIVE STUDY BETWEEN UCQEA AND TCQEA ON 0-1 KNAPSACK PROBLEM WITH NO. OF ITEMS 10000 ON UNCORRELATED DATA INSTANCES

| % of Total Weight | Canonical QEA | | | | | | Tuned-QEA | | | | | |
|---|---|---|---|---|---|---|---|---|---|---|---|---|
| | Best | Worst | Average | Median | Std | Av. Gen | Best | Worst | Average | Median | Std | Av. Gen |
| 1% | 51480.36 | 51263.46 | 51356.21 | 51351.30 | **41.04** | **497313.33** | **51639.86** | **51337.12** | **51440.05** | **51443.58** | 66.73 | 497828.10 |
| 5% | 254514.46 | **254188.82** | 254310.69 | 254299.47 | **85.19** | **497960.00** | **254659.32** | 254149.05 | **254420.41** | **254430.36** | 128.50 | 498075.60 |
| 10% | 507210.74 | 506788.58 | 506998.11 | 506981.74 | **109.10** | **498386.67** | **507610.71** | **506852.40** | **507301.69** | **507325.85** | 190.47 | 498613.50 |
| 20% | 1012173.75 | **1011518.17** | 1011774.31 | 1011761.50 | **152.75** | **497621.67** | **1012822.26** | 1011501.42 | **1012375.03** | **1012419.11** | 313.10 | 497983.20 |
| 50% | 2524717.33 | 2523997.82 | 2524404.46 | 2524434.51 | **152.20** | 498496.67 | **2525302.57** | **2524211.82** | **2524793.80** | **2524818.19** | 243.62 | **497864.40** |

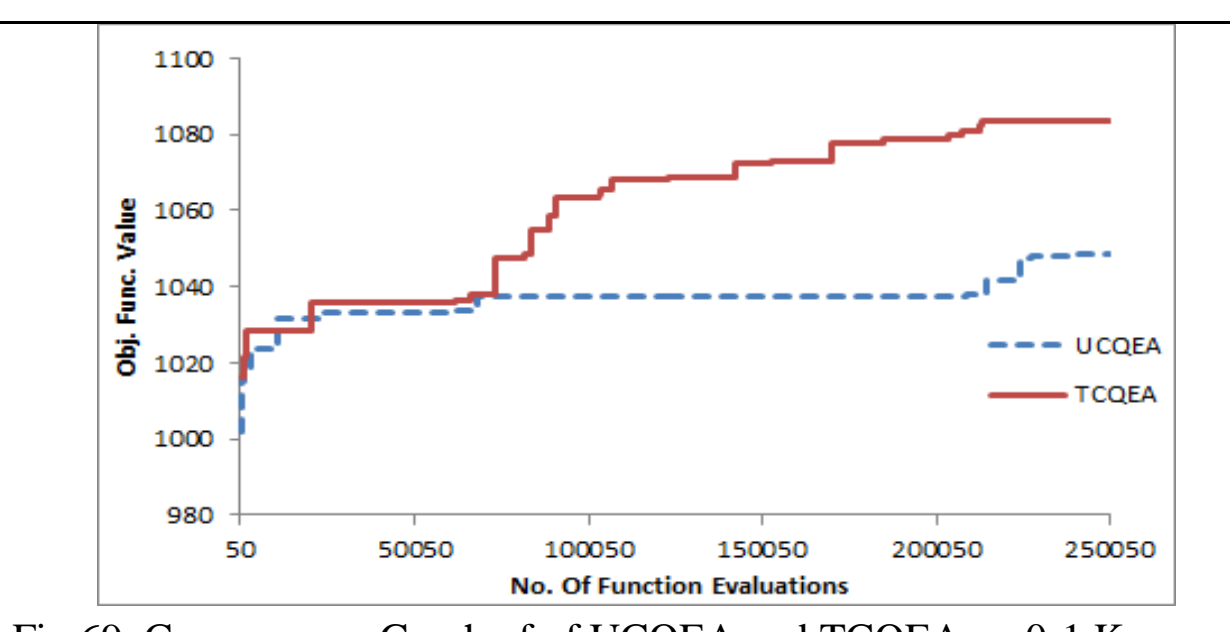

Fig 69. Convergence Graph of of UCQEA and TCQEA on 0-1 Knapsack problem with Uncorrelated Data Instances having No. of Items as 200 and Capacity as 1% of Total Weight

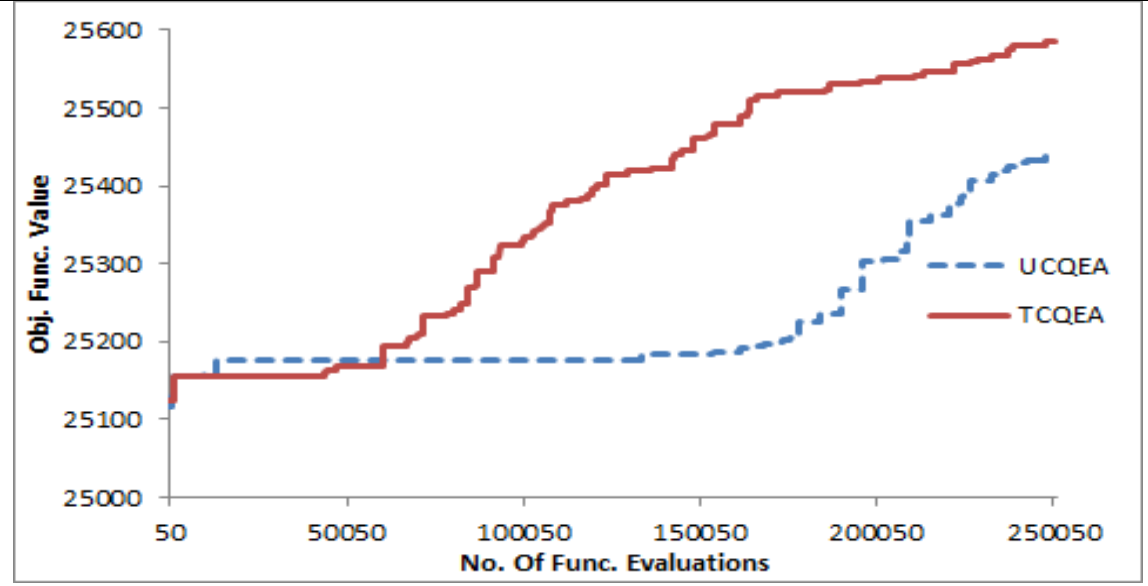

Fig 70. Convergence Graph of UCQEA and TCQEA on 0-1 Knapsack problem with Uncorrelated Data Instances having No. of Items as 5000 and Capacity as 1% of Total Weight

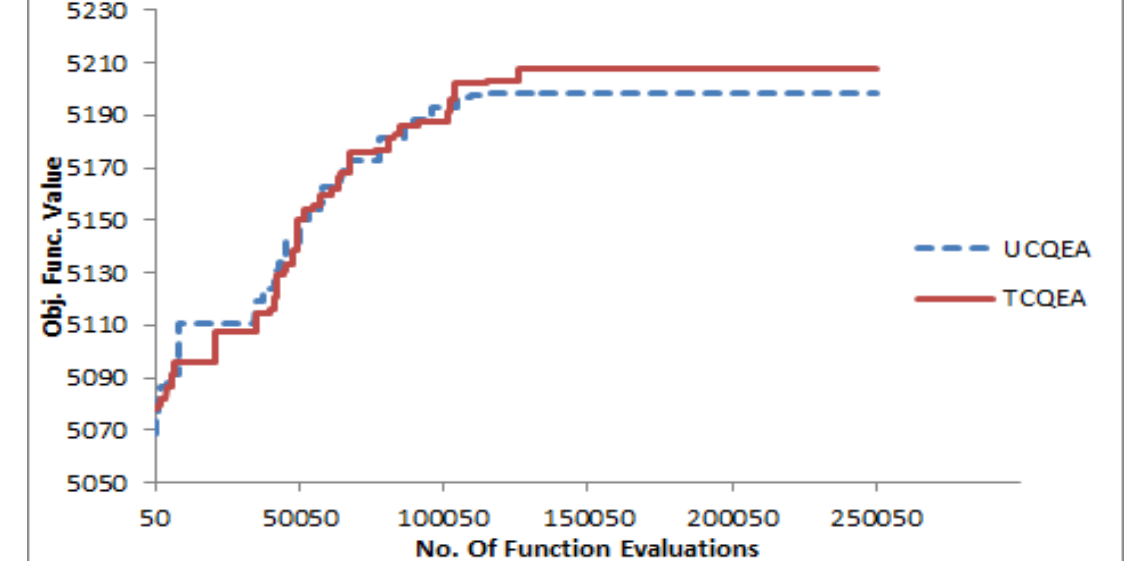

Fig 71. Convergence Graph of UCQEA and TCQEA on 0-1 Knapsack problem with Uncorrelated Data Instances having No. of Items as 200 and Capacity as 5% of Total Weight

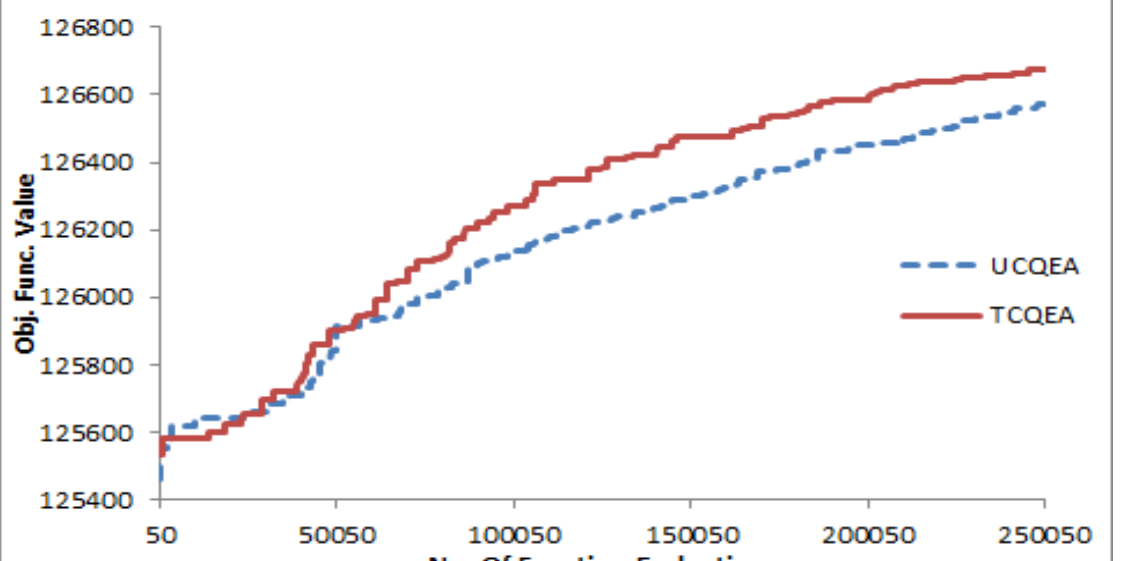

Fig 72. Convergence Graph of UCQEA and TCQEA on 0-1 Knapsack problem with Uncorrelated Data Instances having No. of Items as 5000 and Capacity as 5% of Total Weight

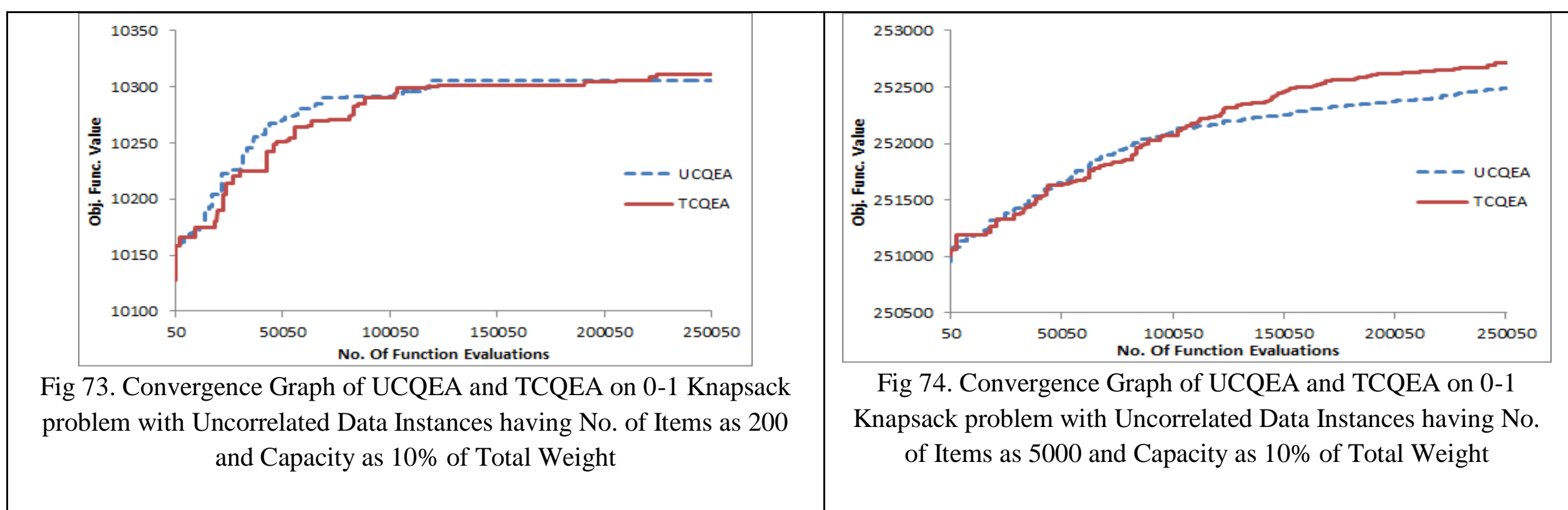

Fig 73. Convergence Graph of UCQEA and TCQEA on 0-1 Knapsack problem with Uncorrelated Data Instances having No. of Items as 200 and Capacity as 10% of Total Weight

Fig 74. Convergence Graph of UCQEA and TCQEA on 0-1 Knapsack problem with Uncorrelated Data Instances having No. of Items as 5000 and Capacity as 10% of Total Weight

In order to confirm the findings in Table 13 to 19, multi-problem non-parametric Wilcoxon's Signed Rank Test [Der2011] was performed on average objective function value (OFV) and average number of generation (Av. Gen.) of Tuned QEA and Canonical QEA on all the instances of Uncorrelated Knapsack problem at a significance level of 5%. In case of comparison on average OFV, the null hypothesis for comparison was average OFV of Tuned QEA $\mu_1$ is less than or equal to the average OFV of Canonical QEA $\mu_2$ i.e. $H_0$: $\mu_1 \leq \mu_2$ and the alternate hypothesis is H1: $\mu_1 > \mu_2$. The result of test is presented in Table 20, which shows that null hypothesis can be safely rejected as Wilcoxon's Signed Rank test statistic is zero and less than the critical value of 303 at significance level of α = 5%. The sample size n= 41 is larger than 25 so z statistic has also been computed under the normal distribution and p value obtained is 0.000, which is less than 0.05, so null hypothesis can be rejected safely. This indicates the success of the proposed tuning method for tuning QEA on problems like Uncorrelated Knapsack problem.

**Table 20: Wilcoxon's Signed Rank Test on Uncorrelated KP Instances (OFV)**

| | TCQEA | UCQEA |
|---|---|---|
| 1 | 52 | 52 |
| 2 | 257.4 | 254.9 |
| 3 | 506.4 | 499.9 |
| 4 | 1001.7 | 986.6 |
| 5 | 1983.4 | 1966.4 |
| 6 | 2475.9 | 2461.8 |
| 7 | 116.1 | 113.3 |
| 8 | 539.2 | 523.3 |
| 9 | 1056.6 | 1033.9 |
| 10 | 2094.4 | 2052 |
| 11 | 4146.6 | 4107.7 |
| 12 | 5175.5 | 5145.6 |
| 13 | 270 | 261.6 |
| 14 | 1295 | 1257.5 |
| 15 | 2559.5 | 2499.8 |
| 16 | 5063.9 | 4981.2 |
| 17 | 10058.7 | 9969.8 |
| 18 | 12545.4 | 12473.3 |
| 19 | 527.2 | 513.6 |
| 20 | 2574 | 2507.7 |
| 21 | 5098.1 | 4999.3 |
| 22 | 10120.4 | 9975.7 |
| 23 | 20104.5 | 19969.3 |
| 24 | 25094.5 | 24975.6 |

| | |
|---|---|
| Σ(+) | 861 |
| Σ(–) | 0 |
| *n* | 41 |

| Null Hypothesis | **Test Statistic** *T* | **Critical Value** At an α of 5% |
|---|---|---|
| $H_0$: $\mu_1 <= \mu_2$ | 0 | 303 |

**For Large Samples (*n* > 25)**

Test Statistic

| | | | |
|---|---|---|---|
| *z* | -5.57857 | *E[T]* | 430.5 |
| | | *σ(T)* | 77.17027 |

| Null Hypothesis | *p*-value | At an α of 5% |
|---|---|---|
| $H_0$: $\mu_1 <= \mu_2$ | 0.0000 | **Reject** |
| 25 | 1060.5 | 1024.5 |
| 26 | 5157.9 | 5052.4 |
| 27 | 10224.5 | 10083.2 |
| 28 | 20349 | 20140.5 |
| 29 | 40512.4 | 40293.9 |
| 30 | 50577 | 50393.6 |
| 31 | 2601.1 | 2537 |
| 32 | 12772.8 | 12599.7 |
| 33 | 25415.9 | 25169.4 |
| 34 | 50636.4 | 50300.5 |
| 35 | 100973 | 100620 |
| 36 | 126134 | 125816 |
| 37 | 5167.7 | 5073.3 |
| 38 | 25521.8 | 25276.7 |
| 39 | 50838.7 | 50509.4 |
| 40 | 101434 | 100970 |
| 41 | 202490 | 201959 |
| 42 | 252953 | 252497 |

In case of comparison on Av. Gen., the null hypothesis for comparison was Av. Gen. of Tuned QEA $\mu_1$ is greater than or equal to the Av. Gen. of Canonical QEA $\mu_2$ i.e. $H_0$: $\mu_1 \geq \mu_2$ and the alternate hypothesis is H1: $\mu_1 < \mu_2$. The result of test is presented in Table 21, which shows that null hypothesis cannot be rejected as Wilcoxon's Signed Rank test statistic is 778 and more than the critical value of 319 at significance level of $\alpha$ = 5%, which indicates that Tuned QEA requires more number of generations as compared to Canonical QEA. The sample size n= 42 is larger than 25 so z statistic has also been computed under the normal distribution and p value obtained is 1.000, which is more than 0.05, so null hypothesis cannot be rejected. However, Tuned QEA is performing significantly better than Canonical QEA as shown by Table. 20. This indicates the success of the proposed tuning method for tuning QEA on problems like Uncorrelated Knapsack instance even though it requires more number of generations and function evaluations as compared to Canonical QEA.

**Table 21: Wilcoxon's Signed Rank Test on Uncorrelated KP Instances (Av. Gen.)**

| | TCQEA | UCQEA |
|---|---|---|
| 1 | 2.3 | 2.5 |
| 2 | 483.3 | 428.3 |
| 3 | 557.9 | 494.8 |
| 4 | 548.3 | 541.6 |
| 5 | 697 | 918.4 |
| 6 | 710.6 | 922.3 |
| 7 | 366.8 | 487 |
| 8 | 533.5 | 514.9 |
| 9 | 723.1 | 531.8 |
| 10 | 831.6 | 659 |
| 11 | 867.9 | 941.6 |
| 12 | 906.9 | 966.8 |
| 13 | 487.9 | 519.5 |
| 14 | 817.8 | 431.1 |
| 15 | 908.5 | 551.2 |
| 16 | 896 | 524 |
| 17 | 960 | 959.4 |

| | |
|---|---|
| Σ(+) | 778 |
| Σ(–) | 125 |
| *n* | 42 |

| Null Hypothesis | **Test Statistic** *T* | **Critical Value** At an α of 5% |
|---|---|---|
| $H_0$: $\mu_1$ >= $\mu_2$ | 778 | 319 |

**For Large Samples (*n* > 25)**

Test Statistic

| | | | |
|---|---|---|---|
| *z* | -4.08245 | *E[T]* | 451.5 |
| | | *σ(T)* | 79.97655907 |

| Null Hypothesis | *p*-value | At an α of 5% |
|---|---|---|
| $H_0$: $\mu_1$ >= $\mu_2$ | 1.000 | |
| 18 | 963.2 | 964.3 |
| 19 | 664.6 | 464.6 |
| 20 | 917.3 | 434.5 |
| 21 | 933.1 | 470.8 |
| 22 | 952.9 | 684.9 |
| 23 | 964.1 | 949.9 |
| 24 | 962.4 | 974.1 |
| 25 | 850.2 | 520.5 |
| 26 | 957.2 | 466.5 |
| 27 | 959.8 | 427.8 |
| 28 | 977.4 | 614.9 |
| 29 | 972.3 | 957.1 |
| 30 | 975.4 | 965.8 |
| 31 | 910.8 | 571.2 |
| 32 | 952.3 | 626.7 |
| 33 | 963.9 | 414 |
| 34 | 982.3 | 581 |
| 35 | 982 | 959.8 |
| 36 | 975.8 | 972.1 |
| 37 | 901.3 | 468.9 |
| 38 | 961.2 | 541.2 |
| 39 | 973.4 | 560.4 |
| 40 | 978.8 | 626.8 |
| 41 | 969.9 | 935.3 |
| 42 | 978.9 | 971 |

### *2) Weakly correlated instances*

The result of comparative study between Tuned QEA (TCQEA) and Canonical QEA (UCQEA) are given in Tables 22 to 28. The performance of TCQEA and UCQEA are similar when the no. of items to choose is 100 or 200 and the capacity of knapsack is 1% or 5% of the total weight of all the items available for selection. However, with the increase in number of items and the capacity of knapsack, the performance of TCQEA has improved over UCQEA in all the instances. The performance of TCQEA was also good on speed of convergence as indicated by average generations and the convergence graphs shown in Fig. 75 to 80, which also compared the speed of convergence of UCQEA and TCQEA.

The convergence graphs have been plotted between objective function value and number of generations for both TCQEA and UCQEA for problem instances having No. of Items as 200 and 5000 and capacity as 1%, 5% and 10% of the total weight of items available for selection. The convergence graph clearly establishes the superiority of Tuning as the TCQEA is faster than UCQEA in most of the graphs. Generally, the difference in performance between TCQEA and UCQEA increases with the capacity size and number of items in the knapsack problem.

TABLE 22

COMPARATIVE RESULTS FOR 0-1 KNAPSACK PROBLEM WITH NO. OF ITEMS 100 ON WEAKLY CORRELATED DATA INSTANCES

| % of Total Weight | Canonical QEA | | | | | | Tuned-QEA | | | | | |
|---|---|---|---|---|---|---|---|---|---|---|---|---|
| | Best | Worst | Average | Median | Std | Av. Gen | Best | Worst | Average | Median | Std | Av. Gen |
| 1% | **782.1** | 723.0 | 754.5 | 753.2 | 21.8 | 216200 | **782.1** | **779.1** | **782.0** | **782.1** | **0.5** | **88998** |
| 5% | **3328.6** | **3235.0** | 3287.5 | 3290.6 | 25.2 | 101503 | **3328.6** | 3227.7 | **3296.3** | **3294.7** | **23.0** | **77402** |
| 10% | **6201.2** | 6086.6 | 6147.9 | 6153.6 | 28.1 | **69562** | **6201.2** | **6099.3** | **6170.9** | **6170.7** | **23.8** | 100581 |
| 20% | **11713.5** | 11625.7 | 11679.8 | 11681.0 | **24.8** | 97480 | **11713.5** | **11631.2** | **11685.2** | **11693.9** | 26.6 | **93535** |

| 50% | **27204.7** | 27128.8 | **27180.7** | **27182.4** | **18.0** | **118498** | **27204.7** | **27143.1** | **27180.0** | 27179.3 | 18.6 | 144078 |
|---|---|---|---|---|---|---|---|---|---|---|---|---|

TABLE 23

COMPARATIVE RESULTS FOR 0-1 KNAPSACK PROBLEM WITH NO. OF ITEMS 200 ON WEAKLY CORRELATED DATA INSTANCES

| % of Total Weight | Canonical QEA | | | | | | Tuned-QEA | | | | | |
|---|---|---|---|---|---|---|---|---|---|---|---|---|
| | Best | Worst | Average | Median | Std | Av. Gen | Best | Worst | Average | Median | Std | Av. Gen |
| 1% | **1648.8** | 1523.3 | 1588.8 | 1585.1 | 34.4 | 358858 | **1648.8** | **1540.2** | **1612.6** | **1613.4** | **32.8** | **144005** |
| 5% | 6602.5 | 6361.0 | 6499.2 | 6508.2 | 65.5 | 197923 | **6608.6** | **6457.6** | **6543.1** | **6544.9** | **44.1** | **174926** |
| 10% | 12245.5 | 12034.7 | 12171.6 | 12180.3 | 47.9 | **193305** | **12261.4** | **12122.7** | **12198.2** | **12198.7** | **33.3** | 205996 |
| 20% | 23106.0 | 22898.1 | 23018.6 | 23018.7 | 51.3 | **187658** | **23115.2** | **22948.3** | **23051.4** | **23057.4** | **37.3** | 277952 |
| 50% | 53795.7 | **53694.2** | 53749.7 | **53754.8** | **25.8** | **320998** | **53811.6** | 53653.6 | **53753.2** | 53753.6 | 32.8 | 329522 |

TABLE 24

COMPARATIVE RESULTS FOR 0-1 KNAPSACK PROBLEM WITH NO. OF ITEMS 500 ON WEAKLY CORRELATED DATA INSTANCES

| % of Total Weight | Canonical QEA | | | | | | Tuned-QEA | | | | | |
|---|---|---|---|---|---|---|---|---|---|---|---|---|
| | Best | Worst | Average | Median | Std | Av. Gen | Best | Worst | Average | Median | Std | Av. Gen |
| 1% | 4372.8 | 4071.0 | 4240.6 | 4243.5 | 71.4 | 356395 | **4400.7** | **4151.8** | **4276.1** | **4271.0** | **64.2** | **233756** |
| 5% | 16258.0 | **16013.3** | 16150.5 | 16148.6 | **59.7** | **336637** | **16331.0** | 15972.5 | **16209.7** | **16228.2** | 80.6 | 339768 |
| 10% | 30119.9 | 29732.4 | 29938.1 | 29964.0 | 104.0 | **360972** | **30153.7** | **29773.8** | **30004.1** | **30021.3** | **81.0** | 372276 |
| 20% | **56861.2** | 56346.1 | 56620.4 | 56649.1 | 97.9 | **407635** | 56799.8 | **56495.2** | **56658.5** | **56660.9** | **79.8** | 440111 |
| 50% | 131718.1 | 131493.6 | 131637.2 | 131644.9 | 62.4 | **444695** | **131732.3** | **131539.9** | **131657.0** | **131671.0** | **49.5** | 475599 |

TABLE 25

COMPARATIVE RESULTS FOR 0-1 KNAPSACK PROBLEM WITH NO. OF ITEMS 1000 ON WEAKLY CORRELATED DATA INSTANCES

| % of Total Weight | Canonical QEA | | | | | | Tuned-QEA | | | | | |
|---|---|---|---|---|---|---|---|---|---|---|---|---|
| | Best | Worst | Average | Median | Std | Av. Gen | Best | Worst | Average | Median | Std | Av. Gen |
| 1% | 8565 | 8018 | 8387 | 8404 | **137** | 460040 | **8842** | **8281** | **8529** | **8497** | 139 | **425614** |
| 5% | 32904 | 32190 | 32614 | 32650 | 183 | **485763** | **32996** | **32299** | **32752** | **32775** | **153** | 489077 |
| 10% | **60978** | 60284 | 60554 | 60555 | **144** | **484903** | 60932 | **60412** | **60723** | **60753** | 124 | 491363 |
| 20% | 114733 | 114055 | 114324 | 114303 | 137 | **492153** | **114772** | **114116** | **114388** | **114378** | **131** | 493660 |
| 50% | **266815** | **266168** | **266467** | **266463** | 129 | **492188** | 266565 | 266140 | 266359 | 266362 | **108** | 494947 |

TABLE 26

COMPARATIVE RESULTS FOR 0-1 KNAPSACK PROBLEM WITH NO. OF ITEMS 2000 ON WEAKLY CORRELATED DATA INSTANCES

| % of Total Weight | Canonical QEA | | | | | | Tuned-QEA | | | | | |
|---|---|---|---|---|---|---|---|---|---|---|---|---|
| | Best | Worst | Average | Median | Std | Av. Gen | Best | Worst | Average | Median | Std | Av. Gen |

| 1% | 16779 | 15715 | 16345 | 16391 | 264 | 495787 | **17176** | **16407** | **16844** | **16855** | **188** | **494383** |
|---|---|---|---|---|---|---|---|---|---|---|---|---|
| 5% | 65934 | 64417 | 65158 | 65138 | 323 | **497398** | **66129** | **65090** | **65591** | **65576** | **225** | 497782 |
| 10% | 122227 | 120955 | 121586 | 121646 | 314 | **496470** | **122809** | **121756** | **122248** | **122224** | **290** | 497373 |
| 20% | 231199 | 230263 | 230724 | 230721 | **216** | 497360 | **231798** | **230861** | **231351** | **231329** | 310 | **496917** |
| 50% | 541536 | 540028 | 540727 | 540715 | 341 | **497678** | **541728** | **540490** | **540882** | **540827** | **266** | 498155 |

TABLE 27

COMPARATIVE RESULTS FOR 0-1 KNAPSACK PROBLEM WITH NO. OF ITEMS 5000 ON WEAKLY CORRELATED DATA INSTANCES

| % of Total Weight | Canonical QEA | | | | | | Tuned-QEA | | | | | |
|---|---|---|---|---|---|---|---|---|---|---|---|---|
| | **Best** | **Worst** | **Average** | **Median** | **Std** | **Av. Gen** | **Best** | **Worst** | **Average** | **Median** | **Std** | **Av. Gen** |
| 1% | 37315 | 35030 | 36159 | 36215 | 473 | **498163** | **39218** | **37691** | **38496** | **38527** | **387** | 498185 |
| 5% | 156536 | 154255 | 155626 | 155692 | **555** | **498043** | **158988** | **156515** | **157849** | **157806** | 681 | 498115 |
| 10% | 297691 | 295154 | 296500 | 296501 | **684** | **498063** | **301216** | **297751** | **299327** | **299361** | 757 | 499125 |
| 20% | 571675 | 567385 | 569852 | 569887 | **999** | **498507** | **575293** | **571320** | **573574** | **573746** | 1047 | 498713 |
| 50% | 1350656 | 1346669 | 1348631 | 1348604 | **1017** | 498580 | **1353792** | **1349705** | **1351542** | **1351552** | 1068 | **498251** |

TABLE 28

COMPARATIVE RESULTS FOR 0-1 KNAPSACK PROBLEM WITH NO. OF ITEMS 10000 ON WEAKLY CORRELATED DATA INSTANCES

| % of Total Weight | Canonical QEA | | | | | | Tuned-QEA | | | | | |
|---|---|---|---|---|---|---|---|---|---|---|---|---|
| | **Best** | **Worst** | **Average** | **Median** | **Std** | **Av. Gen** | **Best** | **Worst** | **Average** | **Median** | **Std** | **Av. Gen** |
| 1% | 68351 | 65701 | 67102 | 67099 | **657** | **497990** | **72852** | **70102** | **71579** | **71709** | 729 | 498350 |
| 5% | 304149 | 300038 | 301944 | 301912 | **800** | **498317** | **308068** | **302875** | **306375** | **306446** | 1134 | 498647 |
| 10% | 584452 | 579449 | 581863 | 581856 | **1385** | **498583** | **593064** | **584387** | **587999** | **587925** | 2085 | 498689 |
| 20% | 1129281 | 1120745 | 1125317 | 1125765 | **1838** | **498608** | **1137725** | **1128635** | **1133825** | **1133602** | 2268 | 499208 |
| 50% | 2685985 | 2675108 | 2680616 | 2679893 | **2922** | **498768** | **2691332** | **2679382** | **2685224** | **2685211** | 3002 | 499300 |

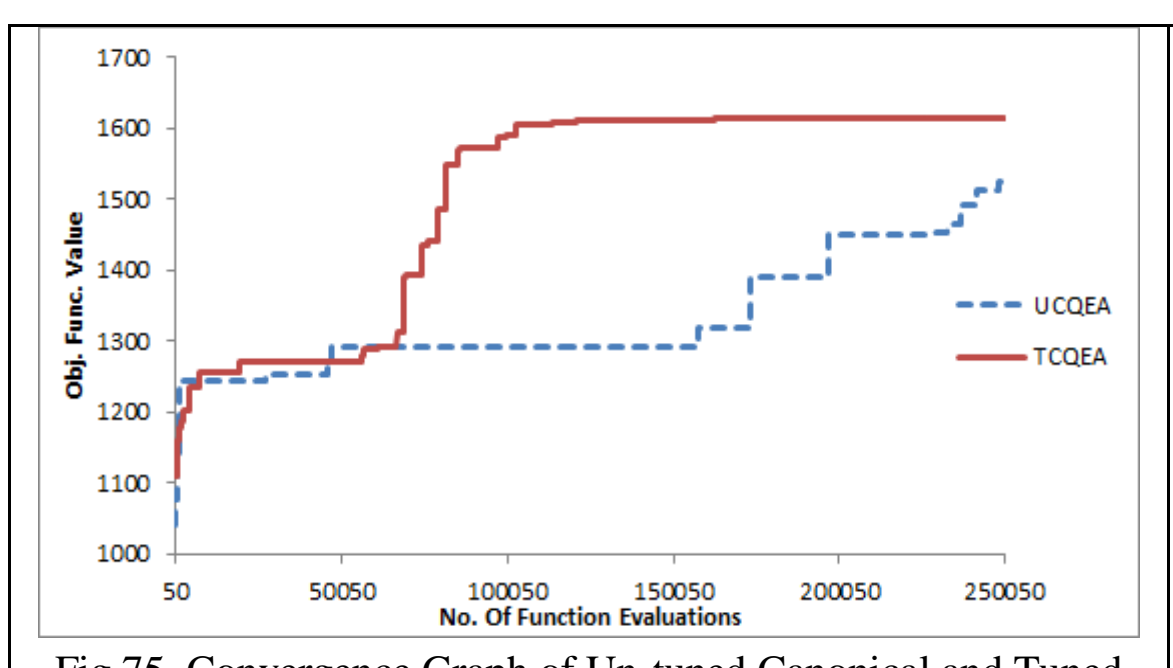

Fig 75. Convergence Graph of Un-tuned Canonical and Tuned Canonical QEA on 0-1 Knapsack problem with Weakly correlated Data Instances having No. of Items as 200 and Capacity as 1% of Total Capacity

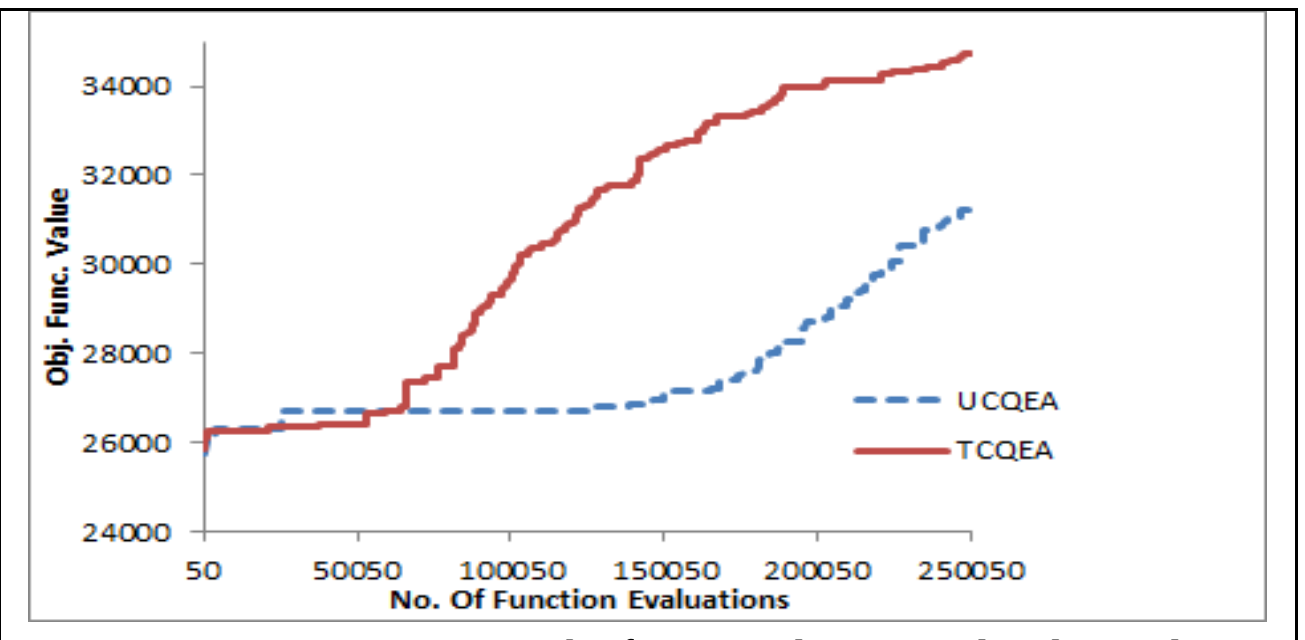

Fig 76. Convergence Graph of Un-tuned Canonical and Tuned Canonical QEA on 0-1 Knapsack problem with Weakly correlated Data Instances having No. of Items as 5000 and Capacity as 1% of Total Capacity

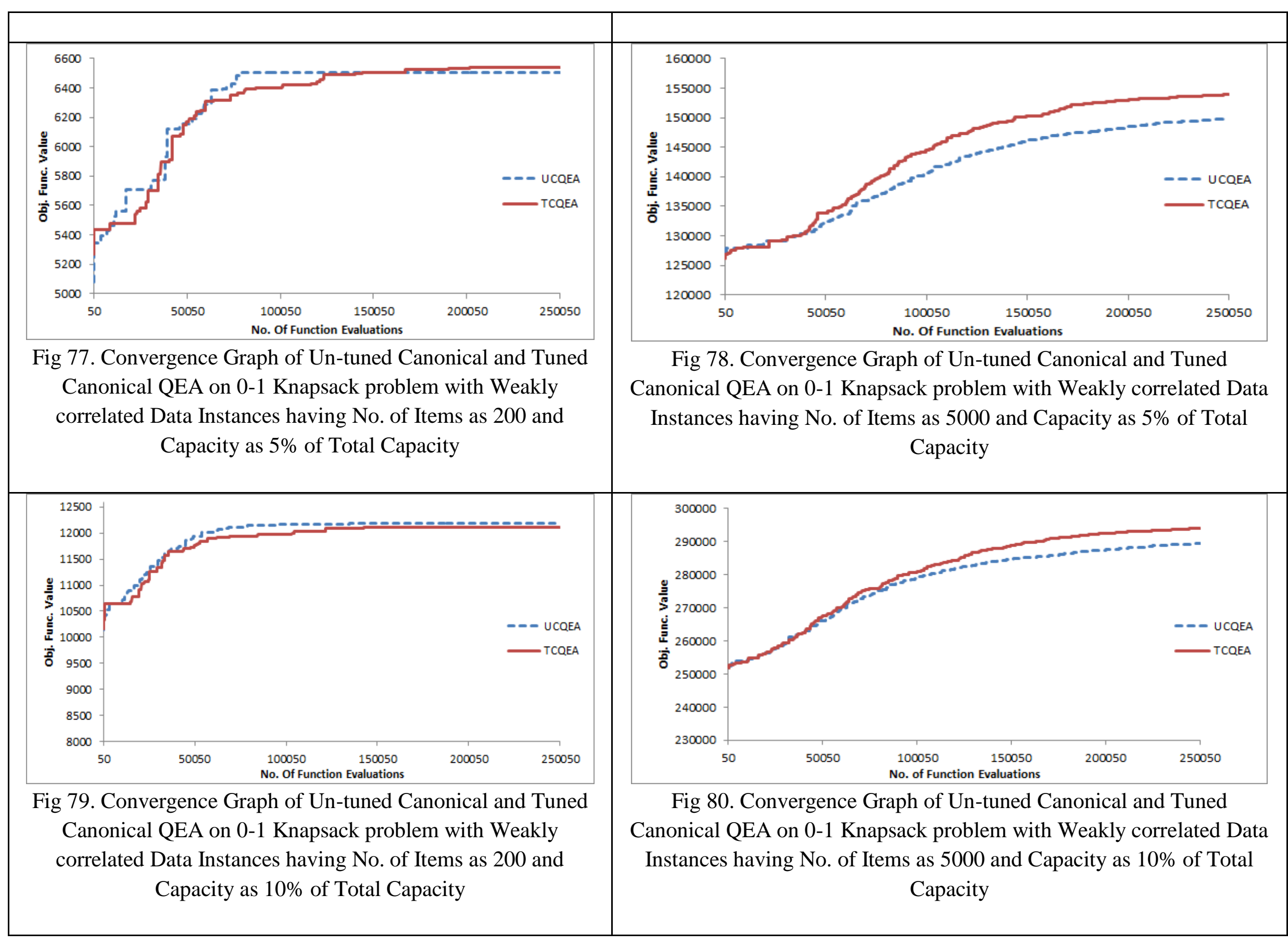

Fig 77. Convergence Graph of Un-tuned Canonical and Tuned Canonical QEA on 0-1 Knapsack problem with Weakly correlated Data Instances having No. of Items as 200 and Capacity as 5% of Total Capacity

Fig 78. Convergence Graph of Un-tuned Canonical and Tuned Canonical QEA on 0-1 Knapsack problem with Weakly correlated Data Instances having No. of Items as 5000 and Capacity as 5% of Total Capacity

Fig 79. Convergence Graph of Un-tuned Canonical and Tuned Canonical QEA on 0-1 Knapsack problem with Weakly correlated Data Instances having No. of Items as 200 and Capacity as 10% of Total Capacity

Fig 80. Convergence Graph of Un-tuned Canonical and Tuned Canonical QEA on 0-1 Knapsack problem with Weakly correlated Data Instances having No. of Items as 5000 and Capacity as 10% of Total Capacity

In order to confirm the findings in Table 22 to 28, multi-problem non-parametric Wilcoxon's Signed Rank Test [Der2011] was performed on average objective function value (OFV) and average number of generation (Av. Gen.) of Tuned QEA and Canonical QEA on all the instances of Weakly Correlated Knapsack problem at a significance level of 5%. In case of comparison on average OFV, the null hypothesis for comparison was average OFV of Tuned QEA $\mu_1$ is less than or equal to the average OFV of Canonical QEA $\mu_2$ i.e. $H_0$: $\mu_1 \leq \mu_2$ and the alternate hypothesis is H1: $\mu_1 > \mu_2$. The result of test is presented in Table 29, which shows that null hypothesis can be safely rejected as Wilcoxon's Signed Rank test statistic is 37 and less than the critical value of 303 at significance level of $\alpha$ = 5%. The sample size n= 41 (distinct) is larger than 25 so z statistic has also been computed under the normal distribution and p value obtained is 0.000, which is less than 0.05, so null hypothesis can be rejected safely. This indicates the success of the proposed tuning method for tuning QEA on problems like Weakly Correlated Knapsack problem.

**Table 29: Wilcoxon's Signed Rank Test on Weakly Correlated KP Instances (OFV)**

| | TCQEA | UCQEA |
|---|---|---|
| | $X_1$ | $X_2$ |
| 1 | 35 | 35 |
| 2 | 424 | 411 |
| 3 | 777 | 723 |
| 4 | 1441 | 1306 |
| 5 | 2669 | 2543 |
| 6 | 3265 | 3218 |
| 7 | 186 | 182 |
| 8 | 838 | 726 |
| 9 | 1560 | 1308 |
| 10 | 2879 | 2390 |
| 11 | 5309 | 4824 |

| | |
|---|---|
| Σ(+) | 824 |
| Σ(−) | 37 |
| *n* | 41 |

| Null Hypothesis | Test Statistic *T* | Critical Value At an α of 5% |
|---|---|---|
| $H_0$: $\mu_1$ <= $\mu_2$ | 37 | 303 |

**For Large Samples (*n* > 25)**

Test Statistic

| | | | |
|---|---|---|---|
| *Z* | -5.09911 | *E[T]* | 430.5 |
| | | *σ(T)* | 77.17026629 |

At an α of

| | | |
|---|---|---|
| 12 | 6475 | 6105 |
| 13 | 706 | 455 |
| 14 | 2331 | 1556 |
| 15 | 4070 | 2895 |
| 16 | 7071 | 5488 |
| 17 | 12742 | 11123 |
| 18 | 15533 | 14057 |
| 19 | 1295 | 741 |
| 20 | 4429 | 2911 |
| 21 | 7571 | 5530 |
| 22 | 13470 | 10722 |
| 23 | 24759 | 21691 |
| 24 | 30210 | 27390 |
| 25 | 2448 | 1311 |
| 26 | 7876 | 5618 |
| 27 | 13933 | 10856 |
| 28 | 25546 | 21238 |
| 29 | 47847 | 42886 |
| 30 | 58845 | 54073 |
| 31 | 4511 | 2968 |
| 32 | 16971 | 13551 |
| 33 | 31457 | 26474 |
| 34 | 59526 | 52200 |
| 35 | 114153 | 104956 |
| 36 | 141200 | 132070 |
| 37 | 7466 | 5693 |
| 38 | 31341 | 26575 |
| 39 | 59704 | 52387 |
| 40 | 114723 | 103683 |
| 41 | 221941 | 207979 |
| 42 | 252953 | 260839 |

| Null Hypothesis | *p*-value | 5% |
|---|---|---|
| $H_0: \mu_1 <= \mu_2$ | 0.0000 | **Reject** |

In case of comparison on Av. Gen., the null hypothesis for comparison was Av. Gen. of Tuned QEA $\mu_1$ is greater than or equal to the Av. Gen. of Canonical QEA $\mu_2$ i.e. $H_0$: $\mu_1 \geq \mu_2$ and the alternate hypothesis is H1: $\mu_1 < \mu_2$. The result of test is presented in Table 30, which shows that null hypothesis cannot be rejected as Wilcoxon's Signed Rank test statistic is 741 and more than the critical value of 319 at significance level of $\alpha = 5\%$, which indicates that Tuned QEA requires more number of generations as compared to Canonical QEA. The sample size n= 42 (distinct) is larger than 25 so z statistic has also been computed under the normal distribution and p value obtained is 0.9999, which is more than 0.05, so null hypothesis cannot be rejected. However, Tuned QEA is performing significantly better than Canonical QEA as shown by Table 29. This indicates the success of the proposed tuning method for tuning QEA on problems like Weakly Correlated Knapsack instance even though it requires more number of generations and function evaluations as compared to Canonical QEA.

**Table 30: Wilcoxon's Signed Rank Test on Weakly Corr. KP Instances (Av. Gen.)**

| | TCQEA | UCQEA |
|---|---|---|
| | **$X_1$** | **$X_2$** |
| 1 | 2 | 2 |
| 2 | 336 | 193 |
| 3 | 349 | 582 |
| 4 | 432 | 718 |
| 5 | 562 | 947 |

| | |
|---|---|
| $\Sigma(+)$ | 741 |
| $\Sigma(-)$ | 162 |
| *n* | 42 |

| Null Hypothesis | **Test Statistic** *T* | **Critical Value** At an $\alpha$ of 5% |
|---|---|---|
| $H_0: \mu_1 >= \mu_2$ | 741 | 319 |
| 6 | 471 | 962 |
| 7 | 360 | 303 |
| 8 | 483 | 559 |
| 9 | 690 | 633 |
| 10 | 725 | 684 |
| 11 | 848 | 959 |
| 12 | 889 | 973 |
| 13 | 673 | 402 |
| 14 | 900 | 523 |
| 15 | 922 | 476 |
| 16 | 954 | 631 |
| 17 | 961 | 954 |
| 18 | 977 | 970 |
| 19 | 833 | 570 |
| 20 | 953 | 540 |
| 21 | 977 | 532 |
| 22 | 973 | 804 |
| 23 | 981 | 965 |
| 24 | 985 | 979 |
| 25 | 937 | 529 |
| 26 | 979 | 511 |
| 27 | 976 | 585 |
| 28 | 977 | 567 |
| 29 | 975 | 974 |
| 30 | 983 | 980 |
| 31 | 972 | 616 |
| 32 | 966 | 565 |
| 33 | 983 | 489 |
| 34 | 983 | 698 |
| 35 | 987 | 978 |
| 36 | 981 | 980 |
| 37 | 936 | 439 |
| 38 | 969 | 528 |
| 39 | 975 | 579 |
| 40 | 982 | 704 |
| 41 | 987 | 961 |
| 42 | 979 | 976 |

**For Large Samples ($n > 25$)**

Test Statistic

| | |
|---|---|
| $z$ | -3.61981 |
| $E[T]$ | 451.5 |
| $\sigma(T)$ | 79.97655907 |

| Null Hypothesis | $p$-value | At an $\alpha$ of 5% |
|---|---|---|
| $H_0$: $\mu_1 >= \mu_2$ | 0.9999 | |

### *3) Strongly correlated instances:*

The result of comparative study between Tuned QEA (TCQEA) and Canonical QEA (UCQEA) are given in Tables 31 to 37. The performance of TCQEA and UCQEA are similar when the no. of items to choose is 100 to 500 and the capacity of knapsack is 0.1 % or 0.5% of the total capacity. However, with the increase in number of items and the capacity of knapsack, the performance of TCQEA has improved over UCQEA in all the instances. The performance of TCQEA was also good on speed of convergence as indicated by average generations and the convergence graphs shown in Fig. 81to 86, which also compared the speed of convergence of TCQEA to UCQEA.

TABLE 31

COMPARATIVE RESULTS FOR 0-1 KNAPSACK PROBLEM WITH NO. OF ITEMS 100 ON STRONGLY CORRELATED DATA INSTANCES

| % of Total Weight | Canonical QEA | | | | | | Tuned-QEA | | | | | |
|---|---|---|---|---|---|---|---|---|---|---|---|---|
| | Best | Worst | Average | Median | Std | Av. Gen | Best | Worst | Average | Median | Std | Av. Gen |

| % of Total Weight | Best | Worst | Average | Median | Std | Av. Gen | Best | Worst | Average | Median | Std | Av. Gen |
|---|---|---|---|---|---|---|---|---|---|---|---|---|
| 1% | 1127 | 976 | 1058 | 1062 | 29 | 152248 | **1278** | **1273** | **1277** | **1278** | **1** | **110642** |
| 5% | **4592** | 4291 | 4485 | 4491 | 74 | 104033 | **4592** | **4392** | **4527** | **4492** | **55** | **80540** |
| 10% | **8082** | 7783 | 7959 | **7983** | 73 | 85902 | **8082** | **7883** | **7976** | **7983** | **52** | **85421** |
| 20% | 14166 | 13966 | 14082 | 14066 | 69 | **71273** | **14167** | **14066** | **14143** | **14166** | **43** | 89100 |
| 50% | **31016** | **30816** | 30899 | 30916 | 46 | **33372** | **31016** | **30816** | **30916** | **30916** | **26** | 60997 |

TABLE 32

COMPARATIVE RESULTS FOR 0-1 KNAPSACK PROBLEM WITH NO. OF ITEMS 200 ON STRONGLY CORRELATED DATA INSTANCES

| % of Total Weight | Canonical QEA | | | | | | Tuned-QEA | | | | | |
|---|---|---|---|---|---|---|---|---|---|---|---|---|
| | Best | Worst | Average | Median | Std | Av. Gen | Best | Worst | Average | Median | Std | Av. Gen |
| 1% | **3004** | 2603 | 2927 | 3003 | 116 | 297800 | **3004** | **2803** | **2983** | **3003** | **48** | **142540** |
| 5% | **9618** | 9218 | 9434 | 9418 | 115 | 101815 | **9618** | **9418** | **9541** | **9518** | **77** | 155826 |
| 10% | 16534 | 16136 | 16356 | 16336 | 92 | 109627 | **16536** | **16336** | **16419** | **16435** | **65** | 112768 |
| 20% | 28972 | 28472 | 28749 | 28772 | 119 | 132833 | **29072** | **28572** | **28859** | **28872** | **111** | **107425** |
| 50% | 64081 | 63681 | 63934 | 63931 | 97 | 63392 | **64180** | **63781** | **64011** | **63981** | **75** | 92869 |

TABLE 33

COMPARATIVE RESULTS FOR 0-1 KNAPSACK PROBLEM WITH NO. OF ITEMS 500 ON STRONGLY CORRELATED DATA INSTANCES

| % of Total Weight | Canonical QEA | | | | | | Tuned-QEA | | | | | |
|---|---|---|---|---|---|---|---|---|---|---|---|---|
| | Best | Worst | Average | Median | Std | Av. Gen | Best | Worst | Average | Median | Std | Av. Gen |
| 1% | **7246** | 6346 | 6893 | 6946 | 250 | 343017 | **7246** | **6846** | **7093** | **7146** | **128** | **238841** |
| 5% | 23032 | 22331 | 22765 | 22732 | 167 | 245223 | **23331** | **22731** | **22978** | **22932** | **153** | **233218** |
| 10% | 40063 | 39163 | 39646 | 39663 | 241 | **249597** | **40163** | **39363** | **39893** | **39913** | **174** | 260588 |
| 20% | 71026 | 70326 | 70706 | 70726 | 185 | 220642 | **71226** | **70626** | **70976** | **71026** | **155** | **199947** |
| 50% | 157315 | 156715 | 157052 | 157115 | 154 | **151160** | **157515** | **156915** | **157222** | **157215** | **136** | 169419 |

TABLE 34

COMPARATIVE RESULTS FOR 0-1 KNAPSACK PROBLEM WITH NO. OF ITEMS 1000 ON STRONGLY CORRELATED DATA INSTANCES

| % of Total Weight | Canonical QEA | | | | | | Tuned-QEA | | | | | |
|---|---|---|---|---|---|---|---|---|---|---|---|---|
| | Best | Worst | Average | Median | Std | Av. Gen | Best | Worst | Average | Median | Std | Av. Gen |
| 1% | 14199 | 12813 | 13500 | 13513 | 359 | 432160 | **14313** | **13213** | **13820** | **13813** | **284** | **377418** |
| 5% | 45467 | 44267 | 45017 | 45117 | 343 | 369815 | **45867** | **44865** | **45457** | **45467** | **268** | **339676** |
| 10% | 79534 | 78434 | 79021 | 79034 | **325** | 391293 | **79834** | **78534** | **79347** | **79384** | 360 | **380853** |
| 20% | 141768 | 140568 | 141164 | 141168 | 355 | 347170 | **142168** | **141168** | **141635** | **141618** | **237** | **319757** |
| 50% | 315070 | 314070 | 314513 | 314520 | 263 | 292710 | **315470** | **314570** | **314999** | **314970** | **247** | **282200** |

TABLE 35

COMPARATIVE RESULTS FOR 0-1 KNAPSACK PROBLEM WITH NO. OF ITEMS 2000 ON STRONGLY CORRELATED DATA INSTANCES

| % of Total Weight | Canonical QEA | | | | | | Tuned-QEA | | | | | |
|---|---|---|---|---|---|---|---|---|---|---|---|---|
| | Best | Worst | Average | Median | Std | Av. Gen | Best | Worst | Average | Median | Std | Av. Gen |
| 1% | 26636 | 24536 | 26044 | 26184 | 550 | 485218 | **27936** | **25936** | **27021** | **27136** | **488** | **464323** |
| 5% | 90580 | 88478 | 89590 | 89611 | **561** | 482052 | **91681** | **88681** | **90344** | **90326** | 563 | **476966** |
| 10% | 159060 | 155961 | 157833 | 157861 | **558** | **459438** | **159561** | **156934** | **158430** | **158411** | 616 | 460809 |
| 20% | 284223 | 282378 | 283328 | 283343 | 495 | 474455 | **285023** | **283123** | **284083** | **284023** | **486** | **429360** |
| 50% | 634507 | 632607 | 633671 | 633707 | **386** | 411388 | **635007** | **633407** | **634307** | **634307** | 509 | **370082** |

TABLE 36

COMPARATIVE RESULTS FOR 0-1 KNAPSACK PROBLEM WITH NO. OF ITEMS 5000 ON STRONGLY CORRELATED DATA INSTANCES

| % of Total Weight | Canonical QEA | | | | | | Tuned-QEA | | | | | |
|---|---|---|---|---|---|---|---|---|---|---|---|---|
| | Best | Worst | Average | Median | Std | Av. Gen | Best | Worst | Average | Median | Std | Av. Gen |
| 1% | 57646 | 54617 | 55957 | 56030 | **653** | **498330** | **62425** | **59441** | **61222** | **61256** | 612 | 498600 |
| 5% | 215226 | 210481 | 212526 | 212453 | **954** | 498720 | **218626** | **213111** | **216405** | **216316** | 1360 | **498703** |
| 10% | 385552 | 381032 | 382674 | 382671 | **1144** | **498375** | **389252** | **383688** | **386671** | **386698** | 1361 | 499109 |
| 20% | 701310 | 694992 | 698197 | 698200 | **1359** | **498918** | **705469** | **699535** | **701953** | **701512** | 1538 | 499231 |
| 50% | 1583076 | 1579892 | 1581201 | 1581123 | **707** | 495382 | **1584680** | **1580667** | **1582584** | **1582527** | 1131 | **486334** |

TABLE 37

COMPARATIVE RESULTS FOR 0-1 KNAPSACK PROBLEM WITH NO. OF ITEMS 10000 ON STRONGLY CORRELATED DATA INSTANCES

| % of Total Weight | Canonical QEA | | | | | | Tuned-QEA | | | | | |
|---|---|---|---|---|---|---|---|---|---|---|---|---|
| | Best | Worst | Average | Median | Std | Av. Gen | Best | Worst | Average | Median | Std | Av. Gen |
| 1% | 99496 | 95503 | 97953 | 98144 | 1046 | **498612** | 110303 | 105860 | 108394 | 108416 | **1039** | 499072 |
| 5% | 402203 | 396513 | 399375 | 399436 | **1514** | **499013** | 410179 | 399941 | 406530 | 406730 | 2525 | 499521 |
| 10% | 737645 | **730830** | 734270 | 734480 | **1801** | **499345** | 754851 | 729128 | 742078 | 743126 | 5181 | 499584 |
| 20% | 1373675 | 1360215 | 1365712 | 1365587 | **3243** | **498960** | 1384688 | 1367158 | 1375294 | 1375172 | 4695 | 499349 |
| 50% | 3159184 | 3149522 | 3156006 | 3156534 | **2444** | **498442** | 3163924 | 3152654 | 3159966 | 3160254 | 2627 | 499056 |

The convergence graphs have been plotted between objective function value and number of generations for both TCQEA and UCQEA for all the problem instances having No. of Items as 200 and 5000 for the median run. The convergence graph clearly establishes the superiority of Tuning as the TCQEA is faster than UCQEA in all the graphs except in problems with no. of items 200 and knapsack capacity as 0.1% & 0.5%, where TCQEA matched UCQEA. The difference in performance between TCQEA and UCQEA increases with the capacity size and number of items in the knapsack problem.

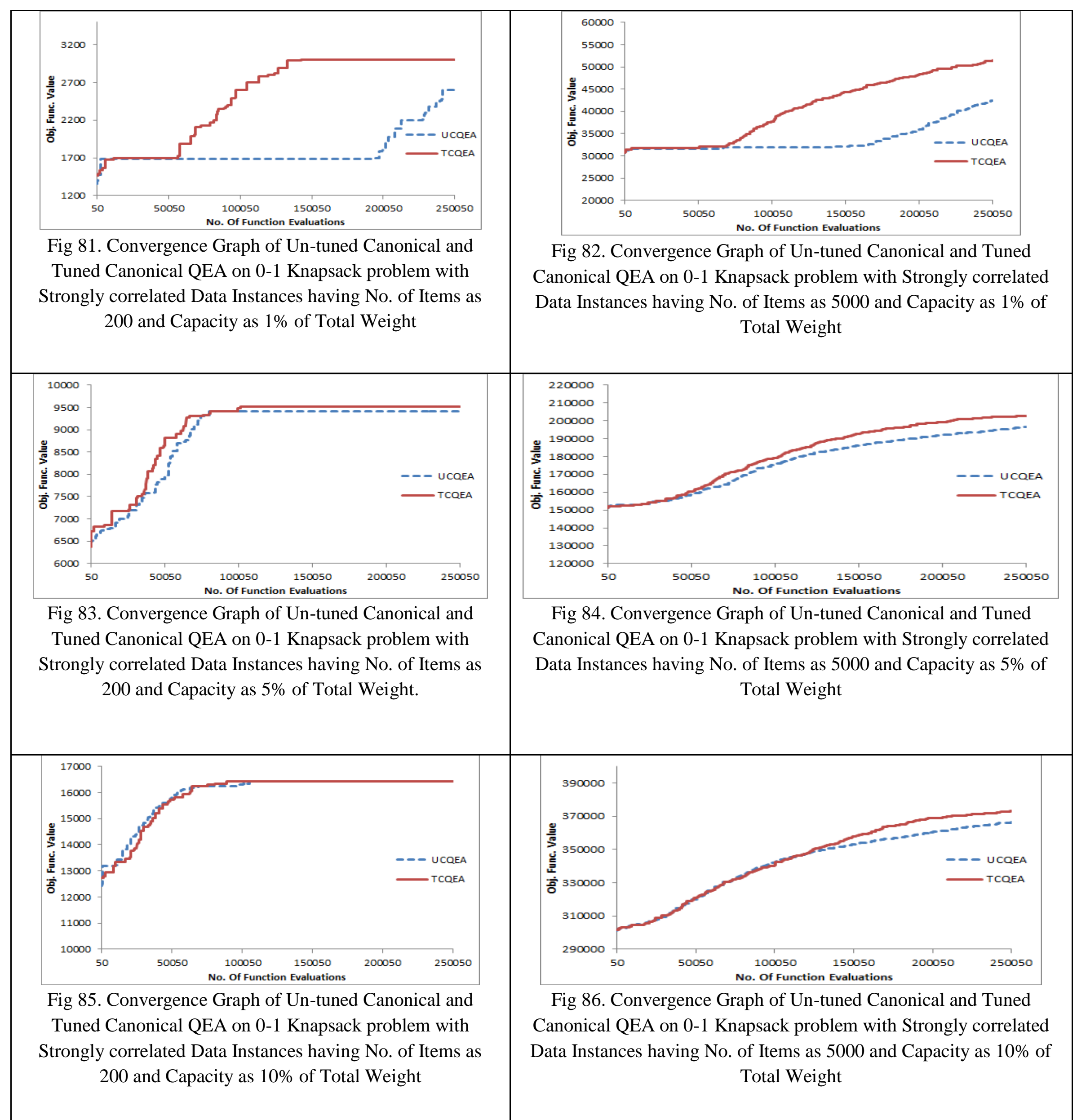

Fig 81. Convergence Graph of Un-tuned Canonical and Tuned Canonical QEA on 0-1 Knapsack problem with Strongly correlated Data Instances having No. of Items as 200 and Capacity as 1% of Total Weight

Fig 82. Convergence Graph of Un-tuned Canonical and Tuned Canonical QEA on 0-1 Knapsack problem with Strongly correlated Data Instances having No. of Items as 5000 and Capacity as 1% of Total Weight

Fig 83. Convergence Graph of Un-tuned Canonical and Tuned Canonical QEA on 0-1 Knapsack problem with Strongly correlated Data Instances having No. of Items as 200 and Capacity as 5% of Total Weight.

Fig 84. Convergence Graph of Un-tuned Canonical and Tuned Canonical QEA on 0-1 Knapsack problem with Strongly correlated Data Instances having No. of Items as 5000 and Capacity as 5% of Total Weight

Fig 85. Convergence Graph of Un-tuned Canonical and Tuned Canonical QEA on 0-1 Knapsack problem with Strongly correlated Data Instances having No. of Items as 200 and Capacity as 10% of Total Weight

Fig 86. Convergence Graph of Un-tuned Canonical and Tuned Canonical QEA on 0-1 Knapsack problem with Strongly correlated Data Instances having No. of Items as 5000 and Capacity as 10% of Total Weight

In order to confirm the findings in Table 31 to 37, multi-problem non-parametric Wilcoxon's Signed Rank Test [Der2011] was performed on average objective function value (OFV) and average number of generation (Av. Gen.) of Tuned QEA and Canonical QEA on all the instances of Strongly Correlated Knapsack problem at a significance level of 5%. In case of comparison on average OFV, the null hypothesis for comparison was average OFV of Tuned QEA $\mu_1$ is less than or equal to the average OFV of Canonical QEA $\mu_2$ i.e. $H_0$: $\mu_1 \leq \mu_2$ and the alternate hypothesis is H1: $\mu_1 > \mu_2$. The result of test is presented in Table 38, which shows that null hypothesis can be safely rejected as Wilcoxon's Signed Rank test statistic is zero and less than the critical value of 242 at significance level of $\alpha$ = 5%. The sample size n= 37 (distinct) is larger than 25 so z statistic has also been computed under the normal distribution and p value obtained is 0.000, which is less than 0.05, so null hypothesis can be rejected safely. This indicates the success of the proposed tuning method for tuning QEA on problems like Strongly Correlated Knapsack problem.

**Table 38: Wilcoxon's Signed Rank Test on Strongly Correlated KP Instances (OFV)**

| | TCQEA | UCQEA |
|---|---|---|
| | $X_1$ | $X_2$ |
| 1 | 0 | 0 |

| Test Statistic | Critical Value |
|---|---|

| 2 | 12.6 | 12.6 |
|---|---|---|
| 3 | 20.3 | 20.3 |
| 4 | 41.6 | 38.6 |
| 5 | 79.2 | 75.2 |
| 6 | 96.2 | 94.9 |
| 7 | 6.1 | 6.1 |
| 8 | 25.3 | 24.6 |
| 9 | 49.5 | 40.7 |
| 10 | 93.2 | 69.7 |
| 11 | 173.2 | 147.5 |
| 12 | 208.1 | 190.3 |
| 13 | 12.7 | 12.7 |
| 14 | 61.3 | 47.3 |
| 15 | 120.3 | 80.6 |
| 16 | 226 | 144.7 |
| 17 | 405.3 | 306.4 |
| 18 | 488.3 | 396.4 |
| 19 | 25.7 | 24 |
| 20 | 126.4 | 78.9 |
| 21 | 238.3 | 142.4 |
| 22 | 432.9 | 263.5 |
| 23 | 765.7 | 546.3 |
| 24 | 909.1 | 708.4 |
| 25 | 51.4 | 40.3 |
| 26 | 243.5 | 141.7 |
| 27 | 447.9 | 259.1 |
| 28 | 796.3 | 493.2 |
| 29 | 1378.1 | 1014 |
| 30 | 1634.7 | 1297.2 |
| 31 | 126 | 80.4 |
| 32 | 548.1 | 318.2 |
| 33 | 971.3 | 600.5 |
| 34 | 1687.2 | 1164.9 |
| 35 | 2966.2 | 2360.6 |
| 36 | 3559.1 | 2992 |
| 37 | 234.8 | 142.2 |
| 38 | 966.4 | 600.8 |
| 39 | 1697.4 | 1154.5 |
| 40 | 2988.3 | 2258.2 |
| 41 | 5372.7 | 4549.9 |
| 42 | 6520.8 | 5730.4 |

| | |
|---|---|
| Σ(+) | 703 |
| Σ(−) | 0 |
| *n* | 37 |

| Null Hypothesis | *T* | At an α of 5% |
|---|---|---|
| $H_0$: $\mu_1$ <= $\mu_2$ | 0 | 242 |

**For Large Samples (*n* > 25)**

Test Statistic

| | | | |
|---|---|---|---|
| *z* | -5.30283 | *E[T]* | 351.5 |
| | | *σ(T)* | 66.28536792 |

| Null Hypothesis | *p*-value | At an α of 5% |
|---|---|---|
| $H_0$: $\mu_1$ <= $\mu_2$ | 0.0000 | **Reject** |

In case of comparison on Av. Gen., the null hypothesis for comparison was Av. Gen. of Tuned QEA $\mu_1$ is greater than or equal to the Av. Gen. of Canonical QEA $\mu_2$ i.e. $H_0$: $\mu_1 \geq \mu_2$ and the alternate hypothesis is H1: $\mu_1 < \mu_2$. The result of test is presented in Table 39, which shows that null hypothesis cannot be rejected as Wilcoxon's Signed Rank test statistic is 356 and more than the critical value of 303 at significance level of $\alpha = 5\%$, which indicates that Tuned QEA requires more number of generations as compared to Canonical QEA. The sample size n= 41 (distinct) is larger than 25 so z statistic has also been computed under the normal distribution and p value obtained is 0.8328, which is more than 0.05, so null hypothesis cannot be rejected. However, Tuned QEA is performing significantly better than Canonical QEA as shown by Table 38. This indicates the success of the proposed tuning method for tuning QEA on problems like Strongly Correlated Knapsack instance even though it requires more number of generations and function evaluations as compared to Canonical QEA.

**Table 39: Wilcoxon's Signed Rank Test on Strongly Corr. KP Instances (Av. Gen.)**

| | TCQEA | UCQEA |
|---|---|---|
| | **$X_1$** | **$X_2$** |
| 1 | 0 | 0 |
| 2 | 61.1 | 83.2 |
| 3 | 262.3 | 463.3 |
| 4 | 297.9 | 528.1 |
| 5 | 393.9 | 947.5 |
| 6 | 404.9 | 951.6 |
| 7 | 3.8 | 3.2 |
| 8 | 240.4 | 523.3 |
| 9 | 362.2 | 503.2 |
| 10 | 455.8 | 607.7 |
| 11 | 634.1 | 961.7 |
| 12 | 622 | 979 |
| 13 | 260.6 | 461.7 |
| 14 | 426.7 | 493.2 |
| 15 | 680.6 | 609.3 |
| 16 | 815 | 616.8 |
| 17 | 879.4 | 968.2 |
| 18 | 942.2 | 977.1 |
| 19 | 405.6 | 481.5 |
| 20 | 687.5 | 513.2 |
| 21 | 876.1 | 584.5 |
| 22 | 948.1 | 710.1 |
| 23 | 959.5 | 975.5 |
| 24 | 977.2 | 986.2 |
| 25 | 533 | 515.8 |
| 26 | 825 | 486.5 |
| 27 | 952.8 | 644.2 |
| 28 | 985.7 | 652.9 |
| 29 | 988.5 | 975.9 |
| 30 | 989.9 | 979.2 |
| 31 | 743.4 | 535.8 |
| 32 | 959.5 | 419.1 |
| 33 | 981.6 | 525.2 |
| 34 | 987.8 | 703.3 |
| 35 | 988 | 971.8 |
| 36 | 993.4 | 982.7 |
| 37 | 816 | 555.7 |
| 38 | 976.1 | 519.7 |
| 39 | 986 | 497.1 |
| 40 | 988 | 741.7 |
| 41 | 989.6 | 977.6 |
| 42 | 989.2 | 980.5 |

| | |
|---|---|
| $\Sigma(+)$ | 505 |
| $\Sigma(-)$ | 356 |
| $n$ | 41 |

| Null Hypothesis | **Test Statistic** $T$ | **Critical Value** At an $\alpha$ of 5% |
|---|---|---|
| $H_0: \mu_1 <= \mu_2$ | 356 | 303 |

**For Large Samples ($n$ > 25)**

Test Statistic

| | | | |
|---|---|---|---|
| $z$ | -0.9654 | $E[T]$ | 430.5 |
| | | $\sigma(T)$ | 77.17026629 |

| Null Hypothesis | $p$-value | At an $\alpha$ of 5% |
|---|---|---|
| $H_0: \mu_1 = \mu_2$ | 0.3343 | |
| $H_0: \mu_1 >= \mu_2$ | 0.8328 | |
| $H_0: \mu_1 <= \mu_2$ | 0.1672 | |

#### *4) Inverse strongly correlated instances:*

The result of comparative study between Tuned QEA (TCQEA) and Canonical QEA (UCQEA) are given in Table 40 to 46. The performance of TCQEA and UCQEA are similar when the no. of items to choose is 100 and 200 for different capacity of knapsack. In fact, when the capacity of knapsack was 0.1% of the total capacity and number of items 100, both the algorithm could not find even a single item for the knapsack. The performance of TCQEA and UCQEA are also similar when the capacity of knapsack is 0.1 % of the total capacity. However, with the increase in number of items and the capacity of knapsack, the performance of TCQEA has improved over UCQEA in rest of the instances. The performance of TCQEA was also good on speed of convergence as indicated by average generations and the convergence graphs shown in figures 87 to 92, which also compared the speed of convergence of TCQEA to UCQEA.

TABLE 40
COMPARATIVE RESULTS FOR 0-1 KNAPSACK PROBLEM WITH NO. OF ITEMS 100 ON INVERSE STRONGLY CORRELATED DATA INSTANCES

| % of Total Weight | Canonical QEA | | | | | | Tuned-QEA | | | | | |
|---|---|---|---|---|---|---|---|---|---|---|---|---|
| | Best | Worst | Average | Median | Std | Av. Gen | Best | Worst | Average | Median | Std | Av. Gen |
| 1% | 461 | 461 | 461 | 461 | 0 | 95 | 461 | 461 | 461 | 461 | 0 | 116 |
| 5% | 2591 | 2591 | 2591 | 2591 | 0 | 23148 | 2591 | 2591 | 2591 | 2591 | 0 | 37356 |
| 10% | 5183 | 5182 | 5183 | 5183 | 0 | 76880 | 5183 | 5181 | 5183 | 5183 | 0 | 51770 |
| 20% | 10366 | 10364 | 10365 | 10366 | 0 | 75228 | 10366 | 10364 | 10366 | 10366 | 0 | 78180 |
| 50% | 25715 | 25612 | 25707 | 25714 | 26 | 67112 | 25715 | 25615 | 25711 | 25714 | 18 | 83411 |

TABLE 41
COMPARATIVE RESULTS FOR 0-1 KNAPSACK PROBLEM WITH NO. OF ITEMS 200 ON INVERSE STRONGLY CORRELATED DATA INSTANCES

| % of Total Weight | Canonical QEA | | | | | | Tuned-QEA | | | | | |
|---|---|---|---|---|---|---|---|---|---|---|---|---|
| | Best | Worst | Average | Median | Std | Av. Gen | Best | Worst | Average | Median | Std | Av. Gen |
| 1% | 1004 | 1004 | 1004 | 1004 | 0 | 16635 | 1004 | 1004 | 1004 | 1004 | 0 | 15302 |
| 5% | 5418 | 5417 | 5418 | 5418 | 0 | 146885 | 5418 | 5417 | 5418 | 5418 | 0 | 76808 |
| 10% | 10835 | 10835 | 10835 | 10835 | 0 | 96693 | 10835 | 10835 | 10835 | 10835 | 0 | 48913 |
| 20% | 21671 | 21471 | 21607 | 21571 | 56 | 66927 | 21671 | 21571 | 21647 | 21671 | 43 | 88892 |
| 50% | 53677 | 53477 | 53554 | 53577 | 57 | 115242 | 53677 | 53477 | 53564 | 53577 | 63 | 116157 |

TABLE 42
COMPARATIVE RESULTS FOR 0-1 KNAPSACK PROBLEM WITH NO. OF ITEMS 500 ON INVERSE STRONGLY CORRELATED DATA INSTANCES

| % of Total Weight | Canonical QEA | | | | | | Tuned-QEA | | | | | |
|---|---|---|---|---|---|---|---|---|---|---|---|---|
| | Best | Worst | Average | Median | Std | Av. Gen | Best | Worst | Average | Median | Std | Av. Gen |
| 1% | 2646 | 2646 | 2646 | 2646 | 0 | 157297 | 2646 | 2646 | 2646 | 2646 | 0 | 87592 |
| 5% | 13330 | 13230 | 13237 | 13230 | 25 | 94818 | 13330 | 13230 | 13250 | 13231 | 41 | 70600 |
| 10% | 26561 | 26361 | 26474 | 26461 | 57 | 122140 | 26561 | 26461 | 26498 | 26461 | 49 | 95406 |
| 20% | 52922 | 52622 | 52819 | 52822 | 76 | 150982 | 53022 | 52722 | 52899 | 52922 | 63 | 127552 |
| 50% | 131206 | 130906 | 131040 | 131056 | 84 | 260080 | 131206 | 130907 | 131096 | 131106 | 80 | 256846 |

Table 43

Comparative Results for 0-1 Knapsack problem with No. of Items 1000 on Inverse Strongly Correlated Data Instances

| % of Total Weight | Canonical QEA | | | | | | Tuned-QEA | | | | | |
|---|---|---|---|---|---|---|---|---|---|---|---|---|
| | Best | Worst | Average | Median | Std | Av. Gen | Best | Worst | Average | Median | Std | Av. Gen |
| 1% | **5313** | **5313** | **5313** | **5313** | **0** | 237190 | **5313** | **5313** | **5313** | **5313** | 0 | **119526** |
| 5% | **26665** | **26565** | **26578** | **26565** | **35** | 167943 | **26665** | **26565** | **26588** | **26565** | 43 | **115018** |
| 10% | **53230** | **52930** | 53063 | 53030 | 92 | 173987 | **53330** | **52930** | **53160** | **53130** | **75** | **144847** |
| 20% | 106160 | 105661 | 105949 | 105960 | **111** | 266106 | **106260** | **105860** | **106070** | **106060** | 112 | **260037** |
| 50% | **262853** | **262453** | **262705** | **262734** | **107** | 406118 | 263053 | **262453** | **262726** | **262753** | 128 | **400392** |

Table 44

Comparative Results for 0-1 Knapsack problem with No. of Items 2000 on Inverse Strongly Correlated Data Instances

| % of Total Weight | Canonical QEA | | | | | | Tuned-QEA | | | | | |
|---|---|---|---|---|---|---|---|---|---|---|---|---|
| | Best | Worst | Average | Median | Std | Av. Gen | Best | Worst | Average | Median | Std | Av. Gen |
| 1% | **10735** | **10735** | **10735** | **10735** | 0 | 241557 | **10735** | **10735** | **10735** | **10735** | **0** | **115282** |
| 5% | **53777** | **53377** | **53610** | **53577** | 80 | 268987 | **53777** | **53577** | **53707** | **53677** | **65** | **185206** |
| 10% | **107353** | 106854 | 107133 | 107154 | 132 | 320445 | **107453** | **107054** | **107260** | **107253** | **117** | **222582** |
| 20% | 214107 | 213508 | 213780 | 213708 | 181 | 366832 | **214407** | **213608** | **214057** | **214057** | **161** | **325496** |
| 50% | 530373 | 529470 | 529968 | 529956 | 222 | **479302** | **530585** | **529774** | **530124** | **530074** | **210** | 479602 |

Table 45

Comparative Results for 0-1 Knapsack problem with No. of Items 5000 on Inverse Strongly Correlated Data Instances

| % of Total Weight | Canonical QEA | | | | | | Tuned-QEA | | | | | |
|---|---|---|---|---|---|---|---|---|---|---|---|---|
| | Best | Worst | Average | Median | Std | Av. Gen | Best | Worst | Average | Median | Std | Av. Gen |
| 1% | **26943** | 26744 | 26837 | **26844** | 52 | 358185 | **26943** | **26844** | **26884** | **26844** | **50** | **223945** |
| 5% | 134318 | 133818 | 134011 | 134018 | 131 | 384282 | **134418** | **133918** | **134141** | **134118** | **128** | **264647** |
| 10% | 267837 | **267237** | 267530 | 267537 | **168** | 397700 | **268136** | **267237** | **267840** | **267837** | 226 | **321938** |
| 20% | 534374 | 533076 | 533738 | 533775 | **283** | 482327 | **535174** | **533675** | **534458** | **534424** | 396 | **462921** |
| 50% | 1320628 | 1318645 | 1319680 | 1319711 | **471** | 498748 | **1323057** | **1319198** | **1321242** | **1321151** | 756 | **497874** |

Table 46

Comparative Results for 0-1 Knapsack problem with No. of Items 10000 on Inverse Strongly Correlated Data Instances

| % of Total Capacity | Canonical QEA | | | | | | Tuned-QEA | | | | | |
|---|---|---|---|---|---|---|---|---|---|---|---|---|
| | Best | Worst | Average | Median | Std | Av. Gen | Best | Worst | Average | Median | Std | Av. Gen |
| 1% | 54002 | **53702** | 53865 | **53902** | **67** | 416330 | **54102** | **53702** | **53909** | **53902** | 87 | **248417** |
| 5% | 268910 | 268311 | 268580 | 268560 | **159** | 438535 | **269310** | **268511** | **268957** | **268910** | 227 | **394202** |

| | | | | | | | | | | | |
|---|---|---|---|---|---|---|---|---|---|---|---|
| 10% | 536822 | 535823 | 536327 | 536322 | 229 | 484127 | **537821** | **536421** | **537051** | **537035** | 353 | **447516** |
| 20% | 1068748 | 1066653 | 1067726 | 1067776 | **513** | 498317 | **1071838** | **1068942** | **1070400** | **1070424** | 755 | **493779** |
| 50% | 2636292 | 2627651 | 2632323 | 2632402 | **1950** | **498798** | **2641744** | **2632155** | **2637346** | **2637676** | 2681 | 499234 |

The convergence graphs have been plotted between objective function value and number of generations for both TCQEA and UCQEA for all the problem instances having No. of Items as 200 and 5000 for the median run. The convergence graph of TCQEA & UCQEA for all the problem instances having No. of Items as 200 is almost similar with TCQEA minutely outperforming UCQEA, however, problem instances having No. of Items as 5000 establishes the superiority of Tuning as the TCQEA is faster than UCQEA in all the graphs. The difference in performance between TCQEA and UCQEA increases with the capacity size and number of items in the knapsack problem.

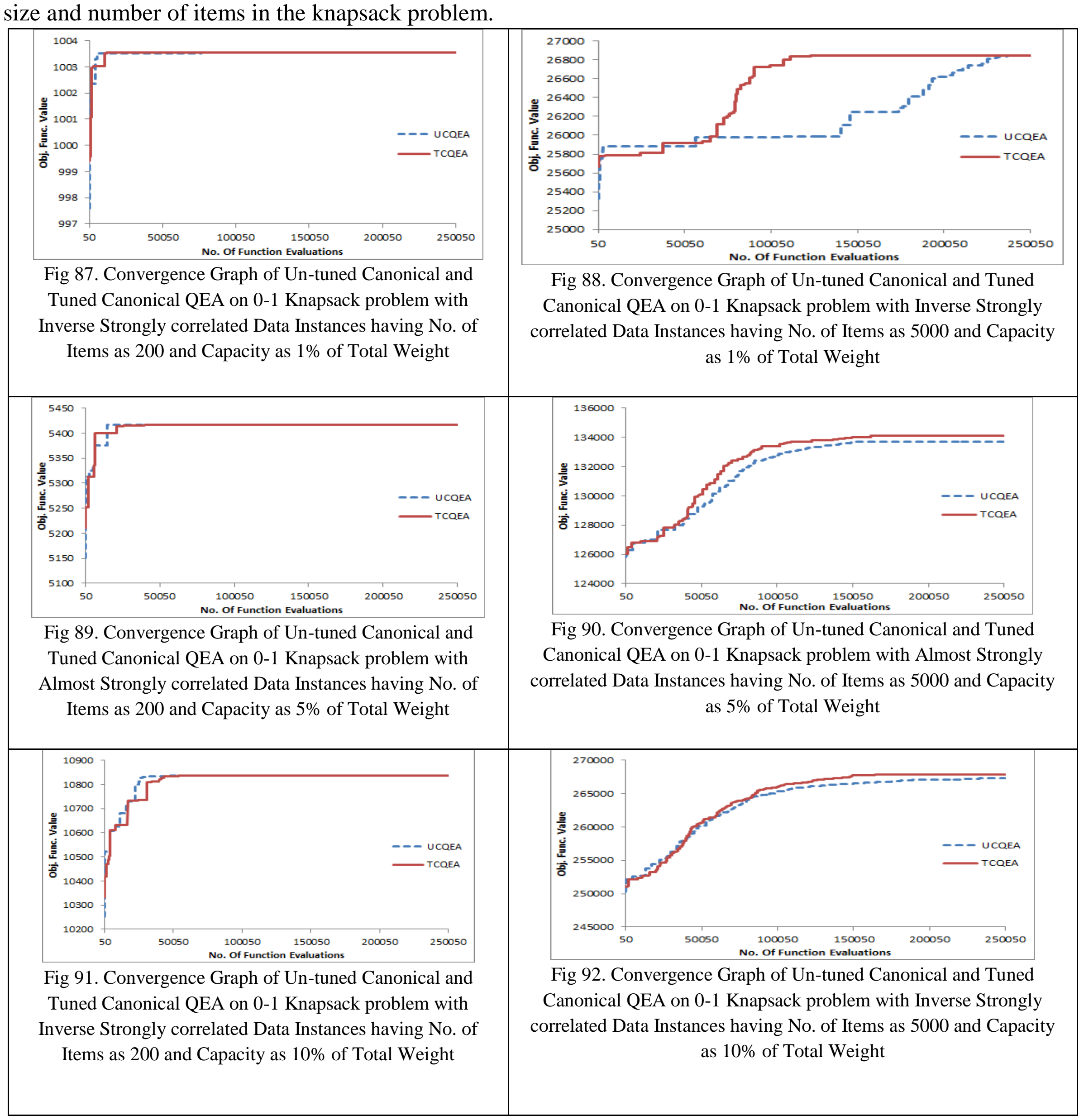

Fig 87. Convergence Graph of Un-tuned Canonical and Tuned Canonical QEA on 0-1 Knapsack problem with Inverse Strongly correlated Data Instances having No. of Items as 200 and Capacity as 1% of Total Weight

Fig 88. Convergence Graph of Un-tuned Canonical and Tuned Canonical QEA on 0-1 Knapsack problem with Inverse Strongly correlated Data Instances having No. of Items as 5000 and Capacity as 1% of Total Weight

Fig 89. Convergence Graph of Un-tuned Canonical and Tuned Canonical QEA on 0-1 Knapsack problem with Almost Strongly correlated Data Instances having No. of Items as 200 and Capacity as 5% of Total Weight

Fig 90. Convergence Graph of Un-tuned Canonical and Tuned Canonical QEA on 0-1 Knapsack problem with Almost Strongly correlated Data Instances having No. of Items as 5000 and Capacity as 5% of Total Weight

Fig 91. Convergence Graph of Un-tuned Canonical and Tuned Canonical QEA on 0-1 Knapsack problem with Inverse Strongly correlated Data Instances having No. of Items as 200 and Capacity as 10% of Total Weight

Fig 92. Convergence Graph of Un-tuned Canonical and Tuned Canonical QEA on 0-1 Knapsack problem with Inverse Strongly correlated Data Instances having No. of Items as 5000 and Capacity as 10% of Total Weight

In order to confirm the findings in Table 40 to 46, multi-problem non-parametric Wilcoxon's Signed Rank Test [Der2011] was performed on average objective function value (OFV) and average number of generation (Av. Gen.) of Tuned QEA and Canonical QEA on all the instances of Inverse Strongly Correlated Knapsack problem at a significance level of 5%. In case of comparison on average OFV, the null hypothesis for comparison was average OFV of Tuned QEA $\mu_1$ is less than or equal to the average OFV of Canonical QEA $\mu_2$ i.e. $H_0$: $\mu_1 \leq \mu_2$ and the alternate hypothesis is H1: $\mu_1 > \mu_2$. The result of test is presented in Table 47, which shows that null hypothesis can be rejected safely as Wilcoxon's Signed Rank test statistic is 2 and less than the critical value of 130 at significance level of $\alpha = 5\%$. The sample size n= 28 (distinct) is larger than 25 so z statistic has also been computed under the normal distribution and p value obtained is 0.000, which is less than 0.05, so null hypothesis can be safely rejected. This indicates the success of the proposed tuning method for tuning QEA on problems like Inverse Strongly Correlated Knapsack problem.

**Table 47: Wilcoxon's Signed Rank Test on Inv. Strongly Corr. KP Instances (OFV)**

| | TCQEA | UCQEA |
|---|---|---|
| | **$X_1$** | **$X_2$** |
| 1 | 0 | 0 |
| 2 | 167.9 | 167.9 |
| 3 | 460.7 | 460.7 |
| 4 | 966.9 | 966.9 |
| 5 | 2013.1 | 2013.2 |
| 6 | 2591.2 | 2591.2 |
| 7 | 18.01 | 18.01 |
| 8 | 476.8 | 476.8 |
| 9 | 1003.6 | 1003.6 |
| 10 | 2107.1 | 2107.1 |
| 11 | 4314 | 4313.8 |
| 12 | 5417.4 | 5417.3 |
| 13 | 194.26 | 194.26 |
| 14 | 1273.1 | 1273.1 |
| 15 | 2645.9 | 2645.9 |
| 16 | 5292 | 5285.3 |
| 17 | 10588 | 10538 |
| 18 | 13237 | 13202 |
| 19 | 488.62 | 488.62 |
| 20 | 2656.4 | 2656.2 |
| 21 | 5312.8 | 5293.7 |
| 22 | 10626 | 10446 |
| 23 | 21255 | 20927 |
| 24 | 26575 | 26246 |
| 25 | 999.39 | 999.39 |
| 26 | 5367.5 | 5348.2 |
| 27 | 10735 | 10538 |
| 28 | 21467 | 20849 |
| 29 | 42885 | 41801 |
| 30 | 53598 | 52426 |
| 31 | 2684.3 | 2684 |
| 32 | 13422 | 13130 |
| 33 | 26830 | 25934 |
| 34 | 53599 | 51439 |
| 35 | 106673 | 103064 |
| 36 | 133103 | 129298 |
| 37 | 5390.1 | 5362.4 |
| 38 | 26899 | 26059 |

| | |
|---|---|
| $\Sigma(+)$ | 404 |
| $\Sigma(-)$ | 2 |
| *n* | 28 |

| Null Hypothesis | **Test Statistic** *T* | **Critical Value** At an $\alpha$ of 5% |
|---|---|---|
| $H_0: \mu_1 <= \mu_2$ | 2 | 130 |

**For Large Samples (*n* > 25)**

Test Statistic

| | | | |
|---|---|---|---|
| *z* | -4.57706 | *E[T]* | 203 |
| | | *σ(T)* | 43.91469003 |

| Null Hypothesis | *p*-value | At an $\alpha$ of 5% |
|---|---|---|
| $H_0: \mu_1 <= \mu_2$ | 0.0000 | **Reject** |
| 39 | 53686 | 51588 |
| 40 | 106950 | 102408 |
| 41 | 212276 | 205306 |
| 42 | 264602 | 257093 |

In case of comparison on Av. Gen., the null hypothesis for comparison was Av. Gen. of Tuned QEA $\mu_1$ is greater than or equal to the Av. Gen. of Canonical QEA $\mu_2$ i.e. $H_0$: $\mu_1 \geq \mu_2$ and the alternate hypothesis is H1: $\mu_1 < \mu_2$. The result of test is presented in Table 48, which shows that null hypothesis can be rejected as Wilcoxon's Signed Rank test statistic is 297 and less than the critical value of 303 at significance level of $\alpha = 5\%$, which indicates that Tuned QEA requires less number of generations as compared to Canonical QEA. The sample size n= 41 (distinct) is larger than 25 so z statistic has also been computed under the normal distribution and p value obtained is 0.0418, which is less than 0.05, so null hypothesis can be rejected. Moreover, Tuned QEA is performing significantly better than Canonical QEA as shown by Table. 47. This indicates the success of the proposed tuning method for tuning QEA on problems like Inversely Strongly Correlated Knapsack instance and it even requires less number of generations and function evaluations as compared to Canonical QEA.

**Table 48: Wilcoxon's Signed Rank Test on Strongly Corr. KP Instances (Av. Gen.)**

| | TCQEA | UCQEA |
|---|---|---|
| | **$X_1$** | **$X_2$** |
| 1 | 0 | 0 |
| 2 | 1.9 | 1.5 |
| 3 | 1.6 | 1.9 |
| 4 | 1.8 | 2.1 |
| 5 | 470.9 | 545.8 |
| 6 | 270.5 | 406.4 |
| 7 | 4.1 | 3.5 |
| 8 | 2.1 | 2.2 |
| 9 | 316 | 237 |
| 10 | 401 | 459 |
| 11 | 254 | 739 |
| 12 | 421 | 854 |
| 13 | 10.2 | 194.26 |
| 14 | 435 | 1273.1 |
| 15 | 343 | 2646.1 |
| 16 | 383 | 5292.1 |
| 17 | 447 | 10584 |
| 18 | 498 | 13231 |
| 19 | 23.8 | 11.7 |
| 20 | 391 | 448 |
| 21 | 344 | 552 |
| 22 | 582 | 662 |
| 23 | 668 | 937 |
| 24 | 738 | 962 |
| 25 | 54.8 | 45.2 |
| 26 | 343 | 472 |
| 27 | 524 | 568 |
| 28 | 738 | 637 |
| 29 | 888 | 963 |
| 30 | 900 | 970 |

| | |
|---|---|
| Σ(+) | 297 |
| Σ(–) | 564 |
| *n* | 41 |

| Null Hypothesis | **Test Statistic** *T* | **Critical Value** At an α of 5% |
|---|---|---|
| $H_0$: $\mu_1$ >= $\mu_2$ | 297 | 303 |

**For Large Samples (*n* > 25)**

Test Statistic

| | | | |
|---|---|---|---|
| *z* | 1.729941 | *E[T]* | 430.5 |
| | | *σ(T)* | 77.17026629 |

| Null Hypothesis | *p*-value | At an α of 5% |
|---|---|---|
| $H_0$: $\mu_1$ >= $\mu_2$ | 0.0418 | **Reject** |

| | | | |
|---|---|---|---|
| 31 | 493 | 560 | |
| 32 | 629 | 425 | |
| 33 | 737 | 483 | |
| 34 | 924 | 703 | |
| 35 | 974 | 966 | |
| 36 | 966 | 960 | |
| 37 | 580 | 535 | |
| 38 | 749 | 487 | |
| 39 | 874 | 504 | |
| 40 | 949 | 579 | |
| 41 | 985 | 950 | |
| 42 | 983 | 972 | |

***5) Almost strongly correlated instances:***

The result of comparative study between Tuned QEA (TCQEA) and Canonical QEA (UCQEA) are given in Tables 49 to 55. The performance of TCQEA and UCQEA are similar when the no. of items to choose is 100 and the capacity of knapsack is 0.1 % of the total capacity. However, with the increase in number of items and the capacity of knapsack, the performance of TCQEA has improved over UCQEA in all the instances. The performance of TCQEA was also good on speed of convergence as indicated by average generations and the convergence graphs shown in figures 93 to 98, which also compared the speed of convergence of TCQEA to UCQEA.

TABLE 49

COMPARATIVE RESULTS FOR 0-1 KNAPSACK PROBLEM WITH NO. OF ITEMS 100 ON ALMOST STRONGLY CORRELATED DATA INSTANCES

| % of Total Weight | Canonical QEA | | | | | | Tuned-QEA | | | | | |
|---|---|---|---|---|---|---|---|---|---|---|---|---|
| | Best | Worst | Average | Median | Std | Av. Gen | Best | Worst | Average | Median | Std | Av. Gen |
| 1% | 1241 | 1082 | 1119 | 1096 | 45 | 236653 | **1500** | **1399** | **1490** | **1500** | **30** | **121166** |
| 5% | **4896** | 4593 | 4768 | 4795 | 65 | 91090 | **4896** | **4793** | **4806** | **4796** | **30** | **80589** |
| 10% | **8383** | 8182 | 8326 | 8330 | 57 | **70587** | 8382 | **8278** | **8347** | **8378** | **47** | 87707 |
| 20% | 14644 | 14337 | 14569 | 14546 | 73 | **94975** | **14646** | **14447** | **14597** | **14640** | **56** | 99469 |
| 50% | 31819 | 31711 | 31762 | 31721 | 51 | **95642** | **31823** | **31716** | **31800** | **31816** | **37** | 114593 |

TABLE 50

COMPARATIVE RESULTS FOR 0-1 KNAPSACK PROBLEM WITH NO. OF ITEMS 200 ON ALMOST STRONGLY CORRELATED DATA INSTANCES

| % of Total Weight | Canonical QEA | | | | | | Tuned-QEA | | | | | |
|---|---|---|---|---|---|---|---|---|---|---|---|---|
| | Best | Worst | Average | Median | Std | Av. Gen | Best | Worst | Average | Median | Std | Av. Gen |
| 1% | 2988 | 2391 | 2761 | 2792 | 153 | 400905 | **3090** | **2892** | **3034** | **2993** | **56** | **150886** |
| 5% | 9445 | 9141 | 9372 | 9437 | 88 | 172093 | **9545** | **9337** | **9465** | **9444** | **63** | **141818** |
| 10% | **16365** | 16057 | 16206 | 16258 | **78** | **131100** | 16363 | **16061** | **16230** | **16260** | 88 | 187100 |
| 20% | 28777 | 28384 | 28576 | 28586 | **79** | **140243** | **28786** | **28486** | **28625** | **28594** | 81 | 174474 |
| 50% | 63238 | 62939 | 63110 | 63135 | **72** | **146637** | **63333** | **63034** | **63131** | **63137** | 73 | 174035 |

TABLE 51

COMPARATIVE RESULTS FOR 0-1 KNAPSACK PROBLEM WITH NO. OF ITEMS 500 ON ALMOST STRONGLY CORRELATED DATA INSTANCES

| % of Total | Canonical QEA | Tuned-QEA |
|---|---|---|

| Weight | Best | Worst | Average | Median | Std | Av. Gen | Best | Worst | Average | Median | Std | Av. Gen |
|---|---|---|---|---|---|---|---|---|---|---|---|---|
| 1% | 7622 | 7122 | 7442 | 7428 | **118** | 405557 | **7723** | **7225** | **7543** | **7527** | 120 | **296670** |
| 5% | 23496 | 22679 | 23176 | 23195 | 187 | **335337** | **23497** | **22889** | **23247** | **23292** | **134** | 367419 |
| 10% | 40031 | 39442 | 39742 | 39732 | **142** | **405793** | **40133** | **39443** | **39864** | **39841** | 165 | 408778 |
| 20% | 70275 | 69494 | 69943 | 69976 | 164 | **409250** | **70382** | **69583** | **70106** | **70131** | 162 | 462442 |
| 50% | 154330 | 153646 | 154075 | 154135 | 151 | **424750** | **154434** | **154022** | **154207** | **154227** | **111** | 454351 |

TABLE 52

COMPARATIVE RESULTS FOR 0-1 KNAPSACK PROBLEM WITH NO. OF ITEMS 1000 ON ALMOST STRONGLY CORRELATED DATA INSTANCES

| % of Total Weight | Canonical QEA | | | | | | Tuned-QEA | | | | | |
|---|---|---|---|---|---|---|---|---|---|---|---|---|
| | Best | Worst | Average | Median | Std | Av. Gen | Best | Worst | Average | Median | Std | Av. Gen |
| 1% | 14814 | 13925 | 14444 | 14465 | 242 | 489228 | **14918** | **14320** | **14680** | **14715** | **172** | **477929** |
| 5% | 46432 | 45738 | 46145 | 46142 | **198** | **488230** | **46830** | **45724** | **46344** | **46375** | 242 | 491446 |
| 10% | 79842 | 78948 | 79514 | 79500 | **232** | **490285** | **80128** | **79023** | **79666** | **79631** | 233 | 491915 |
| 20% | 141567 | 140576 | 141042 | 141068 | 241 | **488678** | **141663** | **140858** | **141246** | **141251** | **194** | 493132 |
| 50% | 311738 | 310958 | 311305 | 311352 | 207 | **490723** | **311944** | **311243** | **311583** | **311600** | **172** | 492598 |

TABLE 53

COMPARATIVE RESULTS FOR 0-1 KNAPSACK PROBLEM WITH NO. OF ITEMS 2000 ON ALMOST STRONGLY CORRELATED DATA INSTANCES

| % of Total Weight | Canonical QEA | | | | | | Tuned-QEA | | | | | |
|---|---|---|---|---|---|---|---|---|---|---|---|---|
| | Best | Worst | Average | Median | Std | Av. Gen | Best | Worst | Average | Median | Std | Av. Gen |
| 1% | 27976 | 26427 | 27150 | 27202 | 380 | **496143** | **29140** | **27736** | **28369** | **28347** | **351** | 496412 |
| 5% | **92522** | 90156 | 91439 | 91421 | 516 | **496280** | 92522 | **91427** | **91944** | **91974** | **313** | 496393 |
| 10% | 160354 | 157866 | 159470 | 159458 | 541 | **495527** | **161026** | **159116** | **160151** | **160139** | **443** | 496205 |
| 20% | 285647 | 283651 | 284616 | 284687 | 453 | 496972 | **285889** | **284078** | **285220** | **285259** | **397** | **496911** |
| 50% | 633200 | 631600 | 632440 | 632433 | **395** | **493363** | **633936** | **632315** | **633279** | **633318** | 415 | 494700 |

TABLE 54

COMPARATIVE RESULTS FOR 0-1 KNAPSACK PROBLEM WITH NO. OF ITEMS 5000 ON ALMOST STRONGLY CORRELATED DATA INSTANCES

| % of Total Weight | Canonical QEA | | | | | | Tuned-QEA | | | | | |
|---|---|---|---|---|---|---|---|---|---|---|---|---|
| | Best | Worst | Average | Median | Std | Av. Gen | Best | Worst | Average | Median | Std | Av. Gen |
| 1% | 57894 | 55665 | 56961 | 56967 | **621** | **498495** | **63938** | **60714** | **62249** | **62291** | 802 | 499019 |
| 5% | 216320 | 212387 | 214464 | 214510 | 980 | **498615** | **219877** | **216056** | **217976** | **218161** | **933** | 499326 |
| 10% | 387621 | 381723 | 384629 | 384410 | **1294** | **498467** | **391901** | **386206** | **388919** | **389040** | 1540 | 499208 |
| 20% | 702807 | 698149 | 700531 | 700498 | **1219** | **498440** | **707706** | **700570** | **704892** | **705132** | 1707 | 499442 |

| 50% | 1585045 | **1581945** | 1583469 | 1583527 | **828** | **496813** | **158719 6** | 1581382 | **1584560** | **1584633** | 1343 | 497323 |
|---|---|---|---|---|---|---|---|---|---|---|---|---|

TABLE 55

COMPARATIVE RESULTS FOR 0-1 KNAPSACK PROBLEM WITH NO. OF ITEMS 10000 ON ALMOST STRONGLY CORRELATED DATA INSTANCES

| % of Total Weight | Canonical QEA | | | | | | Tuned-QEA | | | | | |
|---|---|---|---|---|---|---|---|---|---|---|---|---|
| | **Best** | **Worst** | **Average** | **Median** | **Std** | **Av. Gen** | **Best** | **Worst** | **Average** | **Median** | **Std** | **Av. Gen** |
| 1% | 101672 | 96669 | 98592 | 98390 | **1153** | **498482** | **111412** | **105861** | **108720** | **108706** | 1232 | 499478 |
| 5% | 405353 | 398364 | 401242 | 401271 | **1619** | **498843** | **411606** | **403020** | **407270** | **407194** | 2043 | 499577 |
| 10% | 740740 | 731839 | 736249 | 736151 | **2567** | 499432 | **752363** | **737389** | **744688** | **744251** | 3824 | **499330** |
| 20% | 1372606 | 1361099 | 1367754 | 1367855 | **3320** | **499310** | **1393697** | **1366535** | **1378514** | **1379438** | 5993 | 499551 |
| 50% | 3163094 | 3153536 | 3159419 | 3159617 | **2560** | **498792** | **3170687** | **3159761** | **3164171** | **3164275** | 2811 | 499013 |

The convergence graphs have been plotted between objective function value and number of generations for both TCQEA and UCQEA for all the problem instances having No. of Items as 200 and 5000 for the median run. The convergence graph clearly establishes the superiority of Tuning as the TCQEA is faster than UCQEA in all the graphs. The difference in performance between TCQEA and UCQEA increases with the capacity size and number of items in the knapsack problem.

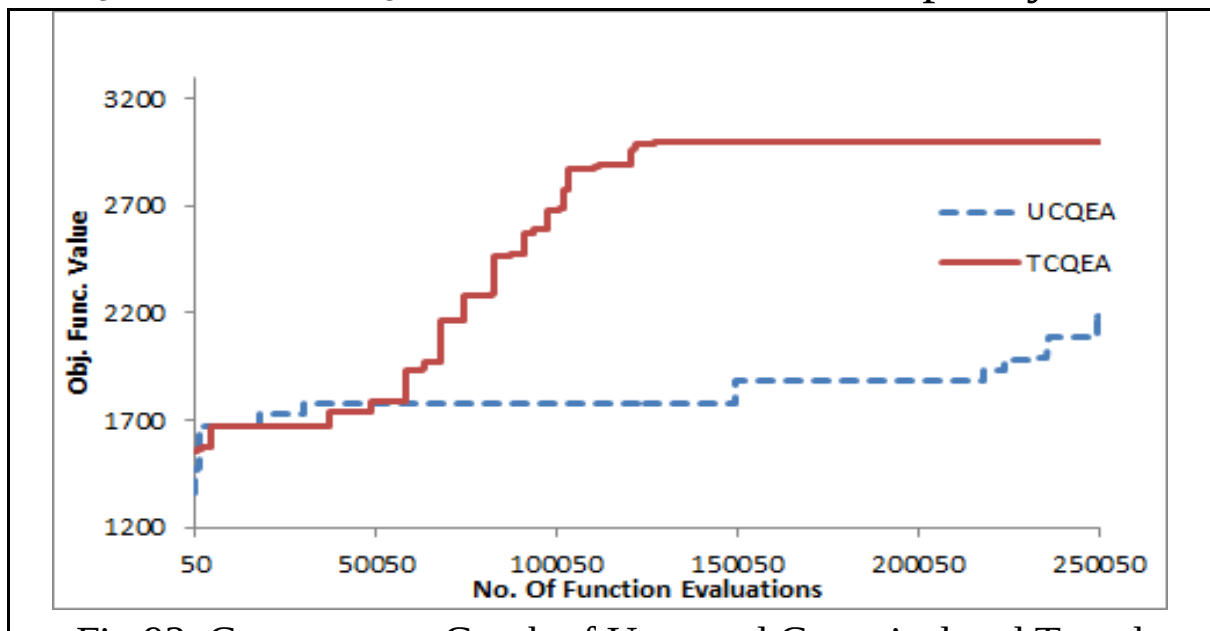

Fig 93. Convergence Graph of Un-tuned Canonical and Tuned Canonical QEA on 0-1 Knapsack problem with Almost Strongly correlated Data Instances having No. of Items as 200 and Capacity as 1% of Total Weight

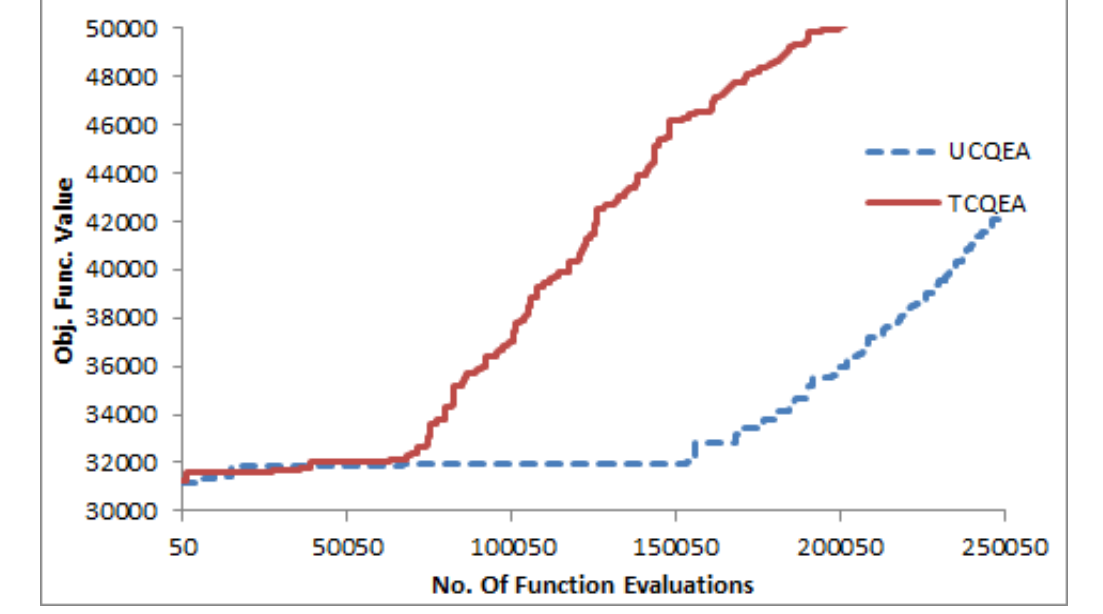

Fig 94. Convergence Graph of Un-tuned Canonical and Tuned Canonical QEA on 0-1 Knapsack problem with Almost Strongly correlated Data Instances having No. of Items as 5000 and Capacity as 1% of Total Weight

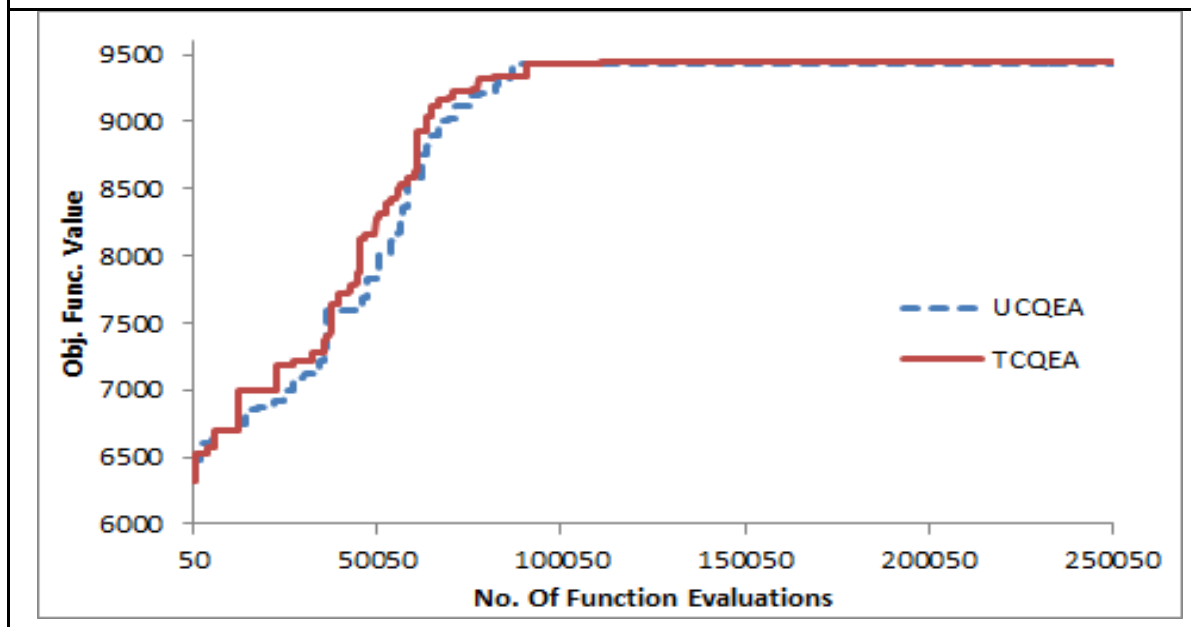

Fig 95. Convergence Graph of Un-tuned Canonical and Tuned Canonical QEA on 0-1 Knapsack problem with Almost Strongly correlated Data Instances having No. of Items as 200 and Capacity as 5% of Total Weight

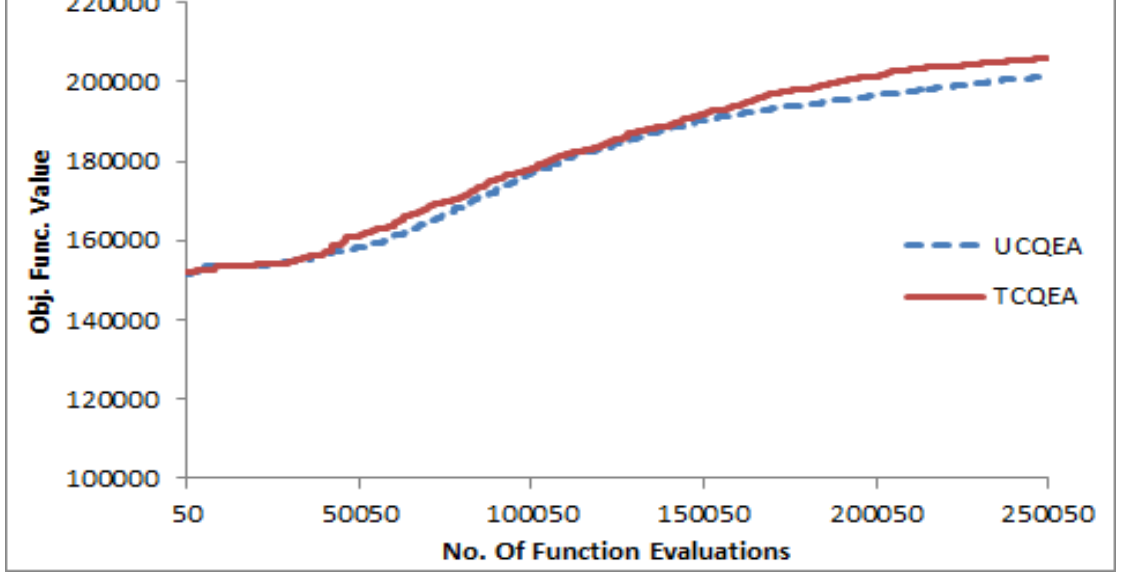

Fig 96. Convergence Graph of Un-tuned Canonical and Tuned Canonical QEA on 0-1 Knapsack problem with Almost Strongly correlated Data Instances having No. of Items as 5000 and Capacity as 5% of Total Weight

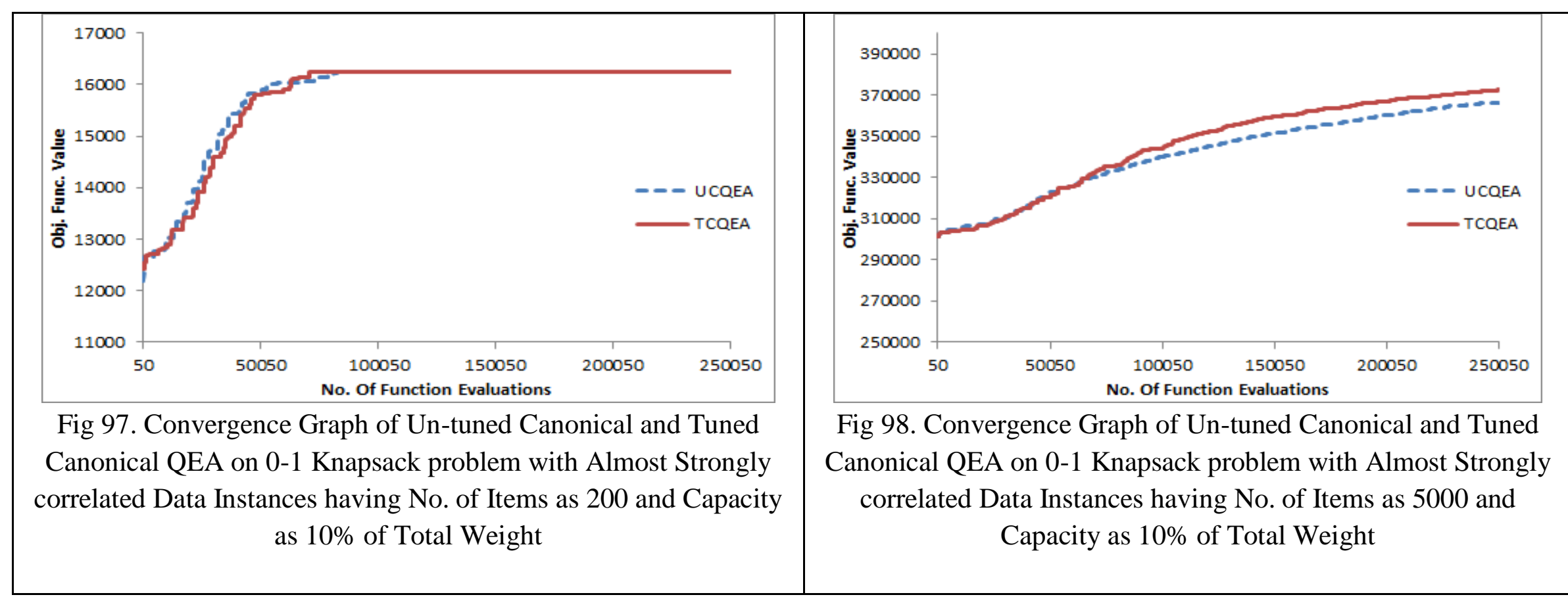

Fig 97. Convergence Graph of Un-tuned Canonical and Tuned Canonical QEA on 0-1 Knapsack problem with Almost Strongly correlated Data Instances having No. of Items as 200 and Capacity as 10% of Total Weight

Fig 98. Convergence Graph of Un-tuned Canonical and Tuned Canonical QEA on 0-1 Knapsack problem with Almost Strongly correlated Data Instances having No. of Items as 5000 and Capacity as 10% of Total Weight

In order to confirm the findings in Table 49 to 55, multi-problem non-parametric Wilcoxon's Signed Rank Test [Der2011] was performed on average objective function value (OFV) and average number of generation (Av. Gen.) of Tuned QEA and Canonical QEA on all the instances of Almost Strongly Correlated Knapsack problem at a significance level of 5%. In case of comparison on average OFV, the null hypothesis for comparison was average OFV of Tuned QEA $\mu_1$ is less than or equal to the average OFV of Canonical QEA $\mu_2$ i.e. $H_0$: $\mu_1 \leq \mu_2$ and the alternate hypothesis is H1: $\mu_1 > \mu_2$. The result of test is presented in Table 56, which shows that null hypothesis can be rejected safely as Wilcoxon's Signed Rank test statistic is 0 and less than the critical value of 303 at significance level of $\alpha$ = 5%. The sample size n= 41 (distinct) is larger than 25 so z statistic has also been computed under the normal distribution and p value obtained is 0.000, which is less than 0.05, so null hypothesis can be safely rejected. This indicates the success of the proposed tuning method for tuning QEA on problems like Almost Strongly Correlated Knapsack problem.

**Table 56: Wilcoxon's Signed Rank Test on Al. Strongly Corr. KP Instances (OFV)**

| | TCQEA | UCQEA |
|---|---|---|
| | $X_1$ | $X_2$ |
| 1 | 246 | 246 |
| 2 | 856 | 666.7 |
| 3 | 1450 | 1070.7 |
| 4 | 2396.2 | 1823.9 |
| 5 | 4033.1 | 3588.1 |
| 6 | 4765.3 | 4576.4 |
| 7 | 687.4 | 419.1 |
| 8 | 1869.8 | 1062 |
| 9 | 2920.3 | 1758.4 |
| 10 | 4663.6 | 3127.7 |
| 11 | 7825.4 | 6448.1 |
| 12 | 9252.5 | 8294.7 |
| 13 | 1633.3 | 681.7 |
| 14 | 4384.7 | 2069.5 |
| 15 | 6868.4 | 3666.2 |
| 16 | 10953 | 6864.4 |
| 17 | 18273.8 | 14091.2 |
| 18 | 21657.2 | 18108.8 |
| 19 | 2774.3 | 1039.1 |
| 20 | 7493.2 | 3667.3 |

| | |
|---|---|
| Σ(+) | 861 |
| Σ(−) | 0 |
| *n* | 41 |

| Null Hypothesis | Test Statistic *T* | Critical Value At an α of 5% |
|---|---|---|
| $H_0$: $\mu_1$ <= $\mu_2$ | 0 | 303 |

**For Large Samples (*n* > 25)**

Test Statistic

| | | | |
|---|---|---|---|
| *z* | -5.57857 | *E[T]* | 430.5 |
| | | *σ(T)* | 77.17026629 |

| Null Hypothesis | *p*-value | At an α of 5% |
|---|---|---|
| $H_0$: $\mu_1$ <= $\mu_2$ | 0.0000 | **Reject** |
| 21 | 11953.5 | 6803.3 |
| 22 | 19652.8 | 13018.2 |
| 23 | 33820.7 | 26692.8 |
| 24 | 40259.8 | 34111.4 |
| 25 | 4392 | 1753.4 |
| 26 | 12521.6 | 6870.9 |
| 27 | 20608.1 | 13149.2 |
| 28 | 35268.4 | 25637.7 |
| 29 | 62129.6 | 52046.6 |
| 30 | 75107.2 | 65762 |
| 31 | 7628.2 | 3725.3 |
| 32 | 24620.3 | 16342 |
| 33 | 43261.3 | 31842 |
| 34 | 77221 | 62615.4 |
| 35 | 142088 | 125899 |
| 36 | 173517 | 158469 |
| 37 | 11545.1 | 6939.3 |
| 38 | 42686.2 | 31809.6 |
| 39 | 77188.4 | 62492.7 |
| 40 | 142590 | 123709 |
| 41 | 269378 | 248335 |
| 42 | 331196 | 311913 |

In case of comparison on Av. Gen., the null hypothesis for comparison was Av. Gen. of Tuned QEA $\mu_1$ is greater than or equal to the Av. Gen. of Canonical QEA $\mu_2$ i.e. $H_0$: $\mu_1 \geq \mu_2$ and the alternate hypothesis is H1: $\mu_1 < \mu_2$. The result of test is presented in Table 57, which shows that null hypothesis cannot be rejected as Wilcoxon's Signed Rank test statistic is 725 and more than the critical value of 319 at significance level of $\alpha$ = 5%, which indicates that Tuned QEA requires more number of generations as compared to Canonical QEA. The sample size n= 42 (distinct) is larger than 25 so z statistic has also been computed under the normal distribution and p value obtained is 0.9997, which is more than 0.05, so null hypothesis cannot be rejected. However, Tuned QEA is performing significantly better than Canonical QEA as shown by Table. 56. This indicates the success of the proposed tuning method for tuning QEA on problems like Almost Strongly Correlated Knapsack instance even though it requires more number of generations and function evaluations as compared to Canonical QEA.

**Table 57: Wilcoxon's Signed Rank Test on Al. Strongly Corr. KP Instances (Av. Gen.)**

| | TCQEA | UCQEA |
|---|---|---|
| | **$X_1$** | **$X_2$** |
| 1 | 42.5 | 77 |
| 2 | 275.8 | 431.9 |
| 3 | 384.2 | 468.2 |
| 4 | 424.9 | 576.7 |
| 5 | 558 | 964.9 |
| 6 | 551.1 | 973.5 |
| 7 | 342.8 | 503.2 |
| 8 | 590.6 | 495.8 |
| 9 | 743.2 | 571.7 |
| 10 | 836.7 | 607.5 |
| 11 | 874.7 | 953.7 |
| 12 | 765.8 | 979.6 |
| 13 | 789.1 | 550.2 |

| | |
|---|---|
| $\Sigma(+)$ | 725 |
| $\Sigma(-)$ | 178 |
| $n$ | 42 |

| Null Hypothesis | **Test Statistic** $T$ | **Critical Value** At an $\alpha$ of 5% |
|---|---|---|
| $H_0$: $\mu_1$ >= $\mu_2$ | 725 | 319 |

**For Large Samples ($n$ > 25)**

Test Statistic

| | | | |
|---|---|---|---|
| $Z$ | -3.41975 | $E[T]$ | 451.5 |
| | | $\sigma(T)$ | 79.97655907 |

| Null Hypothesis | $p$-value | At an $\alpha$ of 5% |
|---|---|---|
| $H_0$: $\mu_1$ >= $\mu_2$ | 0.9997 | |
| 14 | 898.7 | 464.4 |
| 15 | 909.8 | 459.9 |
| 16 | 955.1 | 670.5 |
| 17 | 949.7 | 972.6 |
| 18 | 974 | 981.3 |
| 19 | 890.4 | 482.3 |
| 20 | 956.3 | 570.4 |
| 21 | 969.5 | 514.6 |
| 22 | 963.7 | 558.5 |
| 23 | 985 | 969.1 |
| 24 | 985.4 | 984.4 |
| 25 | 958.6 | 471.3 |
| 26 | 965.1 | 468.1 |
| 27 | 982.2 | 576.7 |
| 28 | 986.4 | 807.3 |
| 29 | 980.9 | 973.4 |
| 30 | 989.4 | 976.6 |
| 31 | 946.3 | 558.8 |
| 32 | 980 | 547.1 |
| 33 | 988 | 412.7 |
| 34 | 989.5 | 643.7 |
| 35 | 986 | 964.3 |
| 36 | 990.2 | 976.2 |
| 37 | 958.5 | 485 |
| 38 | 968 | 472.1 |
| 39 | 987.4 | 534.9 |
| 40 | 990.2 | 661.2 |
| 41 | 988.5 | 964.2 |
| 42 | 988.7 | 984.6 |

***6) Subset sum instances:***

The result of comparative study between Tuned QEA (TCQEA) and Canonical QEA (UCQEA) are given in Tables 58 to 64. The performance of TCQEA and UCQEA are similar in all the problem instances. The performance of TCQEA was also good on speed of convergence as indicated by average generations and the convergence graphs shown in figures 99 to 104, which also compared the speed of convergence of TCQEA to UCQEA.

TABLE 58

COMPARATIVE RESULTS FOR 0-1 KNAPSACK PROBLEM WITH NO. OF ITEMS 100 ON SUBSET SUM DATA INSTANCES

| % of Total Weight | Canonical QEA | | | | | | Tuned-QEA | | | | | |
|---|---|---|---|---|---|---|---|---|---|---|---|---|
| | Best | Worst | Average | Median | Std | Av. Gen | Best | Worst | Average | Median | Std | Av. Gen |
| 1% | 478 | 478 | 478 | 478 | 0 | 158857 | 478 | 478 | 478 | 478 | 0 | 30601 |
| 5% | 2392 | 2392 | 2392 | 2392 | 0 | 40402 | 2392 | 2392 | 2392 | 2392 | 0 | 30743 |
| 10% | 4783 | 4783 | 4783 | 4783 | 0 | 43490 | 4783 | 4783 | 4783 | 4783 | 0 | 48434 |
| 20% | 9567 | 9566 | 9567 | 9567 | 0 | 62422 | 9567 | 9566 | 9567 | 9567 | 0 | 87810 |
| 50% | 23916 | 23916 | 23916 | 23916 | 0 | 130042 | 23916 | 23916 | 23916 | 23916 | 0 | 129396 |

TABLE 59

COMPARATIVE RESULTS FOR 0-1 KNAPSACK PROBLEM WITH NO. OF ITEMS 200 ON SUBSET SUM DATA INSTANCES

| % of Total Weight | Canonical QEA | | | | | | Tuned-QEA | | | | | |
|---|---|---|---|---|---|---|---|---|---|---|---|---|
| | Best | Worst | Average | Median | Std | Av. Gen | Best | Worst | Average | Median | Std | Av. Gen |
| 1% | 1004 | 1004 | 1004 | 1004 | 0 | 107100 | 1004 | 1004 | 1004 | 1004 | 0 | 51655 |
| 5% | 5018 | 5018 | 5018 | 5018 | 0 | 55590 | 5018 | 5018 | 5018 | 5018 | 0 | 25014 |
| 10% | 10036 | 10036 | 10036 | 10036 | 0 | 128418 | 10036 | 10036 | 10036 | 10036 | 0 | 66551 |
| 20% | 20072 | 20072 | 20072 | 20072 | 0 | 140185 | 20072 | 20072 | 20072 | 20072 | 0 | 117035 |
| 50% | 50181 | 50181 | 50181 | 50181 | 0 | 146673 | 50181 | 50181 | 50181 | 50181 | 0 | 120661 |

TABLE 60

COMPARATIVE RESULTS FOR 0-1 KNAPSACK PROBLEM WITH NO. OF ITEMS 500 ON SUBSET SUM DATA INSTANCES

| % of Total Weight | Canonical QEA | | | | | | Tuned-QEA | | | | | |
|---|---|---|---|---|---|---|---|---|---|---|---|---|
| | Best | Worst | Average | Median | Std | Av. Gen | Best | Worst | Average | Median | Std | Av. Gen |
| 1% | 2446 | 2446 | 2446 | 2446 | 0 | 82852 | 2446 | 2446 | 2446 | 2446 | 0 | 40082 |
| 5% | 12232 | 12232 | 12232 | 12232 | 0 | 148558 | 12232 | 12232 | 12232 | 12232 | 0 | 104244 |
| 10% | 24463 | 24463 | 24463 | 24463 | 0 | 93957 | 24463 | 24463 | 24463 | 24463 | 0 | 143567 |
| 20% | 48926 | 48926 | 48926 | 48926 | 0 | 110112 | 48926 | 48926 | 48926 | 48926 | 0 | 123281 |
| 50% | 122315 | 122315 | 122315 | 122315 | 0 | 121532 | 122315 | 122315 | 122315 | 122315 | 0 | 151935 |

TABLE 61

COMPARATIVE RESULTS FOR 0-1 KNAPSACK PROBLEM WITH NO. OF ITEMS 1000 ON SUBSET SUM DATA INSTANCES

| % of Total Weight | Canonical QEA | | | | | | Tuned-QEA | | | | | |
|---|---|---|---|---|---|---|---|---|---|---|---|---|
| | Best | Worst | Average | Median | Std | Av. Gen | Best | Worst | Average | Median | Std | Av. Gen |
| 1% | 4913 | 4913 | 4913 | 4913 | 0 | 76727 | 4913 | 4913 | 4913 | 4913 | 0 | 72567 |
| 5% | 24567 | 24567 | 24567 | 24567 | 0 | 133743 | 24567 | 24567 | 24567 | 24567 | 0 | 136221 |
| 10% | 49134 | 49134 | 49134 | 49134 | 0 | 200485 | 49134 | 49134 | 49134 | 49134 | 0 | 148767 |
| 20% | 98268 | 98268 | 98268 | 98268 | 0 | 179258 | 98268 | 98268 | 98268 | 98268 | 0 | 152655 |
| 50% | 245670 | 245670 | 245670 | 245670 | 0 | 161037 | 245670 | 245670 | 245670 | 245670 | 0 | 136112 |

TABLE 62

COMPARATIVE RESULTS FOR 0-1 KNAPSACK PROBLEM WITH NO. OF ITEMS 2000 ON SUBSET SUM DATA INSTANCES

| % of Total Weight | Canonical QEA | | | | | | Tuned-QEA | | | | | |
|---|---|---|---|---|---|---|---|---|---|---|---|---|
| | Best | Worst | Average | Median | Std | Av. Gen | Best | Worst | Average | Median | Std | Av. Gen |
| 1% | 9936 | 9936 | 9936 | 9936 | 0 | 181055 | 9936 | 9936 | 9936 | 9936 | 0 | 102967 |
| 5% | 49681 | 49681 | 49681 | 49681 | 0 | 167240 | 49681 | 49681 | 49681 | 49681 | 0 | 144395 |
| 10% | 99361 | 99361 | 99361 | 99361 | 0 | 153303 | 99361 | 99361 | 99361 | 99361 | 0 | 191984 |
| 20% | 198723 | 198723 | 198723 | 198723 | 0 | 165315 | 198723 | 198723 | 198723 | 198723 | 0 | 183579 |
| 50% | 496807 | 496807 | 496807 | 496807 | 0 | 240813 | 496807 | 496807 | 496807 | 496807 | 0 | 186295 |

TABLE 63

COMPARATIVE RESULTS FOR 0-1 KNAPSACK PROBLEM WITH NO. OF ITEMS 5000 ON SUBSET SUM DATA INSTANCES

| % of Total Weight | Canonical QEA | | | | | | Tuned-QEA | | | | | |
|---|---|---|---|---|---|---|---|---|---|---|---|---|
| | Best | Worst | Average | Median | Std | Av. Gen | Best | Worst | Average | Median | Std | Av. Gen |
| 1% | 24846 | 24846 | 24846 | 24846 | 0 | 227532 | 24846 | 24846 | 24846 | 24846 | 0 | 200957 |
| 5% | 124228 | 124228 | 124228 | 124228 | 0 | 282130 | 124228 | 124228 | 124228 | 124228 | 0 | 200835 |
| 10% | 248456 | 248456 | 248456 | 248456 | 0 | 205327 | 248456 | 248456 | 248456 | 248456 | 0 | 212048 |
| 20% | 496912 | 496912 | 496912 | 496912 | 0 | 297685 | 496912 | 496912 | 496912 | 496912 | 0 | 225195 |
| 50% | 1242280 | 1242280 | 1242280 | 1242280 | 0 | 188198 | 1242280 | 1242280 | 1242280 | 1242280 | 0 | 167822 |

TABLE 64

COMPARATIVE RESULTS FOR 0-1 KNAPSACK PROBLEM WITH NO. OF ITEMS 10000 ON SUBSET SUM DATA INSTANCES

| % of Total Weight | Canonical QEA | | | | | | Tuned-QEA | | | | | |
|---|---|---|---|---|---|---|---|---|---|---|---|---|
| | Best | Worst | Average | Median | Std | Av. Gen | Best | Worst | Average | Median | Std | Av. Gen |
| 1% | 49906 | 49906 | 49906 | 49906 | 0 | 270943 | 49906 | 49906 | 49906 | 49906 | 0 | 214365 |
| 5% | 249530 | 249530 | 249530 | 249530 | 0 | 240035 | 249530 | 249530 | 249530 | 249530 | 0 | 221704 |
| 10% | 499060 | 499060 | 499060 | 499060 | 0 | 259060 | 499060 | 499060 | 499060 | 499060 | 0 | 218945 |
| 20% | 998119 | 998119 | 998119 | 998119 | 0 | 219655 | 998119 | 998119 | 998119 | 998119 | 0 | 233086 |
| 50% | 2495299 | 2495298 | 2495298 | 2495298 | 0 | 227995 | 2495300 | 2495298 | 2495298 | 2495298 | 0 | 181688 |

The convergence graphs have been plotted between objective function value and number of generations for both TCQEA and UCQEA for all the problem instances having No. of Items as 200 and 5000 for the median run. The convergence graph shows that TCQEA is as fast as UCQEA in all the graphs.

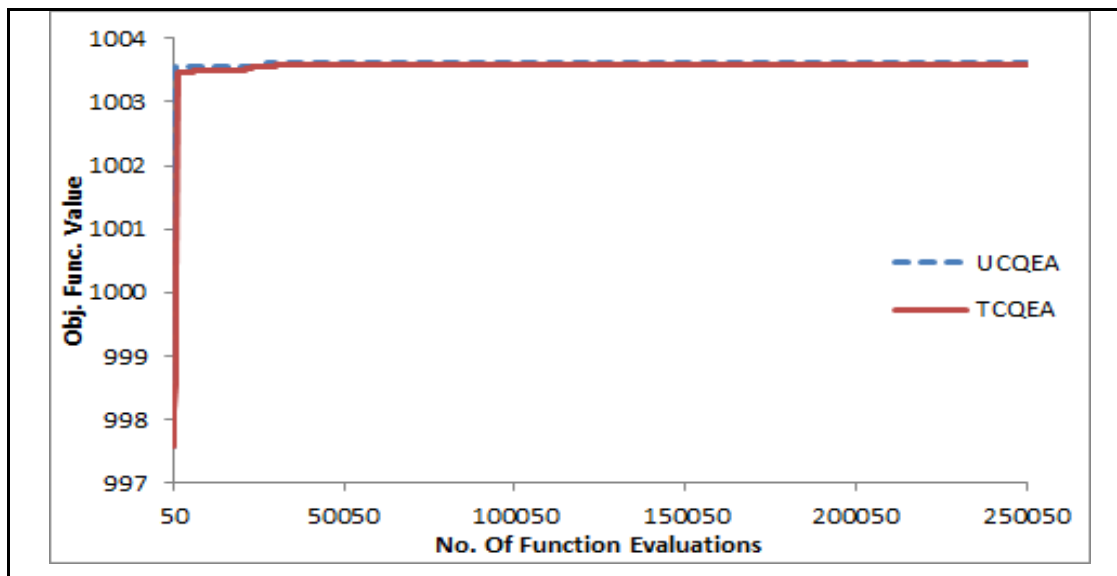

Fig 99. Convergence Graph of Un-tuned Canonical and Tuned Canonical QEA on 0-1 Knapsack problem with Subset sum Data Instances having No. of Items as 200 and Capacity as 1% of Total Weight

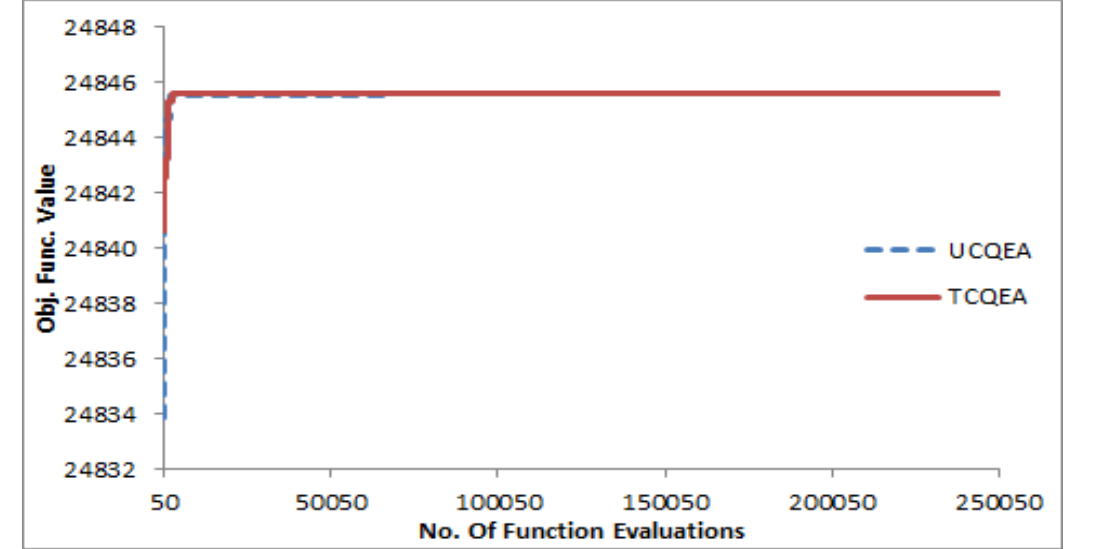

Fig 100. Convergence Graph of Un-tuned Canonical and Tuned Canonical QEA on 0-1 Knapsack problem with Subset sum Data Instances having No. of Items as 5000 and Capacity as 1% of Total Weight

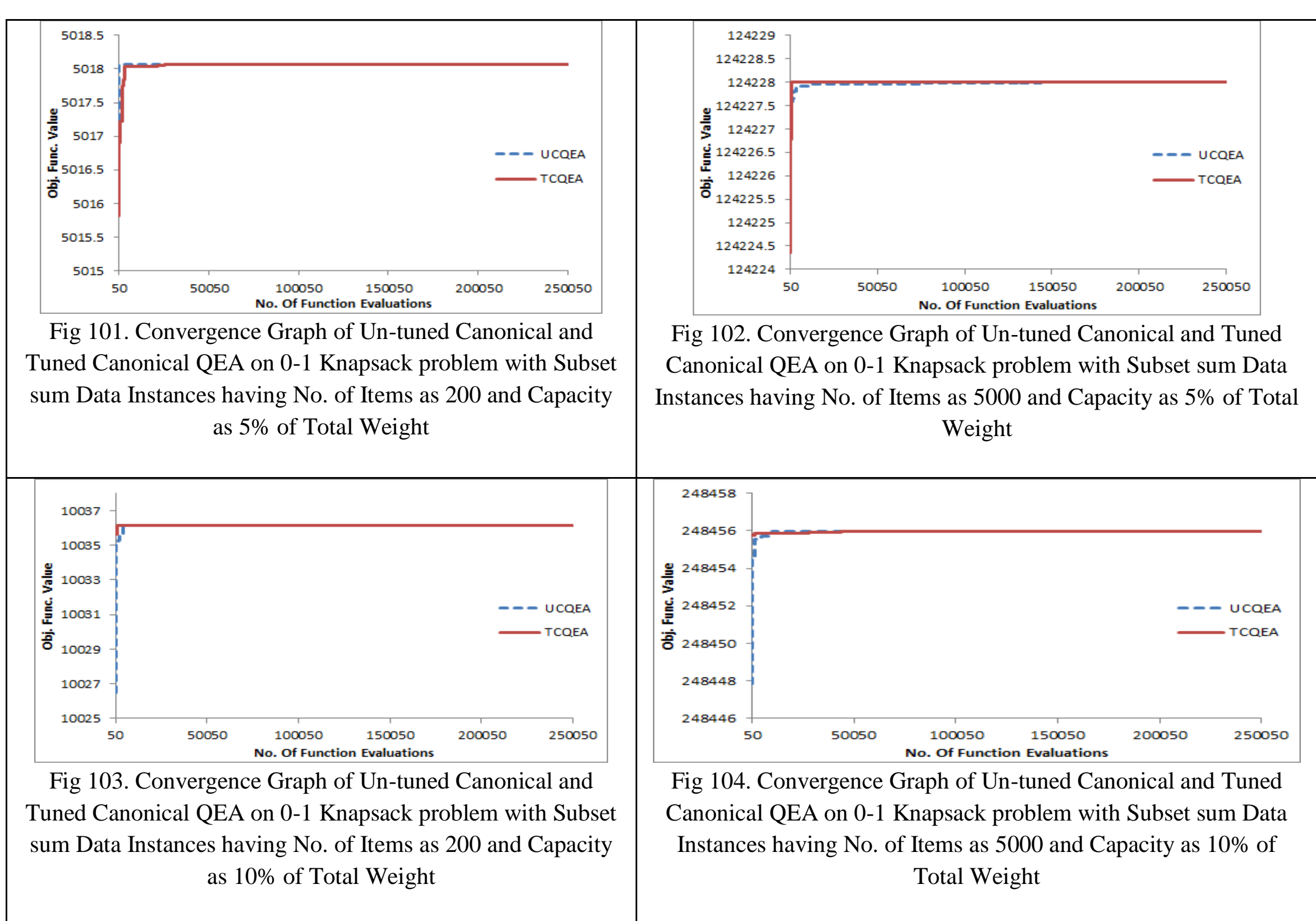

Fig 101. Convergence Graph of Un-tuned Canonical and Tuned Canonical QEA on 0-1 Knapsack problem with Subset sum Data Instances having No. of Items as 200 and Capacity as 5% of Total Weight

Fig 102. Convergence Graph of Un-tuned Canonical and Tuned Canonical QEA on 0-1 Knapsack problem with Subset sum Data Instances having No. of Items as 5000 and Capacity as 5% of Total Weight

Fig 103. Convergence Graph of Un-tuned Canonical and Tuned Canonical QEA on 0-1 Knapsack problem with Subset sum Data Instances having No. of Items as 200 and Capacity as 10% of Total Weight

Fig 104. Convergence Graph of Un-tuned Canonical and Tuned Canonical QEA on 0-1 Knapsack problem with Subset sum Data Instances having No. of Items as 5000 and Capacity as 10% of Total Weight

In order to confirm the findings in Table 58 to 64, multi-problem non-parametric Wilcoxon's Signed Rank Test [Der2011] was performed on average objective function value (OFV) and average number of generation (Av. Gen.) of Tuned QEA and Canonical QEA on all the instances of Subset Sum Knapsack problem at a significance level of 5%. In case of comparison on average OFV, the null hypothesis for comparison was average OFV of Tuned QEA $\mu_1$ is less than or equal to the average OFV of Canonical QEA $\mu_2$ i.e. $H_0$: $\mu_1 \leq \mu_2$ and the alternate hypothesis is H1: $\mu_1 > \mu_2$. The result of test is presented in Table 65, which shows that only two samples have different values so no conclusion regarding superiority of tuning method can be drawn on the basis of the available data. A comparative study on average number of generations is required to determine the relative performance of TCQEA and UCQEA.

**Table 65: Wilcoxon's Signed Rank Test on Subset Sum KP Instances (OFV)**

| | TCQEA | UCQEA |
|---|---|---|
| | $X_1$ | $X_2$ |
| 1 | 47 | 47 |
| 2 | 239 | 239 |
| 3 | 478.3 | 478.3 |
| 4 | 956.6 | 956.6 |
| 5 | 1913.3 | 1913.3 |
| 6 | 2391.6 | 2391.6 |
| 7 | 100.3 | 100.3 |
| 8 | 501.7 | 501.7 |
| 9 | 1003.6 | 1003.6 |
| 10 | 2007.2 | 2007.2 |
| 11 | 4014.4 | 4014.5 |
| 12 | 5018.1 | 5018.1 |
| 13 | 244.5 | 244.5 |
| 14 | 1223.1 | 1223.1 |
| 15 | 2446.3 | 2446.3 |

| | |
|---|---|
| Σ(+) | 0 |
| Σ(−) | 3 |
| *n* | 2 |

| Null Hypothesis | **Test Statistic** *T* | **Critical Value** At an α of 5% |
|---|---|---|
| $H_0$: $\mu_1 = \mu_2$ | 0 | 279 |
| $H_0$: $\mu_1 >= \mu_2$ | 0 | 303 |
| $H_0$: $\mu_1 <= \mu_2$ | 3 | 303 |
| 16 | 4892.6 | 4892.6 |
| 17 | 9785.2 | 9785.2 |
| 18 | 12231.5 | 12231.5 |
| 19 | 491.3 | 491.3 |
| 20 | 2456.7 | 2456.7 |
| 21 | 4913.4 | 4913.4 |
| 22 | 9826.8 | 9826.8 |
| 23 | 19653.6 | 19653.6 |
| 24 | 24567 | 24567 |
| 25 | 993.6 | 993.6 |
| 26 | 4968.1 | 4968.1 |
| 27 | 9936.1 | 9936.1 |
| 28 | 19872.2 | 19872.3 |
| 29 | 39744.5 | 39744.5 |
| 30 | 49680.7 | 49680.7 |
| 31 | 2484.6 | 2484.6 |
| 32 | 12422.8 | 12422.8 |
| 33 | 24845.6 | 24845.6 |
| 34 | 49691.2 | 49691.2 |
| 35 | 99382.4 | 99382.4 |
| 36 | 124228 | 124228 |
| 37 | 4990.6 | 4990.6 |
| 38 | 24953 | 24953 |
| 39 | 49906 | 49906 |
| 40 | 99811.9 | 99811.9 |
| 41 | 199624 | 199624 |
| 42 | 249530 | 249530 |

In case of comparison on Av. Gen., the null hypothesis for comparison was Av. Gen. of Tuned QEA $\mu_1$ is greater than or equal to the Av. Gen. of Canonical QEA $\mu_2$ i.e. $H_0$: $\mu_1 \geq \mu_2$ and the alternate hypothesis is H1: $\mu_1 < \mu_2$. The result of test is presented in Table 66, which shows that null hypothesis cannot be rejected as Wilcoxon's Signed Rank test statistic is 475.5 and greater than the critical value of 319 at significance level of $\alpha = 5\%$, which indicates that Tuned QEA requires more number of generations as compared to Canonical QEA. The sample size n= 42 (distinct) is larger than 25 so z statistic has also been computed under the normal distribution and p value obtained is 0.6179, which is more than 0.05, so null hypothesis cannot be rejected. Further, if we consider the null hypothesis as $H_0$: $\mu_1 \leq \mu_2$ and the alternate hypothesis as H1: $\mu_1 > \mu_2$, Table 66 shows that null hypothesis cannot be rejected as Wilcoxon's Signed Rank test statistic is 427.5 and greater than the critical value of 319 at significance level of $\alpha = 5\%$, so no conclusion can be drawn on the success of the proposed tuning method for tuning QEA on problems like Subset Sum Knapsack instance.

**Table 66: Wilcoxon's Signed Rank Test on Subset Sum KP Instances (Av. Gen.)**

| | TCQEA | UCQEA |
|---|---|---|
| | **$X_1$** | **$X_2$** |
| 1 | 2.3 | 2.5 |
| 2 | 82 | 101.3 |
| 3 | 380.9 | 299.6 |
| 4 | 360.3 | 344.2 |
| 5 | 491.6 | 418.9 |
| 6 | 406.4 | 418.6 |
| 7 | 261.5 | 251.7 |
| 8 | 301.4 | 237 |

| | |
|---|---|
| $\Sigma(+)$ | 475.5 |
| $\Sigma(-)$ | 427.5 |
| *n* | 42 |

| Null Hypothesis | **Test Statistic** *T* | **Critical Value** At an $\alpha$ of 5% |
|---|---|---|
| $H_0$: $\mu_1$ >= $\mu_2$ | 475.5 | 319 |
| $H_0$: $\mu_1$ <= $\mu_2$ | 427.5 | 319 |

**For Large Samples (*n* > 25)**

Test Statistic

| | | |
|---|---|---|
| 9 | 478.2 | 484.8 |
| 10 | 421.2 | 376.9 |
| 11 | 434.8 | 489.6 |
| 12 | 423.1 | 459.2 |
| 13 | 406 | 383.2 |
| 14 | 449.3 | 506.8 |
| 15 | 517.7 | 530.2 |
| 16 | 482 | 519.1 |
| 17 | 476.8 | 587.3 |
| 18 | 475.7 | 476.3 |
| 19 | 468.6 | 475.4 |
| 20 | 540.6 | 429.5 |
| 21 | 489.6 | 440.4 |
| 22 | 498.8 | 472.3 |
| 23 | 493.5 | 536.9 |
| 24 | 525.6 | 539.4 |
| 25 | 530.6 | 492 |
| 26 | 490.6 | 490.3 |
| 27 | 515.4 | 509.1 |
| 28 | 493.4 | 569.4 |
| 29 | 488.7 | 614.5 |
| 30 | 537.3 | 430.8 |
| 31 | 537.4 | 413.3 |
| 32 | 538.1 | 507 |
| 33 | 528.1 | 542.1 |
| 34 | 523.4 | 528.9 |
| 35 | 481 | 427 |
| 36 | 523.7 | 434 |
| 37 | 482.6 | 500.3 |
| 38 | 423.8 | 548.6 |
| 39 | 508.3 | 405.5 |
| 40 | 436.6 | 505.2 |
| 41 | 475.2 | 534.2 |
| 42 | 531.5 | 517.5 |

| | | | |
|---|---|---|---|
| *Z* | -0.30009 | *E[T]* | 451.5 |
| | | *σ(T)* | 79.97655907 |

At an α of

| Null Hypothesis | *p*-value | 5% |
|---|---|---|
| $H_0$: $\mu_1 >= \mu_2$ | 0.6179 | |
| $H_0$: $\mu_1 <= \mu_2$ | 0.3821 | |

### 7) *Uncorrelated instances with similar weights:*

The result of comparative study between Tuned QEA (TCQEA) and Canonical QEA (UCQEA) are given in Table 67 to 73. The performance of TCQEA and UCQEA are similar when the no. of items to choose is 100 for different capacity of knapsack. In fact, when the capacity of knapsack was 0.1% of the total capacity and number of items 100 / 200 / 500, both the algorithm could not find even a single item for the knapsack. The performance of TCQEA and UCQEA are also similar when the capacity of knapsack is 0.1 % of the total capacity. However, with the increase in number of items and the capacity of knapsack, the performance of TCQEA has improved over UCQEA in rest of the instances. The performance of TCQEA was also good on speed of convergence as indicated by average generations and the convergence graphs shown in figures 105 to 110, which also compared the speed of convergence of TCQEA to UCQEA.

TABLE 67

COMPARATIVE RESULTS FOR 0-1 KNAPSACK PROBLEM WITH NO. OF ITEMS 100 ON UNCORRELATED INSTANCES WITH SIMILAR WEIGHTS DATA INSTANCES

| % of Total Weight | Canonical QEA | | | | | | Tuned-QEA | | | | | |
|---|---|---|---|---|---|---|---|---|---|---|---|---|
| | Best | Worst | Average | Median | Std | Av. Gen | Best | Worst | Average | Median | Std | Av. Gen |

| 1% | **981** | **981** | **981** | **981** | **0** | **88** | **981** | **981** | **981** | **981** | **0** | 109 |
|---|---|---|---|---|---|---|---|---|---|---|---|---|
| 5% | **4818** | **4818** | **4818** | **4818** | **0** | 43063 | **4818** | **4818** | **4818** | **4818** | **0** | **39778** |
| 10% | **9522** | **9489** | **9516** | **9513** | **7** | **35565** | **9522** | **9513** | **9520** | **9522** | **4** | 37445 |
| 20% | **18194** | **18194** | **18194** | **18194** | **0** | **31225** | **18194** | **18194** | **18194** | **18194** | **0** | 37947 |
| 50% | 37189 | 37134 | 37178 | 37188 | **17** | **36655** | 37189 | 37134 | 37180 | 37188 | **16** | 48061 |

TABLE 68

COMPARATIVE RESULTS FOR 0-1 KNAPSACK PROBLEM WITH NO. OF ITEMS 200 ON UNCORRELATED INSTANCES WITH SIMILAR WEIGHTS DATA INSTANCES

| **% of Total Weight** | **Canonical QEA** | | | | | | **Tuned-QEA** | | | | | |
|---|---|---|---|---|---|---|---|---|---|---|---|---|
| | **Best** | **Worst** | **Average** | **Median** | **Std** | **Av. Gen** | **Best** | **Worst** | **Average** | **Median** | **Std** | **Av. Gen** |
| 1% | **1965** | **1965** | **1965** | **1965** | **0** | **12505** | **1965** | **1965** | **1965** | **1965** | **0** | 12817 |
| 5% | **9633** | **9609** | **9624** | **9625** | 8 | 71378 | **9633** | **9617** | **9631** | **9633** | **5** | **62420** |
| 10% | **18937** | **18937** | **18937** | **18937** | **0** | **58918** | **18937** | **18937** | **18937** | **18937** | **0** | 63885 |
| 20% | **35749** | **35749** | **35749** | **35749** | **0** | **64773** | **35749** | **35749** | **35749** | **35749** | **0** | 80292 |
| 50% | 73674 | 73667 | 73671 | 73674 | 4 | **82448** | 73674 | 73667 | 73673 | 73674 | **3** | 107765 |

TABLE 69

COMPARATIVE RESULTS FOR 0-1 KNAPSACK PROBLEM WITH NO. OF ITEMS 500 ON UNCORRELATED INSTANCES WITH SIMILAR WEIGHTS DATA INSTANCES

| **% of Total Weight** | **Canonical QEA** | | | | | | **Tuned-QEA** | | | | | |
|---|---|---|---|---|---|---|---|---|---|---|---|---|
| | **Best** | **Worst** | **Average** | **Median** | **Std** | **Av. Gen** | **Best** | **Worst** | **Average** | **Median** | **Std** | **Av. Gen** |
| 1% | **4969** | **4969** | **4969** | **4969** | **0** | 204835 | **4969** | **4969** | **4969** | **4969** | **0** | **97584** |
| 5% | **24409** | **24395** | **24406** | **24407** | 4 | **128695** | **24409** | **24395** | **24406** | **24405** | **4** | 131086 |
| 10% | **47709** | **47709** | **47709** | **47709** | **0** | **138478** | **47709** | **47709** | **47709** | **47709** | **0** | 154176 |
| 20% | **90773** | **90773** | **90773** | **90773** | **0** | **199540** | **90773** | **90773** | **90773** | **90773** | **0** | 247939 |
| 50% | 188641 | 188327 | 188618 | 188637 | 60 | **318540** | 188641 | 188583 | 188632 | 188637 | **16** | 364317 |

TABLE 70

COMPARATIVE RESULTS FOR 0-1 KNAPSACK PROBLEM WITH NO. OF ITEMS 1000 ON UNCORRELATED INSTANCES WITH SIMILAR WEIGHTS DATA INSTANCES

| **% of Total Weight** | **Canonical QEA** | | | | | | **Tuned-QEA** | | | | | |
|---|---|---|---|---|---|---|---|---|---|---|---|---|
| | **Best** | **Worst** | **Average** | **Median** | **Std** | **Av. Gen** | **Best** | **Worst** | **Average** | **Median** | **Std** | **Av. Gen** |
| 1% | **9947** | **9947** | **9947** | **9947** | **0** | 248633 | **9947** | **9947** | **9947** | **9947** | **0** | **142751** |
| 5% | **48865** | **48854** | **48863** | **48865** | 3 | **302247** | **48865** | **48854** | **48863** | **48865** | **3** | 314985 |
| 10% | **95238** | **95238** | **95238** | **95238** | 0 | **346338** | **95238** | **95238** | **95238** | **95238** | **0** | 362713 |
| 20% | **181223** | **181175** | **181214** | **181217** | 10 | **440893** | **181223** | **181195** | **181212** | **181213** | **7** | 453466 |
| 50% | **378080** | **377926** | **378020** | 378022 | **37** | **471998** | 378067 | 377917 | 377981 | **377979** | 37 | 488525 |

TABLE 71

COMPARATIVE RESULTS FOR 0-1 KNAPSACK PROBLEM WITH NO. OF ITEMS 2000 ON UNCORRELATED INSTANCES WITH SIMILAR WEIGHTS DATA INSTANCES

| % of Total Weight | Canonical QEA | | | | | | Tuned-QEA | | | | | |
|---|---|---|---|---|---|---|---|---|---|---|---|---|
| | **Best** | **Worst** | **Average** | **Median** | **Std** | **Av. Gen** | **Best** | **Worst** | **Average** | **Median** | **Std** | **Av. Gen** |
| 1% | **19862** | **19862** | **19862** | **19862** | **0** | 340618 | **19862** | **19862** | **19862** | **19862** | **0** | **211279** |
| 5% | **97179** | **97172** | **97177** | **97178** | **2** | **464325** | **97179** | **97171** | **97177** | **97177** | 2 | 471237 |
| 10% | 189378 | 189298 | 189344 | 189344 | 18 | **491100** | **189378** | **189271** | **189337** | **189335** | **23** | 495175 |
| 20% | 359084 | 358628 | 358921 | 358940 | 101 | 497517 | **359088** | **358818** | **358961** | **358950** | **67** | **497142** |
| 50% | **747072** | 746030 | 746617 | 746627 | 216 | **498212** | 746885 | **746190** | **746559** | **746538** | **191** | 497878 |

TABLE 72

COMPARATIVE RESULTS FOR 0-1 KNAPSACK PROBLEM WITH NO. OF ITEMS 5000 ON UNCORRELATED INSTANCES WITH SIMILAR WEIGHTS DATA INSTANCES

| % of Total Weight | Canonical QEA | | | | | | Tuned-QEA | | | | | |
|---|---|---|---|---|---|---|---|---|---|---|---|---|
| | **Best** | **Worst** | **Average** | **Median** | **Std** | **Av. Gen** | **Best** | **Worst** | **Average** | **Median** | **Std** | **Av. Gen** |
| 1% | 49664 | 49585 | 49638 | 49639 | 16 | 496283 | **49685** | **49680** | **49683** | **49683** | **1** | **476035** |
| 5% | 241788 | 240832 | 241409 | 241433 | 225 | **498203** | **242785** | **242410** | **242580** | **242564** | 86 | **498303** |
| 10% | 468068 | 465385 | 466727 | 466778 | **574** | 498837 | **471688** | **469442** | **470599** | **470621** | 496 | **498683** |
| 20% | 881986 | 876391 | 879207 | 879106 | **1283** | **498575** | **891382** | **883854** | **887490** | **888015** | 1817 | 499175 |
| 50% | 1840683 | 1830433 | 1835985 | 1836343 | **2907** | 498608 | **1844985** | **1829575** | **1839329** | **1839287** | 3213 | **499326** |

TABLE 73

COMPARATIVE RESULTS FOR 0-1 KNAPSACK PROBLEM WITH NO. OF ITEMS 10000 ON UNCORRELATED INSTANCES WITH SIMILAR WEIGHTS DATA INSTANCES

| % of Total Weight | Canonical QEA | | | | | | Tuned-QEA | | | | | |
|---|---|---|---|---|---|---|---|---|---|---|---|---|
| | **Best** | **Worst** | **Average** | **Median** | **Std** | **Av. Gen** | **Best** | **Worst** | **Average** | **Median** | **Std** | **Av. Gen** |
| 1% | 98471 | 97677 | 98087 | 98068 | **161** | 498178 | **99320** | **99206** | **99270** | **99268** | 30 | **498142** |
| 5% | 470250 | 466888 | 468767 | 468808 | **702** | **498295** | **478616** | **476205** | **477551** | **477732** | 708 | 499330 |
| 10% | 901924 | 894308 | 898408 | 898562 | 2418 | **499050** | **923533** | **909934** | **915208** | **914686** | **2848** | 499076 |
| 20% | 1691270 | 1666278 | 1675694 | 1674899 | **6106** | **499297** | **1714164** | **1668640** | **1699855** | **1700715** | 9838 | 499584 |
| 50% | 3554710 | 3500113 | 3530838 | 3530308 | **11773** | **499440** | **3558266** | **3507462** | **3533864** | **3534393** | 13062 | 499620 |

The convergence graphs have been plotted between objective function value and number of generations for both TCQEA and UCQEA for all the problem instances having No. of Items as 200 and 5000 for the median run except for 0.1% capacity (No. of items is 1000 instead of 200). The convergence graph of TCQEA & UCQEA for all the problem instances having No. of Items as 200 is almost similar, however, problem instances having No. of Items as 5000 establishes the superiority of Tuning as the TCQEA is faster than UCQEA in all the graphs. The difference in performance between TCQEA and UCQEA increases with the capacity size and number of items in the knapsack problem.

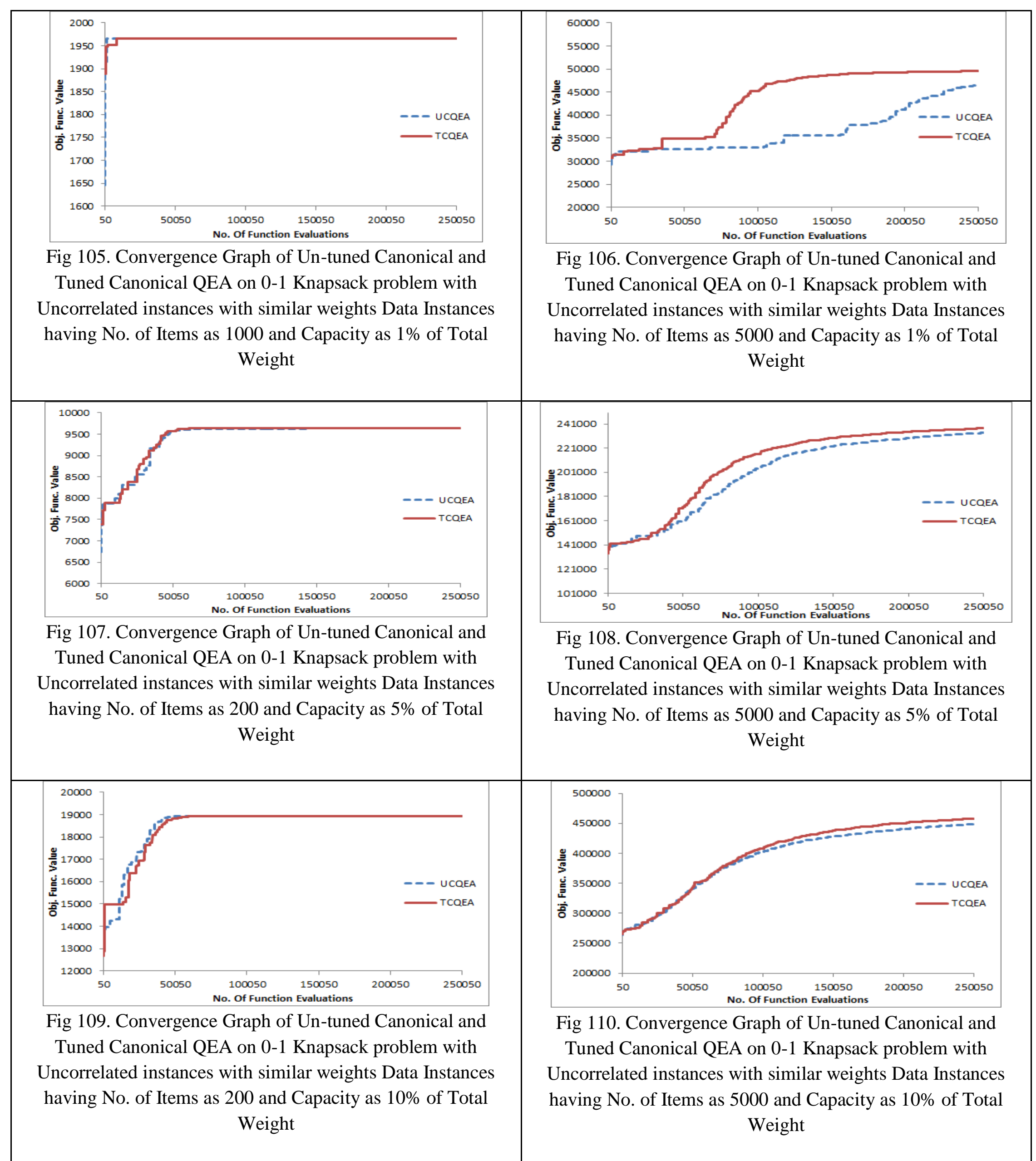

Fig 105. Convergence Graph of Un-tuned Canonical and Tuned Canonical QEA on 0-1 Knapsack problem with Uncorrelated instances with similar weights Data Instances having No. of Items as 1000 and Capacity as 1% of Total Weight

Fig 106. Convergence Graph of Un-tuned Canonical and Tuned Canonical QEA on 0-1 Knapsack problem with Uncorrelated instances with similar weights Data Instances having No. of Items as 5000 and Capacity as 1% of Total Weight

Fig 107. Convergence Graph of Un-tuned Canonical and Tuned Canonical QEA on 0-1 Knapsack problem with Uncorrelated instances with similar weights Data Instances having No. of Items as 200 and Capacity as 5% of Total Weight

Fig 108. Convergence Graph of Un-tuned Canonical and Tuned Canonical QEA on 0-1 Knapsack problem with Uncorrelated instances with similar weights Data Instances having No. of Items as 5000 and Capacity as 5% of Total Weight

Fig 109. Convergence Graph of Un-tuned Canonical and Tuned Canonical QEA on 0-1 Knapsack problem with Uncorrelated instances with similar weights Data Instances having No. of Items as 200 and Capacity as 10% of Total Weight

Fig 110. Convergence Graph of Un-tuned Canonical and Tuned Canonical QEA on 0-1 Knapsack problem with Uncorrelated instances with similar weights Data Instances having No. of Items as 5000 and Capacity as 10% of Total Weight

In order to confirm the findings in Table 67 to 73, multi-problem non-parametric Wilcoxon's Signed Rank Test [Der2011] was performed on average objective function value (OFV) and average number of generation (Av. Gen.) of Tuned QEA and Canonical QEA on all the instances of Uncorrelated Instances with Similar Weights Data Knapsack problem at a significance level of 5%. In case of comparison on average OFV, the null hypothesis for comparison was average OFV of Tuned QEA $\mu_1$ is less than or equal to the average OFV of Canonical QEA $\mu_2$ i.e. $H_0$: $\mu_1 \leq \mu_2$ and the alternate hypothesis is H1: $\mu_1 > \mu_2$. The result of test is presented in Table 74, which shows that null hypothesis can be rejected safely as Wilcoxon's Signed Rank test statistic is zero and less than the critical value of 175 at significance level of $\alpha = 5\%$. The sample size n= 32 (distinct) is larger than 25 so z statistic has also been computed under the normal distribution and p value obtained is 0.000, which is less than 0.05, so null hypothesis can be safely rejected. This indicates the success of the proposed tuning method for tuning QEA on problems like Uncorrelated Instances with Similar Weights Data Knapsack problem.

**Table 74: Wilcoxon's Signed Rank Test on UnCorr. Instances Sim. Wt. data (OFV)**

| | TCQEA | UCQEA |
|---|---|---|

| | $X_1$ | $X_2$ |
|---|---|---|
| 1 | 0 | 0 |
| 2 | 0 | 0 |
| 3 | 980.5 | 980.5 |
| 4 | 1948.2 | 1948.2 |
| 5 | 3854.6 | 3854.6 |
| 6 | 4817.8 | 4817 |
| 7 | 0 | 0 |
| 8 | 984.4 | 984.4 |
| 9 | 1964.9 | 1964.9 |
| 10 | 3894.1 | 3813.6 |
| 11 | 7728.5 | 7599.7 |
| 12 | 9626.1 | 9543 |
| 13 | 0 | 0 |
| 14 | 1994.6 | 1993.6 |
| 15 | 4969.1 | 4714.2 |
| 16 | 9879.4 | 8650.4 |
| 17 | 19592.4 | 17536 |
| 18 | 24372.4 | 22289.3 |
| 19 | 996.2 | 996.2 |
| 20 | 4981.8 | 4716.3 |
| 21 | 9944 | 8500.9 |
| 22 | 19787 | 15549.8 |
| 23 | 38954.3 | 31794.2 |
| 24 | 48122.5 | 40745.8 |
| 25 | 1997.5 | 1989.4 |
| 26 | 9955.4 | 8457.2 |
| 27 | 19803.1 | 15096.8 |
| 28 | 38955.9 | 27893.6 |
| 29 | 74916.3 | 56840.3 |
| 30 | 91815.6 | 73367.4 |
| 31 | 4993.5 | 4694.7 |
| 32 | 24659.7 | 18321.1 |
| 33 | 48255.7 | 33589.4 |
| 34 | 92342.6 | 62912.3 |
| 35 | 171550 | 128051 |
| 36 | 207397 | 163230 |
| 37 | 9950.6 | 8467.4 |
| 38 | 47695.9 | 33495.7 |
| 39 | 91580.6 | 62606.4 |
| 40 | 171197 | 119073 |
| 41 | 310775 | 241830 |
| 42 | 376210 | 305588 |

| | |
|---|---|
| Σ(+) | 528 |
| Σ(−) | 0 |
| *n* | 32 |

| Null Hypothesis | Test Statistic *T* | Critical Value At an α of 5% |
|---|---|---|
| $H_0$: $\mu_1 <= \mu_2$ | 0 | 175 |

**For Large Samples (*n* > 25)**

Test Statistic

| | | | |
|---|---|---|---|
| *Z* | -4.93652 | *E[T]* | 264 |
| | | *σ(T)* | 53.47896783 |

| Null Hypothesis | *p*-value | At an α of 5% |
|---|---|---|
| $H_0$: $\mu_1 <= \mu_2$ | 0.0000 | **Reject** |

In case of comparison on Av. Gen., the null hypothesis for comparison was Av. Gen. of Tuned QEA $\mu_1$ is greater than or equal to the Av. Gen. of Canonical QEA $\mu_2$ i.e. $H_0$: $\mu_1 \geq \mu_2$ and the alternate hypothesis is H1: $\mu_1 < \mu_2$. The result of test is presented in Table 75, which shows that null hypothesis cannot be rejected as Wilcoxon's Signed Rank test statistic is 425.5 and more than the critical value of 256 at significance level of $\alpha = 5\%$, which indicates that Tuned QEA requires more number of generations as compared to Canonical QEA. The sample size n= 38 (distinct) is larger than 25 so z statistic has also been computed under the normal distribution and p value obtained is 0.7875, which is more than 0.05, so null hypothesis cannot be rejected. However, Tuned QEA is performing significantly better than Canonical QEA as shown by Table. 74. This indicates the success of the proposed tuning method for tuning QEA on problems like Uncorrelated Instances with Similar Weights Data even though it requires more number of generations and function evaluations as compared to Canonical QEA.

**Table 75: Wilcoxon's Signed Rank Test on Uncorr. Instances Sim. Wt. (Av. Gen.)**

| | TCQEA | UCQEA |
|---|---|---|
| | $X_1$ | $X_2$ |
| 1 | 0 | 0 |
| 2 | 0 | 0 |
| 3 | 2 | 1.8 |
| 4 | 35.2 | 67.1 |
| 5 | 124 | 802.8 |
| 6 | 158.2 | 851.9 |
| 7 | 0 | 0 |
| 8 | 3.6 | 3.8 |
| 9 | 56.3 | 250.1 |
| 10 | 181.4 | 693.1 |
| 11 | 397 | 961.3 |
| 12 | 441.9 | 974.2 |
| 13 | 0 | 0 |
| 14 | 148.8 | 594.7 |
| 15 | 462.6 | 572.6 |
| 16 | 721.2 | 572.2 |
| 17 | 841.7 | 969.7 |
| 18 | 930.4 | 983.4 |
| 19 | 16.5 | 17.9 |
| 20 | 587.1 | 591.4 |
| 21 | 797.2 | 493.7 |
| 22 | 957.6 | 636.8 |
| 23 | 984.4 | 962.9 |
| 24 | 985.7 | 984.9 |
| 25 | 372.4 | 555.5 |
| 26 | 871.6 | 528.4 |
| 27 | 976.3 | 622.5 |
| 28 | 982.4 | 595.6 |
| 29 | 991.3 | 961.7 |
| 30 | 990.1 | 976.2 |
| 31 | 804.9 | 479 |
| 32 | 974.3 | 525.9 |
| 33 | 985.9 | 554.9 |
| 34 | 988.4 | 594.7 |
| 35 | 990.1 | 972.1 |
| 36 | 992.8 | 980.8 |
| 37 | 925.5 | 426.7 |
| 38 | 981.7 | 510.6 |
| 39 | 986.9 | 607.4 |
| 40 | 988.5 | 576.7 |
| 41 | 990.7 | 966.5 |
| 42 | 990.2 | 977.4 |

| | |
|---|---|
| Σ(+) | 425.5 |
| Σ(–) | 315.5 |
| *n* | 38 |

| Null Hypothesis | **Test Statistic** *T* | **Critical Value** At an α of 5% |
|---|---|---|
| $H_0: \mu_1 >= \mu_2$ | 425.5 | 256 |

**For Large Samples (*n* > 25)**

Test Statistic

| | | | |
|---|---|---|---|
| *Z* | -0.79763 | *E[T]* | 370.5 |
| | | *σ(T)* | 68.95469527 |

| Null Hypothesis | *p*-value | At an α of 5% |
|---|---|---|
| $H_0: \mu_1 >= \mu_2$ | 0.7875 | |

### *8) Spanner instances span (v, m):*

The result of comparative study between Tuned QEA (TCQEA) and Canonical QEA (UCQEA) are given in Table 76 to 82. The performance of TCQEA and UCQEA are similar when the no. of items to choose is 100 and 200 for different capacity of knapsack. In fact, when the capacity of knapsack was 0.1% of the total capacity and number of items 100, both the algorithm could not find even a single item for the knapsack. The performance of TCQEA and UCQEA are also similar when the capacity of knapsack is 0.1 % of the total capacity. However, with the increase in number of items and the capacity of knapsack, the performance of TCQEA has improved over UCQEA in rest of the instances. The performance of TCQEA was also good on speed of convergence as indicated by average generations and the convergence graphs shown in Fig. 111 to 116, which also compared the speed of convergence of TCQEA to UCQEA.

TABLE 76

COMPARATIVE RESULTS FOR 0-1 KNAPSACK PROBLEM WITH NO. OF ITEMS 100 ON SPANNER INSTANCES

| % of Total Weight | Canonical QEA | | | | | | Tuned-QEA | | | | | |
|---|---|---|---|---|---|---|---|---|---|---|---|---|
| | Best | Worst | Average | Median | Std | Av. Gen | Best | Worst | Average | Median | Std | Av. Gen |
| 1% | 167 | 167 | 167 | 167 | 0 | 103545 | 167 | 167 | 167 | 167 | 0 | 9092 |
| 5% | 834 | 834 | 834 | 834 | 0 | 121172 | 834 | 834 | 834 | 834 | 0 | 64000 |
| 10% | 1669 | 1669 | 1669 | 1669 | 0 | 70615 | 1669 | 1669 | 1669 | 1669 | 0 | 50513 |
| 20% | 3337 | 3337 | 3337 | 3337 | 0 | 71262 | 3337 | 3337 | 3337 | 3337 | 0 | 98287 |
| 50% | 7423 | 7423 | 7423 | 7423 | 0 | 71257 | 7423 | 7423 | 7423 | 7423 | 0 | 97004 |

TABLE 77

COMPARATIVE RESULTS FOR 0-1 KNAPSACK PROBLEM WITH NO. OF ITEMS 200 ON SPANNER INSTANCES

| % of Total Weight | Canonical QEA | | | | | | Tuned-QEA | | | | | |
|---|---|---|---|---|---|---|---|---|---|---|---|---|
| | Best | Worst | Average | Median | Std | Av. Gen | Best | Worst | Average | Median | Std | Av. Gen |
| 1% | 352 | 352 | 352 | 352 | 0 | 191482 | 352 | 352 | 352 | 352 | 0 | 73052 |
| 5% | 1761 | 1761 | 1761 | 1761 | 0 | 116910 | 1761 | 1761 | 1761 | 1761 | 0 | 60911 |
| 10% | 3522 | 3522 | 3522 | 3522 | 0 | 119887 | 3522 | 3522 | 3522 | 3522 | 0 | 165597 |
| 20% | 7044 | 7044 | 7044 | 7044 | 0 | 186107 | 7044 | 7044 | 7044 | 7044 | 0 | 193393 |
| 50% | 15452 | 15452 | 15452 | 15452 | 0 | 157733 | 15452 | 15452 | 15452 | 15452 | 0 | 174425 |

TABLE 78

COMPARATIVE RESULTS FOR 0-1 KNAPSACK PROBLEM WITH NO. OF ITEMS 500 ON SPANNER INSTANCES

| % of Total Weight | Canonical QEA | | | | | | Tuned-QEA | | | | | |
|---|---|---|---|---|---|---|---|---|---|---|---|---|
| | Best | Worst | Average | Median | Std | Av. Gen | Best | Worst | Average | Median | Std | Av. Gen |
| 1% | 853 | 853 | 853 | 853 | 0 | 266513 | 853 | 853 | 853 | 853 | 0 | 117064 |
| 5% | 4267 | 4267 | 4267 | 4267 | 0 | 212348 | 4267 | 4267 | 4267 | 4267 | 0 | 164314 |
| 10% | 8533 | 8533 | 8533 | 8533 | 0 | 226458 | 8533 | 8533 | 8533 | 8533 | 0 | 240451 |
| 20% | 17067 | 17067 | 17067 | 17067 | 0 | 211193 | 17067 | 17067 | 17067 | 17067 | 0 | 233881 |
| 50% | 37590 | 37583 | 37590 | 37590 | 2 | 295165 | 37590 | 37584 | 37590 | 37590 | 1 | 280137 |

TABLE 79

COMPARATIVE RESULTS FOR 0-1 KNAPSACK PROBLEM WITH NO. OF ITEMS 1000 ON SPANNER INSTANCES

| % of Total Weight | Canonical QEA | | | | | | Tuned-QEA | | | | | |
|---|---|---|---|---|---|---|---|---|---|---|---|---|
| | Best | Worst | Average | Median | Std | Av. Gen | Best | Worst | Average | Median | Std | Av. Gen |
| 1% | **1707** | **1707** | **1707** | **1707** | **0** | 303383 | **1707** | **1707** | **1707** | **1707** | **0** | **153140** |
| 5% | **8537** | **8537** | **8537** | **8537** | **0** | 263253 | **8537** | **8537** | **8537** | **8537** | **0** | **256153** |
| 10% | **17073** | **17073** | **17073** | **17073** | **0** | **225295** | **17073** | **17073** | **17073** | **17073** | **0** | 257341 |
| 20% | **34147** | **34147** | **34147** | **34147** | **0** | **264945** | **34147** | **34147** | **34147** | **34147** | **0** | 299297 |
| 50% | **75217** | **75196** | **75213** | **75217** | **6** | 353255 | **75217** | 75187 | 75209 | 75211 | 9 | **348460** |

TABLE 80
COMPARATIVE RESULTS FOR 0-1 KNAPSACK PROBLEM WITH NO. OF ITEMS 2000 ON SPANNER INSTANCES

| % of Total Weight | Canonical QEA | | | | | | Tuned-QEA | | | | | |
|---|---|---|---|---|---|---|---|---|---|---|---|---|
| | Best | Worst | Average | Median | Std | Av. Gen | Best | Worst | Average | Median | Std | Av. Gen |
| 1% | **3448** | **3448** | **3448** | **3448** | **0** | 315240 | **3448** | **3448** | **3448** | **3448** | **0** | **237059** |
| 5% | **17241** | **17241** | **17241** | **17241** | **0** | 295948 | **17241** | **17241** | **17241** | **17241** | **0** | **243194** |
| 10% | **34481** | **34481** | **34481** | **34481** | **0** | **248563** | **34481** | **34481** | **34481** | **34481** | **0** | 294050 |
| 20% | **68962** | **68962** | **68962** | **68962** | **0** | **358537** | **68962** | **68962** | **68962** | **68962** | **0** | 364568 |
| 50% | **151820** | 151765 | 151799 | **151801** | 16 | 445622 | **151820** | **151780** | **151802** | **151801** | **10** | **429409** |

TABLE 81
COMPARATIVE RESULTS FOR 0-1 KNAPSACK PROBLEM WITH NO. OF ITEMS 5000 ON SPANNER INSTANCES

| % of Total Weight | Canonical QEA | | | | | | Tuned-QEA | | | | | |
|---|---|---|---|---|---|---|---|---|---|---|---|---|
| | Best | Worst | Average | Median | Std | Av. Gen | Best | Worst | Average | Median | Std | Av. Gen |
| 1% | **8668** | **8668** | **8668** | **8668** | **0** | 339485 | **8668** | **8668** | **8668** | **8668** | **0** | **262291** |
| 5% | **43341** | **43341** | **43341** | **43341** | **0** | 322480 | **43341** | **43341** | **43341** | **43341** | **0** | **321555** |
| 10% | **86682** | **86682** | **86682** | **86682** | **0** | 363507 | **86682** | **86682** | **86682** | **86682** | **0** | **354866** |
| 20% | 173216 | 172939 | 173103 | 173100 | 69 | 498175 | **173364** | **173207** | **173329** | **173334** | **34** | **495439** |
| 50% | 379949 | 379352 | 379635 | 379650 | 137 | 497555 | **380099** | **379847** | **379956** | **379946** | **79** | **497330** |

TABLE 82
COMPARATIVE RESULTS FOR 0-1 KNAPSACK PROBLEM WITH NO. OF ITEMS 10000 ON SPANNER INSTANCES

| % of Total Weight | Canonical QEA | | | | | | Tuned-QEA | | | | | |
|---|---|---|---|---|---|---|---|---|---|---|---|---|
| | Best | Worst | Average | Median | Std | Av. Gen | Best | Worst | Average | Median | Std | Av. Gen |
| 1% | **17439** | **17439** | **17439** | **17439** | **0** | 371338 | **17439** | **17439** | **17439** | **17439** | **0** | **261439** |
| 5% | **87193** | **87193** | **87193** | **87193** | **0** | 390920 | **87193** | **87193** | **87193** | **87193** | **0** | **341263** |

| | | | | | | | | | | | |
|---|---|---|---|---|---|---|---|---|---|---|---|
| 10% | **174387** | **174259** | **174338** | **174348** | 36 | 494885 | **174387** | **174387** | **174387** | **174387** | **0** | **441052** |
| 20% | 342556 | 339872 | 341204 | 341217 | **734** | **499162** | **346311** | **339970** | **344000** | **344132** | 1367 | 499481 |
| 50% | 759894 | 757444 | 758770 | 758725 | **521** | **498588** | **761392** | **758344** | **760243** | **760362** | 728 | 499211 |

The convergence graphs have been plotted between objective function value and number of generations for both TCQEA and UCQEA for all the problem instances having No. of Items as 200 and 5000 for the median run. The convergence graph of TCQEA & UCQEA for all the problem instances having No. of Items as 200 is almost similar with TCQEA minutely outperforming UCQEA, however, problem instances having No. of Items as 5000 establishes the superiority of Tuning as the TCQEA is faster than UCQEA in all the graphs. The difference in performance between TCQEA and UCQEA increases with the capacity size and number of items in the knapsack problem.

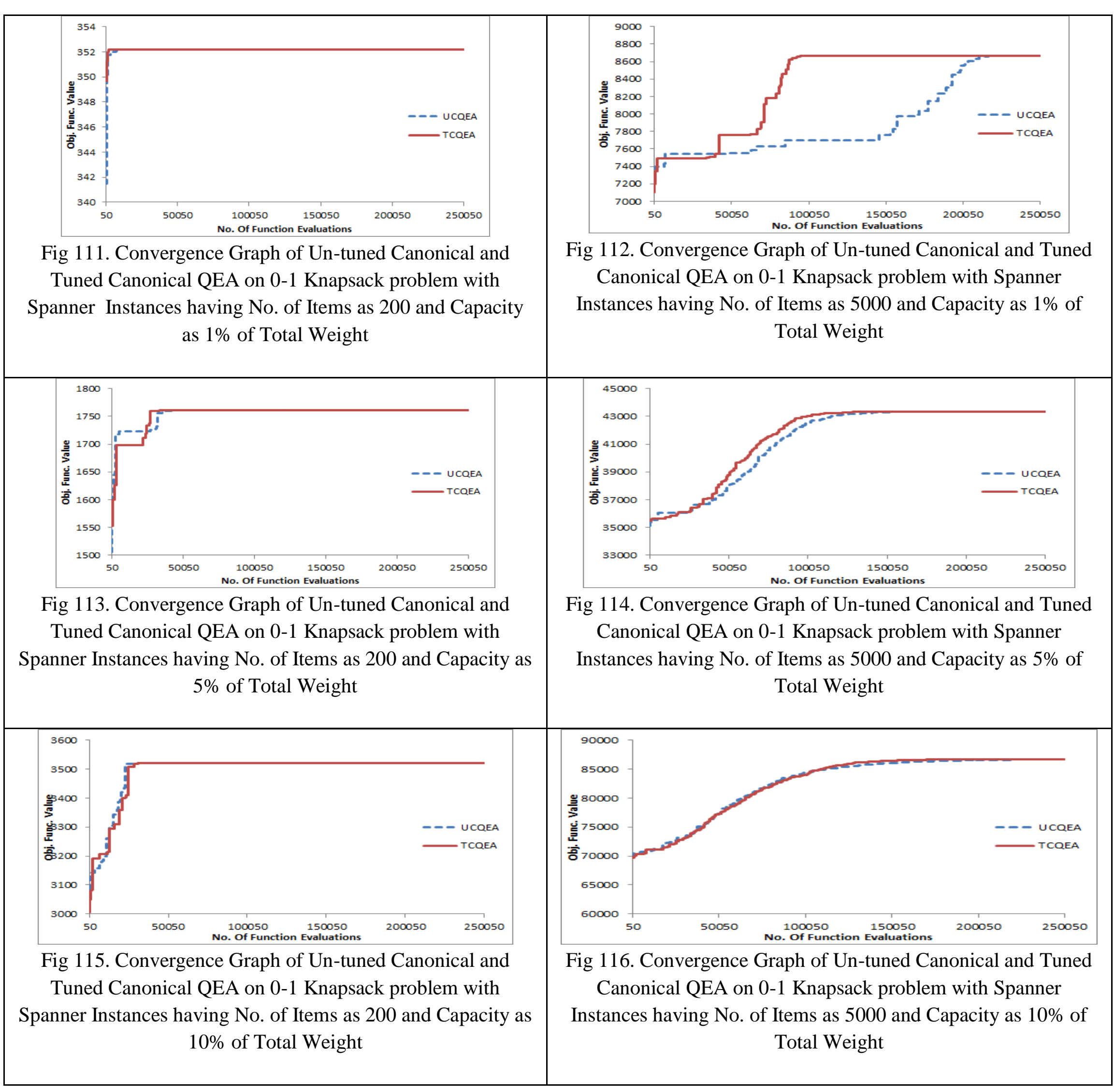

Fig 111. Convergence Graph of Un-tuned Canonical and Tuned Canonical QEA on 0-1 Knapsack problem with Spanner Instances having No. of Items as 200 and Capacity as 1% of Total Weight

Fig 112. Convergence Graph of Un-tuned Canonical and Tuned Canonical QEA on 0-1 Knapsack problem with Spanner Instances having No. of Items as 5000 and Capacity as 1% of Total Weight

Fig 113. Convergence Graph of Un-tuned Canonical and Tuned Canonical QEA on 0-1 Knapsack problem with Spanner Instances having No. of Items as 200 and Capacity as 5% of Total Weight

Fig 114. Convergence Graph of Un-tuned Canonical and Tuned Canonical QEA on 0-1 Knapsack problem with Spanner Instances having No. of Items as 5000 and Capacity as 5% of Total Weight

Fig 115. Convergence Graph of Un-tuned Canonical and Tuned Canonical QEA on 0-1 Knapsack problem with Spanner Instances having No. of Items as 200 and Capacity as 10% of Total Weight

Fig 116. Convergence Graph of Un-tuned Canonical and Tuned Canonical QEA on 0-1 Knapsack problem with Spanner Instances having No. of Items as 5000 and Capacity as 10% of Total Weight

In order to confirm the findings in Table 76 to 82, multi-problem non-parametric Wilcoxon's Signed Rank Test [Der2011] was performed on average objective function value (OFV) and average number of generation (Av. Gen.) of Tuned QEA and Canonical QEA on all the instances of Spanner Knapsack problem at a significance level of 5%. In case of comparison on average OFV, the null hypothesis for comparison was average OFV of Tuned QEA $\mu_1$ is less than or equal to the average OFV of Canonical QEA $\mu_2$ i.e. $H_0$: $\mu_1 \leq \mu_2$ and the alternate hypothesis is H1: $\mu_1 > \mu_2$. The result of test is presented in Table 83, which shows that null hypothesis can be rejected safely as Wilcoxon's Signed Rank test statistic is 0 and less than the critical value of 152 at significance level of $\alpha = 5\%$. The sample size

n= 30 (distinct) is larger than 25 so z statistic has also been computed under the normal distribution and p value obtained is 0.000, which is less than 0.05, so null hypothesis can be safely rejected. This indicates the success of the proposed tuning method for tuning QEA on problems like Spanner Instances Knapsack problem.

**Table 83: Wilcoxon's Signed Rank Test on Spanner Instances KP data (OFV)**

| | TCQEA | UCQEA |
|---|---|---|
| | $X_1$ | $X_2$ |
| 1 | 0 | 0 |
| 2 | 83.4 | 83.4 |
| 3 | 166.8 | 166.8 |
| 4 | 333.7 | 333.7 |
| 5 | 667.5 | 667.4 |
| 6 | 834.3 | 834.3 |
| 7 | 34.1 | 34.1 |
| 8 | 176.1 | 176.1 |
| 9 | 352.2 | 352.2 |
| 10 | 704.4 | 704.2 |
| 11 | 1408.8 | 1408.6 |
| 12 | 1761 | 1760.9 |
| 13 | 85.3 | 85.3 |
| 14 | 426.6 | 426.6 |
| 15 | 853.3 | 852.8 |
| 16 | 1706.6 | 1684 |
| 17 | 3413.3 | 3379.2 |
| 18 | 4266.6 | 4258.3 |
| 19 | 170.7 | 170.7 |
| 20 | 853.6 | 853.2 |
| 21 | 1707.3 | 1676 |
| 22 | 3414.6 | 3187.4 |
| 23 | 6829.3 | 6500.7 |
| 24 | 8536.6 | 8284.3 |
| 25 | 344.8 | 344.8 |
| 26 | 1724 | 1688.5 |
| 27 | 3448.1 | 3194.7 |
| 28 | 6896.2 | 6153.8 |
| 29 | 13792.4 | 12531.2 |
| 30 | 17240.5 | 15920.9 |
| 31 | 866.8 | 866.1 |
| 32 | 4334.1 | 3940.3 |
| 33 | 8668.1 | 7599.1 |
| 34 | 17336.3 | 14820.6 |
| 35 | 34476.2 | 29889.2 |
| 36 | 42549.3 | 37829.9 |
| 37 | 1743.8 | 1690.2 |
| 38 | 8719.2 | 7619.8 |
| 39 | 17438.1 | 14848.3 |
| 40 | 34632.1 | 29122 |
| 41 | 66178.4 | 58531 |
| 42 | 81052.5 | 73736.1 |

| | |
|---|---|
| Σ(+) | 465 |
| Σ(−) | 0 |
| *n* | 30 |

| Null Hypothesis | **Test Statistic** *T* | **Critical Value** At an α of 5% |
|---|---|---|
| $H_0$: $\mu_1$ <= $\mu_2$ | 0 | 152 |

**For Large Samples (*n* > 25)**

Test Statistic

| | | | |
|---|---|---|---|
| *Z* | -4.78214 | *E[T]* | 232.5 |
| | | *σ(T)* | 48.61841215 |

| Null Hypothesis | *p*-value | At an α of 5% |
|---|---|---|
| $H_0$: $\mu_1$ <= $\mu_2$ | 0.0000 | **Reject** |

In case of comparison on Av. Gen., the null hypothesis for comparison was Av. Gen. of Tuned QEA $\mu_1$ is greater than or equal to the Av. Gen. of Canonical QEA $\mu_2$ i.e. $H_0$: $\mu_1 \geq \mu_2$ and the alternate hypothesis is H1: $\mu_1 < \mu_2$. The result of test is presented in Table 84, which shows that null hypothesis cannot be rejected as Wilcoxon's Signed Rank test statistic is 385 and more than the critical value of 319 at significance level of $\alpha = 5\%$, which indicates that Tuned QEA requires more number of generations as compared to Canonical QEA. The sample size n= 42 (distinct) is larger than 25 so z statistic has also been computed under the normal distribution and p value obtained is 0.2028, which is more than 0.05, so null hypothesis cannot be rejected. However, Tuned QEA is performing significantly better than Canonical QEA as shown by Table. 83. This indicates the success of the proposed tuning method for tuning QEA on problems like Spanner Instances even though it requires more number of generations and function evaluations as compared to Canonical QEA.

**Table 84: Wilcoxon's Signed Rank Test on Spanner Instances KP data (Av. Gen.)**

| | TCQEA | UCQEA |
|---|---|---|
| | **$X_1$** | **$X_2$** |
| 1 | 426.7 | 0 |
| 2 | 510.6 | 2.1 |
| 3 | 607.4 | 154.1 |
| 4 | 576.7 | 428 |
| 5 | 966.5 | 579.1 |
| 6 | 977.4 | 729.9 |
| 7 | 3.2 | 3.9 |
| 8 | 398.5 | 416.4 |
| 9 | 528.1 | 484.1 |
| 10 | 358.7 | 478.2 |
| 11 | 357.4 | 847.8 |
| 12 | 377.9 | 893.2 |
| 13 | 220.5 | 207.2 |
| 14 | 451.4 | 521.7 |
| 15 | 291.4 | 623.8 |
| 16 | 371.1 | 673.5 |
| 17 | 599.2 | 956.8 |
| 18 | 507.1 | 972 |
| 19 | 288.9 | 225.4 |
| 20 | 440.4 | 571.1 |
| 21 | 476 | 450.6 |
| 22 | 550.7 | 706.5 |
| 23 | 547.9 | 964 |
| 24 | 661.1 | 976 |
| 25 | 399.4 | 583.6 |
| 26 | 448 | 579.1 |
| 27 | 580.1 | 534.7 |
| 28 | 599.6 | 742.9 |
| 29 | 718.2 | 964.4 |
| 30 | 753.4 | 982.5 |
| 31 | 370.2 | 545.8 |
| 32 | 532.2 | 631.4 |
| 33 | 645.2 | 516.7 |
| 34 | 802.5 | 702.2 |
| 35 | 987.5 | 965 |
| 36 | 988.7 | 983.8 |
| 37 | 544.4 | 550.4 |

| | |
|---|---|
| Σ(+) | 385 |
| Σ(–) | 518 |
| *n* | 42 |

| Null Hypothesis | **Test Statistic** *T* | **Critical Value** At an α of 5% |
|---|---|---|
| $H_0$: $\mu_1$ >= $\mu_2$ | 385 | 319 |

**For Large Samples (*n* > 25)**

Test Statistic

| | | | |
|---|---|---|---|
| *Z* | 0.831494 | *E[T]* | 451.5 |
| | | *σ(T)* | 79.97655907 |

| Null Hypothesis | *p*-value | At an α of 5% |
|---|---|---|
| $H_0$: $\mu_1$ >= $\mu_2$ | 0.2028 | |
| 38 | 651.9 | 489.5 |
| 39 | 849.8 | 590.3 |
| 40 | 988 | 649.5 |
| 41 | 990.8 | 963.4 |
| 42 | 989.3 | 980 |

***9) multiple strongly correlated instances mstr( k1, k2,d) :***

The result of comparative study between Tuned QEA (TCQEA) and Canonical QEA (UCQEA) are given in Tables 85 to 91. The performance of TCQEA and UCQEA are similar when the no. of items to choose is 100 and the capacity of knapsack is 0.1% of the total capacity. However, with the increase in number of items and the capacity of knapsack, the performance of TCQEA has improved over UCQEA in rest of the instances. The performance of TCQEA was also good on speed of convergence as indicated by average generations and the convergence graphs shown in Fig. 117 to 122, which also compared the speed of convergence of TCQEA to UCQEA.

Table 85

Comparative Results for 0-1 Knapsack problem with No. of Items 100 on Multiple Strongly correlated Data Instances

| % of Total Weight | Canonical QEA | | | | | | Tuned-QEA | | | | | |
|---|---|---|---|---|---|---|---|---|---|---|---|---|
| | Best | Worst | Average | Median | Std | Av. Gen | Best | Worst | Average | Median | Std | Av. Gen |
| 1% | 1959 | 1650 | 1790 | 1774 | **79** | 226247 | **2276** | **2074** | **2212** | **2276** | 85 | **102627** |
| 5% | **7186** | 6882 | 7095 | 7086 | 60 | 91500 | **7186** | **7085** | **7141** | **7181** | **49** | **82051** |
| 10% | **11975** | **11676** | 11821 | 11870 | 72 | **65737** | **11975** | 11675 | **11848** | **11875** | **58** | 75266 |
| 20% | **19856** | **19653** | 19750 | 19753 | **66** | 92455 | **19856** | **19654** | **19766** | **19754** | 72 | **73187** |
| 50% | **39501** | 39298 | 39423 | 39401 | 73 | **48160** | **39501** | **39300** | **39470** | **39500** | **53** | 77613 |

Table 86

Comparative Results for 0-1 Knapsack problem with No. of Items 200 on Multiple Strongly correlated Data Instances

| % of Total Weight | Canonical QEA | | | | | | Tuned-QEA | | | | | |
|---|---|---|---|---|---|---|---|---|---|---|---|---|
| | Best | Worst | Average | Median | Std | Av. Gen | Best | Worst | Average | Median | Std | Av. Gen |
| 1% | **5497** | 4899 | 5334 | 5396 | 139 | 297235 | **5497** | **5199** | **5404** | **5399** | **78** | **149351** |
| 5% | **15108** | **14608** | 14871 | **14907** | 135 | **121588** | 15107 | **14607** | **14911** | **14908** | **113** | 145840 |
| 10% | 24319 | 23722 | 23991 | 24021 | 136 | 127532 | **24321** | **23822** | **24121** | **24121** | **111** | **124964** |
| 20% | 39753 | **39450** | 39615 | **39652** | **100** | **104052** | **39852** | 39353 | **39656** | **39653** | 106 | 119965 |
| 50% | 81050 | 80650 | 80887 | 80850 | 103 | **74313** | **81149** | **80750** | **80936** | **80950** | **90** | 127925 |

Table 87

Comparative Results for 0-1 Knapsack problem with No. of Items 500 on Multiple Strongly correlated Data Instances

| % of Total Weight | Canonical QEA | | | | | | Tuned-QEA | | | | | |
|---|---|---|---|---|---|---|---|---|---|---|---|---|
| | Best | Worst | Average | Median | Std | Av. Gen | Best | Worst | Average | Median | Std | Av. Gen |
| 1% | 12836 | 11737 | 12476 | 12536 | 235 | 334448 | **12936** | **12136** | **12686** | **12736** | **201** | **240722** |
| 5% | 36307 | 35308 | 35874 | 35908 | 242 | 301152 | **36407** | **35494** | **36063** | **36107** | **237** | **287097** |

| % of Total Weight | Best | Worst | Average | Median | Std | Av. Gen | Best | Worst | Average | Median | Std | Av. Gen |
|---|---|---|---|---|---|---|---|---|---|---|---|---|
| 10% | **59128** | 58129 | 58629 | 58629 | 221 | **293907** | **59128** | **58429** | **58842** | **58878** | **167** | 299812 |
| 20% | **98476** | 97477 | 97897 | 97877 | 212 | **240680** | **98476** | **97577** | **98140** | **98177** | **207** | 266056 |
| 50% | 200437 | 199538 | 199864 | 199887 | 193 | **187797** | **200437** | **199838** | **200134** | **200137** | **161** | 229261 |

TABLE 88

COMPARATIVE RESULTS FOR 0-1 KNAPSACK PROBLEM WITH NO. OF ITEMS 1000 ON MULTIPLE STRONGLY CORRELATED DATA INSTANCES

| % of Total Weight | Canonical QEA | | | | | | Tuned-QEA | | | | | |
|---|---|---|---|---|---|---|---|---|---|---|---|---|
| | Best | Worst | Average | Median | Std | Av. Gen | Best | Worst | Average | Median | Std | Av. Gen |
| 1% | 25792 | 24094 | 25142 | 25193 | 417 | 458267 | **26390** | **24494** | **25649** | **25692** | **356** | **409464** |
| 5% | 72419 | 70821 | 71546 | 71520 | **371** | 434415 | **72719** | **71220** | **72104** | **72019** | 423 | **412546** |
| 10% | 117965 | 116167 | 117135 | 117066 | 439 | **416072** | **118065** | **116765** | **117568** | **117565** | **319** | 441646 |
| 20% | 195571 | 194272 | 194964 | 194971 | **341** | **425442** | **195970** | **194472** | **195351** | **195420** | 398 | 430950 |
| 50% | **400015** | 398917 | 399518 | 399616 | 286 | 413777 | **400415** | **399416** | **399878** | **399916** | **216** | **409055** |

TABLE 89

COMPARATIVE RESULTS FOR 0-1 KNAPSACK PROBLEM WITH NO. OF ITEMS 2000 ON MULTIPLE STRONGLY CORRELATED DATA INSTANCES

| % of Total Weight | Canonical QEA | | | | | | Tuned-QEA | | | | | |
|---|---|---|---|---|---|---|---|---|---|---|---|---|
| | Best | Worst | Average | Median | Std | Av. Gen | Best | Worst | Average | Median | Std | Av. Gen |
| 1% | 48685 | **45797** | 47712 | 47666 | 592 | 494990 | **50383** | **48389** | **49458** | **49487** | **474** | **489591** |
| 5% | 143147 | 139787 | 141506 | 141618 | 774 | 495940 | **143986** | **141587** | **142685** | **142808** | **612** | **494218** |
| 10% | 233926 | 231629 | 232718 | 232763 | 668 | 496428 | **234514** | **232213** | **233545** | **233556** | **569** | **494314** |
| 20% | 391129 | 388333 | 389802 | 389753 | 702 | 496078 | **391629** | **388869** | **390683** | **390821** | **659** | 496102 |
| 50% | 803899 | 801902 | 802993 | 803150 | **584** | **488330** | **804899** | **802298** | **803725** | **803700** | 639 | 492188 |

TABLE 90

COMPARATIVE RESULTS FOR 0-1 KNAPSACK PROBLEM WITH NO. OF ITEMS 5000 ON MULTIPLE STRONGLY CORRELATED DATA INSTANCES

| % of Total Weight | Canonical QEA | | | | | | Tuned-QEA | | | | | |
|---|---|---|---|---|---|---|---|---|---|---|---|---|
| | Best | Worst | Average | Median | Std | Av. Gen | Best | Worst | Average | Median | Std | Av. Gen |
| 1% | 100787 | 95661 | 98035 | 98054 | **1226** | 498928 | **112856** | **107765** | **110302** | **110402** | 1262 | **498343** |
| 5% | 330892 | 325231 | 327619 | 327652 | **1525** | **498883** | **341498** | **331288** | **335872** | **335831** | 2601 | 498960 |
| 10% | 557944 | 550326 | 554411 | 554773 | 2016 | **498905** | **566415** | **559850** | **563193** | **562786** | **1876** | 499257 |
| 20% | 956181 | 947965 | 952023 | 951851 | **1990** | **499105** | **964102** | **950046** | **959209** | **959640** | 3412 | 499188 |
| 50% | 1998191 | 1992733 | 1994900 | 1994753 | **1503** | 498720 | **2001816** | **1994017** | **1998843** | **1999038** | 1776 | **498409** |

TABLE 91

COMPARATIVE RESULTS FOR 0-1 KNAPSACK PROBLEM WITH NO. OF ITEMS 10000 ON MULTIPLE STRONGLY CORRELATED DATA INSTANCES

| % of Total | Canonical QEA | Tuned-QEA |
|---|---|---|

| Weight | Best | Worst | Average | Median | Std | Av. Gen | Best | Worst | Average | Median | Std | Av. Gen |
|---|---|---|---|---|---|---|---|---|---|---|---|---|
| 1% | 166560 | 157984 | 161856 | 161939 | 2175 | **499018** | **191838** | **184093** | **188491** | **188785** | **2000** | 499171 |
| 5% | 600677 | 589762 | 595880 | 597146 | **2962** | **499280** | **622415** | **603041** | **611780** | **611599** | 5003 | 499547 |
| 10% | 1046490 | 1026935 | 1036658 | 1035996 | **5043** | **499270** | **1066115** | **1039265** | **1051718** | **1052376** | 6790 | 499425 |
| 20% | 1848462 | 1811336 | 1829781 | 1829247 | **7707** | **499192** | **1862683** | **1822019** | **1846970** | **1848248** | 9210 | 499544 |
| 50% | 3970550 | **3949471** | 3960685 | 3961109 | **4825** | 499077 | **3982008** | 3949092 | **3964796** | **3965426** | 6382 | 499498 |

The convergence graphs have been plotted between objective function value and number of generations for both TCQEA and UCQEA for all the problem instances having No. of Items as 200 and 5000 for the median run. The convergence graph clearly establishes the superiority of Tuning as the TCQEA is faster than UCQEA in all the graphs. The difference in performance between TCQEA and UCQEA increases with the capacity size and number of items in the knapsack problem.

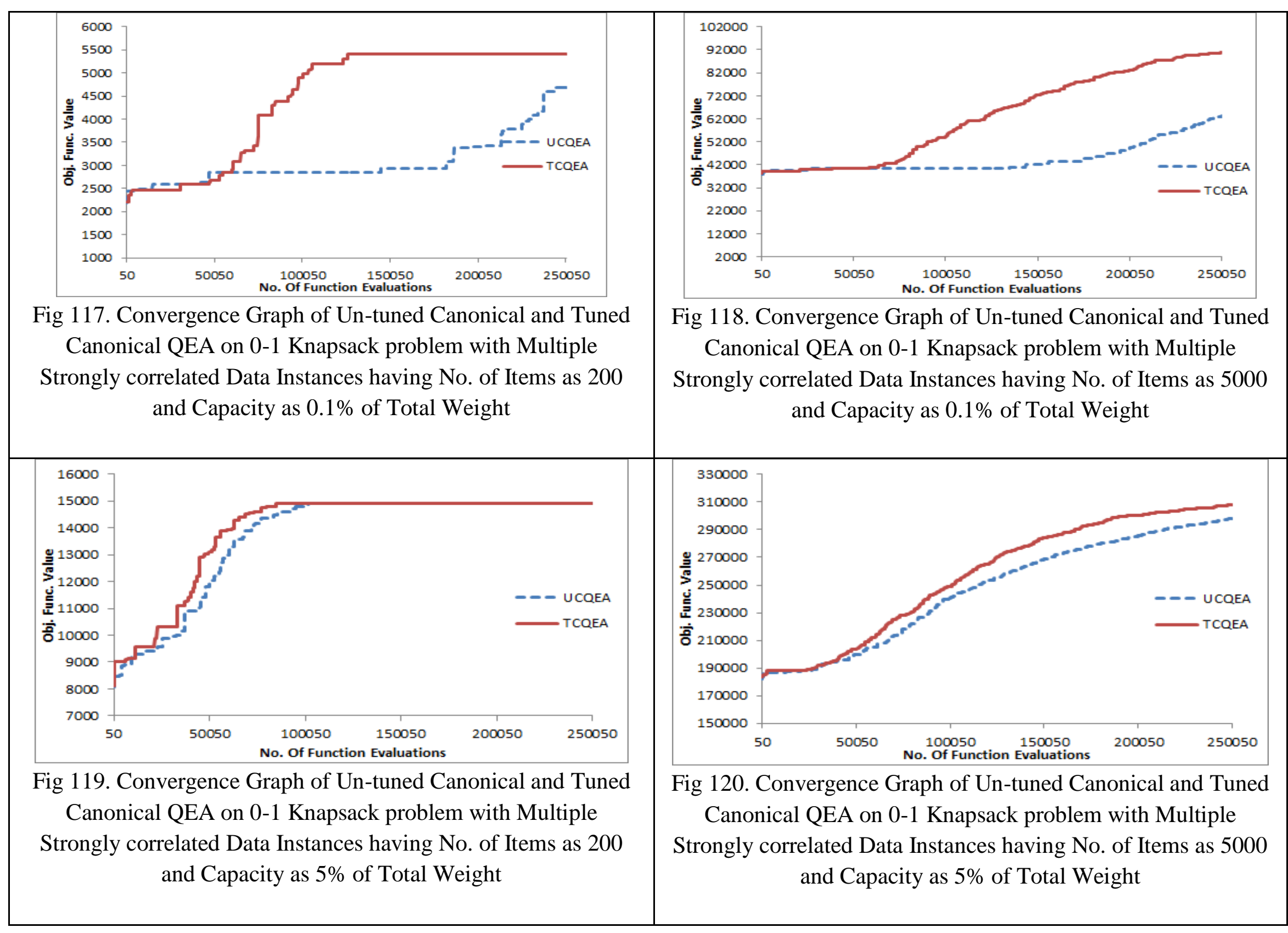

Fig 117. Convergence Graph of Un-tuned Canonical and Tuned Canonical QEA on 0-1 Knapsack problem with Multiple Strongly correlated Data Instances having No. of Items as 200 and Capacity as 0.1% of Total Weight

Fig 118. Convergence Graph of Un-tuned Canonical and Tuned Canonical QEA on 0-1 Knapsack problem with Multiple Strongly correlated Data Instances having No. of Items as 5000 and Capacity as 0.1% of Total Weight

Fig 119. Convergence Graph of Un-tuned Canonical and Tuned Canonical QEA on 0-1 Knapsack problem with Multiple Strongly correlated Data Instances having No. of Items as 200 and Capacity as 5% of Total Weight

Fig 120. Convergence Graph of Un-tuned Canonical and Tuned Canonical QEA on 0-1 Knapsack problem with Multiple Strongly correlated Data Instances having No. of Items as 5000 and Capacity as 5% of Total Weight

In order to confirm the findings in Table 85 to 91, multi-problem non-parametric Wilcoxon's Signed Rank Test [Der2011] was performed on average objective function value (OFV) and average number of generation (Av. Gen.) of Tuned QEA and Canonical QEA on all the instances of Multiple Strongly Correlated Data Instances of Knapsack problem at a significance level of 5%. In case of comparison on average OFV, the null hypothesis for comparison was average OFV of Tuned QEA $\mu_1$ is less than or equal to the average OFV of Canonical QEA $\mu_2$ i.e. $H_0$: $\mu_1 \leq \mu_2$ and the alternate hypothesis is H1: $\mu_1 > \mu_2$. The result of test is presented in Table 92, which shows that null hypothesis can be rejected safely as Wilcoxon's Signed Rank test statistic is zero and less than the critical value of 303 at significance level of $\alpha = 5\%$. The sample size n= 41 (distinct) is larger than 25 so z statistic has also been computed under the normal distribution and p value obtained is 0.000, which is less than 0.05, so null hypothesis can be safely rejected. This indicates the success of the proposed tuning method for tuning QEA on problems like Multiple Strongly Correlated Data Instances of Knapsack problem.

**Table 92: Wilcoxon's Signed Rank Test on Mult. Str. Corr. Data Instances (OFV)**

| | TCQEA | UCQEA |
|---|---|---|
| | **$X_1$** | **$X_2$** |
| 1 | 246.8 | 246.8 |
| 2 | 1168.3 | 1078.3 |
| 3 | 2178.7 | 1640.2 |
| 4 | 3727.7 | 2692.9 |
| 5 | 6033.2 | 5347.1 |
| 6 | 7058.4 | 6713.8 |
| 7 | 1260.7 | 790.589 |
| 8 | 3383.8 | 1768 |
| 9 | 5353.1 | 2756.34 |
| 10 | 8123.5 | 4569.51 |
| 11 | 12627 | 9691.69 |
| 12 | 14673 | 12460.2 |
| 13 | 3295.5 | 1217.19 |
| 14 | 7810.3 | 3113.42 |
| 15 | 11786 | 5192.57 |
| 16 | 18147 | 9453.66 |
| 17 | 28521 | 19916 |
| 18 | 33093 | 25818.6 |
| 19 | 6128.3 | 1771.17 |
| 20 | 13736 | 5194.76 |
| 21 | 20492 | 9312.04 |
| 22 | 32013 | 17340.1 |
| 23 | 51090 | 36042.2 |
| 24 | 60172 | 46525.2 |
| 25 | 9130.1 | 2696.4 |
| 26 | 21811 | 9357.86 |
| 27 | 34095 | 17179.9 |
| 28 | 54775 | 32660 |
| 29 | 90465 | 67004.4 |
| 30 | 106968 | 85885.3 |
| 31 | 14287 | 5263.17 |
| 32 | 40527 | 21049.8 |
| 33 | 66048 | 40048.1 |
| 34 | 110871 | 77818.5 |
| 35 | 193216 | 157688 |
| 36 | 232850 | 199352 |
| 37 | 20137 | 9322.22 |
| 38 | 66045 | 40162.1 |
| 39 | 111394 | 77550.8 |
| 40 | 195672 | 152154 |
| 41 | 354378 | 306007 |
| 42 | 431151 | 385901 |

| | |
|---|---|
| Σ(+) | 861 |
| Σ(–) | 0 |
| *n* | 41 |

| Null Hypothesis | **Test Statistic** *T* | **Critical Value** At an α of 5% |
|---|---|---|
| $H_0$: $\mu_1$ <= $\mu_2$ | 0 | 303 |

**For Large Samples (*n* > 25)**

Test Statistic

| | | | |
|---|---|---|---|
| *z* | -5.57857 | *E[T]* | 430.5 |
| | | *σ(T)* | 77.17026629 |

| Null Hypothesis | *p*-value | At an α of 5% |
|---|---|---|
| $H_0$: $\mu_1$ <= $\mu_2$ | 0.0000 | **Reject** |

In case of comparison on Av. Gen., the null hypothesis for comparison was Av. Gen. of Tuned QEA $\mu_1$ is greater than or equal to the Av. Gen. of Canonical QEA $\mu_2$ i.e. $H_0$: $\mu_1 \geq \mu_2$ and the alternate hypothesis is H1: $\mu_1 < \mu_2$. The result of test is presented in Table 93, which shows that null hypothesis cannot be rejected as Wilcoxon's Signed Rank test statistic is 385 and more than the critical value of 319 at significance level of α = 5%, which indicates that Tuned QEA requires more number of generations as compared to Canonical QEA. The sample size n= 42 (distinct) is larger than 25 so z statistic has also been computed under the normal distribution and p value obtained is 0.2028, which is more than 0.05, so null hypothesis cannot be rejected. However, Tuned QEA is performing significantly better than Canonical QEA as shown by Table 92. This indicates the success of the proposed tuning method for tuning QEA on problems like Multiple Strongly Correlated Data Instances, even though it requires more number of generations and function evaluations as compared to Canonical QEA.

**Table 93: Wilcoxon's Signed Rank Test on Mult. Str. Cor. Data Inst. (Av. Gen.)**

| | TCQEA | UCQEA |
|---|---|---|
| | **$X_1$** | **$X_2$** |
| 1 | 426.7 | 0 |
| 2 | 510.6 | 2.1 |
| 3 | 607.4 | 154.1 |
| 4 | 576.7 | 428 |
| 5 | 966.5 | 579.1 |
| 6 | 977.4 | 729.9 |
| 7 | 3.2 | 3.9 |
| 8 | 398.5 | 416.4 |
| 9 | 528.1 | 484.1 |
| 10 | 358.7 | 478.2 |
| 11 | 357.4 | 847.8 |
| 12 | 377.9 | 893.2 |
| 13 | 220.5 | 207.2 |
| 14 | 451.4 | 521.7 |
| 15 | 291.4 | 623.8 |
| 16 | 371.1 | 673.5 |
| 17 | 599.2 | 956.8 |
| 18 | 507.1 | 972 |
| 19 | 288.9 | 225.4 |
| 20 | 440.4 | 571.1 |
| 21 | 476 | 450.6 |
| 22 | 550.7 | 706.5 |
| 23 | 547.9 | 964 |
| 24 | 661.1 | 976 |
| 25 | 399.4 | 583.6 |
| 26 | 448 | 579.1 |
| 27 | 580.1 | 534.7 |
| 28 | 599.6 | 742.9 |
| 29 | 718.2 | 964.4 |
| 30 | 753.4 | 982.5 |
| 31 | 370.2 | 545.8 |
| 32 | 532.2 | 631.4 |
| 33 | 645.2 | 516.7 |
| 34 | 802.5 | 702.2 |
| 35 | 987.5 | 965 |
| 36 | 988.7 | 983.8 |
| 37 | 544.4 | 550.4 |
| 38 | 651.9 | 489.5 |
| 39 | 849.8 | 590.3 |
| 40 | 988 | 649.5 |
| 41 | 990.8 | 963.4 |

| | |
|---|---|
| $\Sigma(+)$ | 385 |
| $\Sigma(-)$ | 518 |
| $n$ | 42 |

| Null Hypothesis | **Test Statistic** $T$ | **Critical Value** At an $\alpha$ of 5% |
|---|---|---|
| $H_0$: $\mu_1 >= \mu_2$ | 385 | 319 |

**For Large Samples ($n$ > 25)**

Test Statistic

| | |
|---|---|
| $z$ | 0.831494 |
| $E[T]$ | 451.5 |
| $\sigma(T)$ | 79.97655907 |

| Null Hypothesis | $p$-value | At an $\alpha$ of 5% |
|---|---|---|
| $H_0$: $\mu_1 >= \mu_2$ | 0.2028 | |

| 42 | 989.3 | 980 | |
|---|---|---|---|

*10) profit ceiling instances pceil(d) :*

The result of comparative study between Tuned QEA (TCQEA) and Canonical QEA (UCQEA) are given in Table 94 to 100. The performance of TCQEA and UCQEA are similar when the no. of items to choose is 100 and the capacity of knapsack is 0.1% to 2% of the total capacity. In case of problem instance with no. of items as 200 and the capacity of knapsack is 5% of the total capacity, the best result of UCQEA is slightly better than that of TCQEA, but TCQEA is better than UCQEA on average & median result and matches on worst result for objective function value. In case of problem instance with no. of items as 1000 and the capacity of knapsack is 0.1% of the total capacity, the best result of UCQEA is slightly better than that of TCQEA, but TCQEA is better than UCQEA on worst, average & median result. However, the performance of TCQEA is better than UCQEA in rest of the instances. The performance of TCQEA is also good on speed of convergence as indicated by average generations and the convergence graphs shown in Fig. 123 to 128, which also compared the speed of convergence of TCQEA to UCQEA.

TABLE 94
COMPARATIVE RESULTS FOR 0-1 KNAPSACK PROBLEM WITH NO. OF ITEMS 100 ON PROFIT CEILING INSTANCES

| % of Total Weight | Canonical QEA | | | | | | Tuned-QEA | | | | | |
|---|---|---|---|---|---|---|---|---|---|---|---|---|
| | **Best** | **Worst** | **Average** | **Median** | **Std** | **Av. Gen** | **Best** | **Worst** | **Average** | **Median** | **Std** | **Av. Gen** |
| 1% | 486 | **483** | 486 | **486** | **1** | 126227 | **489** | **483** | **485** | **486** | 2 | **53635** |
| 5% | **2421** | 2406 | 2412 | 2412 | 3 | 133537 | 2418 | **2409** | **2413** | **2415** | **3** | **78669** |
| 10% | 4824 | **4812** | 4819 | **4818** | 4 | 139967 | **4827** | **4812** | **4819** | **4818** | **4** | **128561** |
| 20% | 9630 | **9618** | 9624 | 9624 | **3** | 180320 | **9633** | 9615 | **9625** | **9627** | 4 | **151470** |
| 50% | **24030** | 24012 | 24021 | 24021 | 5 | **129802** | 24027 | **24015** | **24023** | **24024** | **4** | 193624 |

TABLE 95
COMPARATIVE RESULTS FOR 0-1 KNAPSACK PROBLEM WITH NO. OF ITEMS 200 ON PROFIT CEILING INSTANCES

| % of Total Weight | Canonical QEA | | | | | | Tuned-QEA | | | | | |
|---|---|---|---|---|---|---|---|---|---|---|---|---|
| | **Best** | **Worst** | **Average** | **Median** | **Std** | **Av. Gen** | **Best** | **Worst** | **Average** | **Median** | **Std** | **Av. Gen** |
| 1% | **1026** | **1014** | **1018** | **1017** | **3** | 252298 | 1023 | 1011 | 1017 | **1017** | 3 | **136178** |
| 5% | **5070** | 5049 | 5061 | **5061** | 6 | 198847 | **5070** | **5052** | **5061** | **5061** | **5** | **158638** |
| 10% | **10116** | **10092** | **10105** | **10104** | **5** | 256945 | **10116** | **10092** | **10105** | **10104** | 6 | **194697** |
| 20% | 20193 | 20163 | 20180 | 20181 | **7** | 285912 | **20199** | **20166** | **20182** | **20183** | 8 | **180635** |
| 50% | 50397 | **50373** | **50385** | **50385** | **6** | 245873 | **50403** | **50373** | 50385 | **50385** | 7 | **233591** |

TABLE 96
COMPARATIVE RESULTS FOR 0-1 KNAPSACK PROBLEM WITH NO. OF ITEMS 500 ON PROFIT CEILING INSTANCES

| % of Total Weight | Canonical QEA | | | | | | Tuned-QEA | | | | | |
|---|---|---|---|---|---|---|---|---|---|---|---|---|
| | **Best** | **Worst** | **Average** | **Median** | **Std** | **Av. Gen** | **Best** | **Worst** | **Average** | **Median** | **Std** | **Av. Gen** |
| 1% | **2487** | **2466** | **2475** | **2475** | **4** | **327145** | 2484 | 2463 | 2472 | 2472 | 6 | **192981** |
| 5% | **12345** | **12306** | **12323** | **12323** | 9 | 361763 | 12336 | 12303 | 12317 | 12315 | **8** | **245583** |
| 10% | 24627 | **24600** | **24615** | **24614** | **8** | 357690 | **24633** | 24591 | 24612 | 24612 | 9 | **276887** |
| 20% | **49200** | **49158** | **49179** | **49176** | **11** | 424377 | 49197 | 49152 | 49173 | 49170 | 13 | **276731** |

| 50% | **122814** | **122775** | **122794** | **122792** | 9 | 376148 | 122808 | **122769** | **122789** | 122790 | **9** | **289390** |
|---|---|---|---|---|---|---|---|---|---|---|---|---|

TABLE 97

COMPARATIVE RESULTS FOR 0-1 KNAPSACK PROBLEM WITH NO. OF ITEMS 1000 ON PROFIT CEILING INSTANCES

| % of Total Weight | Canonical QEA | | | | | | Tuned-QEA | | | | | |
|---|---|---|---|---|---|---|---|---|---|---|---|---|
| | Best | Worst | Average | Median | Std | Av. Gen | Best | Worst | Average | Median | Std | Av. Gen |
| 1% | **4986** | **4956** | **4969** | **4968** | 7 | 395845 | 4971 | 4950 | 4961 | 4959 | **6** | **243157** |
| 5% | **24771** | **24729** | **24752** | **24753** | 12 | 412405 | 24753 | 24711 | 24734 | 24738 | **11** | **379210** |
| 10% | **49473** | **49404** | **49436** | **49437** | **15** | 461158 | 49458 | 49383 | 49421 | 49422 | 16 | **388001** |
| 20% | **98796** | **98736** | **98762** | **98763** | **14** | 455122 | 98781 | 98718 | 98749 | 98751 | 16 | **398188** |
| 50% | 246654 | **246579** | **246619** | **246618** | **18** | 445543 | **246657** | 246573 | 246609 | 246608 | 21 | **346196** |

TABLE 98

COMPARATIVE RESULTS FOR 0-1 KNAPSACK PROBLEM WITH NO. OF ITEMS 2000 ON PROFIT CEILING INSTANCES

| % of Total Weight | Canonical QEA | | | | | | Tuned-QEA | | | | | |
|---|---|---|---|---|---|---|---|---|---|---|---|---|
| | Best | Worst | Average | Median | Std | Av. Gen | Best | Worst | Average | Median | Std | Av. Gen |
| 1% | **10050** | **10017** | **10032** | **10032** | 9 | 430522 | 10032 | 9999 | 10014 | 10014 | **9** | **292951** |
| 5% | **50040** | **49968** | **50012** | **50010** | 15 | 456985 | 49995 | 49947 | 49975 | 49977 | **13** | **414962** |
| 10% | **99954** | **99882** | **99916** | **99918** | **16** | 473622 | 99906 | 99831 | 99870 | 99866 | 19 | **429937** |
| 20% | 199689 | **199608** | **199650** | **199652** | **21** | 486162 | **199692** | 199563 | 199615 | 199617 | 28 | **462175** |
| 50% | **498711** | **498624** | **498668** | **498671** | **20** | 475312 | 498669 | 498567 | 498634 | 498639 | 26 | **420826** |

TABLE 99

COMPARATIVE RESULTS FOR 0-1 KNAPSACK PROBLEM WITH NO. OF ITEMS 5000 ON PROFIT CEILING INSTANCES

| % of Total Weight | Canonical QEA | | | | | | Tuned-QEA | | | | | |
|---|---|---|---|---|---|---|---|---|---|---|---|---|
| | Best | Worst | Average | Median | Std | Av. Gen | Best | Worst | Average | Median | Std | Av. Gen |
| 1% | **25077** | **25026** | **25054** | **25053** | **12** | 471443 | 25050 | 25002 | 25025 | 25025 | 13 | **427370** |
| 5% | **125004** | **124929** | **124964** | **124962** | **20** | 482153 | 124935 | 124824 | 124885 | 124880 | 31 | **468877** |
| 10% | **249723** | **249627** | **249688** | **249687** | **22** | 491412 | 249708 | 249510 | 249614 | 249614 | 41 | **480117** |
| 20% | **499104** | **498966** | **499029** | **499025** | **31** | 494853 | 499101 | 498918 | 499002 | 499001 | 48 | **481599** |
| 50% | 1246764 | **1246647** | **1246714** | **1246719** | **33** | 495203 | **1246797** | 1246584 | 1246699 | 1246700 | 63 | **472636** |

TABLE 100

COMPARATIVE RESULTS FOR 0-1 KNAPSACK PROBLEM WITH NO. OF ITEMS 10000 ON PROFIT CEILING INSTANCES

| % of Total Weight | Canonical QEA | | | | | | Tuned-QEA | | | | | |
|---|---|---|---|---|---|---|---|---|---|---|---|---|
| | Best | Worst | Average | Median | Std | Av. Gen | Best | Worst | Average | Median | Std | Av. Gen |
| 1% | **50298** | **50241** | **50273** | **50271** | **14** | 479530 | 50277 | 50184 | 50221 | 50217 | 25 | **469488** |

| | | | | | | | | | | | |
|---|---|---|---|---|---|---|---|---|---|---|---|
| 5% | **250878** | **250788** | **250837** | **250835** | **26** | 492728 | 250809 | 250668 | 250747 | 250748 | 36 | **484641** |
| 10% | 501393 | **501264** | **501315** | **501317** | **34** | 495528 | **501420** | 501168 | 501270 | 501273 | 55 | **492753** |
| 20% | 1002159 | 1001943 | 1002055 | 1002063 | **60** | 495115 | **1002252** | **1001967** | **1002094** | **1002089** | 73 | **487271** |
| 50% | 2504061 | 2503740 | 2503870 | 2503869 | **71** | 496130 | **2504124** | **2503761** | **2503953** | **2503961** | 104 | **487403** |

The convergence graphs have been plotted between objective function value and number of generations for both TCQEA and UCQEA for all the problem instances having No. of Items as 200 and 5000 for the median run. The convergence graph clearly establishes the superiority of Tuning as the TCQEA is faster than UCQEA in most of the graphs. The difference in performance between TCQEA and UCQEA increases with the capacity size and number of items in the knapsack problem.

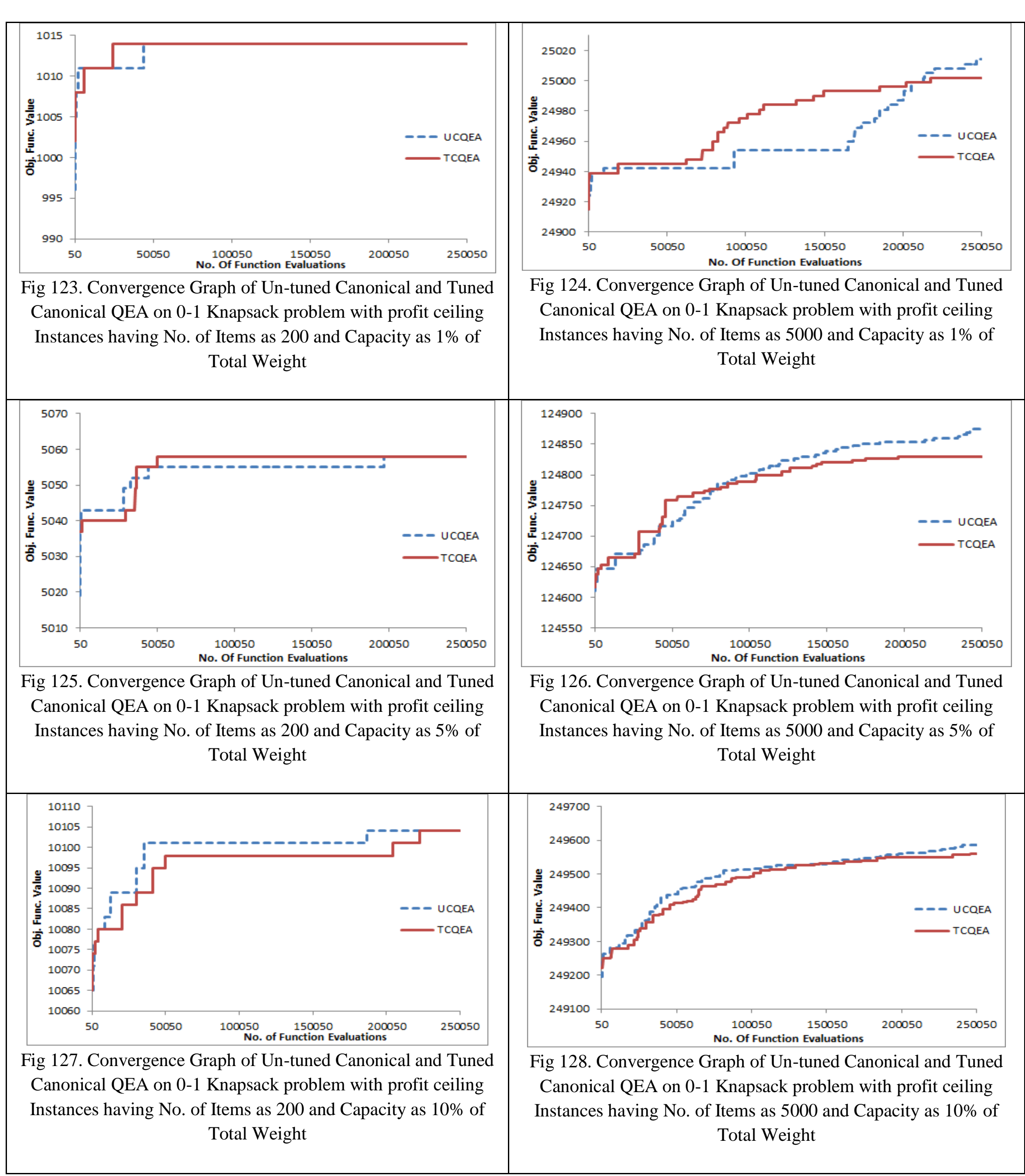

Fig 123. Convergence Graph of Un-tuned Canonical and Tuned Canonical QEA on 0-1 Knapsack problem with profit ceiling Instances having No. of Items as 200 and Capacity as 1% of Total Weight

Fig 124. Convergence Graph of Un-tuned Canonical and Tuned Canonical QEA on 0-1 Knapsack problem with profit ceiling Instances having No. of Items as 5000 and Capacity as 1% of Total Weight

Fig 125. Convergence Graph of Un-tuned Canonical and Tuned Canonical QEA on 0-1 Knapsack problem with profit ceiling Instances having No. of Items as 200 and Capacity as 5% of Total Weight

Fig 126. Convergence Graph of Un-tuned Canonical and Tuned Canonical QEA on 0-1 Knapsack problem with profit ceiling Instances having No. of Items as 5000 and Capacity as 5% of Total Weight

Fig 127. Convergence Graph of Un-tuned Canonical and Tuned Canonical QEA on 0-1 Knapsack problem with profit ceiling Instances having No. of Items as 200 and Capacity as 10% of Total Weight

Fig 128. Convergence Graph of Un-tuned Canonical and Tuned Canonical QEA on 0-1 Knapsack problem with profit ceiling Instances having No. of Items as 5000 and Capacity as 10% of Total Weight

In order to confirm the findings in Table 94 to 100, multi-problem non-parametric Wilcoxon's Signed Rank Test [Der2011] was performed on average objective function value (OFV) and average number of generation (Av. Gen.) of Tuned QEA and Canonical QEA on all the instances of Profit Ceiling Knapsack problem at a significance level of 5%. In case of comparison on average OFV, the null hypothesis for comparison was average OFV of Tuned QEA $\mu_1$ is less than or equal to the average OFV of Canonical QEA $\mu_2$ i.e. $H_0$: $\mu_1 \leq \mu_2$ and the alternate hypothesis is H1: $\mu_1 > \mu_2$. The result of test is presented in Table 101, which shows that null hypothesis can be rejected safely as Wilcoxon's Signed Rank test statistic is zero and less than the critical value of 287 at significance level of $\alpha = 5\%$. The sample size n= 40 (distinct) is larger than 25 so z statistic has also been computed under the normal distribution and p value obtained is 0.000, which is less than 0.05, so null hypothesis can be safely rejected. This indicates the success of the proposed tuning method for tuning QEA on problems like Profit Ceiling Instances of Knapsack problem.

**Table 101: Wilcoxon's Signed Rank Test on Profit Ceiling Instances (OFV)**

| | TCQEA | UCQEA |
|---|---|---|
| | **$X_1$** | **$X_2$** |
| 1 | 48 | 48 |
| 2 | 243 | 243 |
| 3 | 485 | 484 |
| 4 | 966.9 | 966.2 |
| 5 | 1931.2 | 1929.1 |
| 6 | 2412.2 | 2410.3 |
| 7 | 105.2 | 105 |
| 8 | 509.9 | 508.3 |
| 9 | 1015.7 | 1013.5 |
| 10 | 2026.4 | 2021.9 |
| 11 | 4044.9 | 4042.3 |
| 12 | 5054 | 5053.7 |
| 13 | 250.7 | 249.7 |
| 14 | 1237 | 1234 |
| 15 | 2468.9 | 2462.5 |
| 16 | 4929.1 | 4919.1 |
| 17 | 9850.1 | 9839.3 |
| 18 | 12307 | 12301.3 |
| 19 | 499.6 | 498.4 |
| 20 | 2477.8 | 2473.5 |
| 21 | 4953.9 | 4941.6 |
| 22 | 9894.7 | 9875.2 |
| 23 | 19771.7 | 19750 |
| 24 | 24706.8 | 24691.7 |
| 25 | 1005.8 | 1002.9 |
| 26 | 5006.8 | 4994.9 |
| 27 | 10003.2 | 9983.4 |
| 28 | 19991.4 | 19955.8 |
| 29 | 39951 | 39918 |
| 30 | 49926.4 | 49902.6 |
| 31 | 2509.3 | 2501 |
| 32 | 12504 | 12477.8 |
| 33 | 24987.6 | 24945.2 |
| 34 | 49942.4 | 49877.5 |
| 35 | 99831.6 | 99766.1 |
| 36 | 124779 | 124714 |
| 37 | 5033.7 | 5017.2 |

| | |
|---|---|
| $\Sigma(+)$ | 820 |
| $\Sigma(-)$ | 0 |
| *n* | 40 |

| Null Hypothesis | **Test Statistic** *T* | **Critical Value** At an $\alpha$ of 5% |
|---|---|---|
| $H_0$: $\mu_1 <= \mu_2$ | 0 | 287 |

**For Large Samples (*n* > 25)**

Test Statistic

| | | | |
|---|---|---|---|
| *z* | -5.51093 | *E[T]* | 410 |
| | | *σ(T)* | 74.39758061 |

| Null Hypothesis | *p*-value | At an $\alpha$ of 5% |
|---|---|---|
| $H_0$: $\mu_1 <= \mu_2$ | 0.0000 | **Reject** |
| 38 | 25101.7 | 25053.3 |
| 39 | 50163.7 | 50092.5 |
| 40 | 100267 | 100167 |
| 41 | 200464 | 200339 |
| 42 | 250549 | 250438 |

In case of comparison on Av. Gen., the null hypothesis for comparison was Av. Gen. of Tuned QEA $\mu_1$ is greater than or equal to the Av. Gen. of Canonical QEA $\mu_2$ i.e. $H_0$: $\mu_1 \geq \mu_2$ and the alternate hypothesis is H1: $\mu_1 < \mu_2$. The result of test is presented in Table 102, which shows that null hypothesis cannot be rejected as Wilcoxon's Signed Rank test statistic is 638 and more than the critical value of 319 at significance level of $\alpha = 5\%$, which indicates that Tuned QEA requires more number of generations as compared to Canonical QEA. The sample size n= 42 (distinct) is larger than 25 so z statistic has also been computed under the normal distribution and p value obtained is 0.9901, which is more than 0.05, so null hypothesis cannot be rejected. However, Tuned QEA is performing significantly better than Canonical QEA as shown by Table. 101. This indicates the success of the proposed tuning method for tuning QEA on problems like Profit Ceiling Instances, even though it requires more number of generations and function evaluations as compared to Canonical QEA.

**Table 102: Wilcoxon's Signed Rank Test on Profit Ceiling Instances (Av. Gen.)**

| | TCQEA | UCQEA |
|---|---|---|
| | $X_1$ | $X_2$ |
| 1 | 2.3 | 2.5 |
| 2 | 40.8 | 39.8 |
| 3 | 290.2 | 185.1 |
| 4 | 293.1 | 345.6 |
| 5 | 412.1 | 529.1 |
| 6 | 455.1 | 621.7 |
| 7 | 197.8 | 262.8 |
| 8 | 286 | 185.1 |
| 9 | 342.1 | 297.2 |
| 10 | 407.6 | 333.2 |
| 11 | 449.7 | 771.8 |
| 12 | 550.2 | 829.9 |
| 13 | 195.8 | 176.8 |
| 14 | 432.1 | 306.2 |
| 15 | 444 | 276.3 |
| 16 | 528.2 | 380.1 |
| 17 | 739.2 | 889.8 |
| 18 | 780 | 948.1 |
| 19 | 244 | 273.2 |
| 20 | 370.5 | 416.1 |
| 21 | 727.2 | 340.5 |
| 22 | 846.2 | 548.6 |
| 23 | 869.1 | 898.4 |
| 24 | 873.2 | 944.3 |
| 25 | 448.3 | 247.2 |
| 26 | 655.2 | 342.5 |
| 27 | 785.4 | 463.2 |
| 28 | 879.3 | 485.6 |
| 29 | 900.9 | 931.2 |
| 30 | 897.8 | 966.7 |
| 31 | 539.9 | 403.1 |

| | |
|---|---|
| Σ(+) | 638 |
| Σ(–) | 265 |
| *n* | 42 |

| Null Hypothesis | **Test Statistic** *T* | **Critical Value** At an α of 5% |
|---|---|---|
| $H_0$: $\mu_1$ >= $\mu_2$ | 638 | 319 |

**For Large Samples (*n* > 25)**

Test Statistic

| | | | |
|---|---|---|---|
| *z* | -2.33193 | *E[T]* | 451.5 |
| | | *σ(T)* | 79.97655907 |

| Null Hypothesis | *p*-value | At an α of 5% |
|---|---|---|
| $H_0$: $\mu_1$ >= $\mu_2$ | 0.9901 | |
| 32 | 842.2 | 370.3 |
| 33 | 868.5 | 349.4 |
| 34 | 947.3 | 450.6 |
| 35 | 939.2 | 955.1 |
| 36 | 949.1 | 956.8 |
| 37 | 742.2 | 306.1 |
| 38 | 876.6 | 332.1 |
| 39 | 940.4 | 365.7 |
| 40 | 939.5 | 539.9 |
| 41 | 951.9 | 942.9 |
| 42 | 960.9 | 957.7 |

### *11) circle instances circle(d) :*

The result of comparative study between Tuned QEA (TCQEA) and Canonical QEA (UCQEA) are given in Table 103 to 109. The performance of TCQEA and UCQEA are similar when the no. of items to choose is 100 and the capacity of knapsack is 0.1 % of the total capacity. However, with the increase in number of items and the capacity of knapsack, the performance of TCQEA has improved over UCQEA in all the instances. The performance of TCQEA was also good on speed of convergence as indicated by average generations and the convergence graphs shown in Fig. 129 to 134, which also compared the speed of convergence of TCQEA to UCQEA.

TABLE 103
COMPARATIVE RESULTS FOR 0-1 KNAPSACK PROBLEM WITH NO. OF ITEMS 100 ON CIRCLE INSTANCES

| % of Total Weight | Canonical QEA | | | | | | Tuned-QEA | | | | | |
|---|---|---|---|---|---|---|---|---|---|---|---|---|
| | Best | Worst | Average | Median | Std | Av. Gen | Best | Worst | Average | Median | Std | Av. Gen |
| 1% | 2174 | 1981 | 2084 | 2089 | **61** | 219922 | **2528** | **2215** | **2512** | **2528** | 61 | **113375** |
| 5% | **9210** | 8955 | 9168 | 9204 | 78 | 110665 | **9210** | **8981** | **9198** | **9210** | **41** | **98789** |
| 10% | **15838** | 15489 | 15727 | **15830** | 130 | **58628** | **15838** | **15504** | **15743** | **15830** | **117** | 73511 |
| 20% | **25848** | 25760 | 25799 | 25800 | 28 | **65242** | **25848** | **25768** | **25821** | **25821** | **23** | 75230 |
| 50% | **49049** | 48974 | **49009** | **49009** | 21 | **99695** | 49032 | **48979** | 49006 | 49009 | **15** | 105656 |

TABLE 104
COMPARATIVE RESULTS FOR 0-1 KNAPSACK PROBLEM WITH NO. OF ITEMS 200 ON CIRCLE INSTANCES

| % of Total Weight | Canonical QEA | | | | | | Tuned-QEA | | | | | |
|---|---|---|---|---|---|---|---|---|---|---|---|---|
| | Best | Worst | Average | Median | Std | Av. Gen | Best | Worst | Average | Median | Std | Av. Gen |
| 1% | **5802** | 5210 | 5591 | 5613 | 143 | 318865 | **5802** | **5407** | **5671** | **5680** | **82** | **158776** |
| 5% | 19046 | 18462 | 18772 | 18774 | 126 | **138005** | **19047** | **18638** | **18811** | **18778** | **102** | 177933 |
| 10% | **31587** | 30811 | 31265 | 31257 | 206 | **116388** | **31587** | **30923** | **31298** | **31289** | **150** | 147929 |
| 20% | **51224** | 50534 | 51028 | 51100 | **158** | **105522** | 51216 | **50673** | **51091** | **51106** | **93** | 146698 |
| 50% | **99189** | 98567 | 98875 | 98814 | 223 | **159023** | 99182 | **98600** | **99045** | **99177** | **190** | 190826 |

TABLE 105
COMPARATIVE RESULTS FOR 0-1 KNAPSACK PROBLEM WITH NO. OF ITEMS 500 ON CIRCLE INSTANCES

| % of Total Weight | Canonical QEA | | | | | | Tuned-QEA | | | | | |
|---|---|---|---|---|---|---|---|---|---|---|---|---|
| | Best | Worst | Average | Median | Std | Av. Gen | Best | Worst | Average | Median | Std | Av. Gen |

| 1% | 13533 | 12173 | 13192 | 13331 | 321 | 358892 | **13713** | **12907** | **13314** | **13331** | **216** | **299973** |
|---|---|---|---|---|---|---|---|---|---|---|---|---|
| 5% | 45669 | 44058 | 45171 | 45217 | **312** | **313940** | **45694** | **44359** | **45289** | **45370** | 320 | 343570 |
| 10% | 76758 | 74937 | 76162 | 76242 | 370 | **298185** | **76788** | **75465** | **76321** | **76352** | **315** | 332571 |
| 20% | **127820** | 126629 | 127195 | 127189 | 319 | **292023** | 127698 | **126699** | **127393** | **127478** | **247** | 302933 |
| 50% | 247430 | 246599 | 247070 | 247109 | **211** | **266805** | **247701** | **247046** | **247289** | **247173** | 245 | 287437 |

TABLE 106

COMPARATIVE RESULTS FOR 0-1 KNAPSACK PROBLEM WITH NO. OF ITEMS 1000 ON CIRCLE INSTANCES

| % of Total Weight | Canonical QEA | | | | | | Tuned-QEA | | | | | |
|---|---|---|---|---|---|---|---|---|---|---|---|---|
| | Best | Worst | Average | Median | Std | Av. Gen | Best | Worst | Average | Median | Std | Av. Gen |
| 1% | 26811 | 24683 | 25964 | 26026 | 532 | 467325 | **26901** | **25126** | **26190** | **26253** | **492** | **427885** |
| 5% | 90456 | 88152 | 89452 | 89452 | 517 | **456892** | **90773** | **88820** | **89845** | **89872** | **451** | 467795 |
| 10% | 152924 | 150373 | 151642 | 151626 | 569 | **438443** | **152994** | **151388** | **152136** | **152067** | **416** | 452126 |
| 20% | 254180 | 251238 | 253182 | 253279 | 673 | 439062 | **254559** | **252261** | **253432** | **253460** | **519** | **406392** |
| 50% | 495096 | **493346** | 494190 | 494183 | 417 | **435308** | **495118** | 493283 | **494478** | **494545** | **402** | 439280 |

TABLE 107

COMPARATIVE RESULTS FOR 0-1 KNAPSACK PROBLEM WITH NO. OF ITEMS 2000 ON CIRCLE INSTANCES

| % of Total Weight | Canonical QEA | | | | | | Tuned-QEA | | | | | |
|---|---|---|---|---|---|---|---|---|---|---|---|---|
| | Best | Worst | Average | Median | Std | Av. Gen | Best | Worst | Average | Median | Std | Av. Gen |
| 1% | 49991 | 48344 | 49226 | 49264 | **426** | 498490 | **52072** | **49711** | **51015** | **51148** | 548 | **495257** |
| 5% | 176949 | 172392 | 174911 | 174875 | **896** | 497640 | **178307** | **172816** | **176641** | **176779** | 1136 | **497228** |
| 10% | 300645 | 296538 | 298608 | 298674 | 1067 | **497538** | **301895** | **298857** | **300197** | **300152** | **798** | 498399 |
| 20% | 506875 | 503546 | 505441 | 505486 | 1010 | 496865 | **508014** | **505229** | **506829** | **507069** | **777** | **496241** |
| 50% | 992783 | 990446 | 991643 | 991614 | **607** | **487825** | **993478** | **990992** | **992362** | **992218** | 811 | 489565 |

TABLE 108

COMPARATIVE RESULTS FOR 0-1 KNAPSACK PROBLEM WITH NO. OF ITEMS 5000 ON CIRCLE INSTANCES

| % of Total Weight | Canonical QEA | | | | | | Tuned-QEA | | | | | |
|---|---|---|---|---|---|---|---|---|---|---|---|---|
| | Best | Worst | Average | Median | Std | Av. Gen | Best | Worst | Average | Median | Std | Av. Gen |
| 1% | 107044 | 102173 | 104741 | 104807 | 1249 | **498433** | **114531** | **111149** | **113044** | **112954** | **926** | 498610 |
| 5% | 401861 | 393326 | 398179 | 398439 | **1937** | **498628** | **412379** | **404551** | **408455** | **408556** | 1952 | 499260 |
| 10% | 703183 | 692142 | 697450 | 696994 | **2946** | **498935** | **718847** | **703895** | **710138** | **710179** | 4590 | 499036 |
| 20% | 1220793 | 1204028 | 1212524 | 1213070 | **4182** | **498712** | **1235625** | **1219591** | **1228249** | **1228672** | 4297 | 499310 |

| 50% | 2464209 | 2451488 | 2458580 | 2458255 | **2723** | **498255** | **2470755** | **2458187** | **2463903** | **2463763** | 2843 | 498861 |
|---|---|---|---|---|---|---|---|---|---|---|---|---|

TABLE 109

COMPARATIVE RESULTS FOR 0-1 KNAPSACK PROBLEM WITH NO. OF ITEMS 10000 ON CIRCLE INSTANCES

| % of Total Weight | Canonical QEA | | | | | | Tuned-QEA | | | | | |
|---|---|---|---|---|---|---|---|---|---|---|---|---|
| | **Best** | **Worst** | **Average** | **Median** | **Std** | **Av. Gen** | **Best** | **Worst** | **Average** | **Median** | **Std** | **Av. Gen** |
| 1% | 186565 | 179029 | 182039 | 181999 | 1776 | **499033** | **204276** | **197290** | **200602** | **200501** | **1693** | 499382 |
| 5% | 733283 | 716838 | 723383 | 723322 | **4120** | **499132** | **760693** | **728628** | **744025** | **744078** | 7558 | 499376 |
| 10% | 1298905 | 1282621 | 1292184 | 1292971 | **4817** | **499083** | **1344170** | **1295281** | **1316045** | **1317789** | 12332 | 499514 |
| 20% | 2310887 | 2279802 | 2293063 | 2294553 | **7884** | **499178** | **2356125** | **2295050** | **2324957** | **2326246** | 13563 | 499445 |
| 50% | 4847341 | 4817442 | 4833333 | 4836603 | **10274** | **499165** | **4865335** | **4819671** | **4842682** | **4843552** | 12631 | 499660 |

The convergence graphs have been plotted between objective function value and number of generations for both TCQEA and UCQEA for all the problem instances having No. of Items as 200 and 5000 for the median run. The convergence graph clearly establishes the superiority of Tuning as the TCQEA is faster than UCQEA in all the graphs. The difference in performance between TCQEA and UCQEA increases with the capacity size and number of items in the knapsack problem.

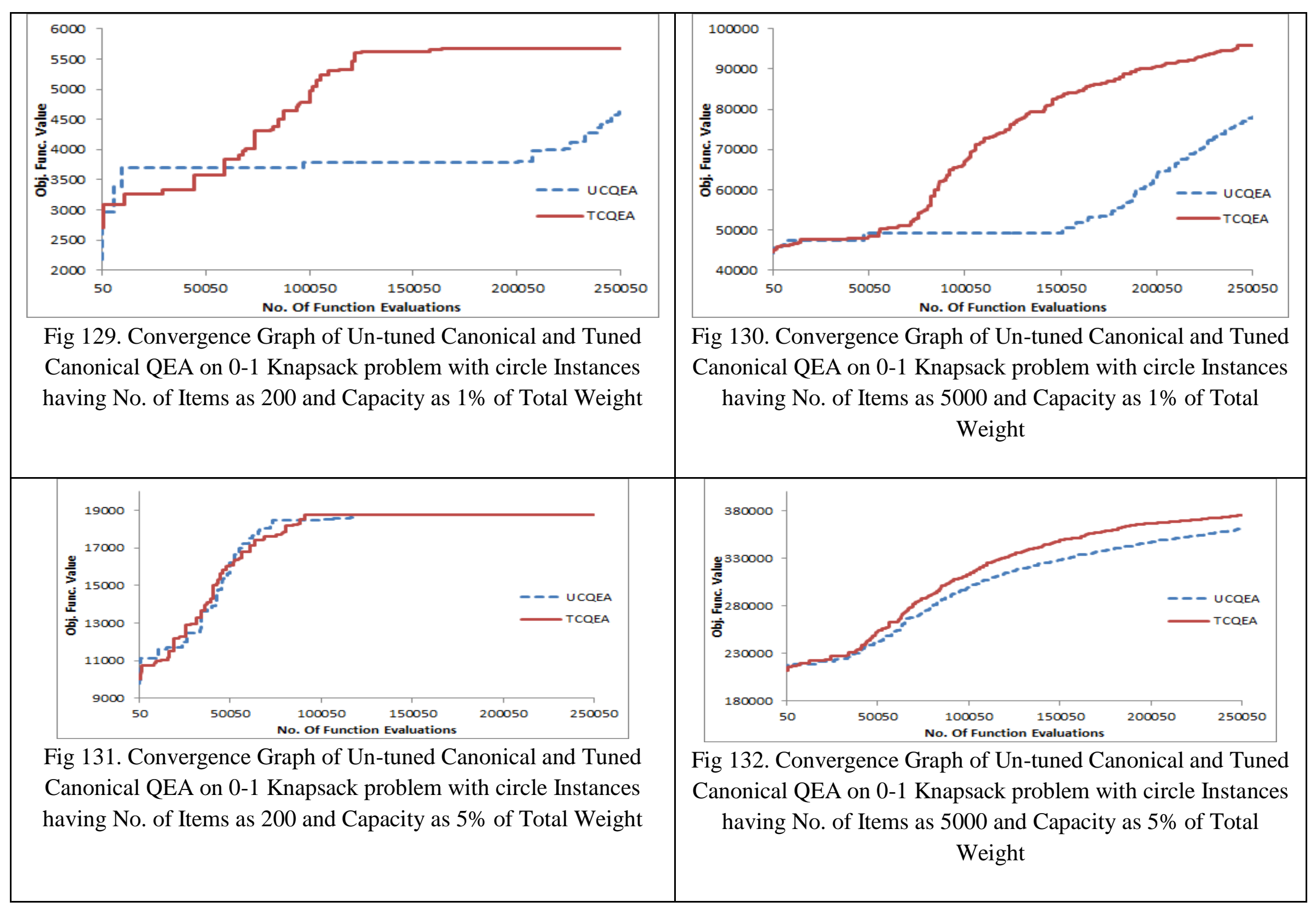

Fig 129. Convergence Graph of Un-tuned Canonical and Tuned Canonical QEA on 0-1 Knapsack problem with circle Instances having No. of Items as 200 and Capacity as 1% of Total Weight

Fig 130. Convergence Graph of Un-tuned Canonical and Tuned Canonical QEA on 0-1 Knapsack problem with circle Instances having No. of Items as 5000 and Capacity as 1% of Total Weight

Fig 131. Convergence Graph of Un-tuned Canonical and Tuned Canonical QEA on 0-1 Knapsack problem with circle Instances having No. of Items as 200 and Capacity as 5% of Total Weight

Fig 132. Convergence Graph of Un-tuned Canonical and Tuned Canonical QEA on 0-1 Knapsack problem with circle Instances having No. of Items as 5000 and Capacity as 5% of Total Weight

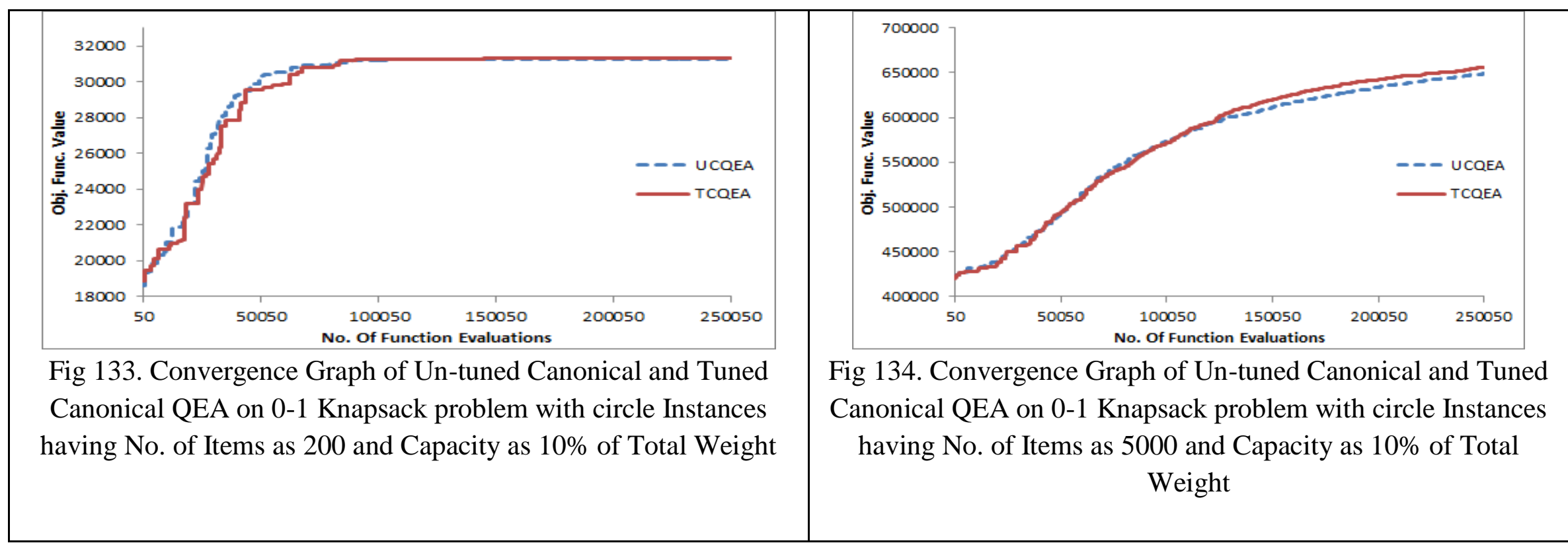

Fig 133. Convergence Graph of Un-tuned Canonical and Tuned Canonical QEA on 0-1 Knapsack problem with circle Instances having No. of Items as 200 and Capacity as 10% of Total Weight

Fig 134. Convergence Graph of Un-tuned Canonical and Tuned Canonical QEA on 0-1 Knapsack problem with circle Instances having No. of Items as 5000 and Capacity as 10% of Total Weight

In order to confirm the findings in Table 103 to 109, multi-problem non-parametric Wilcoxon's Signed Rank Test [Der2011] was performed on average objective function value (OFV) and average number of generation (Av. Gen.) of Tuned QEA and Canonical QEA on all the instances of Circle Knapsack problem at a significance level of 5%. In case of comparison on average OFV, the null hypothesis for comparison was average OFV of Tuned QEA $\mu_1$ is less than or equal to the average OFV of Canonical QEA $\mu_2$ i.e. $H_0$: $\mu_1 \leq \mu_2$ and the alternate hypothesis is H1: $\mu_1 > \mu_2$. The result of test is presented in Table 110, which shows that null hypothesis can be rejected safely as Wilcoxon's Signed Rank test statistic is 0 and less than the critical value of 303 at significance level of $\alpha = 5\%$. The sample size n= 41 (distinct) is larger than 25 so z statistic has also been computed under the normal distribution and p value obtained is 0.000, which is less than 0.05, so null hypothesis can be safely rejected. This indicates the success of the proposed tuning method for tuning QEA on problems like Circle Instances of Knapsack problem.

**Table 110: Wilcoxon's Signed Rank Test on Circle Instances (OFV)**

| | TCQEA | UCQEA |
|---|---|---|
| | **$X_1$** | **$X_2$** |
| 1 | 287.4 | 287.4 |
| 2 | 1332.5 | 1200 |
| 3 | 2496.3 | 2007.4 |
| 4 | 4400.1 | 3378 |
| 5 | 7612.7 | 6775 |
| 6 | 9106.2 | 8567.3 |
| 7 | 880.6 | 697.4 |
| 8 | 3207.5 | 2067.2 |
| 9 | 5522 | 3415.1 |
| 10 | 9362.5 | 5742.4 |
| 11 | 15613 | 12111.6 |
| 12 | 18504.9 | 15767.1 |
| 13 | 2085.2 | 1236.4 |
| 14 | 7086.4 | 3923.8 |
| 15 | 12341.6 | 6563.9 |
| 16 | 20901.3 | 11819.6 |
| 17 | 35527.8 | 25001.6 |
| 18 | 41971.4 | 32429.7 |
| 19 | 3937.8 | 1994.7 |
| 20 | 13039.4 | 6565.1 |
| 21 | 22044 | 11550.9 |
| 22 | 37721.7 | 21244 |
| 23 | 64260.7 | 44709.6 |
| 24 | 76472.1 | 57624.2 |

| | |
|---|---|
| Σ(+) | 861 |
| Σ(−) | 0 |
| *n* | 41 |

| Null Hypothesis | **Test Statistic** *T* | **Critical Value** At an α of 5% |
|---|---|---|
| $H_0$: $\mu_1$ <= $\mu_2$ | 0 | 303 |

**For Large Samples (*n* > 25)**

Test Statistic

| | | | |
|---|---|---|---|
| *z* | -5.57857 | *E[T]* | 430.5 |
| | | *σ(T)* | 77.17026629 |

| Null Hypothesis | *p*-value | At an α of 5% |
|---|---|---|
| $H_0$: $\mu_1$ <= $\mu_2$ | 0.0000 | **Reject** |
| 25 | 6600.8 | 3283.2 |
| 26 | 22644.4 | 11612.7 |
| 27 | 39066.6 | 21192.5 |
| 28 | 66512.3 | 39654.4 |
| 29 | 114633 | 81733.4 |
| 30 | 136515 | 104687 |
| 31 | 13300.1 | 6544.4 |
| 32 | 46979.6 | 25637.3 |
| 33 | 81120.2 | 48347.5 |
| 34 | 139685 | 92543.1 |
| 35 | 243410 | 188210 |
| 36 | 291834 | 238816 |
| 37 | 21354.4 | 11603.7 |
| 38 | 79496.6 | 48200.2 |
| 39 | 140017 | 92013.3 |
| 40 | 245692 | 178704 |
| 41 | 439199 | 360713 |
| 42 | 528850 | 455649 |

In case of comparison on Av. Gen., the null hypothesis for comparison was Av. Gen. of Tuned QEA $\mu_1$ is greater than or equal to the Av. Gen. of Canonical QEA $\mu_2$ i.e. $H_0$: $\mu_1 \geq \mu_2$ and the alternate hypothesis is H1: $\mu_1 < \mu_2$. The result of test is presented in Table 111, which shows that null hypothesis cannot be rejected as Wilcoxon's Signed Rank test statistic is 746 and more than the critical value of 319 at significance level of $\alpha$ = 5%, which indicates that Tuned QEA requires more number of generations as compared to Canonical QEA. The sample size n= 42 (distinct) is larger than 25 so z statistic has also been computed under the normal distribution and p value obtained is 0.9999, which is more than 0.05, so null hypothesis cannot be rejected. However, Tuned QEA is performing significantly better than Canonical QEA as shown by Table 110. This indicates the success of the proposed tuning method for tuning QEA on problems like Circle Instances, even though it requires more number of generations and function evaluations as compared to Canonical QEA.

**Table 111: Wilcoxon's Signed Rank Test on Circle Instances (Av. Gen.)**

| | TCQEA | UCQEA |
|---|---|---|
| | **$X_1$** | **$X_2$** |
| 1 | 2.3 | 2.5 |
| 2 | 322.4 | 561.2 |
| 3 | 287.4 | 490.7 |
| 4 | 392 | 512.6 |
| 5 | 629.2 | 971 |
| 6 | 613 | 965.5 |
| 7 | 293.9 | 513.9 |
| 8 | 554.4 | 445.5 |
| 9 | 727 | 606.3 |
| 10 | 816.4 | 495.4 |
| 11 | 871.3 | 965 |
| 12 | 882 | 971.7 |
| 13 | 746.3 | 479.1 |
| 14 | 913.1 | 545.6 |
| 15 | 939.4 | 521 |
| 16 | 954.9 | 687 |
| 17 | 969.8 | 969.2 |
| 18 | 971.3 | 983.6 |

| | |
|---|---|
| $\Sigma(+)$ | 746 |
| $\Sigma(-)$ | 157 |
| *n* | 42 |

| Null Hypothesis | **Test Statistic** *T* | **Critical Value** At an $\alpha$ of 5% |
|---|---|---|
| $H_0$: $\mu_1$ >= $\mu_2$ | 746 | 319 |

**For Large Samples (*n* > 25)**

Test Statistic

| | | | |
|---|---|---|---|
| *z* | -3.68233 | *E[T]* | 451.5 |
| | | *σ(T)* | 79.97655907 |

| Null Hypothesis | *p*-value | At an $\alpha$ of 5% |
|---|---|---|
| $H_0$: $\mu_1$ >= $\mu_2$ | 0.9999 | |
| 19 | 870.5 | 484.1 |
| 20 | 940.5 | 553.5 |
| 21 | 979.6 | 494.1 |
| 22 | 982.8 | 641.7 |
| 23 | 979.2 | 977.9 |
| 24 | 984.9 | 983 |
| 25 | 917.9 | 478 |
| 26 | 978.1 | 508.2 |
| 27 | 984.4 | 599 |
| 28 | 987.9 | 629 |
| 29 | 988.2 | 972 |
| 30 | 988.4 | 978.3 |
| 31 | 966.1 | 597.4 |
| 32 | 985.7 | 447.2 |
| 33 | 988.4 | 585.6 |
| 34 | 984.7 | 752.7 |
| 35 | 990.2 | 967.7 |
| 36 | 991.3 | 982.9 |
| 37 | 963.2 | 542.6 |
| 38 | 979.1 | 527.8 |
| 39 | 988 | 587.1 |
| 40 | 994.1 | 662.7 |
| 41 | 993.4 | 964.3 |
| 42 | 992.9 | 982 |

### *D. P-PEAKS Problem* [43], [47]

It is a multimodal problem generator, which is an easily parameterizable task with a tunable degree of difficulty. The advantage of using problem generator is that it removes the opportunity to hand-tune algorithms to a particular problem, thus, allows a large fairness while comparing the performance of different algorithms or different instances of same algorithm. It helps in evaluating the algorithms on a high number of random problem instances, so that the predictive power of the results for the problem class as a whole is very high.

The idea of P-PEAKS is to generate P random N-bit strings that represent the location of P peaks in the search space. The fitness value of a string is the hamming distance between this string and the closest peak, divided by N (as shown in Eq. 10). Using a higher (or lower) number of peaks we obtain more (or less) epistatic problems. The maximum fitness value for the problem instances is 1.0

$$f_{P-PEAKS(\vec{x})} = \frac{1}{N} \max_{1 \le i \le p}\{\mathrm{N} - \mathrm{Hamming\ D}(\vec{\mathrm{x}}, \mathrm{Peak_i})\} \qquad (10)$$

The QEA was tuned by using the proposed framework on problem with P=100 and N=1000. The initial range of values for parameters in QEA used for tuning is given in Table 112. The parameter range for magnitude of rotation angles (θ1, θ2, θ4, θ6, θ7 & θ8) is 0 to 0.05π as they change by small magnitude as compared to θ3 & θ5, whose range is from 0 to 0.5π, which is very large as compared to the range suggested by [14]. The direction of rotation depends on the sign of α, β and relative fitness as per Table 1. The range for population size is 10 to 200, and covers the values for similar parameter for most studies in Evolutionary Algorithms. The range for group size is 1 to 20, which is four times bigger than the value suggested in [14]. The range for global migration is 1 to 500, which is again five times the value suggested by [14]. The change in value of each parameter during the tuning process is depicted in Table 113 and shown in Figures 135 to 145. Initially, there were three rounds for exploration stage and one round for exploitation stage after which optimal value was reached in every independent run, however, the average number of function evaluations was 3,45,444.5, which was considered high. It was decided to further tune the QEA for reducing the average number of function evaluation without sacrificing the quality of objective function value, so, it was decided to run the Exploration stage again with the best set of parameter vector found so far as the PIVOT. The average number of function evaluations was reduced to almost half to 1,76,830. Then again exploitation was carried out to further reduce the average number of function evaluations to 1,57,808.934 and it was decided to stop further tuning to conserve resources.

TABLE 112
INITIAL RANGE OF PARAMETERS OF CANONICAL QEA

| Parameter | θ1 (* π) | θ2 (* π) | θ3 (* π) | θ4 (* π) | θ5 (* π) | θ6 (* π) | θ7 (* π) | θ8 (* π) | Pop size | Group size | Global Migration |
|---|---|---|---|---|---|---|---|---|---|---|---|
| Lower Limit | 0 | 0 | 0 | 0 | 0 | 0 | 0 | 0 | 5 | 1 | 1 |
| Upper Limit | 0.05 | 0.05 | 0.5 | 0.05 | 0.5 | 0.05 | 0.05 | 0.05 | 200 | 20 | 500 |

TABLE 113
BEST PARAMETER VECTOR (PIVOT) DURING TUNING PROCESS

| Iter. No. | Best Pivot Parameter Value | | | | | | | | | | | Av. OFV |
|---|---|---|---|---|---|---|---|---|---|---|---|---|
| | θ1 (* π) | θ2 (* π) | θ3 (* π) | θ4 (* π) | θ5 (* π) | θ6 (* π) | θ7 (* π) | θ8 (* π) | Pop. Size | Grp. No. | Glb. Mig. | |
| Explor. – 1 | 0.0035 | 0.0035 | 0.015 | 0.0035 | 0.015 | 0.0035 | 0.0035 | 0.0035 | 72 | 8 | 200 | 0.93 |
| Explor. – 2 | 0.0035 | 0 | 0.015 | 0.00175 | 0.018 | 0.0035 | 0.0035 | 0 | 90 | 10 | 100 | 0.99 |
| Explor. – 3 | 0.0035 | 0 | 0.406 | 0.00133 | 0.0726 | 0 | 0.003 | 0 | 144 | 16 | 383 | 1.0 |
| Exploit. – 4 | 0.0037 | 0 | 0.38 | 0.001 | 0.0739 | 0 | 0.009 | 0.008 | 136 | 8 | 369 | 1.0 |
| Explor. – 5 | 0.0182 | 0 | 0.184 | 0.0764 | 0.0768 | 0 | 0.0107 | 0.0803 | 130 | 5 | 130 | 1.0 |
| Exploit. - 6 | 0.0184 | 0 | 0.169 | 0.0784 | 0.0768 | 0 | 0.0163 | 0.0818 | 132 | 4 | 125 | **1.0** |

The parameter value for θ1 remained constant during initial exploration & exploitation and then increased during final exploration & exploitation stages. The parameter value for θ2 initially decreased to 0 and then remained constant in subsequent stages of exploration & exploitation. The parameter value for θ3 increased during initial exploration and then decreased during subsequent stages. The parameter value for θ4 decreased during initial exploration & exploitation and then increased during subsequent exploration and exploitation stages. The parameter value for θ5 increased during initial exploration and then increased a little during subsequent stages. The parameter value for θ6 initially remained constant and then decreased during exploration and then remained constant during subsequent stages. The parameter value for θ7 remained constant during initial exploration and then kept on increasing during subsequent stages. The parameter value for θ8 decreased during initial exploration and then increased during subsequent stages. The parameter value for Population Size increased during initial exploration and then decreased slightly during subsequent stages. The parameter value for group size increased during initial exploration and then decreased during subsequent stages. The parameter value for Global Migration decreased and then increased during initial exploration and then remained almost constant during initial exploitation and decreased substantially during subsequent stages. The final value of each parameter is given in Table 113 in the last row. The convergence graph given in figure 146 shows convergence to optimal within 2000 generations.

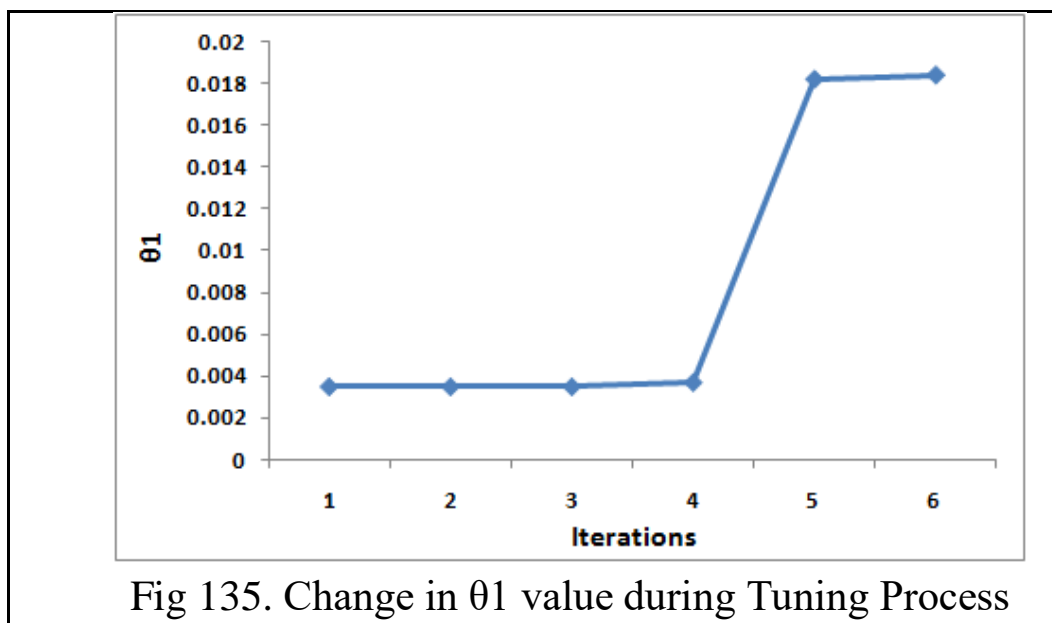

Fig 135. Change in θ1 value during Tuning Process

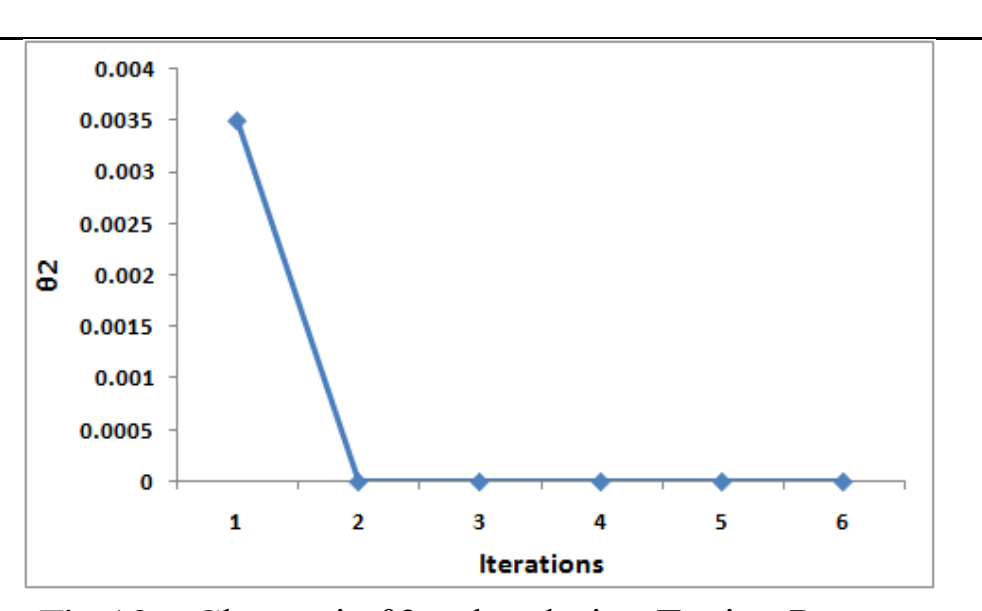

Fig 136. Change in θ2 value during Tuning Process

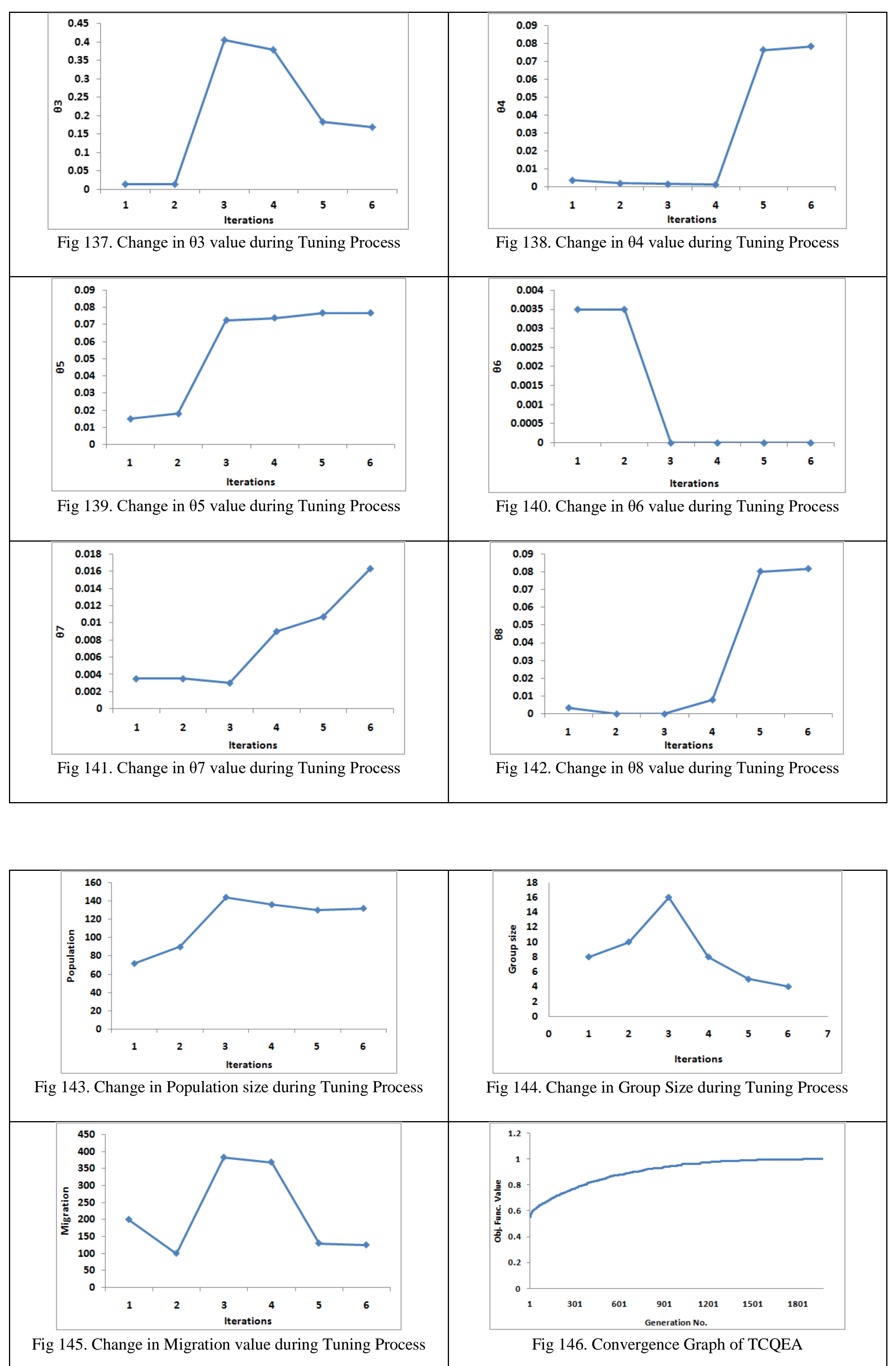

Fig 137. Change in θ3 value during Tuning Process

Fig 138. Change in θ4 value during Tuning Process

Fig 139. Change in θ5 value during Tuning Process

Fig 140. Change in θ6 value during Tuning Process

Fig 141. Change in θ7 value during Tuning Process

Fig 142. Change in θ8 value during Tuning Process

Fig 143. Change in Population size during Tuning Process

Fig 144. Change in Group Size during Tuning Process

Fig 145. Change in Migration value during Tuning Process

Fig 146. Convergence Graph of TCQEA

The deviation from optimal for large variation in parameter setting, which is computed from the average results of fifty runs from the first iteration of exploration stage is 0.09 whereas the deviation from optimal for small variation in parameter setting, which is computed from the average results of twenty seven runs from the first iteration of exploitation stage is zero. Therefore, the tuned QEA is robust to small variation in parameter set, and is also stable for large variation in parameter set. Thus, the deviation from the known optimal at the end of the iteration can be used for deciding the need for continuing the search in that stage. However, if the Optimal is unknown then this strategy cannot be applied with the same confidence, but in case of real world problems, often a best known solution is available, so in such cases, it may be used in place of the Optimal.

The comparative study performed between parameter Tuned QEA (TCQEA) and Canonical QEA (UCQEA) with Parameters given in Table 2 on a set of twenty one instances of P Peaks problem with P = 1 to 1000 peaks of length N = 1000 bits each. The results are given in Table 114. Thirty independent runs were made on each problem size and the comparison has been made on Best, Median, Worst, Mean, percentage of runs in which optimal was achieved i.e. percentage of Success runs, and Average number of function evaluations (NFE). The Canonical QEA was able to reach optimal value till problem size 700, but was not able to find the optimal for rest of the problem instances whereas the Tuned-QEA was able to reach 100% success in all the problem instances. The performance of Tuned QEA was also good on speed of convergence as indicated by average NFE and the convergence graphs shown in figures 147 to 167, which also compared the speed of convergence of Tuned QEA to Canonical QEA.

The convergence graphs have been plotted between objective function value and number of generations for both Tuned QEA and Canonical QEA for all the problem instances of P-PEAKS for the median run. The convergence graph clearly establishes the superiority of Tuning as the Tuned QEA is able to reach the near optimal in less than 2000 generations in all the graphs whereas the Canonical QEA is much slower and takes much larger number of generations to reach near the optimal.

TABLE 114
COMPARATIVE STUDY BETWEEN TCQEA AND UCQEA ON P-PEAKS PROBLEM INSTANCES

| No. Of Peaks | Algo | Best | Worst | Avg. | Median | % Success Runs | Std | Average NFE |
|---|---|---|---|---|---|---|---|---|
| 1 | UCQEA | 1.0000 | 1.0000 | 1.0000 | 1.0000 | 100.0 | 0.0000 | 115592 |
| | **TCQEA** | **1.0000** | **1.0000** | **1.0000** | **1.0000** | **100.0** | **0.0000** | **97106** |
| 5 | UCQEA | 1.0000 | 0.9750 | 0.9961 | 0.9980 | 33.3 | 0.0054 | 146393 |
| | **TCQEA** | **1.0000** | **1.0000** | **1.0000** | **1.0000** | **100.0** | **0.0000** | **99634** |
| 10 | UCQEA | 1.0000 | 0.9690 | 0.9893 | 0.9905 | 16.7 | 0.0095 | 148433 |
| | TCQEA | 1.0000 | 1.0000 | 1.0000 | 1.0000 | 100.0 | 0.0000 | 97085 |
| 20 | UCQEA | 1.0000 | 0.9770 | 0.9878 | 0.9870 | 6.7 | 0.0068 | 149448 |
| | TCQEA | 1.0000 | 0.9990 | 1.0000 | 1.0000 | 96.6 | 0.0002 | 99803 |
| 30 | UCQEA | 1.0000 | 0.9680 | 0.9887 | 0.9890 | 6.7 | 0.0075 | 149817 |
| | TCQEA | 1.0000 | 1.0000 | 1.0000 | 1.0000 | 100.0 | 0.0000 | 98480 |
| 40 | UCQEA | 0.9990 | 0.9720 | 0.9867 | 0.9870 | 0.0 | 0.0072 | 150050 |
| | TCQEA | 1.0000 | 1.0000 | 1.0000 | 1.0000 | 100.0 | 0.0000 | 98674 |
| 50 | UCQEA | 0.9980 | 0.9560 | 0.9798 | 0.9810 | 0.0 | 0.0088 | 150050 |
| | TCQEA | 1.0000 | 1.0000 | 1.0000 | 1.0000 | 100.0 | 0.0000 | 102432 |
| 60 | UCQEA | 1.0000 | 0.9700 | 0.9846 | 0.9835 | 3.3 | 0.0077 | 149988 |
| | TCQEA | 1.0000 | 1.0000 | 1.0000 | 1.0000 | 100.0 | 0.0000 | 100749 |
| 70 | UCQEA | 0.9960 | 0.9690 | 0.9859 | 0.9875 | 0.0 | 0.0074 | 150050 |
| | TCQEA | 1.0000 | 1.0000 | 1.0000 | 1.0000 | 100.0 | 0.0000 | 98008 |

| 80 | UCQEA | 0.9990 | 0.9660 | 0.9848 | 0.9865 | 0.0 | 0.0088 | 150050 |
|---|---|---|---|---|---|---|---|---|
| | TCQEA | 1.0000 | 0.9990 | 1.0000 | 1.0000 | 96.6 | 0.0002 | 101893 |
| 90 | UCQEA | 0.9960 | 0.9580 | 0.9842 | 0.9840 | 0.0 | 0.0084 | 150050 |
| | TCQEA | 1.0000 | 1.0000 | 1.0000 | 1.0000 | 100.0 | 0.0000 | 97146 |
| 100 | UCQEA | 0.9990 | 0.9710 | 0.9847 | 0.9840 | 0.0 | 0.0073 | 150050 |
| | TCQEA | 1.0000 | 1.0000 | 1.0000 | 1.0000 | 100.0 | 0.0000 | 96928 |
| 200 | UCQEA | 0.9910 | 0.9630 | 0.9811 | 0.9835 | 0.0 | 0.0075 | 150050 |
| | TCQEA | 1.0000 | 1.0000 | 1.0000 | 1.0000 | 100.0 | 0.0000 | 98459 |
| 300 | UCQEA | 1.0000 | 0.9690 | 0.9838 | 0.9840 | 3.3 | 0.0068 | 150027 |
| | TCQEA | 1.0000 | 0.9990 | 1.0000 | 1.0000 | 96.6 | 0.0002 | 99822 |
| 400 | UCQEA | 0.9950 | 0.9600 | 0.9827 | 0.9845 | 0.0 | 0.0082 | 150050 |
| | TCQEA | 1.0000 | 0.9990 | 1.0000 | 1.0000 | 96.6 | 0.0002 | 99416 |
| 500 | UCQEA | 0.9940 | 0.9660 | 0.9816 | 0.9825 | 0.0 | 0.0074 | 150050 |
| | TCQEA | 1.0000 | 0.9990 | 1.0000 | 1.0000 | 96.6 | 0.0002 | 100646 |
| 600 | UCQEA | 0.9950 | 0.9630 | 0.9804 | 0.9810 | 0.0 | 0.0088 | 150050 |
| | TCQEA | 1.0000 | 1.0000 | 1.0000 | 1.0000 | 100.0 | 0.0000 | 98568 |
| 700 | UCQEA | 0.9940 | 0.9620 | 0.9802 | 0.9840 | 0.0 | 0.0088 | 150050 |
| | TCQEA | 1.0000 | 1.0000 | 1.0000 | 1.0000 | 100.0 | 0.0000 | 97398 |
| 800 | UCQEA | 0.9940 | 0.9620 | 0.9785 | 0.9790 | 0.0 | 0.0081 | 150050 |
| | TCQEA | 1.0000 | 1.0000 | 1.0000 | 1.0000 | 100.0 | 0.0000 | 98970 |
| 900 | UCQEA | 0.9940 | 0.9620 | 0.9815 | 0.9840 | 0.0 | 0.0092 | 150050 |
| | TCQEA | 1.0000 | 1.0000 | 1.0000 | 1.0000 | 100.0 | 0.0000 | 98654 |
| 1000 | UCQEA | 0.9940 | 0.9690 | 0.9817 | 0.9840 | 0.0 | 0.0058 | 150050 |
| | TCQEA | 1.0000 | 1.0000 | 1.0000 | 1.0000 | 100.0 | 0.0000 | 100277 |

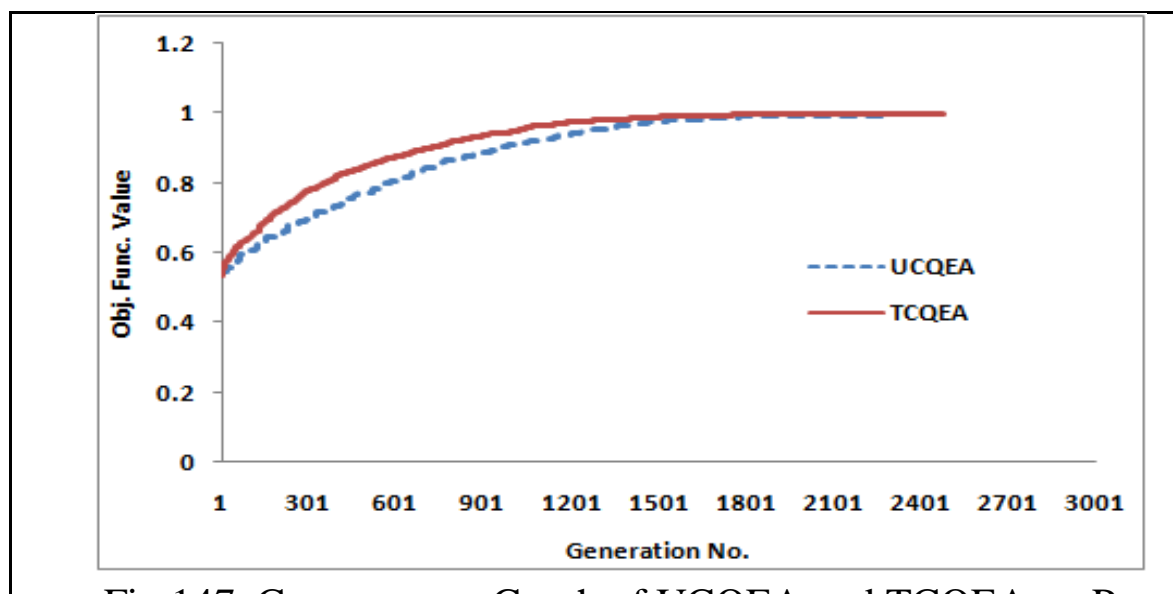

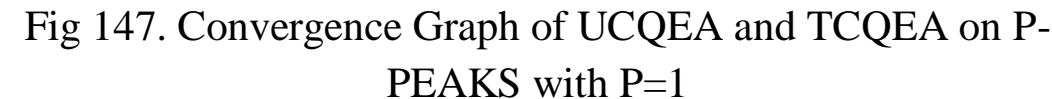
Fig 147. Convergence Graph of UCQEA and TCQEA on P-PEAKS with P=1

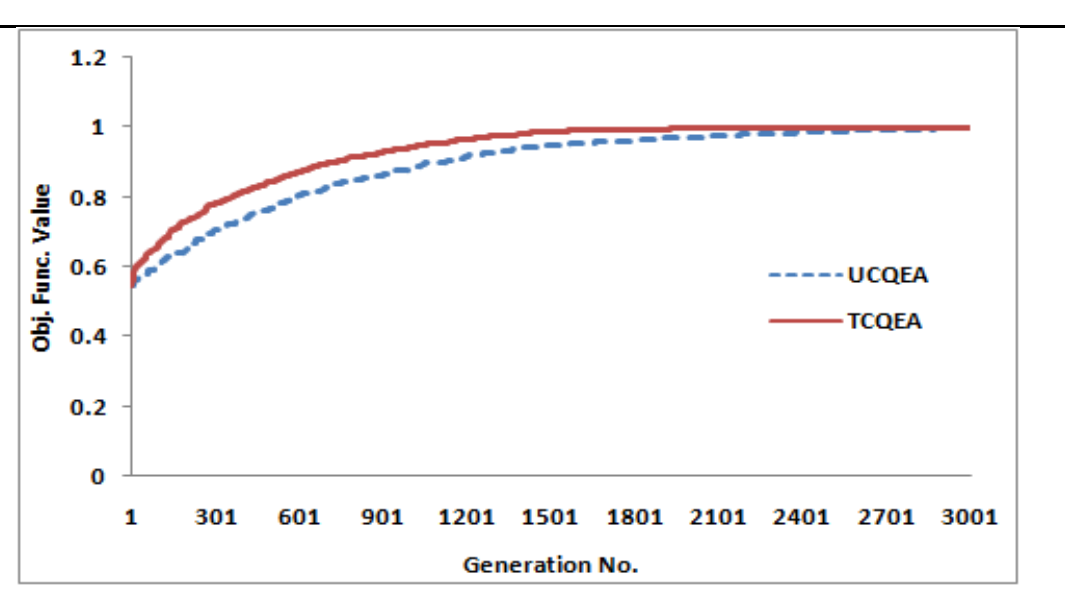

Fig 148. Convergence Graph of UCQEA and TCQEA on P-PEAKS with P=5

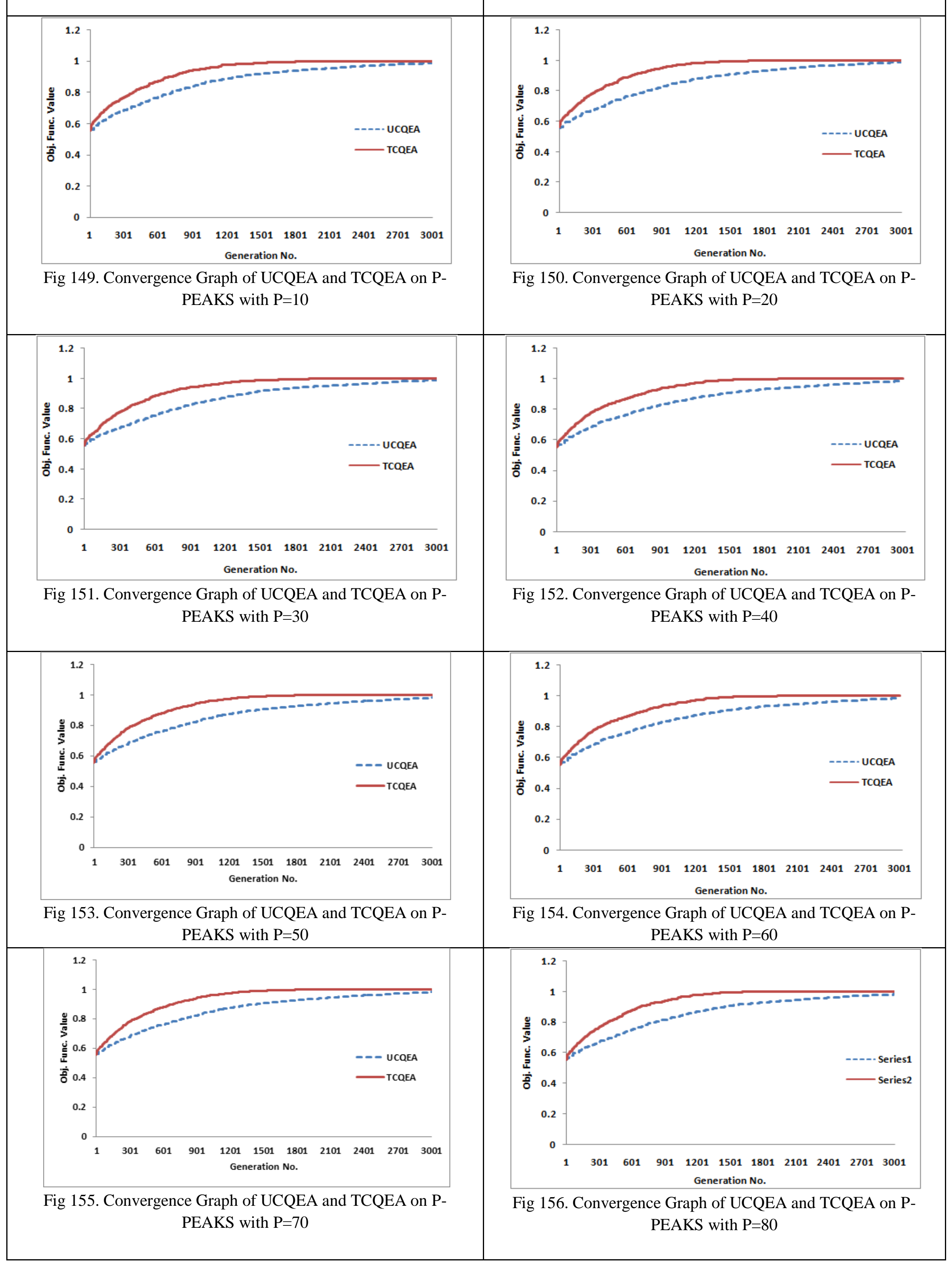

Fig 149. Convergence Graph of UCQEA and TCQEA on P-PEAKS with P=10

Fig 150. Convergence Graph of UCQEA and TCQEA on P-PEAKS with P=20

Fig 151. Convergence Graph of UCQEA and TCQEA on P-PEAKS with P=30

Fig 152. Convergence Graph of UCQEA and TCQEA on P-PEAKS with P=40

Fig 153. Convergence Graph of UCQEA and TCQEA on P-PEAKS with P=50

Fig 154. Convergence Graph of UCQEA and TCQEA on P-PEAKS with P=60

Fig 155. Convergence Graph of UCQEA and TCQEA on P-PEAKS with P=70

Fig 156. Convergence Graph of UCQEA and TCQEA on P-PEAKS with P=80

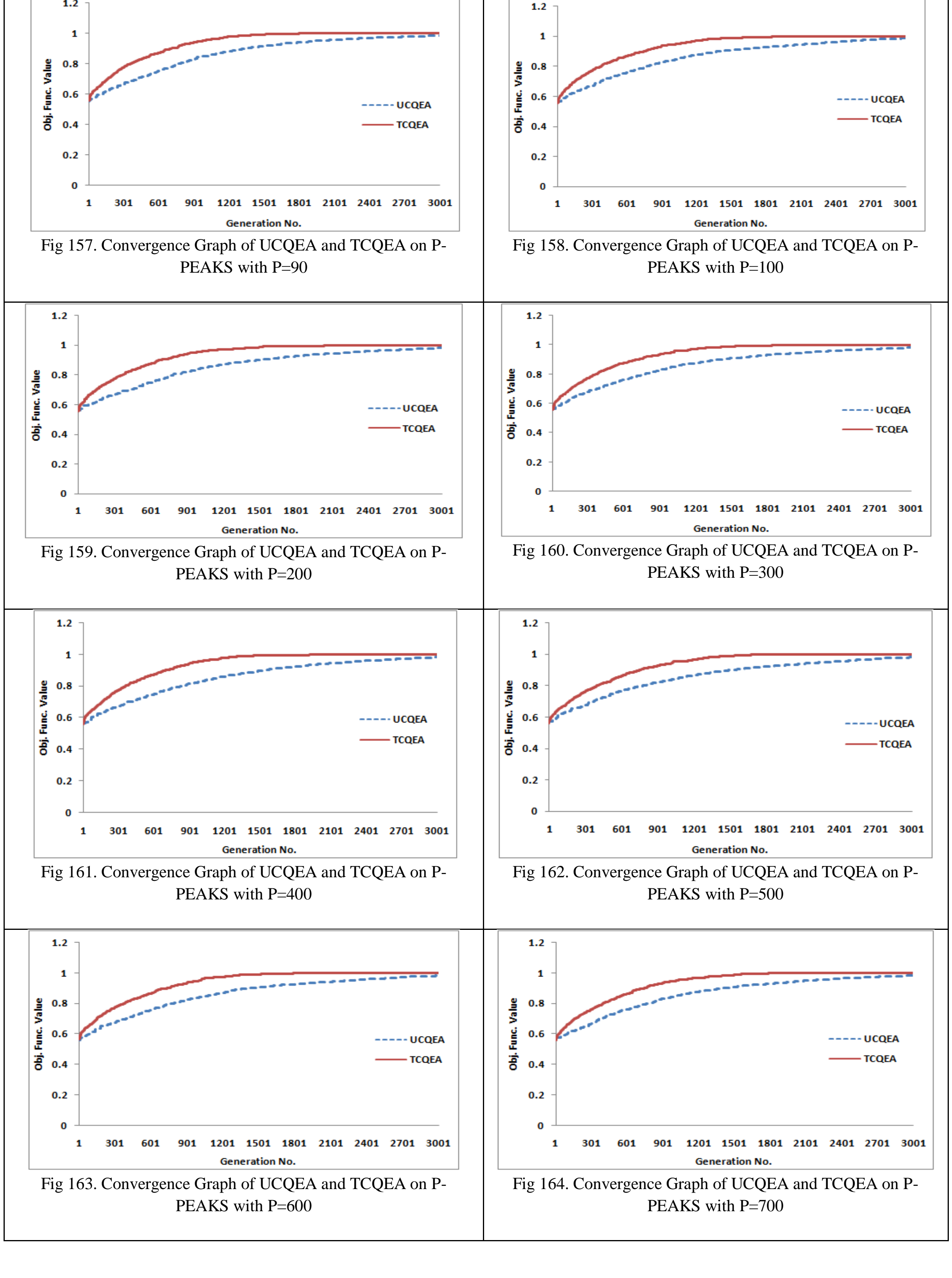

Fig 157. Convergence Graph of UCQEA and TCQEA on P-PEAKS with P=90

Fig 158. Convergence Graph of UCQEA and TCQEA on P-PEAKS with P=100

Fig 159. Convergence Graph of UCQEA and TCQEA on P-PEAKS with P=200

Fig 160. Convergence Graph of UCQEA and TCQEA on P-PEAKS with P=300

Fig 161. Convergence Graph of UCQEA and TCQEA on P-PEAKS with P=400

Fig 162. Convergence Graph of UCQEA and TCQEA on P-PEAKS with P=500

Fig 163. Convergence Graph of UCQEA and TCQEA on P-PEAKS with P=600

Fig 164. Convergence Graph of UCQEA and TCQEA on P-PEAKS with P=700

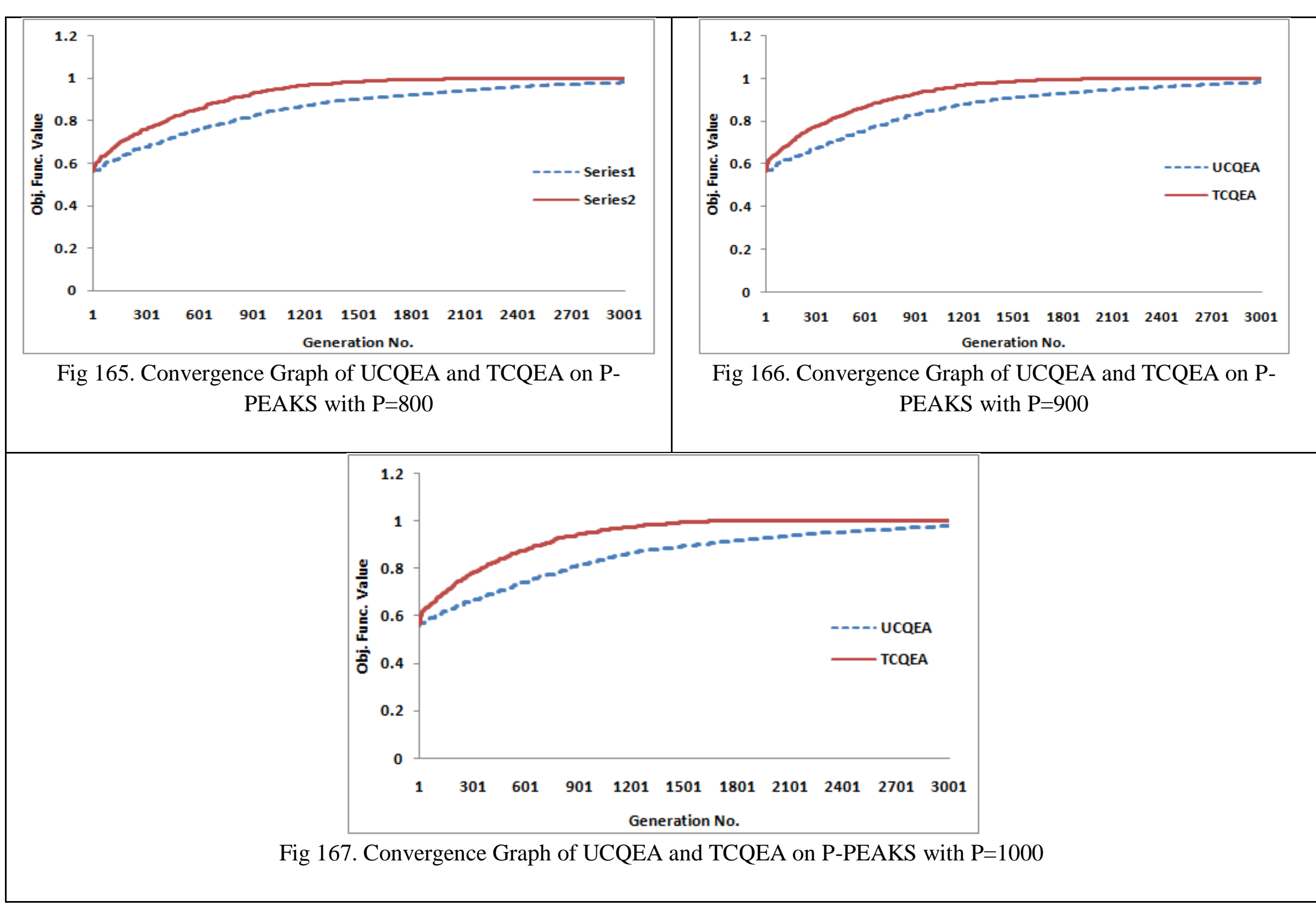

Fig 165. Convergence Graph of UCQEA and TCQEA on P-PEAKS with P=800

Fig 166. Convergence Graph of UCQEA and TCQEA on P-PEAKS with P=900

Fig 167. Convergence Graph of UCQEA and TCQEA on P-PEAKS with P=1000

The performance of Tuned QEA is superior to Canonical QEA for all instances of P-PEAKS Problem used in this work as indicated by the Table 114 and Figures 147 to 167. In order to confirm the findings in Table 114, multi-problem non-parametric Wilcoxon's Signed Rank Test [Der2011] was performed on average objective function value (OFV) and average NFE of Tuned QEA and Canonical QEA on all the instances of P-PEAKS at a significance level of 5%. In case of comparison on average OFV, the null hypothesis for comparison was average OFV of Tuned QEA $\mu_1$ is less than or equal to the average OFV of Canonical QEA $\mu_2$ i.e. $H_0$: $\mu_1 \leq \mu_2$ and the alternate hypothesis is H1: $\mu_1 > \mu_2$. The result of test is presented in Table 115, which shows that null hypothesis can be safely rejected as Wilcoxon's Signed Rank test statistic is zero and less than the critical value of 60 at significance level of $\alpha$ = 5%. This indicates the success of the proposed tuning method for tuning QEA on problems like P-PEAKS.

**Table 115: Wilcoxon's Signed Rank Test on P-PEAKS Instances (OFV)**

| | TCQEA | UCQEA |
|---|---|---|
| | $X_1$ | $X_2$ |
| 1 | 1 | 1 |
| 2 | 1 | 0.9961 |
| 3 | 1 | 0.9893 |
| 4 | 1 | 0.9878 |
| 5 | 1 | 0.9887 |
| 6 | 1 | 0.9867 |
| 7 | 1 | 0.9798 |
| 8 | 1 | 0.9846 |
| 9 | 1 | 0.9859 |
| 10 | 1 | 0.9848 |
| 11 | 1 | 0.9842 |
| 12 | 1 | 0.9847 |
| 13 | 1 | 0.9811 |
| 14 | 1 | 0.9838 |
| 15 | 1 | 0.9827 |
| 16 | 1 | 0.9816 |

| | |
|---|---|
| $\Sigma(+)$ | 210 |
| $\Sigma(-)$ | 0 |
| $n$ | 20 |

| Null Hypothesis | Test Statistic $T$ | Critical Value At an $\alpha$ of 5% |
|---|---|---|
| $H_0$: $\mu_1$ <= $\mu_2$ | 0 | 60 |
| 17 | 1 | 0.9804 |
| 18 | 1 | 0.9802 |
| 19 | 1 | 0.9785 |
| 20 | 1 | 0.9815 |
| 21 | 1 | 0.9817 |

In case of comparison on average NFE, the null hypothesis for comparison was average NFE of Tuned QEA $\mu_1$ is greater than or equal to the average NFE of Canonical QEA $\mu_2$ i.e. $H_0: \mu_1 \geq \mu_2$ and the alternate hypothesis is $\mu_1 < \mu_2$. The result of test is presented in Table 116, which shows that null hypothesis can be safely rejected as Wilcoxon's Signed Rank test statistic is zero and less than the critical value of 60 at significance level of $\alpha = 5\%$. This indicates the success of the proposed tuning method for tuning QEA on problems like P-PEAKS.

**Table 116: Wilcoxon's Signed Rank Test on P-PEAKS Instances (NFE)**

| | TCQEA | UCQEA |
|---|---|---|
| | **$X_1$** | **$X_2$** |
| 1 | 97106 | 115592 |
| 2 | 99634 | 146393 |
| 3 | 97085 | 148433 |
| 4 | 99803 | 149448 |
| 5 | 98480 | 149817 |
| 6 | 98674 | 150050 |
| 7 | 102432 | 150050 |
| 8 | 100749 | 149988 |
| 9 | 98008 | 150050 |
| 10 | 101893 | 150050 |
| 11 | 97146 | 150050 |
| 12 | 96928 | 150050 |
| 13 | 98459 | 150050 |
| 14 | 99822 | 150027 |
| 15 | 99416 | 150050 |
| 16 | 100646 | 150050 |
| 17 | 98568 | 150050 |
| 18 | 97398 | 150050 |
| 19 | 98970 | 150050 |
| 20 | 98654 | 150050 |
| 21 | 100277 | 150050 |

| | |
|---|---|
| $\Sigma(+)$ | 0 |
| $\Sigma(-)$ | 231 |
| $n$ | 21 |

| Null Hypothesis | **Test Statistic** $T$ | **Critical Value** At an $\alpha$ of 5% |
|---|---|---|
| $H_0: \mu_1 >= \mu_2$ | 0 | 60 |

## IV. CONCLUSIONS

This paper focuses on working of Quantum Rotation operator used in QEA for solving combinatorial optimization problem. The Quantum Rotation operator has eight parameters, which are problem dependent and require tuning. There are several methods of parameter tuning available in literature, however, the total number of parameters to tune QEA is identified as eleven, which most of the current techniques are not capable of handling. Therefore, a new heuristic parameter tuning method was proposed for tuning algorithms like QEA with relatively large number of parameters. The proposed parameter tuning method was designed after considering the recommendation made in [27]. It is an improvement on Calibra [34] i.e. it can handle more number of parameters with more numbers of levels without using factorial number of experiments in initial stages of tuning. The proposed tuning method is a multi-stage, iterative, metaheuristic approach that uses Taguchi's approach in evaluating and generating parameter vectors for experiments. The performance of proposed tuning method has been tested by tuning QEA on four set of benchmark combinatorial problems. It required QEA to tune only on one instance of the benchmark problem, whereas in Calibra, a training set of problem instances was required to tune algorithms. After tuning, QEA was able to solve several instance of same class of problems successfully. Thus, proposed tuning method is a novel, simple and effective technique for tuning QEAs for solving combinatorial optimization problems.